%% file: main.tex
\ifdefined\XeTeXversion\else\pdfoutput=1\fi
\PassOptionsToPackage{table}{xcolor}
\documentclass[review]{fcs}
\ifdefined\XeTeXversion
 \microtypesetup{expansion=false}
\fi
\usepackage{multirow}
\usepackage{makecell}
\usepackage{overpic}
\usepackage{xspace}
\input{survey_macros}
\input{layout_macros}

\title{A Survey on Multimodal Benchmarks:\\ In the Era of Large AI Models}
\shorttitle{A Survey on Multimodal Benchmarks}

\author[1,2]{Lin~Li}
\author[3]{Yihan~Shao}
\author[1]{Guikun~Chen}
\author[2,4]{Haochen Shi}
\author[3]{Jiaming~Lei}
\author[3]{Hanrong~Shi}
\author[3]{Jun~Xiao}
\author[1,+]{Long~Chen}
\address[1]{The Hong Kong University of Science and Technology, Hong Kong, China}
\address[2]{AI Chip Center for Emerging Smart Systems, Hong Kong, China}
\address[3]{Zhejiang University, Hangzhou, China}
\address[4]{The Hong Kong Polytechnic University, Hong Kong, China}
\corremail{lllidy@ust.hk (Lin Li); shaoyihan@zju.edu.cn (Yihan Shao);
 guikunchen@gmail.com (Guikun Chen); haochen8.shi@connect.polyu.hk (Haochen Shi);
 jmlei@zju.edu.cn (Jiaming Lei);
 hanrong@zju.edu.cn (Hanrong Shi); junx@zju.edu.cn (Jun Xiao);
 longchen@ust.hk (Long Chen, corresponding author)}

\fcssetup{doi={}, received={}, accepted={}, copyright={}, first-page={1}}
\ExplSyntaxOn
\tl_set:Nn \l__fcs_doi_url_tl { }
\tl_set:Nn \l__fcs_short_detailed_page_tl { Preprint }
\tl_set:Nn \l__fcs_long_detailed_page_tl
 {
 \footnotesize\textorange{Preprint\ $\vert$\ }
 \normalsize\int_use:c { c@page }
 }
\ExplSyntaxOff

\begin{abstract}
The rapid evolution of Multimodal Large Language Models (MLLMs) has brought substantial advancements in artificial intelligence, significantly enhancing the capability to understand and generate multimodal content. As MLLMs advance toward omni-modal perception, reflective self-improvement, unified synthesis, and autonomous agency, evaluation has undergone a fundamental departure from traditional static, task-isolated leaderboards. In this paper, we present a systematic and comprehensive overview of \textbf{358 peer-reviewed multimodal benchmarks} published between 2024 and 2026. Specifically, we structure this rapidly expanding landscape into five core evaluation pillars, namely \textbf{understanding}, \textbf{reasoning}, \textbf{generation}, \textbf{trustworthiness and safety}, and \textbf{application}, examining their task formulations, capability frontiers, and specialized evaluation metrics across diverse modalities and deployment scenarios. Beyond cataloging individual suites, we review end-to-end dataset construction and quality-control pipelines, and critically analyze key methodological pitfalls governing cross-benchmark comparison validity, including unimodal shortcuts, acausal decision protocols, closed-loop sandbox bottlenecks, and cross-stage biases in LLM-based data synthesis and judging. Building on this analysis, we distill actionable benchmark design blueprints and a pre-release validity checklist to guide reliable evaluation design. We hope that this survey will contribute to the ongoing advancement of MLLM research by offering a comprehensive overview of benchmarking practices and identifying promising directions for future work. An associated GitHub repository collecting the latest papers is available at \href{https://github.com/HKUST-LongGroup/Awesome-MLLM-Benchmarks}{https://github.com/HKUST-LongGroup/Awesome-MLLM-Benchmarks}.
\end{abstract}

\begin{document}
\input{sec/intro.tex}
\input{sec/undertanding.tex}

\input{sec/reasoning.tex}

\input{sec/generation.tex}

\input{sec/trustworthiness.tex}

\input{sec/application.tex}

\input{sec/dataset_and_eval.tex}
\input{sec/cross_benchmark.tex}
\input{sec/direction.tex}
\input{sec/conclusion.tex}

\begingroup
\fontsize{8.5pt}{9.85pt}\selectfont
\bibliographystyle{fcs}

\bibliography{references}
\endgroup
\end{document}

%% file: survey_macros.tex
\usepackage{etoolbox}
\definecolor{mygray}{gray}{.85}
\definecolor{mygray1}{gray}{.7}
\definecolor{mygray2}{gray}{.93}
\newcommand{\humantag}[1]{\tikz[baseline=(X.base)]\node[draw=blue,fill=cyan!40,semithick,rectangle,inner sep=1.5pt,rounded corners=3pt] (X) {#1};\allowbreak}
\newcommand{\ruletag}[1]{\tikz[baseline=(X.base)]\node[draw=blue,fill=green!30,semithick,rectangle,inner sep=1.5pt,rounded corners=3pt] (X) {#1};\allowbreak}
\newcommand{\llmtag}[1]{\tikz[baseline=(X.base)]\node[draw=blue,fill=yellow!20,semithick,rectangle,inner sep=1.5pt,rounded corners=3pt] (X) {#1};\allowbreak}
\newcommand{\colltag}[1]{\tikz[baseline=(X.base)]\node[draw=blue,fill=pink!40,semithick,rectangle,inner sep=1.5pt,rounded corners=3pt] (X) {#1};\allowbreak}
\newcommand{\s}{\S\nobreak}

\newcommand{\eg}{\textit{e}.\textit{g}.}
\newcommand{\cf}{\textit{cf.}}
\newcommand{\etal}{\textit{et~al.}}
\newcommand{\etc}{\textit{etc.}}

\makeatletter
\providecommand{\thickhline}{\noalign{\ifnum0=`}\fi\hrule height 1pt\futurelet\reserved@a\@xhline}
\makeatother

\newsavebox{\surveyhumanbox}
\newsavebox{\surveyrulebox}
\newsavebox{\surveyllmbox}
\newsavebox{\surveycollbox}
\newcommand{\prepareSurveyTableTags}{%
  \sbox{\surveyhumanbox}{\humantag{Human}}%
  \def\humantag##1{\usebox{\surveyhumanbox}\allowbreak}%
  \sbox{\surveyrulebox}{\ruletag{Rule}}%
  \def\ruletag##1{\usebox{\surveyrulebox}\allowbreak}%
  \sbox{\surveyllmbox}{\llmtag{LLM}}%
  \def\llmtag##1{\usebox{\surveyllmbox}\allowbreak}%
  \sbox{\surveycollbox}{\colltag{Coll}}%
  \def\colltag##1{\usebox{\surveycollbox}\allowbreak}%
}

%% file: sec/intro.tex
\section{Introduction}\label{sec:intro}

The rapid advancement of Artificial Intelligence (AI) has been fundamentally intertwined with the development of robust benchmarks~\cite{deng2009imagenet,lin2014microsoft,krishna2017visual}. These benchmarks serve as crucial standards, providing objective metrics to evaluate and compare the performance of AI models. As a pioneer in computer vision, ImageNet~\cite{deng2009imagenet}, offering a large-scale and well-annotated dataset, has paved the way for developing models that are both highly accurate and broadly generalizable. Previous development of AI models and benchmarks has been complementary. For instance, as classification benchmarks grew in terms of data volume and class diversity, models trained on them improved significantly, resulting in better real-world performance. This synergy between task-specific benchmarks and model architecture has been a cornerstone of AI's practical applications.

Recent breakthroughs in Large Language Models (LLMs), \eg, ChatGPT~\cite{chatgpt}, have reshaped numerous research and industrial sectors. Harnessing LLMs as their reasoning core, Multimodal Large Language Models (MLLMs), \eg, GPT-4V~\cite{achiam2023gpt,openai2023gpt4v} and Gemini~\cite{team2023gemini}, integrate text, vision, audio, and 3D streams to perceive, reason, synthesize media, and act across digital and physical environments. Despite their rapid evolution, benchmark design has not always kept pace. Traditional benchmarks focused on scaling sample counts or class labels struggle to evaluate the multifaceted capabilities of modern MLLMs, raising a central question: \textit{\textbf{How can we effectively evaluate the capabilities and reliability of these large AI models?}}
\input{sec/misc/fig_timeline}
\input{sec/misc/fig_structure}
\input{sec/misc/fig_release_tree}

Currently, the landscape of multimodal benchmarks for AI models is rich and varied (\cf~Fig.~\ref{fig:timeline}), encompassing a wide range of tasks such as visual question answering and image captioning. This diversity has undoubtedly spurred the development of MLLMs, providing researchers with multiple avenues to explore and refine their models. However, \textbf{\textit{the plethora of benchmarks is a double-edged sword.}} The sheer number of benchmarks makes it difficult to navigate, especially for newcomers: 1) \textit{Disjointed objectives}: The abundance of benchmarks, each with distinct goals, results in a fragmented research landscape, where broad labels often conflate distinct constructs such as content synthesis versus safety auditing, or static question answering versus closed-loop execution. Researchers must invest considerable time determining whether existing benchmarks adequately test MLLMs' capabilities, complicating the development of new, aligned benchmarks.
2) \textit{Task saturation}: The proliferation of tasks, driven by diverse objectives, has led to an overwhelming landscape. This saturation makes it challenging to discern truly innovative or impactful benchmarks, causing valuable resources to be overlooked or underutilized, while rapid successor releases frequently alter distractors, input modalities, and task mixes at once.
3) \textit{Metric evolution \& discrepancies}: Although some studies have proposed well-designed metrics, their adoption is hindered by the rapid evolution and frequent updates of benchmarks. This inconsistency forces researchers to navigate a complex balance between traditional and new metrics, compounded by the widespread reuse of LLMs in both dataset construction and response judging, complicating fair comparisons and hindering the holistic evaluation of MLLMs.

\input{sec/related_scope.tex}

To address this gap, we present a comprehensive survey that reviews the capabilities, task designs, dataset constructions, and evaluation metrics of modern multimodal benchmarks. This survey examines these aspects across five mutually exclusive primary evaluation pillars, pairing each pillar with an observable internal classification axis (Fig.~\ref{fig_structure}):

\noindent$\bullet$~\textbf{Understanding~(\s\ref{sec:understanding}).}
It refers to the ability to extract and integrate features from multimodal data to perform cross-modal analysis. This involves tasks such as interpreting visual representations, identifying key details, grasping semantic meaning, and responding accurately to related questions. Evaluating these capabilities is fundamental, as they form the foundation for MLLMs' broader functionality across diverse tasks and applications. We organize 107 Understanding benchmarks by \textit{perceptual and contextual scope}: single-instance perception (\textbf{U1}), multi-image and long-context understanding (\textbf{U2}), temporal and streaming understanding (\textbf{U3}), and audio-visual or omni-modal understanding (\textbf{U4}).

\noindent$\bullet$~\textbf{Reasoning~(\s\ref{sec:reasoning}).}
It goes beyond basic understanding, encompassing the ability to perform complex inferences and draw logical conclusions across modalities. This includes tasks that require models to process and manipulate information, enabling them to solve problems and make decisions based on cross-modal data. Strong reasoning skills are crucial for MLLMs to handle sophisticated tasks that require deeper cognitive processing. We organize 73 Reasoning benchmarks by \textit{reasoning supervision and grounding}: outcome-based multimodal reasoning (\textbf{R1}), visually grounded reasoning (\textbf{R2}), chain-of-thought and process reasoning (\textbf{R3}), and reflective critique, self-improvement, or knowledge editing (\textbf{R4}).

\noindent$\bullet$~\textbf{Generation~(\s\ref{sec:generation}).}
It involves the creation of new content based on multimodal inputs, such as instruction-following and creative text generation (\textbf{G1}), visual-to-code and programmatic generation (\textbf{G2}), and visual, interleaved, or spoken media generation (\textbf{G3}) across 31 pure synthesis benchmarks. This capability is vital for practical applications where creativity, coherence, and structural fidelity are paramount.

\noindent$\bullet$~\textbf{Trustworthiness and Safety (\s\ref{sec:trustworthiness}).}
Distinct from generative synthesis, this pillar evaluates 48 benchmarks probing whether MLLMs and diagnostic evaluators remain factually grounded, ethically aligned, calibrated, privacy-preserving, and resilient to adversarial or spurious cues. We classify them into hallucination and factual grounding (\textbf{T1}), safety alignment, calibration, and unlearning (\textbf{T2}), adversarial robustness, spurious correlations, and demographic bias (\textbf{T3}), and evaluator, guardrail, and authenticity auditing (\textbf{T4}).

\noindent$\bullet$~\textbf{Application~(\s\ref{sec:application}).}
It explores benchmarks that assess the practical application of MLLMs in real-world scenarios. It includes embodied AI, agent-based tasks, and domain-specific applications like medical diagnostics and autonomous systems, and increasingly covers vision-language-action (VLA) policies, automated scientific research, and multimodal search agents. The benchmarks in this category evaluate how well models can integrate multiple modalities to perform complex tasks in dynamic and interactive environments. Across 99 benchmarks, we distinguish \textit{Regime~I: interactive GUI and embodied agents} (\textbf{A1--A2}) from \textit{Regime~II: knowledge-intensive specialist verticals} (\textbf{A3--A7}), crossed with an orthogonal \textit{track endpoint} (\textbf{Q/L/D/P/E}).

This survey covers 358 peer-reviewed multimodal benchmarks published between 2024 and 2026 across non-text modalities (image, video, audio, and 3D; Fig.~\ref{fig:author-tree}), with representative tasks illustrated across the five chapter maps (Fig.~\ref{fig:map-understanding}--Fig.~\ref{fig:map-application}). Because benchmarks frequently span multiple capabilities, we assign each suite a single mutually exclusive primary pillar under the precedence \textit{Application $\rightarrow$ Trustworthiness and Safety $\rightarrow$ Generation $\rightarrow$ Reasoning $\rightarrow$ Understanding}, qualified by two operational protocol rules: \textit{Application} requires either live interactive environment execution (\eg, WASP~\cite{evtimov2025wasp}, IS-Bench~\cite{lu2026isbench}) or an end-to-end specialist/agentic workflow (\eg, SWE-bench Multimodal~\cite{yang2025swebenchmultimodal}, PaperBench~\cite{starace2025paperbench}), whereas offline safety or hazard probes over static prompts and pre-recorded trajectories (\eg, MMA-SafetyBench~\cite{wang2026mmasafetybench}, AgentSafe~\cite{ying2026safe}) belong to \textit{Trustworthiness and Safety}, and direct single-pass prompt-to-code or document synthesis (\eg, OmniGIRL~\cite{guo2025omnigirl}, BlenderBench~\cite{yin2026blenderbench}, Paper2Poster~\cite{pang2025paper2poster}) belongs to \textit{Generation}.

The structure of this article is presented as follows: \s\ref{sec:understanding} through \s\ref{sec:application} review understanding, reasoning, generation, trustworthiness and safety, and application benchmarks along their observable axes, integrating task designs and evaluation metrics. \s\ref{sec:holistic_evaluation} introduces the general dataset details and construction processes. \s\ref{sec:cross} examines construct validity, protocol validity, pipeline coupling, and design blueprints across the 358-benchmark inventory. Finally, \s\ref{sec:direction} and \s\ref{sec:conclusions} suggest potentially valuable directions for further research and conclude the survey. We hope that this survey will help newcomers and practitioners to navigate the current landscape of benchmarking in the field of MLLMs, as well as provide the MLLM community with background information for generating future research.

%% file: sec/misc/fig_timeline.tex
\begin{figure*}[tp]
 \centering
 \includegraphics[width=0.98\textwidth,height=0.86\textheight,keepaspectratio]{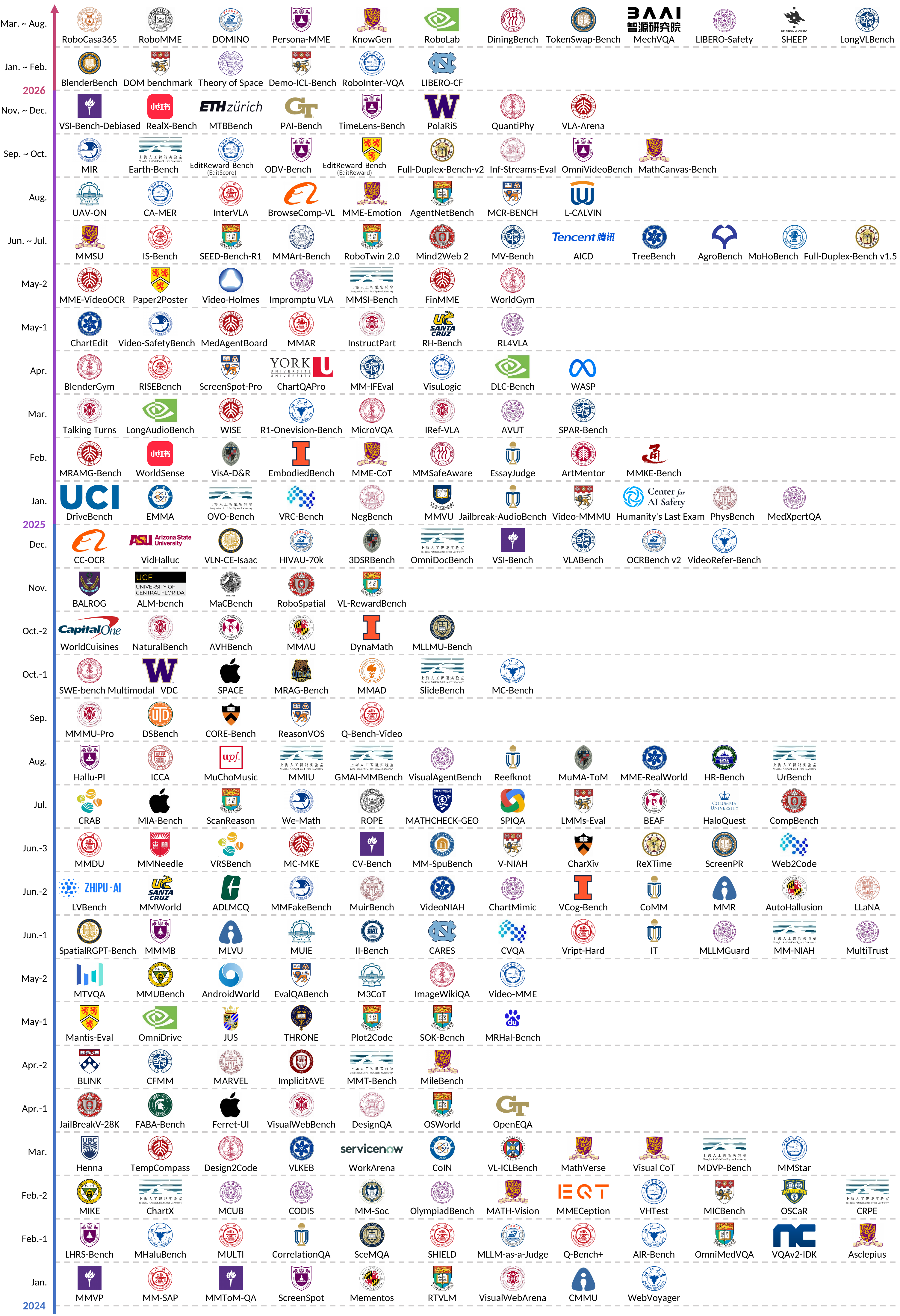}
 \caption{Chronological overview of representative multimodal benchmarks from 2024 to 2026, with icons above each entry indicating the primary affiliating institution or research laboratory.}
 \label{fig:timeline}
\end{figure*}

%% file: sec/misc/fig_structure.tex
\begin{figure*}[tp]
\centering
\chaptergraphic[width=\textwidth]{survey_map}
\caption{Organization of the survey. The left branches lead from the introduction to the five mutually exclusive primary pillars: Understanding (107), Reasoning (73), Generation (31), Trustworthiness and Safety (48), and Application (99). The right branches cover dataset construction and quality control, cross-benchmark comparison validity, open challenges and research directions, and the conclusion. Parenthetical labels indicate the corresponding sections and benchmark counts.}
\label{fig_structure}
\end{figure*}

%% file: sec/misc/fig_release_tree.tex
\begin{figure}[!t]
\centering
\includegraphics[width=\columnwidth]{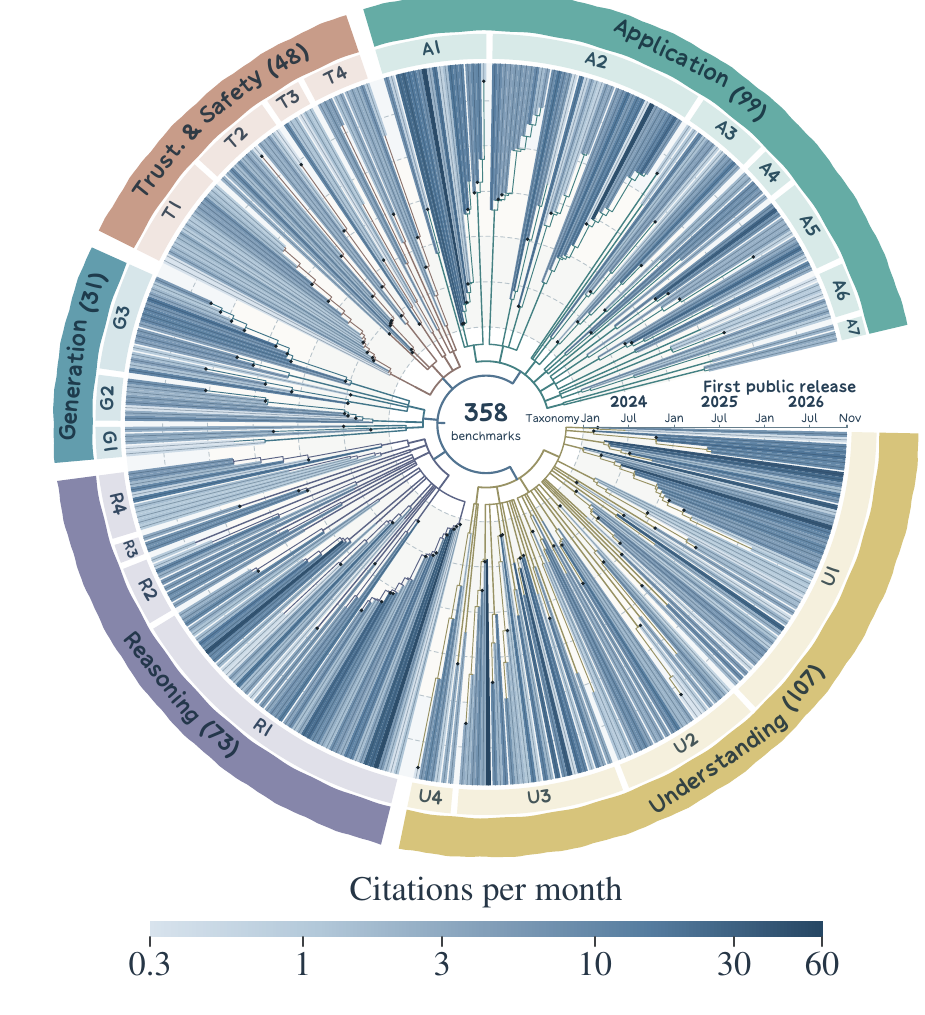}
\caption{Overview of the peer-reviewed multimodal benchmarks ($N=358$) across the five primary evaluation pillars: Understanding (107), Reasoning (73), Generation (31), Trustworthiness and Safety (48), and Application (99). Radial bars span each benchmark's release timeline from 2024 through November~2026, color shading indicates monthly citation rates, and black dots mark the 135 benchmarks incorporating LLM-based evaluation.}
\label{fig:author-tree}
\end{figure}

%% file: sec/related_scope.tex
\begin{table*}[t]\centering
\caption{Comparison between representative multimodal evaluation surveys and this work (\checkmark: systematic taxonomy or corpus-level analysis; $\circ$: partial or qualitative discussion; --: out of scope).}\label{tab:prior_surveys}
\begingroup\footnotesize\linespread{1.02}\selectfont\renewcommand{\arraystretch}{1.14}\setlength{\tabcolsep}{4.0pt}
\begin{tabularx}{\textwidth}{@{}>{\raggedright\arraybackslash}p{.195\textwidth}c>{\raggedright\arraybackslash}p{.145\textwidth}>{\raggedright\arraybackslash}Xcccc@{}}\hline\thickhline
\rowcolor{mygray}\textbf{Survey} & \textbf{Year} & \textbf{Scope \& Scale} & \textbf{Primary Organization} & \makecell{\textbf{Gen. vs.}\\\textbf{Safety}} & \makecell{\textbf{Version}\\\textbf{Audit}} & \makecell{\textbf{Causal}\\\textbf{Protocol}} & \makecell{\textbf{Pipeline}\\\textbf{Coupling}}\\\hline
Yin \etal~\cite{yin2023survey} & 2024 & General MLLM review & Model architectures, training, and early tasks & $\circ$ & -- & -- & --\\
Li \etal~\cite{li2024benchmarkreview} & 2024 & 200+ MLLM benchmarks & Model capabilities and input modalities & $\circ$ & $\circ$ & $\circ$ & --\\
Fu \etal~\cite{fu2024mmesurvey} & 2024 & Evaluation pipeline & Data curation, scoring, and toolkits & $\circ$ & $\circ$ & $\circ$ & --\\
Gu \etal~\cite{gu2024llmasajudge} & 2024 & LLM-as-a-Judge & Judge prompting, biases, and validation & -- & -- & -- & $\circ$\\
Kim \etal~\cite{kim2025vqasurvey} & 2025 & VQA survey & VQA datasets, methods, and metrics & -- & $\circ$ & -- & --\\
Zhang \etal~\cite{zhang2025lmmevalreview} & 2025 & MLLM eval.\ review & General/vertical QA and media generation & \checkmark & $\circ$ & $\circ$ & --\\
Li \etal~\cite{li2026missing} & 2026 & Position paper & Real-world temporal and physical gaps & -- & -- & \checkmark & --\\
Zhu \etal~\cite{zhu2026agentreview} & 2026 & LLM/MLLM agents & Agent capabilities and interactive testbeds & -- & $\circ$ & \checkmark & --\\\hline
\textbf{Ours} & \textbf{2026} & \textbf{358 peer-reviewed suites} & \textbf{Five mutually exclusive pillars + 18-field matrix} & \textbf{\checkmark} & \textbf{\checkmark} & \textbf{\checkmark} & \textbf{\checkmark}\\\hline
\end{tabularx}\endgroup
\end{table*}

Recent surveys examine multimodal and foundation-model evaluation from complementary angles (Table~\ref{tab:prior_surveys}), spanning general MLLM architectures~\cite{yin2023survey}, capability-oriented benchmark lists~\cite{li2024benchmarkreview}, evaluation toolkits~\cite{fu2024mmesurvey}, LLM-as-a-Judge biases~\cite{gu2024llmasajudge}, VQA datasets~\cite{kim2025vqasurvey}, domain and modality tables~\cite{zhang2025lmmevalreview}, temporal/physical testbed gaps~\cite{li2026missing}, and agentic environments~\cite{zhu2026agentreview} (alongside focused reviews of hallucination~\cite{ji2023survey,bai2024hallucination,liu2024survey} and self-improvement~\cite{deng2025selfimprovementsurvey}). In Table~\ref{tab:prior_surveys}, \textit{Gen.\ vs.\ Safety} checks whether open-ended media synthesis is separated from safety/hallucination auditing; \textit{Version Audit} checks paired analysis of predecessor--successor benchmark revisions; \textit{Causal Protocol} checks systematic separation of streaming/closed-loop protocols from offline QA; and \textit{Pipeline Coupling} checks corpus-level cross-tabulation between dataset construction and response judging.

%% file: sec/undertanding.tex
\section{Understanding Benchmarks}
\label{sec:understanding}
\begin{figure*}[tp]
\centering\chaptergraphic[width=\textwidth]{understanding_map}
\caption{Taxonomy and representative examples of Understanding benchmarks ($N=107$) organized by perceptual and contextual scope (\textbf{U1--U4}). Illustrated tasks cover fine-grained visual contrast in MMVP~\cite{tong2024mmvp}, context-dependent disambiguation in CODIS~\cite{luo2024codis}, temporal motion understanding in TempCompass~\cite{liu2024TempCompass}, and audio-visual necessity in AVUT~\cite{yang2025avut}.}\label{fig:map-understanding}
\end{figure*}

\input{sec/misc/tbl_understanding}

\input{sec/understanding/bg}

\input{sec/understanding/task}

%% file: sec/misc/tbl_understanding.tex
\input{sec/misc/tbl_understanding_clean}

%% file: sec/misc/tbl_understanding_clean.tex
\begin{table*}[tp]\centering
\caption{Overview of Understanding benchmarks ($N=107$), spanning \textbf{U1} single-instance perception (50), \textbf{U2} multi-image and long-context understanding (23), \textbf{U3} temporal and streaming understanding (27), and \textbf{U4} audio-visual and omni-modal understanding (7). Output formats: binary (\textbf{BI}), multiple-choice (\textbf{MC}), defined-form (\textbf{DF}), and free-form (\textbf{FREE}); pipeline tags denote collected media (\textbf{Coll}), human, rule-based, and LLM stages.}
\label{tbl:understanding}
\begingroup\linespread{0.99}\fontsize{7.2}{8.1}\selectfont\setlength\tabcolsep{1.25pt}\renewcommand\arraystretch{1.00}
\prepareSurveyTableTags\begin{tabularx}{\textwidth}{@{}>{\raggedright\arraybackslash}p{.133\textwidth}l>{\raggedright\arraybackslash}p{.166\textwidth}>{\raggedright\arraybackslash}X>{\raggedright\arraybackslash}p{.070\textwidth}>{\raggedright\arraybackslash\spaceskip=1pt}p{.089\textwidth}>{\raggedright\arraybackslash\spaceskip=1pt}p{.160\textwidth}>{\raggedright\arraybackslash\spaceskip=1pt}p{.090\textwidth}@{}}
\hline\thickhline\rowcolor{mygray}\textbf{Benchmark} & \textbf{Ref.} & \textbf{Category \& Focus} & \textbf{Task} & \textbf{Output} & \textbf{Evaluation} & \textbf{Construction} & \textbf{Filter} \\ \hline
3D MM-Vet & \cite{qi20243dmmvet} & \textbf{U1}\, 3D Understanding & General recognition, \etc\ (5 categories) & FREE & \llmtag{LLM} & \colltag{Coll} \humantag{Human} & -- \\ \hline
AIR-Bench & \cite{yang2024air} & \textbf{U1}\, Audio Understanding & Speech grounding, \etc\ (19 total) & MC FREE & \llmtag{LLM} & \colltag{Coll} \ruletag{Rule} \llmtag{LLM} & \humantag{Human} \llmtag{LLM} \\ \hline
ALM-bench & \cite{vayani2025almbench} & \textbf{U1}\, Multilingual Perception & Multilingual cultural VQA (100 languages) & BI MC DF FREE & \llmtag{LLM} & \colltag{Coll} \humantag{Human} \llmtag{LLM} & \humantag{Human} \ruletag{Rule} \\ \hline
BLINK & \cite{fu2024blink} & \textbf{U1}\, Comprehensive Perception & Visual correspondence, \etc\ (14 total visual perception tasks) & BI MC & \ruletag{Rule} & \colltag{Coll} \humantag{Human} \ruletag{Rule} \llmtag{LLM} & \humantag{Human} \\ \hline
CC-OCR & \cite{yang2025ccocr} & \textbf{U1}\, Fine-Grained Perception & Multi-scene OCR, \etc\ (39 subsets) & DF & \ruletag{Rule} & \colltag{Coll} \humantag{Human} \ruletag{Rule} & \humantag{Human} \llmtag{LLM} \\ \hline
Cultural\allowbreak{}VQA & \cite{nayak2024culturalvqa} & \textbf{U1}\, Multilingual Perception & Cultural VQA & DF & \llmtag{LLM} & \colltag{Coll} \humantag{Human} & \humantag{Human} \ruletag{Rule} \\ \hline
CV-Bench & \cite{tong2024cvbench} & \textbf{U1}\, Fine-Grained Perception & Object count; spatial relation; depth order (3D); relative distance (3D) & MC & \ruletag{Rule} & \colltag{Coll} \ruletag{Rule} & \humantag{Human} \\ \hline
CVQA & \cite{romero2024cvqa} & \textbf{U1}\, Multilingual Perception & Culturally-diverse multilingual VQA & MC & \ruletag{Rule} & \colltag{Coll} \humantag{Human} & \humantag{Human} \\ \hline
DLC-Bench & \cite{lian2025dlcbench} & \textbf{U1}\, Fine-Grained Perception & Detailed localized captioning & FREE & \llmtag{LLM} & \colltag{Coll} \humantag{Human} \llmtag{LLM} & \humantag{Human} \ruletag{Rule} \\ \hline
Ego\allowbreak{}Text\allowbreak{}VQA & \cite{zhou2025egotextvqa} & \textbf{U1}\, Long Video Understanding & Egocentric Scene-Text Aware Video Question Answering & FREE & \ruletag{Rule} \llmtag{LLM} & \colltag{Coll} \humantag{Human} \llmtag{LLM} & \ruletag{Rule} \humantag{Human} \llmtag{LLM} \\ \hline
FABA-Bench & \cite{li2024facial} & \textbf{U1}\, Higher-order Perception & Emotion recognition; action unit (AU) recognition & FREE & \ruletag{Rule} & \colltag{Coll} \humantag{Human} \llmtag{LLM} & \humantag{Human} \\ \hline
Henna & \cite{alwajih2024peacock} & \textbf{U1}\, Multilingual Perception & Arabic QA across attractions, food, events, and other Arabic-relevant objects & FREE & \llmtag{LLM} & \colltag{Coll} \humantag{Human} \llmtag{LLM} & \humantag{Human} \\ \hline
HR-Bench & \cite{wang2024hrbench} & \textbf{U1}\, Fine-Grained Perception & Single- and cross-instance perception on 4K/8K images & MC & \ruletag{Rule} & \colltag{Coll} \humantag{Human} \llmtag{LLM} & \humantag{Human} \\ \hline
Hue\allowbreak{}Manity & \cite{grover2025huemanity} & \textbf{U1}\, Fine-Grained Perception & Ishihara-style alphanumeric visual pattern recognition & DF & \ruletag{Rule} & \ruletag{Rule} & \humantag{Human} \ruletag{Rule} \\ \hline
II-Bench & \cite{liu2024ii} & \textbf{U1}\, Higher-order Perception & Implication understanding & MC & \ruletag{Rule} & \colltag{Coll} \humantag{Human} & \humantag{Human} \ruletag{Rule} \\ \hline
Image\allowbreak{}Wiki\allowbreak{}QA & \cite{zhang2024imagewikiqa} & \textbf{U1}\, Fine-Grained Perception & Knowledge-intensive object QA & MC & \ruletag{Rule} & \colltag{Coll} \llmtag{LLM} & \humantag{Human} \llmtag{LLM} \\ \hline
Implicit\allowbreak{}AVE & \cite{zou2024implicitave} & \textbf{U1}\, Higher-order Perception & Implication understanding & MC & \ruletag{Rule} & \colltag{Coll} \humantag{Human} \ruletag{Rule} \llmtag{LLM} & \humantag{Human} \ruletag{Rule} \llmtag{LLM} \\ \hline
IT & \cite{pi2024image} & \textbf{U1}\, Comprehensive Perception & Detailed image description; description-to-image; linguistic description & BI FREE & \ruletag{Rule} & \colltag{Coll} \humantag{Human} \llmtag{LLM} & \humantag{Human} \\ \hline
L\allowbreak{}La\allowbreak{}NA & \cite{amaduzzi2024LLaNA} & \textbf{U1}\, 3D Understanding & NeRF captioning; NeRF QA & FREE & \ruletag{Rule} & \colltag{Coll} \llmtag{LLM} & -- \\ \hline
LM\allowbreak{}Ms-Eval & \cite{zhang2025lmmseval} & \textbf{U1}\, Comprehensive Perception & Unified evaluation of 50+ tasks & BI MC DF FREE & \ruletag{Rule} \llmtag{LLM} & \colltag{Coll} \llmtag{LLM} & \humantag{Human} \ruletag{Rule} \llmtag{LLM} \\ \hline
MDVP-Bench & \cite{lin2024MDVP-Bench} & \textbf{U1}\, Fine-Grained Perception & Concise descriptions, elaborate narratives, \etc & FREE & \llmtag{LLM} & \colltag{Coll} \humantag{Human} \llmtag{LLM} & \humantag{Human} \\ \hline
MM-SAP & \cite{wang2024mm-sap} & \textbf{U1}\, Fine-Grained Perception & Basic visual information QA; knowledge-intensive visual information QA; beyond visual information QA & MC & \ruletag{Rule} & \colltag{Coll} \humantag{Human} & -- \\ \hline
MMAR & \cite{ma2025mmar} & \textbf{U1}\, Audio Understanding & Deep audio reasoning (4 layers) & MC & \ruletag{Rule} & \colltag{Coll} \humantag{Human} \llmtag{LLM} & \humantag{Human} \\ \hline
MMAU & \cite{sakshi2025mmau} & \textbf{U1}\, Audio Understanding & Speech, sound and music reasoning (27 tasks) & MC & \ruletag{Rule} & \colltag{Coll} \humantag{Human} \llmtag{LLM} & \humantag{Human} \\ \hline
MME-Emotion & \cite{zhang2026mmeemotion} & \textbf{U1}\, Higher-order Perception & Emotion recognition, \etc\ (8 tasks) & DF & \ruletag{Rule} \llmtag{LLM} & \colltag{Coll} \ruletag{Rule} & -- \\ \hline
MME-Real\allowbreak{}World & \cite{zhang2025mmerealworld} & \textbf{U1}\, Fine-Grained Perception & OCR in the wild, \etc\ (43 subtasks) & MC & \ruletag{Rule} & \colltag{Coll} \humantag{Human} & \humantag{Human} \\ \hline
MME-Video\allowbreak{}OCR & \cite{shi2025mme} & \textbf{U1}\, Video Comprehensive Understanding & Recognition; Visual Text QA & MC DF & \ruletag{Rule} \llmtag{LLM} & \colltag{Coll} \humantag{Human} \llmtag{LLM} & \humantag{Human} \llmtag{LLM} \\ \hline
MMMB & \cite{sun2024parrot} & \textbf{U1}\, Multilingual Perception & Six languages \& fifteen categories VQA & BI MC & \ruletag{Rule} & \colltag{Coll} \humantag{Human} \llmtag{LLM} & \humantag{Human} \\ \hline
MMMU-Pro & \cite{yue2025mmmupro} & \textbf{U1}\, Multi-Discipline Understanding & College-level multi-discipline QA & MC & \ruletag{Rule} & \colltag{Coll} \humantag{Human} \llmtag{LLM} & \humantag{Human} \llmtag{LLM} \\ \hline
MM\allowbreak{}Star & \cite{chen2024MMStar} & \textbf{U1}\, Comprehensive Perception & Six core capabilities \& eighteen detailed axes related tasks & MC & \ruletag{Rule} & \colltag{Coll} & \humantag{Human} \llmtag{LLM} \\ \hline
MMSU & \cite{wang2026mmsu} & \textbf{U1}\, Audio Understanding & Phonology, \etc\ (47 tasks) & MC & \ruletag{Rule} & \colltag{Coll} \humantag{Human} \ruletag{Rule} \llmtag{LLM} & \humantag{Human} \\ \hline
MMT-Bench & \cite{yingmmt} & \textbf{U1}\, Comprehensive Perception & Visual recognition, \etc\ (162 total) & MC & \ruletag{Rule} & \colltag{Coll} \humantag{Human} \ruletag{Rule} \llmtag{LLM} & \humantag{Human} \ruletag{Rule} \\ \hline
MMVP & \cite{tong2024mmvp} & \textbf{U1}\, Fine-Grained Perception & Orientation and direction, \etc\ (9 visual patterns) & BI MC & \ruletag{Rule} & \colltag{Coll} \humantag{Human} \ruletag{Rule} & -- \\ \hline
MTVQA & \cite{tang2024mtvqa} & \textbf{U1}\, Multilingual Perception & Multilingual VQA & DF & \ruletag{Rule} & \colltag{Coll} \humantag{Human} & \humantag{Human} \ruletag{Rule} \\ \hline
MUIE & \cite{zhang2024muie} & \textbf{U1}\, Omni-modal Understanding & Grounded multimodal universal information extraction & DF & \ruletag{Rule} & \colltag{Coll} \humantag{Human} \ruletag{Rule} \llmtag{LLM} & \ruletag{Rule} \\ \hline
Natural\allowbreak{}Bench & \cite{li2024naturalbench} & \textbf{U1}\, Comprehensive Perception & Natural adversarial VQA & BI MC & \ruletag{Rule} & \colltag{Coll} \llmtag{LLM} & \humantag{Human} \\ \hline
\end{tabularx}\endgroup\end{table*}
\begin{table*}[tp]\centering
\ContinuedFloat
\caption[]{Overview of Understanding benchmarks (continued).}
\begingroup\linespread{0.99}\fontsize{7.2}{8.1}\selectfont\setlength\tabcolsep{1.25pt}\renewcommand\arraystretch{1.00}
\prepareSurveyTableTags\begin{tabularx}{\textwidth}{@{}>{\raggedright\arraybackslash}p{.133\textwidth}l>{\raggedright\arraybackslash}p{.166\textwidth}>{\raggedright\arraybackslash}X>{\raggedright\arraybackslash}p{.070\textwidth}>{\raggedright\arraybackslash\spaceskip=1pt}p{.089\textwidth}>{\raggedright\arraybackslash\spaceskip=1pt}p{.160\textwidth}>{\raggedright\arraybackslash\spaceskip=1pt}p{.090\textwidth}@{}}
\hline\thickhline\rowcolor{mygray}\textbf{Benchmark} & \textbf{Ref.} & \textbf{Category \& Focus} & \textbf{Task} & \textbf{Output} & \textbf{Evaluation} & \textbf{Construction} & \textbf{Filter} \\ \hline
Neg\allowbreak{}Bench & \cite{alhamoud2025negbench} & \textbf{U1}\, Fine-Grained Perception & Retrieval with Negation; Multiple Choice Questions with Negated Captions & DF MC & \ruletag{Rule} & \colltag{Coll} \llmtag{LLM} & \ruletag{Rule} \\ \hline
OCR\allowbreak{}Bench v2 & \cite{fu2025ocrbenchv2} & \textbf{U1}\, Fine-Grained Perception & Text recognition, \etc\ (23 total OCR tasks) & MC DF & \ruletag{Rule} & \colltag{Coll} \humantag{Human} \ruletag{Rule} & \humantag{Human} \ruletag{Rule} \\ \hline
ODV-Bench & \cite{zeng2025odvbench} & \textbf{U1}\, Streaming Video Understanding & Real-time traffic perception, \etc\ (12 tasks) & MC & \ruletag{Rule} & \colltag{Coll} \humantag{Human} \ruletag{Rule} \llmtag{LLM} & \humantag{Human} \ruletag{Rule} \\ \hline
Omni\allowbreak{}Doc\allowbreak{}Bench & \cite{ouyang2025omnidocbench} & \textbf{U1}\, Fine-Grained Perception & PDF parsing, \etc\ (9 page types) & DF & \ruletag{Rule} & \colltag{Coll} \humantag{Human} \llmtag{LLM} & \humantag{Human} \ruletag{Rule} \\ \hline
Punch\allowbreak{}Bench & \cite{ouyang2025punchbench} & \textbf{U1}\, Higher-order Perception & Multimodal humor and punchline comprehension & BI MC FREE & \ruletag{Rule} \llmtag{LLM} & \colltag{Coll} \humantag{Human} \llmtag{LLM} & \humantag{Human} \ruletag{Rule} \\ \hline
Q-Bench+ & \cite{zhang2024q-bench} & \textbf{U1}\, Low-Level Perception & Low-level perception for single image; low-level perception for image pair; low-level description & BI MC DF FREE & \ruletag{Rule} \llmtag{LLM} & \colltag{Coll} \humantag{Human} & \humantag{Human} \ruletag{Rule} \\ \hline
Q-Bench-Video & \cite{zhang2025qbenchvideo} & \textbf{U1}\, Low-Level Perception & Video quality QA, \etc\ (4 distortion types) & BI MC FREE & \ruletag{Rule} \llmtag{LLM} & \colltag{Coll} \humantag{Human} & \humantag{Human} \\ \hline
Shot\allowbreak{}Bench & \cite{liu2025shotbench} & \textbf{U1}\, Higher-order Perception & Cinematography understanding (8 dimensions) & MC & \ruletag{Rule} & \colltag{Coll} \humantag{Human} \ruletag{Rule} & \humantag{Human} \ruletag{Rule} \\ \hline
UCIT & \cite{guo2025ucit} & \textbf{U1}\, Multi-Discipline Understanding & Continual instruction tuning on six datasets & DF FREE & \ruletag{Rule} & \colltag{Coll} & \ruletag{Rule} \\ \hline
UNICBench & \cite{rong2026unicbench} & \textbf{U1}\, Fine-Grained Perception & Unified counting across images, documents and audio & DF & \ruletag{Rule} & \colltag{Coll} \humantag{Human} \ruletag{Rule} & \humantag{Human} \ruletag{Rule} \\ \hline
Video\allowbreak{}Refer-Bench & \cite{yuan2025videoreferbench} & \textbf{U1}\, Spatial-Temporal Perception & Video region description and QA & MC FREE & \ruletag{Rule} \llmtag{LLM} & \colltag{Coll} \humantag{Human} \llmtag{LLM} & \humantag{Human} \llmtag{LLM} \\ \hline
Vis\allowbreak{}A-D\&R & \cite{xu2025visadr} & \textbf{U1}\, Fine-Grained Perception & Anomaly detection and description & BI FREE & \ruletag{Rule} \llmtag{LLM} & \colltag{Coll} \humantag{Human} \llmtag{LLM} & \humantag{Human} \\ \hline
Voice\allowbreak{}Bench & \cite{chen2026voicebench} & \textbf{U1}\, Audio Understanding & Spoken instruction following and voice-assistant QA & MC DF FREE & \ruletag{Rule} \llmtag{LLM} & \colltag{Coll} \ruletag{Rule} \llmtag{LLM} & \humantag{Human} \ruletag{Rule} \\ \hline
World\allowbreak{}Cuisines & \cite{winata2025worldcuisines} & \textbf{U1}\, Multilingual Perception & Dish and location prediction VQA & MC DF & \ruletag{Rule} & \colltag{Coll} \humantag{Human} \ruletag{Rule} \llmtag{LLM} & \humantag{Human} \ruletag{Rule} \\ \hline
CODIS & \cite{luo2024codis} & \textbf{U2}\, Context-Dependent Understanding & Context-dependent image disambiguation & MC FREE & \humantag{Human} \llmtag{LLM} & \colltag{Coll} \humantag{Human} & \humantag{Human} \\ \hline
Demo-ICL-Bench & \cite{dong2026demoiclbench} & \textbf{U2}\, Multimodal In-Context Learning & In-context learning on procedural video (3 tasks) & MC & \ruletag{Rule} & \colltag{Coll} \llmtag{LLM} & \humantag{Human} \ruletag{Rule} \llmtag{LLM} \\ \hline
ICCA & \cite{hua2024icca} & \textbf{U2}\, Multimodal In-Context Learning & Conversational adaptation in reference games & MC FREE & \ruletag{Rule} & \colltag{Coll} & -- \\ \hline
Long\allowbreak{}Doc\allowbreak{}URL & \cite{deng2025longdocurl} & \textbf{U2}\, Long-context Understanding & Long-document QA, numerical reasoning and layout locating (20 subtasks) & DF FREE & \ruletag{Rule} \llmtag{LLM} & \colltag{Coll} \humantag{Human} \llmtag{LLM} & \humantag{Human} \ruletag{Rule} \\ \hline
Long\allowbreak{}VL\allowbreak{}Bench & \cite{zhang2026longvlbench} & \textbf{U2}\, Interleaved Image-Text Understanding & Long interleaved image-text QA & BI DF FREE & \llmtag{LLM} & \colltag{Coll} \ruletag{Rule} \llmtag{LLM} & \humantag{Human} \llmtag{LLM} \\ \hline
Mantis-Eval & \cite{Jiang2024MANTISIM} & \textbf{U2}\, Multi-Image Understanding & Multi-image QA & MC DF & \ruletag{Rule} & \humantag{Human} & -- \\ \hline
MC-Bench & \cite{xu2025mcbench} & \textbf{U2}\, Multi-Image Understanding & Multi-context visual grounding & DF & \ruletag{Rule} & \colltag{Coll} \humantag{Human} & \humantag{Human} \\ \hline
Mementos & \cite{wang2024mementos} & \textbf{U2}\, Multi-Image Understanding & Image sequences reasoning & FREE & \ruletag{Rule} & \colltag{Coll} \humantag{Human} \llmtag{LLM} & \humantag{Human} \\ \hline
MIC\allowbreak{}Bench & \cite{wu2024micbench} & \textbf{U2}\, Low-Level Perception & Multi-image quality comparison & BI MC & \ruletag{Rule} & \colltag{Coll} \humantag{Human} & \humantag{Human} \\ \hline
Mile\allowbreak{}Bench & \cite{song2024milebench} & \textbf{U2}\, Long-context Understanding & Needle-in-a-haystack; image retrieval; semantic multi-image QA; temporal multi-image QA & MC DF FREE & \ruletag{Rule} & \colltag{Coll} \ruletag{Rule} \llmtag{LLM} & \humantag{Human} \\ \hline
MIR & \cite{du2025mir} & \textbf{U2}\, Interleaved Image-Text Understanding & Multi-image interleaved reasoning (3 categories) & MC & \llmtag{LLM} & \colltag{Coll} \humantag{Human} \ruletag{Rule} \llmtag{LLM} & \humantag{Human} \\ \hline
MM-NIAH & \cite{wang2024MM-NIAH} & \textbf{U2}\, Long-context Understanding & Needle-in-a-haystack & MC DF FREE & \ruletag{Rule} & \colltag{Coll} \humantag{Human} \ruletag{Rule} \llmtag{LLM} & -- \\ \hline
MMDU & \cite{liu2024mmdu} & \textbf{U2}\, Multi-Image Understanding & Multi-turn multi-image dialogue & FREE & \llmtag{LLM} & \colltag{Coll} \humantag{Human} \llmtag{LLM} & \humantag{Human} \ruletag{Rule} \\ \hline
MMIU & \cite{meng2024MMIU} & \textbf{U2}\, Multi-Image Understanding & Action recognition, \etc\ (52 total) & MC & \ruletag{Rule} & \colltag{Coll} \ruletag{Rule} \llmtag{LLM} & \humantag{Human} \\ \hline
MMLong\allowbreak{}Bench-Doc & \cite{ma2024mmlongbenchdoc} & \textbf{U2}\, Long-context Understanding & Cross-page multimodal PDF document QA and unanswerable detection & DF FREE & \ruletag{Rule} \llmtag{LLM} & \colltag{Coll} \humantag{Human} & \humantag{Human} \\ \hline
MM\allowbreak{}Needle & \cite{wang2024multimodal} & \textbf{U2}\, Long-context Understanding & Needle-in-a-haystack & DF & \ruletag{Rule} & \colltag{Coll} \ruletag{Rule} & -- \\ \hline
MRAG-Bench & \cite{hu2025mragbenchvisioncentricevaluationretrievalaugmented} & \textbf{U2}\, Context-Dependent Understanding & Retrieval-augmented multiple-choice QA & MC & \ruletag{Rule} & \colltag{Coll} \humantag{Human} & \humantag{Human} \ruletag{Rule} \\ \hline
MTMEUR & \cite{hu2025mtmeur} & \textbf{U2}\, Higher-order Perception & Multi-turn emotion QA & MC & \ruletag{Rule} & \colltag{Coll} \llmtag{LLM} & \humantag{Human} \llmtag{LLM} \\ \hline
Muir\allowbreak{}Bench & \cite{wang2024MuirBench} & \textbf{U2}\, Multi-Image Understanding & Action understanding, \etc\ (12 total) & MC & \ruletag{Rule} & \colltag{Coll} \humantag{Human} & \humantag{Human} \ruletag{Rule} \\ \hline
OHRBench & \cite{zhang2025ohrbench} & \textbf{U2}\, Context-Dependent Understanding & Document QA under cascading semantic and formatting OCR noise & DF FREE & \ruletag{Rule} \llmtag{LLM} & \colltag{Coll} \ruletag{Rule} & \humantag{Human} \ruletag{Rule} \\ \hline
Persona-MME & \cite{nie2026personamme} & \textbf{U2}\, Context-Dependent Understanding & Memory, intent, \etc\ (7 aspects) & MC & \ruletag{Rule} & \colltag{Coll} \llmtag{LLM} & \humantag{Human} \\ \hline
Token\allowbreak{}Swap-Bench & \cite{hua2026tokenswapbench} & \textbf{U2}\, Interleaved Image-Text Understanding & Image-interleaved MMLU QA & MC DF & \ruletag{Rule} & \colltag{Coll} \llmtag{LLM} & \humantag{Human} \llmtag{LLM} \\ \hline
\end{tabularx}\endgroup\end{table*}
\begin{table*}[tp]\centering
\ContinuedFloat
\caption[]{Overview of Understanding benchmarks (continued).}
\begingroup\linespread{0.99}\fontsize{7.2}{8.1}\selectfont\setlength\tabcolsep{1.25pt}\renewcommand\arraystretch{1.00}
\prepareSurveyTableTags\begin{tabularx}{\textwidth}{@{}>{\raggedright\arraybackslash}p{.133\textwidth}l>{\raggedright\arraybackslash}p{.166\textwidth}>{\raggedright\arraybackslash}X>{\raggedright\arraybackslash}p{.070\textwidth}>{\raggedright\arraybackslash\spaceskip=1pt}p{.089\textwidth}>{\raggedright\arraybackslash\spaceskip=1pt}p{.160\textwidth}>{\raggedright\arraybackslash\spaceskip=1pt}p{.090\textwidth}@{}}
\hline\thickhline\rowcolor{mygray}\textbf{Benchmark} & \textbf{Ref.} & \textbf{Category \& Focus} & \textbf{Task} & \textbf{Output} & \textbf{Evaluation} & \textbf{Construction} & \textbf{Filter} \\ \hline
VL-ICL\allowbreak{}Bench & \cite{zong2024VL-ICLBench} & \textbf{U2}\, Multimodal In-Context Learning & Open-ended miniImageNet, \etc\ (10 total) & BI DF & \ruletag{Rule} \llmtag{LLM} & \colltag{Coll} \ruletag{Rule} & -- \\ \hline
ADLMCQ & \cite{chakraborty2024ADLMCQ} & \textbf{U3}\, Long Video Understanding & Action recognition and video description & MC FREE & \ruletag{Rule} \llmtag{LLM} & \colltag{Coll} \humantag{Human} & -- \\ \hline
Ego\allowbreak{}Exo\allowbreak{}Bench & \cite{he2025egoexobench} & \textbf{U3}\, Spatial-Temporal Perception & Cross-view egocentric-exocentric video QA (11 subtasks) & MC & \ruletag{Rule} & \colltag{Coll} \humantag{Human} \llmtag{LLM} & \humantag{Human} \ruletag{Rule} \\ \hline
Ego\allowbreak{}Gaze\allowbreak{}VQA & \cite{peng2025eyemllm} & \textbf{U3}\, Video Comprehensive Understanding & Spatial Intent QA; Temporal Intent QA; Causal Intent QA & MC & \ruletag{Rule} & \colltag{Coll} \llmtag{LLM} & \humantag{Human} \\ \hline
FAVOR-Bench & \cite{tu2025favorbench} & \textbf{U3}\, Spatial-Temporal Perception & Fine-grained video motion QA and captioning (6 subtasks) & MC FREE & \ruletag{Rule} \llmtag{LLM} & \colltag{Coll} \humantag{Human} \llmtag{LLM} & \humantag{Human} \\ \hline
HIVAU-70k & \cite{zhang2025hivau70k} & \textbf{U3}\, Video Comprehensive Understanding & Video anomaly captioning, \etc\ (4 instruction types) & DF FREE & \ruletag{Rule} & \colltag{Coll} \humantag{Human} \llmtag{LLM} & \humantag{Human} \\ \hline
Hour\allowbreak{}Video & \cite{chandrasegaran2024hourvideo} & \textbf{U3}\, Long Video Understanding & 1-hour egocentric video summarization, perception and navigation QA & MC & \ruletag{Rule} & \colltag{Coll} \humantag{Human} \llmtag{LLM} & \humantag{Human} \llmtag{LLM} \\ \hline
Inf-Streams-Eval & \cite{xu2026infstreamseval} & \textbf{U3}\, Streaming Video Understanding & Real-time sports commentary captioning & FREE & \llmtag{LLM} & \colltag{Coll} \llmtag{LLM} & \ruletag{Rule} \llmtag{LLM} \\ \hline
Long\allowbreak{}Audio\allowbreak{}Bench & \cite{ghosh2025longaudiobench} & \textbf{U3}\, Audio Understanding & Long-audio captioning and QA & MC FREE & \llmtag{LLM} & \colltag{Coll} \llmtag{LLM} & \humantag{Human} \llmtag{LLM} \\ \hline
Long\allowbreak{}Video\allowbreak{}Bench & \cite{wu2024longvideobench} & \textbf{U3}\, Long Video Understanding & Referring reasoning on long videos & MC & \ruletag{Rule} & \colltag{Coll} \humantag{Human} & \humantag{Human} \ruletag{Rule} \\ \hline
LV\allowbreak{}Bench & \cite{wang2025lvbench} & \textbf{U3}\, Long Video Understanding & Long video QA & MC & \ruletag{Rule} & \colltag{Coll} \humantag{Human} & \humantag{Human} \llmtag{LLM} \\ \hline
MA-Bench & \cite{li2026mabench} & \textbf{U3}\, Video Comprehensive Understanding & Micro-action perception, relational comprehension and reasoning & MC FREE & \ruletag{Rule} \llmtag{LLM} & \colltag{Coll} \humantag{Human} \llmtag{LLM} & \humantag{Human} \\ \hline
MLVU & \cite{zhou2024mlvu} & \textbf{U3}\, Long Video Understanding & Topic reasoning, \etc\ (9 total) & MC FREE & \ruletag{Rule} \llmtag{LLM} & \colltag{Coll} \humantag{Human} \ruletag{Rule} & \\ \hline
Mu\allowbreak{}Cho\allowbreak{}Music & \cite{weck2024MuChoMusic} & \textbf{U3}\, Audio Understanding & Music multiple-choice QA & MC & \ruletag{Rule} & \colltag{Coll} \llmtag{LLM} & \humantag{Human} \\ \hline
OS\allowbreak{}Ca\allowbreak{}R & \cite{nguyen2024oscar} & \textbf{U3}\, Spatial-Temporal Perception & Object states visual captioning; visual dialog reasoning & FREE & \humantag{Human} \ruletag{Rule} \llmtag{LLM} & \colltag{Coll} \humantag{Human} \llmtag{LLM} & \humantag{Human} \ruletag{Rule} \\ \hline
OVO-Bench & \cite{li2025ovobench} & \textbf{U3}\, Streaming Video Understanding & Episodic memory, \etc\ (12 tasks) & BI MC DF & \ruletag{Rule} & \colltag{Coll} \humantag{Human} \ruletag{Rule} \llmtag{LLM} & \humantag{Human} \\ \hline
RTV-Bench & \cite{xun2025rtv} & \textbf{U3}\, Streaming Video Understanding & Multi-timestamp real-time continuous video QA & MC & \ruletag{Rule} & \colltag{Coll} \humantag{Human} \llmtag{LLM} & \humantag{Human} \\ \hline
SPORTU & \cite{zou2024sportu} & \textbf{U3}\, Multi-Discipline Understanding & Slow-motion sports video recognition, foul detection and rule QA & MC & \ruletag{Rule} & \colltag{Coll} \humantag{Human} & \humantag{Human} \\ \hline
Temp\allowbreak{}Compass & \cite{liu2024TempCompass} & \textbf{U3}\, Spatial-Temporal Perception & Temporal multi-choice QA; captioning & BI MC FREE & \ruletag{Rule} \llmtag{LLM} & \colltag{Coll} \humantag{Human} \ruletag{Rule} \llmtag{LLM} & \humantag{Human} \\ \hline
Time\allowbreak{}Lens-Bench & \cite{zhang2026timelensbench} & \textbf{U3}\, Spatial-Temporal Perception & Video temporal grounding (3 datasets) & DF & \ruletag{Rule} & \colltag{Coll} \humantag{Human} & \humantag{Human} \\ \hline
V-NIAH & \cite{zhang2024vniah} & \textbf{U3}\, Long-context Understanding & Needle-in-a-haystack video QA & MC DF & \ruletag{Rule} & \colltag{Coll} \humantag{Human} \ruletag{Rule} \llmtag{LLM} & -- \\ \hline
VDC & \cite{chai2025vdc} & \textbf{U3}\, Video Comprehensive Understanding & Video detailed captioning (5 caption types) & FREE & \llmtag{LLM} & \colltag{Coll} \humantag{Human} \llmtag{LLM} & \humantag{Human} \\ \hline
Vid\allowbreak{}Composition & \cite{tang2025vidcomposition} & \textbf{U3}\, Higher-order Perception & Cinematic camera movement, shot composition and narrative QA & MC & \ruletag{Rule} & \colltag{Coll} \humantag{Human} \llmtag{LLM} & \humantag{Human} \\ \hline
Video-MME & \cite{fu2024Video-MME} & \textbf{U3}\, Video Comprehensive Understanding & Action recognition, \etc\ (12 total) & MC & \ruletag{Rule} & \colltag{Coll} \humantag{Human} & \humantag{Human} \llmtag{LLM} \\ \hline
Video-MMMU & \cite{hu2026videommmu} & \textbf{U3}\, Multi-Discipline Understanding & Educational video QA (30 subjects) & MC DF & \ruletag{Rule} & \colltag{Coll} \humantag{Human} \llmtag{LLM} & \humantag{Human} \llmtag{LLM} \\ \hline
Video\allowbreak{}NIAH & \cite{zhao2024VideoNIAH} & \textbf{U3}\, Spatial-Temporal Perception & Retrieval; ordering; counting & MC & \ruletag{Rule} & \colltag{Coll} \ruletag{Rule} & \ruletag{Rule} \\ \hline
Vript-Hard & \cite{yang2024vripthard} & \textbf{U3}\, Video Comprehensive Understanding & Hallucination evaluation, \etc\ (3 tasks) & MC DF FREE & \ruletag{Rule} \llmtag{LLM} & \colltag{Coll} \humantag{Human} \llmtag{LLM} & \humantag{Human} \\ \hline
VSI-Bench-Debiased & \cite{brown2026vsibenchdebiased} & \textbf{U3}\, 3D Understanding & Debiased spatial video QA & MC DF & \ruletag{Rule} & \colltag{Coll} & \ruletag{Rule} \\ \hline
AVUT & \cite{yang2025avut} & \textbf{U4}\, Audio Understanding & Multiple-choice QA; Open-ended QA & MC FREE & \ruletag{Rule} & \colltag{Coll} \humantag{Human} & \humantag{Human} \llmtag{LLM} \\ \hline
CA-MER & \cite{han2025camer} & \textbf{U4}\, Higher-order Perception & Audio-video emotion conflict reasoning & DF & \ruletag{Rule} \llmtag{LLM} & \colltag{Coll} \llmtag{LLM} & \humantag{Human} \ruletag{Rule} \\ \hline
HLR-AVScene\allowbreak{}QA & \cite{kim2026hlravsceneqa} & \textbf{U4}\, Omni-modal Understanding & Binaural sound recognition, spatial localization and scene reasoning & MC FREE & \ruletag{Rule} \llmtag{LLM} & \colltag{Coll} \ruletag{Rule} \llmtag{LLM} & \humantag{Human} \ruletag{Rule} \\ \hline
MCUB & \cite{chen2024mcub} & \textbf{U4}\, Omni-modal Understanding & Cross-modal commonality identification & MC & \ruletag{Rule} & \colltag{Coll} \llmtag{LLM} & \ruletag{Rule} \\ \hline
Omni\allowbreak{}Bench & \cite{li2024omnibench} & \textbf{U4}\, Omni-modal Understanding & Tri-modal simultaneous image, audio and text QA & MC & \ruletag{Rule} & \colltag{Coll} \humantag{Human} & \humantag{Human} \llmtag{LLM} \\ \hline
Omni\allowbreak{}Video\allowbreak{}Bench & \cite{li2026omnivideobench} & \textbf{U4}\, Omni-modal Understanding & Audio-visual reasoning (13 question types) & MC & \ruletag{Rule} & \colltag{Coll} \humantag{Human} & \humantag{Human} \llmtag{LLM} \\ \hline
World\allowbreak{}Sense & \cite{hong2026worldsense} & \textbf{U4}\, Omni-modal Understanding & Audio-visual understanding, \etc\ (26 tasks) & MC & \ruletag{Rule} & \colltag{Coll} \humantag{Human} & \humantag{Human} \ruletag{Rule} \llmtag{LLM} \\ \hline
\end{tabularx}\endgroup\end{table*}

%% file: sec/understanding/bg.tex
\subsection{Taxonomy by Perceptual and Contextual Scope}
\label{sec:understanding_bg}

Multimodal understanding benchmarks test whether an MLLM can extract, align, and interpret perceptual and semantic evidence from non-text inputs~\cite{alayrac2022flamingo,zhu2024multimodal,ye2024mm,zhou2024mlvu}. Separating visual perception, contextual comprehension, and modality-specific lists creates overlapping buckets because modality and context vary independently. We instead organize all 107 Understanding benchmarks (Fig.~\ref{fig:map-understanding} and Table~\ref{tbl:understanding}) along one observable axis: the \textbf{perceptual and contextual scope} required by the representative track. This yields four canonical categories: single-instance perception (\textbf{U1}, 50 benchmarks), multi-image and long-context understanding (\textbf{U2}, 23), temporal and streaming understanding (\textbf{U3}, 27), and audio-visual or omni-modal understanding (\textbf{U4}, 7).

\subsubsection{Single-Instance Perception}
\label{sec:u1_single_item}
Single-instance benchmarks (\textbf{U1}) test whether a model can extract low-level, fine-grained, affective, semantic, or sensory properties from an individual perceptual unit (an image, document page, short clip, audio recording, or 3D asset)~\cite{li2024densefusion}:

\begin{itemize}
 \item \textbf{Low-Level Perceptual Quality.} Low-level perception involves detecting basic visual attributes (\eg, color, lighting, composition) and distortions (\eg, blur, noise, artifacts) that do not require external knowledge~\cite{zhang2024q-bench}. Q-Bench+~\cite{zhang2024q-bench} extends single-image distortion QA to image-pair comparisons and quality descriptions, applying softmax pooling over ``good'' and ``poor'' token logits to obtain continuous quality scores, while Q-Bench-Video~\cite{zhang2025qbenchvideo} generalizes low-level quality evaluation to video clips.

 \item \textbf{Fine-Grained Perception, OCR, and Perceptual Boundaries.} This layer evaluates recognition of small targets, subtle visual-linguistic distinctions, and dense text. For text and document parsing, OCRBench v2~\cite{fu2025ocrbenchv2} and CC-OCR~\cite{yang2025ccocr} expand scene-text OCR to text grounding, multilingual scripts, and key information extraction; OmniDocBench~\cite{ouyang2025omnidocbench} scores whole-page PDF parsing (layout, tables, formulas) via normalized edit distance and Tree-Edit-Distance-based Similarity (TEDS); and EgoTextVQA~\cite{zhou2025egotextvqa} and MME-VideoOCR~\cite{shi2025mme} extend scene-text reading to egocentric and dynamic video frames. For subtle visual distinctions and high-resolution imagery, MMVP~\cite{tong2024mmvp} pairs CLIP-blind images, CV-Bench~\cite{tong2024cvbench} tests 2D/3D spatial perception, and HR-Bench~\cite{wang2024hrbench} and MME-RealWorld~\cite{zhang2025mmerealworld} probe 4K/8K scenes (using cyclic option permutation in HR-Bench to remove position bias). Region-level description and perceptual boundaries are tested by MDVP-Bench~\cite{lin2024MDVP-Bench}, DLC-Bench~\cite{lian2025dlcbench}, VideoRefer-Bench~\cite{yuan2025videoreferbench}, NegBench~\cite{alhamoud2025negbench} (negation), HueManity~\cite{grover2025huemanity} (Ishihara color plates), and UNICBench~\cite{rong2026unicbench} (cross-modal counting). Domain-focused perception suites cover industrial defects (VisA-D\&R~\cite{xu2025visadr}), entity-linked QA (ImageWikiQA~\cite{zhang2024imagewikiqa}), universal information extraction (MUIE~\cite{zhang2024muie}), online driving perception (ODV-Bench~\cite{zeng2025odvbench}), and perceptual self-awareness requiring abstention on ``known unknowns'' (MM-SAP~\cite{wang2024mm-sap}).

 \item \textbf{Higher-Order Affective, Implicit, and Cinematic Perception.} Higher-order benchmarks~\cite{street2024llms} assess emotional, implicit, and aesthetic understanding: facial action units in FABA-Bench~\cite{li2024facial}, step-scored multimodal emotion recognition in MME-Emotion~\cite{zhang2026mmeemotion}, unstated visual implications and product attributes in II-Bench~\cite{liu2024ii} and ImplicitAVE~\cite{zou2024implicitave}, multimodal humor and punchline comprehension under contrastively edited captions in PunchBench~\cite{ouyang2025punchbench}, and film shot grammar in ShotBench~\cite{liu2025shotbench}.

 \item \textbf{Comprehensive, Multilingual, and Cultural Perception.} General perception suites span diverse visual tasks~\cite{fu2024blink,yingmmt,pi2024image,chen2024MMStar,li2024naturalbench,zhang2025lmmseval}, continual instruction tuning (UCIT~\cite{guo2025ucit}), and blind-filtered multidisciplinary perception (MMMU-Pro~\cite{yue2025mmmupro}). To counter contamination and text-only shortcuts, MMStar~\cite{chen2024MMStar} defines Multi-modal Gain (MG) and Multi-modal Leakage (ML), NaturalBench~\cite{li2024naturalbench} requires joint accuracy across paired image--question quartets, and LMMs-Eval~\cite{zhang2025lmmseval} pairs pruned subsets with live web-sourced items. Multilingual and culturally grounded suites broaden coverage to up to 100 languages and regional traditions~\cite{tang2024mtvqa,romero2024cvqa,alwajih2024peacock,sun2024parrot,nayak2024culturalvqa,vayani2025almbench,winata2025worldcuisines}.

 \item \textbf{Foundational Audio and 3D Perception.} Single-instance perception also covers speech, sound, and native 3D representations. AIR-Bench~\cite{yang2024air} evaluates foundational and chat-oriented audio tasks, MMAU~\cite{sakshi2025mmau} and MMAR~\cite{ma2025mmar} probe expert audio perception and reasoning, MMSU~\cite{wang2026mmsu} covers spoken-language understanding, and VoiceBench~\cite{chen2026voicebench} evaluates voice assistants across real and synthetic spoken instructions. In 3D, LLaNA~\cite{amaduzzi2024LLaNA} directly processes NeRF weights for captioning and QA, and 3D MM-Vet~\cite{qi20243dmmvet} evaluates 3D object comprehension via GPT-4 grading.
\end{itemize}

\subsubsection{Multi-Image and Long-Context Understanding}
\label{sec:u2_contextual}
Multi-image and long-context benchmarks (\textbf{U2}, 23 benchmarks) require integrating evidence across multiple images, retrieved passages, dialogue turns, or interleaved demonstrations, where any single instance is insufficient:

\begin{itemize}
 \item \textbf{Context-Dependent Disambiguation.} CODIS~\cite{luo2024codis} pairs ambiguous images with contrasting textual contexts and measures context awareness; MRAG-Bench~\cite{hu2025mragbenchvisioncentricevaluationretrievalaugmented} quantifies gains from retrieved visual evidence, while OHRBench~\cite{zhang2025ohrbench} measures cascading semantic and formatting OCR noise in retrieval-augmented document pipelines; and Persona-MME~\cite{nie2026personamme} tracks user memory, intent, and preferences over long multimodal dialogues.

 \item \textbf{Multi-Image Comparison and Sequence Tracking.} MIC\allowbreak{}Bench~\cite{wu2024micbench} extends visual quality comparisons to multi-image groups, MTMEUR~\cite{hu2025mtmeur} links multi-turn emotional cues, and Muir\allowbreak{}Bench~\cite{wang2024MuirBench}, Mantis-Eval~\cite{Jiang2024MANTISIM}, MMIU~\cite{meng2024MMIU}, MC-Bench~\cite{xu2025mcbench}, and MMDU~\cite{liu2024mmdu} evaluate cross-image correspondence and multi-turn grounding. Mementos~\cite{wang2024mementos} tracks object and behavioral state transitions across image sequences using GPT-4 keyword standardization for precision, recall, and F1 scoring.

 \item \textbf{Long-Context Needle Retrieval.} Motivated by multi-round QA~\cite{li2022mmcoqa}, action understanding~\cite{wu2star}, and scientific reading~\cite{pramanick2024spiqa}, MMNeedle~\cite{wang2024multimodal} scores sequence-, image-, and sub-image-level needle localization via existence, index, and exact accuracy. MileBench~\cite{song2024milebench} and MM-NIAH~\cite{wang2024MM-NIAH} evaluate retrieval and counting inside long multimodal haystacks, and MMLongBench-Doc~\cite{ma2024mmlongbenchdoc} and LongDocURL~\cite{deng2025longdocurl} extend cross-page comprehension, numerical reasoning, and layout localization to lengthy PDF documents.

 \item \textbf{Interleaved Documents and Multimodal In-Context Learning.} MIR~\cite{du2025mir} and LongVLBench~\cite{zhang2026longvlbench} (alongside CoMM~\cite{chen2024comm}, catalogued under interleaved generation in Table~\ref{tbl:generation}) test comprehension over long interleaved image--text streams, and TokenSwap-Bench~\cite{hua2026tokenswapbench} isolates the modality gap when textual concepts are replaced by images. For multimodal in-context learning, VL-ICLBench~\cite{zong2024VL-ICLBench} spans ten few-shot visual tasks, Demo-ICL-Bench~\cite{dong2026demoiclbench} tests procedural learning from text or video demonstrations, and ICCA~\cite{hua2024icca} measures listener accuracy and communication efficiency in repeated reference games.
\end{itemize}

\subsubsection{Temporal and Streaming Understanding}
\label{sec:u3_temporal}
Temporal benchmarks (\textbf{U3}, 27 benchmarks) evaluate comprehension over time-ordered video or long-form audio streams, where answers depend on temporal order, long-horizon memory, or online deadlines:

\begin{itemize}
 \item \textbf{Temporal Order, Grounding, and Needle Retrieval.} TempCompass~\cite{liu2024TempCompass} pairs videos with identical static appearance but opposite motion or speed, OSCaR~\cite{nguyen2024oscar} tracks object state changes, FAVOR-Bench~\cite{tu2025favorbench} and MA-Bench~\cite{li2026mabench} assess fine-grained video motion and subtle human micro-actions, EgoExoBench~\cite{he2025egoexobench} aligns egocentric and exocentric perspectives, TimeLens-Bench~\cite{zhang2026timelensbench} scores temporal segment localization via IoU, and VideoNIAH~\cite{zhao2024VideoNIAH} and V-NIAH~\cite{zhang2024vniah} insert synthetic visual needles into long videos up to multi-hour durations.

 \item \textbf{Long-Horizon Video and Audio Comprehension.} Long-duration retention is probed by MLVU~\cite{zhou2024mlvu} (extending early memory designs such as MovieChat~\cite{song2024moviechat}), LongVideoBench~\cite{wu2024longvideobench} (referring reasoning over hour-long subtitled videos), HourVideo~\cite{chandrasegaran2024hourvideo} (1-hour egocentric summarization, perception, and navigation), and LVBench~\cite{wang2025lvbench} (filtered against frame-free shortcuts). Comprehensive and domain-focused video suites include Video-MME~\cite{fu2024Video-MME}, educational lecture adaptation in Video-MMMU~\cite{hu2026videommmu}, slow-motion sports foul and rule understanding in SPORTU~\cite{zou2024sportu}, compiled cinematic composition in VidComposition~\cite{tang2025vidcomposition}, daily-living recognition in ADLMCQ~\cite{chakraborty2024ADLMCQ}, gaze-guided QA in EgoGazeVQA~\cite{peng2025eyemllm}, hierarchical anomaly understanding in HIVAU-70k~\cite{zhang2025hivau70k}, fine-grained video captioning in VDC~\cite{chai2025vdc} and Vript-Hard~\cite{yang2024vripthard}, and debiased indoor video-spatial perception in VSI-Bench-Debiased~\cite{brown2026vsibenchdebiased}. In audio, MuChoMusic~\cite{weck2024MuChoMusic} and LongAudioBench~\cite{ghosh2025longaudiobench} evaluate temporal musical structure and long-form spoken recordings.

 \item \textbf{Online Streaming Comprehension.} Whereas offline protocols expose the entire recording upfront, OVO-Bench~\cite{li2025ovobench} queries models causally along arriving frames across backward tracing, real-time perception, and forward active responding, RTV-Bench~\cite{xun2025rtv} evaluates continuous multi-timestamp perception and reasoning on real-time video streams, and Inf-Streams-Eval~\cite{xu2026infstreamseval} scores continuous real-time commentary via pairwise LLM win rates.
\end{itemize}

\subsubsection{Audio-Visual and Omni-Modal Understanding}
\label{sec:u4_cross_signal}
Cross-modal benchmarks (\textbf{U4}, 7 benchmarks) require joint synthesis across complementary or conflicting sensory channels. AVUT~\cite{yang2025avut} enforces audio-visual necessity in video understanding through answer-permutation filtering, while CA-MER~\cite{han2025camer} pairs conflicting audio and visual affect cues to diagnose modality dominance. At the omni-modal level, MCUB~\cite{chen2024mcub} evaluates cross-modal concept commonality, OmniBench~\cite{li2024omnibench} tests simultaneous tri-modal image--audio--text integration, HLR-AVSceneQA~\cite{kim2026hlravsceneqa} benchmarks binaural sound recognition, spatial localization, and scene reasoning, and WorldSense~\cite{hong2026worldsense} and OmniVideoBench~\cite{li2026omnivideobench} construct real-world audio-visual questions unsolvable from either stream alone.

%% file: sec/understanding/task.tex
\subsection{Task Formats, Evaluation Metrics, and Synthesis}
\label{sec:understanding_task}

Across all four perceptual and contextual categories, understanding benchmarks rely on three response formats that govern scoring precision and judge dependence (Table~\ref{tbl:understanding}):

\begin{itemize}
\item \textbf{Binary and Multiple-Choice (BI / MC).} Closed-form options~\cite{zhang2025mmerealworld} enable deterministic rule-based accuracy, precision, and recall. To suppress option-position bias and parsing failures, HR-Bench~\cite{wang2024hrbench} averages accuracy over cyclic permutations, NaturalBench~\cite{li2024naturalbench} requires all four paired image--question items to be solved jointly, and suites such as MMT-Bench~\cite{yingmmt} adopt OpenCompass's cascading regex-and-LLM option extractor~\cite{contributors2023opencompass,yue2024mmmu,hu2025mragbenchvisioncentricevaluationretrievalaugmented,du2025mir}.

\item \textbf{Defined-Form (DF).} Structured outputs match task geometry without open-ended judge variance: coordinate tuples in MMNeedle~\cite{wang2024multimodal}, temporal intervals scored by IoU in TimeLens-Bench~\cite{zhang2026timelensbench}, continuous estimates scored by Mean Relative Accuracy (MRA)~\cite{brown2026vsibenchdebiased}, and layout/table/formula trees scored by normalized edit distance and TEDS~\cite{fu2025ocrbenchv2,yang2025ccocr,ouyang2025omnidocbench}.

\item \textbf{Free-Form (FREE).} Open-ended descriptions~\cite{liu2024mmdu,chai2025vdc} prevent candidate-option leakage at the cost of automated grading sensitivity. Reference n-gram metrics (BLEU, ROUGE-L)~\cite{song2024milebench,nguyen2024oscar,zhang2025hivau70k} are widely paired or replaced with LLM-as-a-Judge~\cite{zheng2024judging}, spanning rubric scoring in 3D MM-Vet and Q-Bench+~\cite{qi20243dmmvet,zhang2024q-bench,liu2023mmbench}, atomic question-decomposition grading in VDC and DLC-Bench~\cite{chai2025vdc,lian2025dlcbench}, step-wise affect scoring in MME-Emotion~\cite{zhang2026mmeemotion}, and pairwise win rates in Inf-Streams-Eval~\cite{xu2026infstreamseval}.
\end{itemize}

\textbf{Analytical synthesis.} Perceptual and contextual scope dictates what an understanding score can establish. Single-instance suites (\textbf{U1}, 50/107 benchmarks) have shifted from coarse four-option recognition toward paired contrast sets and blind-filtered tests: replacing marginal item accuracy with paired consistency causes 40--55 pp drops in NaturalBench~\cite{li2024naturalbench}, and embedding text options into screenshots produces 16.8--26.9 pp drops in MMMU-Pro~\cite{yue2025mmmupro} (alongside visual-necessity controls in MMVP~\cite{tong2024mmvp} and MMStar~\cite{chen2024MMStar}). Multi-image and long-context suites (\textbf{U2}, 23/107) expose steep context-length degradation when distractors or OCR noise accumulate across pages~\cite{ma2024mmlongbenchdoc,zhang2025ohrbench}, whereas online streaming (\textbf{U3}) and audio-visual/omni-modal suites (\textbf{U4}, 7/107) remain the scarcest sub-pillars: only four of 73 video benchmarks in our entire corpus enforce causal streaming deadlines (\s\ref{sec:protocols}), and conflicting audio--visual cues in \textbf{U4} reveal strong visual or textual dominance over acoustic evidence~\cite{han2025camer,hong2026worldsense}.

%% file: sec/reasoning.tex
\section{Reasoning Benchmarks}
\label{sec:reasoning}
\begin{figure*}[tp]
\centering\chaptergraphic[width=\textwidth]{reasoning_map}
\caption{Taxonomy and representative examples of Reasoning benchmarks ($N=73$) organized by supervision granularity and visual grounding (\textbf{R1--R4}). Illustrated tasks cover 3D spatial and quantitative inference in 3DSRBench and MATH-Vision~\cite{ma20253dsrbench,wang2024measuring}, grounded spatial pointing in RefSpatial-Bench~\cite{zhou2025refspatialbench}, intermediate reasoning verification in VRC-Bench~\cite{thawakar2025vrcbench}, and preference judging in MLLM-as-a-Judge~\cite{chen2024mllmasajudge}.}\label{fig:map-reasoning}
\end{figure*}

\input{sec/misc/tbl_reasoning}

\input{sec/reasoning/bg}

\input{sec/reasoning/task}

%% file: sec/misc/tbl_reasoning.tex
\input{sec/misc/tbl_reasoning_clean}

%% file: sec/misc/tbl_reasoning_clean.tex
\begin{table*}[tp]\centering
\caption{Overview of Reasoning benchmarks ($N=73$), spanning \textbf{R1} outcome-based multimodal reasoning (50), \textbf{R2} visually grounded reasoning (9), \textbf{R3} chain-of-thought and process reasoning (3), and \textbf{R4} reflective critique, self-improvement, and editing (11). Output formats (\textbf{BI}, \textbf{MC}, \textbf{DF}, \textbf{FREE}) and pipeline tags follow Table~\ref{tbl:understanding}.}
\label{tbl:reasoning}
\begingroup\linespread{0.96}\fontsize{7.2}{8.1}\selectfont\setlength\tabcolsep{1.25pt}\renewcommand\arraystretch{1.00}
\prepareSurveyTableTags\begin{tabularx}{\textwidth}{@{}>{\raggedright\arraybackslash}p{.133\textwidth}l>{\raggedright\arraybackslash}p{.166\textwidth}>{\raggedright\arraybackslash}X>{\raggedright\arraybackslash}p{.070\textwidth}>{\raggedright\arraybackslash\spaceskip=1pt}p{.089\textwidth}>{\raggedright\arraybackslash\spaceskip=1pt}p{.160\textwidth}>{\raggedright\arraybackslash\spaceskip=1pt}p{.090\textwidth}@{}}
\hline\thickhline\rowcolor{mygray}\textbf{Benchmark} & \textbf{Ref.} & \textbf{Category \& Focus} & \textbf{Task} & \textbf{Output} & \textbf{Evaluation} & \textbf{Construction} & \textbf{Filter} \\ \hline
3DSR\allowbreak{}Bench & \cite{ma20253dsrbench} & \textbf{R1}\, Relational Reasoning; Spatial Reasoning & 3D spatial reasoning (12 question types) & BI MC & \ruletag{Rule} & \colltag{Coll} \humantag{Human} \ruletag{Rule} & \humantag{Human} \\ \hline
CFMM & \cite{li2024eyes} & \textbf{R1}\, Reflective Reasoning & Assumptive VQA (6 question types) & MC & \ruletag{Rule} & \colltag{Coll} \humantag{Human} \llmtag{LLM} & \humantag{Human} \\ \hline
Chart\allowbreak{}QA\allowbreak{}Pro & \cite{masry2025chartqapro} & \textbf{R1}\, Domain-specific Reasoning & Chart QA with diverse question types & BI MC DF & \ruletag{Rule} & \colltag{Coll} \humantag{Human} \llmtag{LLM} & \humantag{Human} \\ \hline
Chart\allowbreak{}X & \cite{xia2024chartx} & \textbf{R1}\, Domain-specific Reasoning & Chart image-QA & DF FREE & \ruletag{Rule} \llmtag{LLM} & \humantag{Human} \ruletag{Rule} \llmtag{LLM} & \humantag{Human} \ruletag{Rule} \llmtag{LLM} \\ \hline
Char\allowbreak{}Xiv & \cite{wang2024charxiv} & \textbf{R1}\, Domain-specific Reasoning & Chart image-QA & DF & \llmtag{LLM} & \colltag{Coll} \humantag{Human} \llmtag{LLM} & \humantag{Human} \ruletag{Rule} \\ \hline
CMMU & \cite{he2024cmmu} & \textbf{R1}\, Domain-specific Reasoning & High school-level Chinese multi-discipline multimodal understanding & MC DF & \ruletag{Rule} \llmtag{LLM} & \colltag{Coll} & \humantag{Human} \\ \hline
Comp\allowbreak{}Bench & \cite{kil2024CompBench} & \textbf{R1}\, Relational Reasoning; Comparative Reasoning & Comparative QA & MC & \ruletag{Rule} & \colltag{Coll} \humantag{Human} \ruletag{Rule} \llmtag{LLM} & \humantag{Human} \ruletag{Rule} \\ \hline
CRPE & \cite{wang2024CRPE} & \textbf{R1}\, Relational Reasoning & Relation conversation & MC & \ruletag{Rule} & \colltag{Coll} \ruletag{Rule} \llmtag{LLM} & \humantag{Human} \\ \hline
Dyna\allowbreak{}Math & \cite{zou2025dynamath} & \textbf{R1}\, Domain-specific Reasoning & Math reasoning under program-generated variants & MC DF & \ruletag{Rule} & \colltag{Coll} \humantag{Human} \ruletag{Rule} & \humantag{Human} \\ \hline
EEE-Bench & \cite{li2025eee-bench} & \textbf{R1}\, Domain-specific Reasoning & Electrical and electronics engineering circuit and system QA (10 subdomains) & MC DF & \ruletag{Rule} & \colltag{Coll} \humantag{Human} & \humantag{Human} \\ \hline
EMMA & \cite{hao2025emma} & \textbf{R1}\, Domain-specific Reasoning & Multimodal STEM reasoning, \etc & MC DF & \ruletag{Rule} & \colltag{Coll} \humantag{Human} \ruletag{Rule} \llmtag{LLM} & \humantag{Human} \llmtag{LLM} \\ \hline
Humanity's Last Exam & \cite{phan2025humanityslastexam} & \textbf{R1}\, Domain-specific Reasoning & Closed-ended expert academic QA & MC DF & \llmtag{LLM} & \humantag{Human} & \humantag{Human} \llmtag{LLM} \\ \hline
M3Co\allowbreak{}T & \cite{chen2024m3cot} & \textbf{R1}\, Multi-step Reasoning & Multimodal chain-of-thought QA & MC & \ruletag{Rule} & \colltag{Coll} \humantag{Human} \ruletag{Rule} \llmtag{LLM} & \humantag{Human} \ruletag{Rule} \\ \hline
MARVEL & \cite{jiang2024MARVEL} & \textbf{R1}\, Relational Reasoning & Abstract visual reasoning & BI MC DF & \ruletag{Rule} & \colltag{Coll} \humantag{Human} \ruletag{Rule} & \humantag{Human} \\ \hline
MATH-Vision & \cite{wang2024measuring} & \textbf{R1}\, Domain-specific Reasoning & Mathematics & MC DF & \ruletag{Rule} & \colltag{Coll} & \humantag{Human} \ruletag{Rule} \\ \hline
MATHCHECK-GEO & \cite{zhou2024your} & \textbf{R1}\, Domain-specific Reasoning & Mathematics & BI DF & \ruletag{Rule} & \colltag{Coll} \ruletag{Rule} \llmtag{LLM} & \humantag{Human} \\ \hline
Math\allowbreak{}Verse & \cite{zhang2024mathverse} & \textbf{R1}\, Domain-specific Reasoning & Mathematics & MC DF & \ruletag{Rule} \llmtag{LLM} & \colltag{Coll} \humantag{Human} & \humantag{Human} \\ \hline
Micro\allowbreak{}VQA & \cite{burgess2025microvqa} & \textbf{R1}\, Domain-specific Reasoning & Multimodal VQA & MC & \ruletag{Rule} & \humantag{Human} & \humantag{Human} \\ \hline
MMSI-Bench & \cite{yang2026mmsibench} & \textbf{R1}\, Relational Reasoning; Spatial Reasoning & Multi-image spatial reasoning (11 task types) & MC & \ruletag{Rule} & \colltag{Coll} \humantag{Human} & \humantag{Human} \\ \hline
MMVU & \cite{zhao2025mmvu} & \textbf{R1}\, Domain-specific Reasoning & Expert-level video reasoning (27 subjects) & MC FREE & \llmtag{LLM} & \colltag{Coll} \humantag{Human} & \humantag{Human} \\ \hline
MM\allowbreak{}World & \cite{he2024MMWorld} & \textbf{R1}\, Domain-specific Reasoning & Multi-discipline multi-faceted QA in videos & MC & \ruletag{Rule} \llmtag{LLM} & \colltag{Coll} \humantag{Human} \llmtag{LLM} & \humantag{Human} \\ \hline
MULTI & \cite{zhu2024multi} & \textbf{R1}\, Domain-specific Reasoning & Multilingual, multimodal, multilevel benchmark & MC DF FREE & \ruletag{Rule} & \colltag{Coll} \humantag{Human} \llmtag{LLM} & \humantag{Human} \\ \hline
Olympiad\allowbreak{}Bench & \cite{he2024olympiadbench} & \textbf{R1}\, Domain-specific Reasoning & Olympiad-level math and physics & DF FREE & \ruletag{Rule} & \colltag{Coll} \humantag{Human} & \humantag{Human} \ruletag{Rule} \\ \hline
PAI-Bench & \cite{zhou2026paibench} & \textbf{R1}\, Relational Reasoning; Physical Reasoning & Physical AI video generation and understanding (3 tracks) & BI MC FREE & \ruletag{Rule} \llmtag{LLM} & \colltag{Coll} \humantag{Human} \llmtag{LLM} & \humantag{Human} \\ \hline
Phys\allowbreak{}Bench & \cite{chow2025physbench} & \textbf{R1}\, Relational Reasoning; Physical Reasoning & Object property, \etc\ (19 subclasses) & MC & \ruletag{Rule} & \colltag{Coll} \humantag{Human} \ruletag{Rule} \llmtag{LLM} & \humantag{Human} \ruletag{Rule} \\ \hline
Quanti\allowbreak{}Phy & \cite{puyin2026quantiphy} & \textbf{R1}\, Relational Reasoning; Physical Reasoning & Physical quantity estimation from video & DF & \ruletag{Rule} & \colltag{Coll} \humantag{Human} \ruletag{Rule} & \humantag{Human} \\ \hline
R1-Onevision-Bench & \cite{yang2025r1onevisionbench} & \textbf{R1}\, Domain-specific Reasoning & Grade-level multi-subject reasoning & MC DF & \llmtag{LLM} & -- & -- \\ \hline
Re\allowbreak{}X\allowbreak{}Time & \cite{chen2024rextime} & \textbf{R1}\, Relational Reasoning; Temporal Reasoning & QA each belong to different time spans & MC DF & \ruletag{Rule} & \colltag{Coll} \humantag{Human} \llmtag{LLM} & \humantag{Human} \ruletag{Rule} \llmtag{LLM} \\ \hline
Robo\allowbreak{}Spatial & \cite{song2025robospatial} & \textbf{R1}\, Relational Reasoning; Spatial Reasoning & Spatial relation QA for robotics & BI DF & \ruletag{Rule} & \colltag{Coll} \humantag{Human} \ruletag{Rule} & \ruletag{Rule} \\ \hline
Sce\allowbreak{}MQA & \cite{liang2024scemqa} & \textbf{R1}\, Domain-specific Reasoning & High school multi-discipline QA & BI MC DF & \ruletag{Rule} \llmtag{LLM} & \colltag{Coll} \humantag{Human} & \humantag{Human} \\ \hline
SEED-Bench-R1 & \cite{chen2026seedbenchr1} & \textbf{R1}\, Multi-step Reasoning & Egocentric next-action planning & MC & \ruletag{Rule} & \colltag{Coll} \ruletag{Rule} & \humantag{Human} \\ \hline
See\allowbreak{}Phys & \cite{xiang2025seephys} & \textbf{R1}\, Domain-specific Reasoning & Diagram-essential physics reasoning from middle school to PhD exams & DF & \ruletag{Rule} \llmtag{LLM} & \colltag{Coll} \humantag{Human} & \humantag{Human} \llmtag{LLM} \\ \hline
SOK-Bench & \cite{wang2024SOK-Bench} & \textbf{R1}\, Relational Reasoning; Temporal Reasoning & Video QA & MC FREE & \ruletag{Rule} & \colltag{Coll} \humantag{Human} \llmtag{LLM} & \humantag{Human} \\ \hline
SPACE & \cite{ramakrishnan2025spatialcognition} & \textbf{R1}\, Relational Reasoning & Mapping; Navigation; Mental rotation; Object layout judgment & MC & \ruletag{Rule} & \humantag{Human} & \ruletag{Rule} \\ \hline
Spa\allowbreak{}CE-10 & \cite{gong2026space10} & \textbf{R1}\, Relational Reasoning; Spatial Reasoning & Atomic and compositional 3D indoor spatial intelligence QA & MC & \ruletag{Rule} & \colltag{Coll} \humantag{Human} \ruletag{Rule} \llmtag{LLM} & \humantag{Human} \\ \hline
\end{tabularx}\endgroup\end{table*}
\begin{table*}[tp]\centering
\ContinuedFloat
\caption[]{Overview of Reasoning benchmarks (continued).}
\begingroup\linespread{0.96}\fontsize{7.2}{8.1}\selectfont\setlength\tabcolsep{1.25pt}\renewcommand\arraystretch{1.00}
\prepareSurveyTableTags\begin{tabularx}{\textwidth}{@{}>{\raggedright\arraybackslash}p{.133\textwidth}l>{\raggedright\arraybackslash}p{.166\textwidth}>{\raggedright\arraybackslash}X>{\raggedright\arraybackslash}p{.070\textwidth}>{\raggedright\arraybackslash\spaceskip=1pt}p{.089\textwidth}>{\raggedright\arraybackslash\spaceskip=1pt}p{.160\textwidth}>{\raggedright\arraybackslash\spaceskip=1pt}p{.090\textwidth}@{}}
\hline\thickhline\rowcolor{mygray}\textbf{Benchmark} & \textbf{Ref.} & \textbf{Category \& Focus} & \textbf{Task} & \textbf{Output} & \textbf{Evaluation} & \textbf{Construction} & \textbf{Filter} \\ \hline
SPAR-Bench & \cite{zhang2025sparbench} & \textbf{R1}\, Relational Reasoning; Spatial Reasoning & Spatial perception and reasoning (20 tasks) & MC DF & \ruletag{Rule} & \colltag{Coll} \ruletag{Rule} \llmtag{LLM} & \humantag{Human} \ruletag{Rule} \\ \hline
Spatial\allowbreak{}RGPT-Bench & \cite{cheng2024spatialrgpt} & \textbf{R1}\, Relational Reasoning & Grounded spatial reasoning & BI DF & \ruletag{Rule} \llmtag{LLM} & \colltag{Coll} \ruletag{Rule} & \humantag{Human} \\ \hline
SPIQA & \cite{pramanick2024spiqa} & \textbf{R1}\, Domain-specific Reasoning & Scientific figure and table QA & FREE & \ruletag{Rule} \llmtag{LLM} & \colltag{Coll} \llmtag{LLM} & \humantag{Human} \\ \hline
V\allowbreak{}Cog-Bench & \cite{cao2024VCog-Bench} & \textbf{R1}\, Relational Reasoning & Pattern relationship & MC & \ruletag{Rule} & \colltag{Coll} \humantag{Human} \llmtag{LLM} & \humantag{Human} \\ \hline
Video\allowbreak{}Espresso & \cite{han2025videoespresso} & \textbf{R1}\, Multi-step Reasoning; Reasoning Process & Narrative, \etc\ (14 video reasoning tasks) & FREE & \ruletag{Rule} \llmtag{LLM} & \colltag{Coll} \llmtag{LLM} & \humantag{Human} \ruletag{Rule} \llmtag{LLM} \\ \hline
Video\allowbreak{}Think\allowbreak{}Bench & \cite{tong2026videothinkbench} & \textbf{R1}\, Multi-step Reasoning; Thinking with Images & Eyeballing puzzles, \etc & MC DF & \humantag{Human} \ruletag{Rule} \llmtag{LLM} & \colltag{Coll} \ruletag{Rule} & -- \\ \hline
Vis\allowbreak{}Num\allowbreak{}Bench & \cite{weng2025visnumbench} & \textbf{R1}\, Domain-specific Reasoning & Visual number sense and numerical estimation QA (7 attributes) & MC & \ruletag{Rule} & \colltag{Coll} \ruletag{Rule} & \humantag{Human} \ruletag{Rule} \\ \hline
Vis\allowbreak{}Res Bench & \cite{tortei2026visresbench} & \textbf{R1}\, Relational Reasoning & Prior-free perceptual completion and multi-attribute visual reasoning & MC & \ruletag{Rule} & \colltag{Coll} \ruletag{Rule} & \humantag{Human} \ruletag{Rule} \\ \hline
Visual Co\allowbreak{}T & \cite{shao2024VisualCoT} & \textbf{R1}\, Multi-step Reasoning & VQA; captioning & DF FREE & \ruletag{Rule} \llmtag{LLM} & \colltag{Coll} \ruletag{Rule} \llmtag{LLM} & \ruletag{Rule} \\ \hline
Visual\allowbreak{}Web\allowbreak{}Bench & \cite{liu2024VisualWebBench} & \textbf{R1}\, Domain-specific Reasoning & Webpage QA & MC DF FREE & \ruletag{Rule} & \colltag{Coll} \humantag{Human} \ruletag{Rule} \llmtag{LLM} & \humantag{Human} \\ \hline
Visu\allowbreak{}Logic & \cite{xu2026visulogic} & \textbf{R1}\, Multi-step Reasoning & Vision-centric logical reasoning (6 categories) & MC & \ruletag{Rule} & \colltag{Coll} \humantag{Human} & \humantag{Human} \ruletag{Rule} \\ \hline
VOILA & \cite{yilmaz2025voila} & \textbf{R1}\, Relational Reasoning & Open-ended multi-image visual analogical reasoning & DF FREE & \ruletag{Rule} \llmtag{LLM} & \colltag{Coll} \ruletag{Rule} \llmtag{LLM} & \humantag{Human} \\ \hline
VSI-Bench & \cite{yang2025vsibench} & \textbf{R1}\, Relational Reasoning; Spatial Reasoning & Video spatial QA (8 tasks) & MC DF & \ruletag{Rule} & \colltag{Coll} \humantag{Human} \ruletag{Rule} & \humantag{Human} \ruletag{Rule} \\ \hline
We-Math & \cite{qiao2025wemath} & \textbf{R1}\, Domain-specific Reasoning & Visual math by knowledge concept (67 concepts) & MC & \ruletag{Rule} & \colltag{Coll} \humantag{Human} & \humantag{Human} \\ \hline
WSI-Bench & \cite{liang2025wsillava} & \textbf{R1}\, Domain-specific Reasoning & Pathological analysis, including morphological understanding, diagnosis & BI MC & \ruletag{Rule} & \colltag{Coll} \llmtag{LLM} & \humantag{Human} \llmtag{LLM} \\ \hline
Ca\allowbreak{}ST-Bench & \cite{zhang2026castbench} & \textbf{R2}\, Relational Reasoning; Temporal Reasoning & Causal chain-grounded spatio-temporal video QA & MC DF & \ruletag{Rule} & \colltag{Coll} \humantag{Human} \llmtag{LLM} & \humantag{Human} \\ \hline
Multihop\allowbreak{}Spatial & \cite{lee2026multihopspatial} & \textbf{R2}\, Relational Reasoning; Spatial Reasoning & 1-to-3-hop compositional spatial reasoning with box grounding (Acc@50IoU) & MC DF & \ruletag{Rule} & \colltag{Coll} \ruletag{Rule} \llmtag{LLM} & \humantag{Human} \ruletag{Rule} \\ \hline
Reason\allowbreak{}VOS & \cite{bai2024tokenseg} & \textbf{R2}\, Multi-step Reasoning & Reasoning-guided video object segmentation & DF & \ruletag{Rule} & \colltag{Coll} \humantag{Human} \llmtag{LLM} & \humantag{Human} \llmtag{LLM} \\ \hline
Ref\allowbreak{}Spatial-Bench & \cite{zhou2025refspatialbench} & \textbf{R2}\, Relational Reasoning; Spatial Reasoning & Spatial referring point prediction & DF & \ruletag{Rule} & \humantag{Human} & -- \\ \hline
Re\allowbreak{}VOS & \cite{yan2024revos} & \textbf{R2}\, Relational Reasoning & Reasoning video object segmentation & DF & \ruletag{Rule} & \colltag{Coll} \humantag{Human} & -- \\ \hline
Scan\allowbreak{}Reason & \cite{zhu2024scanreason} & \textbf{R2}\, Relational Reasoning; Spatial Reasoning & 3D reasoning grounding & DF FREE & \ruletag{Rule} & \colltag{Coll} \llmtag{LLM} & \humantag{Human} \\ \hline
Tree\allowbreak{}Bench & \cite{wang2026treebench} & \textbf{R2}\, Multi-step Reasoning; Thinking with Images & Traceable visual grounded reasoning & MC DF & \ruletag{Rule} & \colltag{Coll} \humantag{Human} \llmtag{LLM} & \humantag{Human} \llmtag{LLM} \\ \hline
Video-Holmes & \cite{cheng2026videoholmes} & \textbf{R2}\, Multi-step Reasoning & Video clue localisation and reasoning (7 task types) & MC & \ruletag{Rule} & \colltag{Coll} \humantag{Human} \llmtag{LLM} & \humantag{Human} \\ \hline
XLRS-Bench & \cite{wang2025xlrsbench} & \textbf{R2}\, Domain-specific Reasoning & VQA; Visual Grounding; Image Captioning & MC DF FREE & \ruletag{Rule} & \colltag{Coll} \humantag{Human} & \humantag{Human} \\ \hline
Math\allowbreak{}Canvas-Bench & \cite{shi2026mathcanvasbench} & \textbf{R3}\, Multi-step Reasoning; Thinking with Images & Interleaved visual-textual math solving & DF & \llmtag{LLM} & \colltag{Coll} & \ruletag{Rule} \llmtag{LLM} \\ \hline
MME-Co\allowbreak{}T & \cite{jiang2025mmecot} & \textbf{R3}\, Multi-step Reasoning; Reasoning Process & CoT quality evaluation (6 domains) & MC DF & \ruletag{Rule} \llmtag{LLM} & \colltag{Coll} \humantag{Human} \llmtag{LLM} & \humantag{Human} \\ \hline
VRC-Bench & \cite{thawakar2025vrcbench} & \textbf{R3}\, Multi-step Reasoning; Reasoning Process & Step-by-step visual reasoning (8 categories) & DF FREE & \llmtag{LLM} & \colltag{Coll} \llmtag{LLM} & \humantag{Human} \ruletag{Rule} \\ \hline
Edit\allowbreak{}Reward-Bench (EditScore) & \cite{luo2026editrewardbench} & \textbf{R4}\, Reflective Reasoning; Critique and Judgment & Image-editing reward ranking (13 subtasks) & DF & \ruletag{Rule} & \colltag{Coll} \humantag{Human} \llmtag{LLM} & \humantag{Human} \\ \hline
Edit\allowbreak{}Reward-Bench (EditReward) & \cite{wu2026editrewardbench} & \textbf{R4}\, Reflective Reasoning; Critique and Judgment & Multi-way image-editing preference ranking & DF & \ruletag{Rule} & \colltag{Coll} \humantag{Human} \llmtag{LLM} & \humantag{Human} \ruletag{Rule} \\ \hline
Eval\allowbreak{}QA\allowbreak{}Bench & \cite{hengyuan2024lova3} & \textbf{R4}\, Reflective Reasoning & Visual question answering, asking and assessment & BI & \ruletag{Rule} & \colltag{Coll} \llmtag{LLM} & \humantag{Human} \ruletag{Rule} \\ \hline
MC-MKE & \cite{zhang2024mc-mke} & \textbf{R4}\, Reflective Reasoning & Image-to-entity sub-rel-obj img-rel-obj & DF & \ruletag{Rule} & \colltag{Coll} \ruletag{Rule} \llmtag{LLM} & \humantag{Human} \ruletag{Rule} \llmtag{LLM} \\ \hline
MIKE & \cite{li2024mike} & \textbf{R4}\, Reflective Reasoning & Vanilla name answering, \etc\ (3 tasks) & BI DF FREE & \ruletag{Rule} & \colltag{Coll} \humantag{Human} \ruletag{Rule} \llmtag{LLM} & \humantag{Human} \ruletag{Rule} \llmtag{LLM} \\ \hline
MLLM-as-a-Judge & \cite{chen2024mllmasajudge} & \textbf{R4}\, Reflective Reasoning; Critique and Judgment & Scoring, pair comparison, \etc\ (3 judging tasks) & MC DF & \ruletag{Rule} & \colltag{Coll} \humantag{Human} \llmtag{LLM} & \ruletag{Rule} \\ \hline
MMKE-Bench & \cite{du2025mmke-bench} & \textbf{R4}\, Reflective Reasoning & Multimodal knowledge editing & DF & \ruletag{Rule} & \colltag{Coll} \humantag{Human} \llmtag{LLM} & \humantag{Human} \\ \hline
Vi\allowbreak{}Crit-Bench & \cite{wang2025vicrit} & \textbf{R4}\, Reflective Reasoning; Self-Improvement & Verifiable visual perception error criticism for RL self-improvement & DF & \ruletag{Rule} & \colltag{Coll} \ruletag{Rule} \llmtag{LLM} & \humantag{Human} \ruletag{Rule} \\ \hline
VISCO & \cite{wu2025visco} & \textbf{R4}\, Reflective Reasoning; Self-Improvement & Step-wise visual critique and self-correction (VISCO-Score) & BI FREE & \ruletag{Rule} \llmtag{LLM} & \colltag{Coll} \humantag{Human} \llmtag{LLM} & \humantag{Human} \\ \hline
Visual\allowbreak{}Process\allowbreak{}Bench & \cite{wang2026visualprm} & \textbf{R4}\, Reflective Reasoning; Self-Improvement & Step-wise multimodal process reward verification for test-time scaling & BI & \ruletag{Rule} & \colltag{Coll} \humantag{Human} \llmtag{LLM} & \humantag{Human} \\ \hline
VLKEB & \cite{huang2024VLKEB} & \textbf{R4}\, Reflective Reasoning & Image-to-entity & DF & \ruletag{Rule} & \colltag{Coll} \humantag{Human} \ruletag{Rule} \llmtag{LLM} & \humantag{Human} \ruletag{Rule} \\ \hline
\end{tabularx}\endgroup\end{table*}

%% file: sec/reasoning/bg.tex
\subsection{Taxonomy by Reasoning Supervision and Grounding}
\label{sec:reasoning_bg}

Reasoning benchmarks evaluate multi-step deduction, quantitative and spatial inference, counterfactual reasoning, and knowledge editing across modalities. Grouping reasoning benchmarks by subject area alone (math, spatial, or causal) conflates tests that score only final answers with tests that verify visual grounding or intermediate derivations. We organize all 73 Reasoning benchmarks (Fig.~\ref{fig:map-reasoning} and Table~\ref{tbl:reasoning}) by \textbf{reasoning supervision and grounding}: outcome-based multimodal reasoning (\textbf{R1}, 50 benchmarks), visually grounded reasoning (\textbf{R2}, 9), chain-of-thought and process reasoning (\textbf{R3}, 3), and reflective critique, self-improvement, and editing (\textbf{R4}, 11).

\subsubsection{Outcome-Based Multimodal Reasoning}
\label{sec:r1_conclusion}
Accounting for 50 of 73 Reasoning benchmarks (68.5\%), \textbf{R1} benchmarks require multi-step inference yet verify only the final outcome (a choice letter, numeric value, or short answer) on their primary leaderboard track, spanning three problem sub-families (24 mathematical/scientific, 19 spatial/temporal/physical, and 7 logic/counterfactual suites):

\begin{itemize}
 \item \textbf{Mathematical, Scientific, and Analytical Reasoning (24 benchmarks).} Visual math benchmarks integrate geometric diagrams and functions into symbolic problem solving~\cite{wang2024measuring,zhou2024your,zhang2024mathverse}, decompose multi-step problems into atomic sub-questions to diagnose knowledge gaps (We-Math~\cite{qiao2025wemath}), instantiate programmatic variants to measure worst-case consistency (DynaMath~\cite{zou2025dynamath}), or probe intuitive visual number sense and estimation (VisNumBench~\cite{weng2025visnumbench}). Multidisciplinary suites span K--12, Olympiad, and Ph.D.-level science across images and videos~\cite{he2024cmmu,zhu2024multi,liang2024scemqa,he2024MMWorld,he2024olympiadbench,phan2025humanityslastexam,yang2025r1onevisionbench,hao2025emma,zhao2025mmvu}, with EMMA~\cite{hao2025emma} screening out text- or caption-solvable items. Domain analytical suites cover research figures (SPIQA~\cite{pramanick2024spiqa}), complex charts (CharXiv~\cite{wang2024charxiv}, ChartX~\cite{xia2024chartx}, ChartQAPro~\cite{masry2025chartqapro}), web pages (VisualWebBench~\cite{liu2024VisualWebBench}), whole-slide pathology (WSI-Bench~\cite{liang2025wsillava}), biological microscopy (MicroVQA~\cite{burgess2025microvqa}), diagram-essential K--12-to-Ph.D. physics reasoning (SeePhys~\cite{xiang2025seephys}), and electrical and electronics engineering schematics (EEE-Bench~\cite{li2025eee-bench}).

 \item \textbf{Relational, Spatial, Temporal, and Physical Reasoning (19 benchmarks).} Relational suites evaluate entity relations, abstract visual patterns, open-ended multi-image analogies, prior-free visual rules, and paired comparisons (CRPE~\cite{wang2024CRPE}, VCog-Bench~\cite{cao2024VCog-Bench}, MARVEL~\cite{jiang2024MARVEL}, CompBench~\cite{kil2024CompBench}, VOILA~\cite{yilmaz2025voila}, VisRes~Bench~\cite{tortei2026visresbench}). Spatial benchmarks probe single-view 3D geometry (3DSRBench~\cite{ma20253dsrbench}, SpatialRGPT-Bench~\cite{cheng2024spatialrgpt}), robotic spatial relations (RoboSpatial~\cite{song2025robospatial}), multi-view correspondence (SPAR-Bench~\cite{zhang2025sparbench}, MMSI-Bench~\cite{yang2026mmsibench}), indoor egocentric video geometry (VSI-Bench~\cite{yang2025vsibench}), compositional 3D indoor spatial intelligence across point clouds and multi-choice QA (SpaCE-10~\cite{gong2026space10}), and cognitive mapping (SPACE~\cite{ramakrishnan2025spatialcognition}). Temporal and commonsense video reasoning are tested by ReXTime~\cite{chen2024rextime} and SOK-Bench~\cite{wang2024SOK-Bench}, while physical reasoning spans physical property dynamics (PhysBench~\cite{chow2025physbench}), quantitative video kinematics (QuantiPhy~\cite{puyin2026quantiphy}), and physical-AI prediction (PAI-Bench~\cite{zhou2026paibench}).

 \item \textbf{Multi-Step Logic, CoT-Prompted, and Counterfactual Conclusions (7 benchmarks).} Several suites elicit multi-step or counterfactual reasoning while grading final answer accuracy on their headline track (distinguishing them from step-scored \textbf{R3}): rule-induction puzzles in VisuLogic~\cite{xu2026visulogic}, multimodal CoT QA in M$^3$CoT~\cite{chen2024m3cot} and Visual CoT~\cite{shao2024VisualCoT} (which supplies auxiliary bounding boxes while scoring QA accuracy), egocentric action reasoning in SEED-Bench-R1~\cite{chen2026seedbenchr1}, key-frame video QA in VideoEspresso~\cite{han2025videoespresso}, visual puzzle solving via video generation models in VideoThinkBench~\cite{tong2026videothinkbench}, and counterfactual assumption VQA in CFMM~\cite{li2024eyes}.
\end{itemize}

\subsubsection{Visually Grounded Reasoning}
\label{sec:r2_grounded}
Nine Reasoning benchmarks (\textbf{R2}) require explicit visual grounding (a bounding box, 2D/3D point, timestamped clue, or segmentation mask) to anchor deduction in the scene; all nine score outputs with deterministic geometric rules. In 2D and remote-sensing imagery, TreeBench~\cite{wang2026treebench} pairs dense small-object localization with multi-step QA via mean Intersection over Union (IoU), MultihopSpatial~\cite{lee2026multihopspatial} jointly scores 1-to-3-hop spatial QA and bounding-box localization via Acc@50IoU, and XLRS-Bench~\cite{wang2025xlrsbench} evaluates grounded reasoning on ultra-high-resolution satellite scenes. In 3D and embodied scenes, ScanReason~\cite{zhu2024scanreason} jointly scores 3D reasoning and 3D bounding-box grounding, while RefSpatial-Bench~\cite{zhou2025refspatialbench} verifies whether predicted target points land inside ground-truth masks. In video and pixel-level reasoning, ReVOS~\cite{yan2024revos} and ReasonVOS~\cite{bai2024tokenseg} score implicit-instruction segmentation masks by region--contour similarity ($\mathcal{J}\&\mathcal{F}$), Video-Holmes~\cite{cheng2026videoholmes} links suspense-film clue localization with multi-clue deduction, and CaST-Bench~\cite{zhang2026castbench} grounds causal video QA in annotated temporal segments and bounding-box tracks.

\subsubsection{Chain-of-Thought and Process Reasoning}
\label{sec:r3_trace}
Three dedicated benchmarks (\textbf{R3}) supervise intermediate chain-of-thought (CoT) derivations to separate step-by-step reasoning from lucky guesses; all three use LLM-assisted process grading. VRC-Bench~\cite{thawakar2025vrcbench} and MME-CoT~\cite{jiang2025mmecot} annotate reference reasoning steps and grade generated chains for logical validity, visual relevance, and efficiency, with MME-CoT quantifying accuracy shifts between direct and CoT responses. MathCanvas-Bench~\cite{shi2026mathcanvasbench} extends process evaluation beyond text by scoring interleaved diagram-and-text mathematical derivations.

\subsubsection{Reflective Critique, Self-Improvement, and Editing}
\label{sec:r4_judgment}
Eleven Reasoning benchmarks (\textbf{R4}) evaluate meta-cognitive critique, iterative self-correction toward recursive self-improvement (RSI)~\cite{deng2025selfimprovementsurvey}, process-reward verification, and post-hoc multimodal knowledge editing, with 10 of the 11 pairing LLM-assisted construction with deterministic rule-based verification. To keep the boundary with evaluator auditing in \textbf{T4} (\cf~\s\ref{sec:t4_diagnostics}) explicit, \textbf{R4} groups capability-oriented critique, step-verification, and task/editing reward benchmarks aimed at improving reasoning or instruction-following fidelity, whereas \textbf{T4} isolates safety guardrails, factual hallucination detectors, alignment preference rewards, and judge-bias audits:

\begin{itemize}
 \item \textbf{Self-Correction, Process Verification, and Critique Judges.} Driven by test-time scaling and multimodal self-improvement~\cite{deng2025selfimprovementsurvey}, VISCO~\cite{wu2025visco} benchmarks fine-grained step-wise critique and iterative self-correction in visual reasoning via hierarchical F1 (VISCO-Score), ViCrit-Bench~\cite{wang2025vicrit} evaluates verifiable visual-perception error criticism designed as a reinforcement-learning self-improvement proxy, and VisualProcessBench~\cite{wang2026visualprm} measures step-level error localization for multimodal process reward models (PRMs) guiding best-of-$N$ self-verification. Complementing self-correction, MLLM-as-a-Judge~\cite{chen2024mllmasajudge} benchmarks general scoring, pairwise, and batch-ranking utility across vision--language tasks, while the concurrent ICLR~2026 papers introducing EditScore~\cite{luo2026editrewardbench} and EditReward~\cite{wu2026editrewardbench} each present a benchmark named EditReward-Bench to evaluate reward models that rank instruction-guided image edits for online reinforcement learning and human preference alignment.

 \item \textbf{Analytical Questioning and Multimodal Knowledge Editing.} LOVA$^3$ (EvalQABench)~\cite{hengyuan2024lova3} tests whether models can pose and verify analytical visual questions. Multimodal knowledge-editing suites, including MC-MKE~\cite{zhang2024mc-mke}, MIKE~\cite{li2024mike}, VLKEB~\cite{huang2024VLKEB} (one-hop and multi-hop edits), and MMKE-Bench~\cite{du2025mmke-bench} (spanning entity, semantic, and user-specific visual edits), test whether MLLMs update obsolete facts without degrading unrelated knowledge.
\end{itemize}

%% file: sec/reasoning/task.tex
\subsection{Evaluation Metrics and Analytical Synthesis}
\label{sec:reasoning_task}

Across all 73 Reasoning benchmarks (Table~\ref{tbl:reasoning}), evaluation metrics align with the four reasoning-supervision categories:

\begin{itemize}
 \item \textbf{Outcome-Based Accuracy and Quantitative Metrics (R1).} Discrete outcomes use exact-match or multiple-choice accuracy hardened by CircularEval~\cite{wang2024CRPE}, hierarchical sub-problem diagnostics in We-Math~\cite{qiao2025wemath}, and worst-case program-variant accuracy in DynaMath~\cite{zou2025dynamath}. Continuous spatial and kinematic predictions (distances, object dimensions, velocities) use Mean Relative Accuracy (MRA)~\cite{yang2025vsibench,zhang2025sparbench,puyin2026quantiphy}, whereas open-ended Olympiad and Ph.D. answers increasingly rely on LLM answer extractors and graders~\cite{yang2025r1onevisionbench,phan2025humanityslastexam,zhao2025mmvu,zhou2026paibench}.

 \item \textbf{Visual Grounding Metrics (R2).} Visual and 3D evidence predictions are verified deterministically via 2D/3D bounding-box IoU~\cite{zhu2024scanreason,wang2026treebench,wang2025xlrsbench}, point-in-mask hit rate~\cite{zhou2025refspatialbench}, and region--contour similarity ($\mathcal{J}\&\mathcal{F}$)~\cite{yan2024revos,bai2024tokenseg}.

 \item \textbf{Process and CoT Metrics (R3).} Process evaluators score step recall, precision, logical coherence, CoT accuracy delta, and token efficiency across textual and interleaved visual chains~\cite{thawakar2025vrcbench,jiang2025mmecot,shi2026mathcanvasbench}.

 \item \textbf{Reflective Self-Improvement, Critique, and Editing Metrics (R4).} Self-correction and process-reward benchmarks evaluate step-error localization F1, critique quality, and pre-to-post revision gain (VISCO-Score)~\cite{wu2025visco,wang2025vicrit,wang2026visualprm}; knowledge-editing benchmarks measure reliability, generality, locality, portability, and multi-hop consistency~\cite{zhang2024mc-mke,li2024mike,huang2024VLKEB,du2025mmke-bench}; and judge/reward benchmarks quantify human agreement via Pearson/Spearman correlation and pairwise preference accuracy~\cite{chen2024mllmasajudge,luo2026editrewardbench,wu2026editrewardbench}.
\end{itemize}

\textbf{Analytical synthesis.} The steep skew between outcome-only suites (\textbf{R1}, 50/73) and process-supervised suites (\textbf{R3}, 3/73) reflects an annotation bottleneck rather than taxonomic granularity: collecting verified intermediate reasoning chains or spatial boxes (\textbf{R2}, 9/73) is far costlier than scraping exam answer keys. Across horizontal comparisons, frontier models that perform strongly on static textbook QA drop sharply when required to maintain worst-case consistency across programmatic diagram seeds (DynaMath~\cite{zou2025dynamath}), solve prior-free visual rule induction (VisuLogic~\cite{xu2026visulogic}), or jointly localize supporting visual regions (\textbf{R2}~\cite{wang2026treebench,lee2026multihopspatial}). Grouping by supervision level also explains why Reasoning is the sole pillar with a negative descriptive gap ($-7.8$ pp) between LLM-assisted construction and LLM judging (\s\ref{sec:crossstats}): all nine \textbf{R2} suites use deterministic geometric rules, and 10 of 11 \textbf{R4} suites combine LLM construction with deterministic verification.

%% file: sec/generation.tex
\section{Generation Benchmarks}
\label{sec:generation}
\begin{figure*}[tp]
\centering\chaptergraphic[width=\textwidth]{generation_map}
\caption{Taxonomy and representative examples of Generation benchmarks ($N=31$) organized by target synthesis modality (\textbf{G1--G3}). Illustrated tasks cover constrained instruction-following text generation in MM-IFEval~\cite{ding2025mmifeval}, chart-to-code synthesis in ChartMimic~\cite{yang2025chartmimic}, reasoning-guided visual editing in RISEBench~\cite{zhao2025risebench}, and academic poster generation in Paper2Poster~\cite{pang2025paper2poster}.}\label{fig:map-generation}
\end{figure*}

\input{sec/misc/tbl_generation}

\input{sec/generation/bg}

\input{sec/generation/task}

%% file: sec/misc/tbl_generation.tex
\input{sec/misc/tbl_generation_clean}

%% file: sec/misc/tbl_generation_clean.tex
\begin{table*}[tp]\centering
\caption{Overview of Generation benchmarks ($N=31$), spanning \textbf{G1} instruction-following and creative text generation (5), \textbf{G2} visual-to-code and programmatic generation (8), and \textbf{G3} visual, interleaved, and spoken media generation (18). Output modalities: image (\textbf{I}), text (\textbf{T}), and audio (\textbf{A}).}
\label{tbl:generation}
\begingroup\linespread{0.99}\fontsize{7.2}{8.1}\selectfont\setlength\tabcolsep{1.25pt}\renewcommand\arraystretch{1.00}
\prepareSurveyTableTags\begin{tabularx}{\textwidth}{@{}>{\raggedright\arraybackslash}p{.133\textwidth}l>{\raggedright\arraybackslash}p{.166\textwidth}>{\raggedright\arraybackslash}X>{\raggedright\arraybackslash}p{.070\textwidth}>{\raggedright\arraybackslash\spaceskip=1pt}p{.089\textwidth}>{\raggedright\arraybackslash\spaceskip=1pt}p{.160\textwidth}>{\raggedright\arraybackslash\spaceskip=1pt}p{.090\textwidth}@{}}
\hline\thickhline\rowcolor{mygray}\textbf{Benchmark} & \textbf{Ref.} & \textbf{Category \& Focus} & \textbf{Task} & \textbf{Output} & \textbf{Evaluation} & \textbf{Construction} & \textbf{Filter} \\ \hline
Asso\allowbreak{}Ci\allowbreak{}Am & \cite{liu2025assocciam} & \textbf{G1}\, Creative & Visual Association QA & T & \ruletag{Rule} & \colltag{Coll} \ruletag{Rule} \llmtag{LLM} & \humantag{Human} \ruletag{Rule} \\ \hline
Co\allowbreak{}IN & \cite{chen2024coin} & \textbf{G1}\, Instruction Following & Knowledge-grounded image QA, \etc\ (8 tasks) & T & \ruletag{Rule} \llmtag{LLM} & \colltag{Coll} \ruletag{Rule} & -- \\ \hline
GOD\allowbreak{}Bench & \cite{lei2025godbench} & \textbf{G1}\, Creative & Video comment discrimination and generation & T & \ruletag{Rule} & \colltag{Coll} & \ruletag{Rule} \humantag{Human} \\ \hline
MIA-Bench & \cite{qian2024mia} & \textbf{G1}\, Instruction Following & Instruction Following & T & \llmtag{LLM} & \colltag{Coll} \humantag{Human} & -- \\ \hline
MM-IF\allowbreak{}Eval & \cite{ding2025mmifeval} & \textbf{G1}\, Instruction Following & Multimodal instruction following & T & \ruletag{Rule} \llmtag{LLM} & \colltag{Coll} \humantag{Human} \llmtag{LLM} & \humantag{Human} \llmtag{LLM} \\ \hline
Blender\allowbreak{}Bench & \cite{yin2026blenderbench} & \textbf{G2}\, Code Generation & Camera and scene editing via Blender code (3 tracks) & T & \ruletag{Rule} \llmtag{LLM} & -- & -- \\ \hline
Chart\allowbreak{}Edit & \cite{zhao2025chartedit} & \textbf{G2}\, Code Generation & Chart editing via code generation & T & \ruletag{Rule} \llmtag{LLM} \humantag{Human} & \colltag{Coll} \humantag{Human} \llmtag{LLM} & \ruletag{Rule} \humantag{Human} \\ \hline
Chart\allowbreak{}Mimic & \cite{yang2025chartmimic} & \textbf{G2}\, Code Generation & Direct Mimic, \etc\ (2 chart-to-code tasks) & T & \ruletag{Rule} \llmtag{LLM} & \colltag{Coll} \humantag{Human} & \humantag{Human} \ruletag{Rule} \\ \hline
Design2Code & \cite{si2025design2code} & \textbf{G2}\, Code Generation & Webpage screenshot-to-code & T & \humantag{Human} \ruletag{Rule} & \colltag{Coll} & \humantag{Human} \ruletag{Rule} \\ \hline
MMCode & \cite{li2024mmcode} & \textbf{G2}\, Code Generation & Algorithmic code synthesis from visually rich programming problems & T & \ruletag{Rule} & \colltag{Coll} \humantag{Human} & \humantag{Human} \ruletag{Rule} \\ \hline
Omni\allowbreak{}GIRL & \cite{guo2025omnigirl} & \textbf{G2}\, Code Generation & GitHub issue resolution across multiple programming languages & T & \ruletag{Rule} & \colltag{Coll} & \humantag{Human} \ruletag{Rule} \\ \hline
Plot2Code & \cite{wu2024plot2code} & \textbf{G2}\, Code Generation & Plot-to-code generation (6 plot types) & T & \ruletag{Rule} \llmtag{LLM} & \colltag{Coll} \llmtag{LLM} & \humantag{Human} \ruletag{Rule} \\ \hline
Web2Code & \cite{yun2024web2code} & \textbf{G2}\, Code Generation & HTML code generation and webpage QA & T & \ruletag{Rule} \llmtag{LLM} & \colltag{Coll} \ruletag{Rule} \llmtag{LLM} & \humantag{Human} \ruletag{Rule} \llmtag{LLM} \\ \hline
Chrono\allowbreak{}Magic-Bench & \cite{yuan2024chronomagicbench} & \textbf{G3}\, Unified Multimodal Generation & Metamorphic time-lapse text-to-video synthesis (75 subcategories) & I & \ruletag{Rule} \humantag{Human} & \colltag{Coll} \humantag{Human} \llmtag{LLM} & \humantag{Human} \ruletag{Rule} \\ \hline
Co\allowbreak{}MM & \cite{chen2024comm} & \textbf{G3}\, Interleaved Image-Text Generation & Image-to-text sequence generation; text-to-image sequence generation; interleaved image-text content continuation; question-based interleaved image-text generation & I; T; I+T & \ruletag{Rule} \llmtag{LLM} & \colltag{Coll} & \ruletag{Rule} \llmtag{LLM} \\ \hline
Full-Duplex-Bench v1.5 & \cite{lin2026fullduplexbenchv15} & \textbf{G3}\, Speech Generation & Interruption handling, \etc\ (4 scenarios) & A & \ruletag{Rule} \llmtag{LLM} & \ruletag{Rule} & -- \\ \hline
Full-Duplex-Bench-v2 & \cite{lin2026fullduplexbenchv2} & \textbf{G3}\, Speech Generation & Full-duplex multi-turn dialogue (4 task families) & A & \llmtag{LLM} & \llmtag{LLM} & -- \\ \hline
Img\allowbreak{}Edit & \cite{ye2025imgedit} & \textbf{G3}\, Unified Multimodal Generation & Single-turn and multi-turn instruction-guided image editing & I & \llmtag{LLM} & \colltag{Coll} \ruletag{Rule} \llmtag{LLM} & \humantag{Human} \ruletag{Rule} \llmtag{LLM} \\ \hline
Interleaved\allowbreak{}Bench & \cite{liu2024interleavedbench} & \textbf{G3}\, Interleaved Image-Text Generation & Open-ended interleaved text-and-image generation across real-world tasks & I+T & \llmtag{LLM} \humantag{Human} & \colltag{Coll} \humantag{Human} & \humantag{Human} \\ \hline
Know\allowbreak{}Gen & \cite{feng2026knowgen} & \textbf{G3}\, Unified Multimodal Generation & Search-grounded text-to-image generation & I & \llmtag{LLM} & \colltag{Coll} \llmtag{LLM} & \humantag{Human} \ruletag{Rule} \llmtag{LLM} \\ \hline
MMGR & \cite{cai2026mmgr} & \textbf{G3}\, Unified Multimodal Generation & Physical, logical and spatial generative reasoning across image and video & I & \ruletag{Rule} \llmtag{LLM} & \colltag{Coll} \humantag{Human} \ruletag{Rule} & \humantag{Human} \ruletag{Rule} \\ \hline
MRAMG-Bench & \cite{yu2025mramgbench} & \textbf{G3}\, Interleaved Image-Text Generation & Retrieval-augmented multimodal answer generation & I+T & \ruletag{Rule} \llmtag{LLM} & \colltag{Coll} \humantag{Human} \llmtag{LLM} & \humantag{Human} \ruletag{Rule} \llmtag{LLM} \\ \hline
Paper2Poster & \cite{pang2025paper2poster} & \textbf{G3}\, Interleaved Image-Text Generation & Academic poster generation & I+T & \ruletag{Rule} \llmtag{LLM} & \colltag{Coll} \llmtag{LLM} & \ruletag{Rule} \\ \hline
RISE\allowbreak{}Bench & \cite{zhao2025risebench} & \textbf{G3}\, Unified Multimodal Generation & Reasoning-informed visual editing (4 categories) & I & \humantag{Human} \llmtag{LLM} & \humantag{Human} & -- \\ \hline
T2V-Comp\allowbreak{}Bench & \cite{sun2025t2vcompbench} & \textbf{G3}\, Unified Multimodal Generation & Compositional text-to-video generation across 7 binding categories & I & \ruletag{Rule} \llmtag{LLM} & \humantag{Human} \llmtag{LLM} & \humantag{Human} \\ \hline
Talking Turns & \cite{arora2025talkingturns} & \textbf{G3}\, Speech Generation & Turn-taking, \etc\ (5 abilities) & T; A & \ruletag{Rule} & \colltag{Coll} \humantag{Human} \ruletag{Rule} & \humantag{Human} \ruletag{Rule} \\ \hline
UEval & \cite{li2026ueval} & \textbf{G3}\, Interleaved Image-Text Generation & Unified image-and-text generation scored on 10,417 expert rubrics & I+T & \llmtag{LLM} & \humantag{Human} \llmtag{LLM} & \humantag{Human} \\ \hline
Uni-MMMU & \cite{zou2026unimmmu} & \textbf{G3}\, Unified Multimodal Generation & Bidirectionally coupled visual generation and multi-discipline reasoning & I+T & \ruletag{Rule} \llmtag{LLM} & \colltag{Coll} \humantag{Human} \ruletag{Rule} & \humantag{Human} \ruletag{Rule} \\ \hline
UReason & \cite{yang2026ureason} & \textbf{G3}\, Unified Multimodal Generation & Reasoning-guided text-to-image alignment across 5 reasoning tasks & I+T & \ruletag{Rule} \llmtag{LLM} & \humantag{Human} & \humantag{Human} \\ \hline
WISE & \cite{niu2026wise} & \textbf{G3}\, Unified Multimodal Generation & World-knowledge text-to-image generation & I & \llmtag{LLM} & \colltag{Coll} \humantag{Human} \llmtag{LLM} & \humantag{Human} \\ \hline
World\allowbreak{}Score & \cite{duan2025worldscore} & \textbf{G3}\, Unified Multimodal Generation & Camera-controlled sequential 3D, 4D and video world generation & I & \ruletag{Rule} & \colltag{Coll} \ruletag{Rule} \llmtag{LLM} & \humantag{Human} \ruletag{Rule} \\ \hline
\end{tabularx}\endgroup\end{table*}

%% file: sec/generation/bg.tex
\subsection{Taxonomy by Target Generation Modality}
\label{sec:generation_bg}

Unlike standalone text-to-image diffusion benchmarks~\cite{qu2024unified}, MLLM generation benchmarks evaluate unified models that synthesize constrained text, executable visual-rendering code, edited imagery, interleaved documents, and spoken dialogue (whereas multi-step repository or workspace agents sit in \textbf{A5}, \s\ref{sec:application_bg}, and safety/hallucination audits sit in \s\ref{sec:trustworthiness}). Here we organize all 31 pure multimodal synthesis benchmarks (Fig.~\ref{fig:map-generation} and Table~\ref{tbl:generation}) along three target generation modalities: instruction-following and creative text generation (\textbf{G1}, 5 benchmarks), visual-to-code and programmatic generation (\textbf{G2}, 8), and visual, interleaved, and spoken media generation (\textbf{G3}, 18).

\subsubsection{Instruction-Following and Creative Text Generation}
\label{sec:g1_verbal}
Five Generation benchmarks (\textbf{G1}) evaluate MLLMs producing open-ended text governed by explicit instruction constraints or divergent creative criteria:

\begin{itemize}
 \item \textbf{Multimodal Instruction Following.} Extending text-only instruction suites such as IFEval~\cite{zhou2023instruction} and InfoBench (DRFR)~\cite{qin2024infobench}, MM-IFEval~\cite{ding2025mmifeval} pairs verifiable output-format constraints with visual-perception requirements, MIA-Bench~\cite{qian2024mia} tests layered compositional directives, and CoIN~\cite{chen2024coin} measures instruction retention across continual tuning stages.

 \item \textbf{Creative and Associative Generation.} Moving beyond literal description, GODBench~\cite{lei2025godbench} evaluates humorous and rhetorical video commentary across five dimensions, and AssoCiAm~\cite{liu2025assocciam} benchmarks divergent visual associative thinking through a hybrid computational rubric.
\end{itemize}

\subsubsection{Visual-to-Code and Programmatic Generation}
\label{sec:g2_executable}
Eight benchmarks (\textbf{G2}) translate visual specifications into executable programs, extending text-based code synthesis~\cite{lu1codexglue,balog2017deepcoder,chen2021evaluating} to visual-to-code engineering. Web2Code~\cite{yun2024web2code} and Design2Code~\cite{si2025design2code} reconstruct HTML/CSS webpages from UI screenshots; Plot2Code~\cite{wu2024plot2code}, ChartEdit~\cite{zhao2025chartedit}, and ChartMimic~\cite{yang2025chartmimic} generate or modify plotting scripts to match target charts; BlenderBench~\cite{yin2026blenderbench} writes Python scripts in Blender to reproduce 3D lighting, camera, and scene edits; MMCode~\cite{li2024mmcode} tests algorithmic problem solving with visually rich programming specifications, and OmniGIRL~\cite{guo2025omnigirl} evaluates multilingual GitHub issue repair grounded in UI bug screenshots.

\subsubsection{Visual, Interleaved, and Spoken Media Generation}
\label{sec:g3_media}
Eighteen benchmarks (\textbf{G3}) require synthesizing non-text visual, interleaved, or spoken audio streams:

\begin{itemize}
 \item \textbf{Interleaved Image--Text and Document Synthesis.} CoMM~\cite{chen2024comm} evaluates narrative and stylistic coherence in interleaved image--text continuation and question-driven sequence generation; MRAMG-Bench~\cite{yu2025mramgbench} interleaves retrieved document images into long-form answers; InterleavedBench~\cite{liu2024interleavedbench} and UEval~\cite{li2026ueval} evaluate open-ended interleaved text-and-image synthesis via reference-free GPT-4o and 10,417 expert-validated rubric criteria; and Paper2Poster~\cite{pang2025paper2poster} condenses full scientific papers into single-page academic posters.

 \item \textbf{Knowledge- and Reasoning-Driven Visual Synthesis.} Testing unified understanding--generation models beyond surface aesthetics, WISE~\cite{niu2026wise} probes world-knowledge text-to-image synthesis, RISEBench~\cite{zhao2025risebench} evaluates causal, spatial, and temporal reasoning in visual editing, ImgEdit~\cite{ye2025imgedit} extends evaluation to single- and multi-turn instruction-guided editing, KnowGen~\cite{feng2026knowgen} benchmarks search-augmented knowledge image generation, and Uni-MMMU~\cite{zou2026unimmmu}, UReason~\cite{yang2026ureason}, and MMGR~\cite{cai2026mmgr} probe bidirectional reasoning--generation synergy, reasoning-to-image alignment, and physical/logical generative reasoning across images and videos. In dynamic video and 3D/4D world synthesis, ChronoMagic-Bench~\cite{yuan2024chronomagicbench} evaluates metamorphic time-lapse generation, T2V-CompBench~\cite{sun2025t2vcompbench} tests seven-category compositional text-to-video binding, and WorldScore~\cite{duan2025worldscore} unifies camera-controlled multi-scene world generation.

 \item \textbf{Spoken and Full-Duplex Dialogue Synthesis.} Spoken-interaction suites jointly evaluate semantic content, prosody, and conversational timing: Talking Turns~\cite{arora2025talkingturns} assesses turn-taking and backchanneling, Full-Duplex-Bench v1.5~\cite{lin2026fullduplexbenchv15} measures interruption handling and response latency under overlapping speech, and Full-Duplex-Bench-v2~\cite{lin2026fullduplexbenchv2} conducts live examiner-paced multi-turn spoken tasks.
\end{itemize}

%% file: sec/generation/task.tex
\subsection{Modality-Specific Metrics and Analytical Synthesis}
\label{sec:generation_task}

Because generative deliverables admit multiple valid realizations across text, code, pixels, layouts, and waveforms, Table~\ref{tbl:generation} combines deterministic execution or signal checks with MLLM-based semantic judges across four output regimes:

\begin{itemize}
 \item \textbf{Constrained Text (G1).} Reference n-gram metrics (ROUGE, METEOR) are largely superseded by hybrid rule-and-LLM verification: MM-IFEval~\cite{ding2025mmifeval} pairs deterministic format verifiers with contrastive LLM grading, while MIA-Bench~\cite{qian2024mia}, GODBench~\cite{lei2025godbench}, and AssoCiAm~\cite{liu2025assocciam} apply multi-dimensional computational and LLM rubrics.

 \item \textbf{Render-and-Compare Code Verification (G2).} Visual-to-code benchmarks execute generated scripts and compare rendered outputs against reference targets via execution pass rates, DOM/text/color matching, CLIP similarity, and GPT-4V/4o visual fidelity~\cite{yun2024web2code,wu2024plot2code,si2025design2code,yang2025chartmimic,zhao2025chartedit,yin2026blenderbench}, complemented by unit-test resolution in OmniGIRL~\cite{guo2025omnigirl}.

 \item \textbf{Visual and Interleaved Media (G3).} Signal-level metrics such as FID~\cite{heusel2017fid}, IS~\cite{salimans2016improved}, SSIM, PSNR, and CLIP scores~\cite{chen2024comm} quantify perceptual fidelity but miss factual or causal errors. Unified generation benchmarks therefore weight MLLM-judge scores toward world-knowledge accuracy in WISE and KnowGen~\cite{niu2026wise,feng2026knowgen} or enforce strict all-dimension pass criteria in RISEBench~\cite{zhao2025risebench}. Interleaved suites combine image-ordering precision/recall with GPT-4o relevance ratings in MRAMG-Bench~\cite{yu2025mramgbench} and VLM reader-quiz accuracy on rendered posters in Paper2Poster~\cite{pang2025paper2poster}.

 \item \textbf{Spoken Timing and Prosody (G3).} Audio dialogue benchmarks pair transcript grading with temporal interaction metrics, including human-calibrated turn-taking classifiers~\cite{arora2025talkingturns}, interruption latency and prosodic adaptation~\cite{lin2026fullduplexbenchv15}, and multi-turn task tracking under live examiner pacing~\cite{lin2026fullduplexbenchv2}.
\end{itemize}

\textbf{Analytical synthesis.} Isolating the 31 pure synthesis benchmarks from trustworthiness uncovers three cross-benchmark patterns (\s\ref{sec:protocols}--\s\ref{sec:crossstats}). First, pure multimodal generation expanded more than threefold from 5 peer-reviewed benchmarks in 2024 to 17 in 2025 (plus 9 in partial 2026). Second, horizontal comparisons expose a wide gap between perceptual surface quality and compositional or executable precision: frontier generators achieve high visual aesthetics yet struggle on strict all-constraint visual editing (RISEBench~\cite{zhao2025risebench}), exact chart/3D script reproduction (ChartMimic~\cite{yang2025chartmimic}, BlenderBench~\cite{yin2026blenderbench}), and overlapping spoken turn-taking~\cite{lin2026fullduplexbenchv2}. Third, Generation has the highest LLM-judge reliance of any pillar (23/31, 74.2\%; 80.0\% among LLM-constructed suites), making cross-family judge auditing (Blueprint~3) essential.

%% file: sec/trustworthiness.tex
\section{Trustworthiness and Safety Benchmarks}
\label{sec:trustworthiness}
\begin{figure*}[tp]
\centering\chaptergraphic[width=\textwidth]{trustworthiness_map}
\nocite{kaul2024throne,wang2025mmsafeaware}
\caption{Taxonomy and representative examples of Trustworthiness and Safety benchmarks ($N=48$) organized by risk dimension and evaluator role (\textbf{T1--T4}). Illustrated tasks cover object hallucination probing in THRONE~\cite{kaul2024throne}, unanswerable-query abstention in MoHoBench, adversarial misleading-prompt robustness in MMR~\cite{liu2024MMR}, and pairwise reward evaluation in VL-RewardBench~\cite{li2025vlrewardbench}.}\label{fig:map-trustworthiness}
\end{figure*}

\input{sec/misc/tbl_trustworthiness}

\input{sec/trustworthiness/bg}

\input{sec/trustworthiness/task}

%% file: sec/misc/tbl_trustworthiness.tex
\input{sec/misc/tbl_trustworthiness_clean}

%% file: sec/misc/tbl_trustworthiness_clean.tex
\begin{table*}[tp]\centering
\caption{Overview of Trustworthiness and Safety benchmarks ($N=48$), spanning \textbf{T1} hallucination and factual grounding (18), \textbf{T2} safety alignment, calibration, and unlearning (14), \textbf{T3} adversarial robustness, spurious correlations, and demographic bias (6), and \textbf{T4} evaluator, guardrail, and authenticity auditing (10). Output modality: text (\textbf{T}).}
\label{tbl:trustworthiness}
\begingroup\linespread{1.03}\fontsize{7.2}{8.1}\selectfont\setlength\tabcolsep{1.25pt}\renewcommand\arraystretch{1.00}
\prepareSurveyTableTags\begin{tabularx}{\textwidth}{@{}>{\raggedright\arraybackslash}p{.133\textwidth}l>{\raggedright\arraybackslash}p{.166\textwidth}>{\raggedright\arraybackslash}X>{\raggedright\arraybackslash}p{.070\textwidth}>{\raggedright\arraybackslash\spaceskip=1pt}p{.089\textwidth}>{\raggedright\arraybackslash\spaceskip=1pt}p{.160\textwidth}>{\raggedright\arraybackslash\spaceskip=1pt}p{.090\textwidth}@{}}
\hline\thickhline\rowcolor{mygray}\textbf{Benchmark} & \textbf{Ref.} & \textbf{Category \& Focus} & \textbf{Task} & \textbf{Output} & \textbf{Evaluation} & \textbf{Construction} & \textbf{Filter} \\ \hline
Auto\allowbreak{}Hallusion & \cite{wu2024autohallusion} & \textbf{T1}\, Hallucination Mitigation & Scene generation; image manipulation; question construction; hallucination detection & T & \llmtag{LLM} & \colltag{Coll} \ruletag{Rule} \llmtag{LLM} & \humantag{Human} \\ \hline
AV-Confuse\allowbreak{}Bench & \cite{ye2025eyesearsdisagree} & \textbf{T1}\, Hallucination Mitigation & Audio-muted QA; Audio-modified Description & T & \ruletag{Rule} & \colltag{Coll} \ruletag{Rule} & -- \\ \hline
AVH\allowbreak{}Bench & \cite{kim2025avhbench} & \textbf{T1}\, Hallucination Mitigation & Audio-driven Video Hallucination; Video-driven Audio Hallucination; AV Matching & T & \ruletag{Rule} & \colltag{Coll} \ruletag{Rule} \llmtag{LLM} & \humantag{Human} \\ \hline
BEAF & \cite{ye2024beaf} & \textbf{T1}\, Hallucination Mitigation & Image manipulation & T & \ruletag{Rule} & \colltag{Coll} \humantag{Human} \ruletag{Rule} & \humantag{Human} \\ \hline
Cross\allowbreak{}Check-Bench & \cite{tian2025crosscheckbench} & \textbf{T1}\, Hallucination Mitigation & Contradiction detection and conflict resolution across image-text pairs & T & \ruletag{Rule} & \colltag{Coll} \humantag{Human} \ruletag{Rule} \llmtag{LLM} & \humantag{Human} \ruletag{Rule} \\ \hline
Hallu-PI & \cite{ding2024hallu} & \textbf{T1}\, Hallucination Mitigation & Generative and discriminative hallucination & T & \ruletag{Rule} & \colltag{Coll} \humantag{Human} \ruletag{Rule} & \humantag{Human} \\ \hline
Halo\allowbreak{}Quest & \cite{wang2024haloquest} & \textbf{T1}\, Hallucination Mitigation & Visual hallucination QA & T & \humantag{Human} \llmtag{LLM} & \colltag{Coll} \humantag{Human} \llmtag{LLM} & \humantag{Human} \llmtag{LLM} \\ \hline
MCR-BENCH & \cite{wang2025audiotextdisagree} & \textbf{T1}\, Hallucination Mitigation & Audio QA; Speech Emotion Recognition; Vocal Sound Classification & T & \ruletag{Rule} & \colltag{Coll} \llmtag{LLM} & \ruletag{Rule} \\ \hline
MME\allowbreak{}Ception & \cite{cao2023genception} & \textbf{T1}\, Hallucination Mitigation & MLLMs' tendency to hallucinate & T & \ruletag{Rule} & \colltag{Coll} & -- \\ \hline
MR\allowbreak{}Hal-Bench & \cite{zhang2024automated} & \textbf{T1}\, Hallucination Mitigation & Multi-size expert generation; incremental generation & T & \llmtag{LLM} & \llmtag{LLM} & \humantag{Human} \\ \hline
Reefknot & \cite{zheng2024reefknot} & \textbf{T1}\, Hallucination Mitigation & Relation hallucination & T & \ruletag{Rule} & \colltag{Coll} \ruletag{Rule} \llmtag{LLM} & \humantag{Human} \ruletag{Rule} \\ \hline
RH-Bench & \cite{liu2025rhbench} & \textbf{T1}\, Hallucination Mitigation & Reasoning-hallucination balance QA & T & \ruletag{Rule} \llmtag{LLM} & \colltag{Coll} & \humantag{Human} \\ \hline
ROPE & \cite{chen2024multi} & \textbf{T1}\, Hallucination Mitigation & Recognition-based object probing & T & \ruletag{Rule} & \colltag{Coll} \ruletag{Rule} & \ruletag{Rule} \\ \hline
Simple\allowbreak{}VQA & \cite{cheng2025simplevqa} & \textbf{T1}\, Hallucination Mitigation & Factuality short-answer VQA, \etc\ (9 tasks) & T & \llmtag{LLM} & \colltag{Coll} \humantag{Human} \llmtag{LLM} & \humantag{Human} \llmtag{LLM} \\ \hline
THRONE & \cite{kaul2024throne} & \textbf{T1}\, Hallucination Mitigation & Type I hallucinations & T & \llmtag{LLM} & \colltag{Coll} & -- \\ \hline
VH\allowbreak{}Test & \cite{huang2024VHTest} & \textbf{T1}\, Hallucination Mitigation & Visual hallucination (8 modes) & T & \humantag{Human} \ruletag{Rule} & \colltag{Coll} \humantag{Human} \ruletag{Rule} \llmtag{LLM} & \humantag{Human} \ruletag{Rule} \\ \hline
Vid\allowbreak{}Halluc & \cite{li2024vidhalluc} & \textbf{T1}\, Hallucination Mitigation & Action Hallucination; Temporal Sequence Sorting; Scene Transition Detection & T & \ruletag{Rule} & \colltag{Coll} \humantag{Human} \llmtag{LLM} & \ruletag{Rule} \humantag{Human} \llmtag{LLM} \\ \hline
VQ\allowbreak{}Av2-IDK & \cite{cha2024visually} & \textbf{T1}\, Hallucination Mitigation & Unanswerable VQA (``I don't know'') & T & \ruletag{Rule} & \colltag{Coll} \ruletag{Rule} & -- \\ \hline
JUS & \cite{groot2024jus} & \textbf{T2}\, Trustworthiness & Verbalized uncertainty on hard queries & T & \ruletag{Rule} & \humantag{Human} & -- \\ \hline
MLLM\allowbreak{}Guard & \cite{gu2024mllmguard} & \textbf{T2}\, Safety & Personal privacy, \etc\ (12 total across five dimensions) & T & \humantag{Human} \llmtag{LLM} & \colltag{Coll} \humantag{Human} & \humantag{Human} \\ \hline
MLLMU-Bench & \cite{liu2025mllmubench} & \textbf{T2}\, Trustworthiness & Multimodal unlearning, \etc\ (3 task types) & T & \ruletag{Rule} \llmtag{LLM} & \colltag{Coll} \llmtag{LLM} & \humantag{Human} \\ \hline
MMA-Safety\allowbreak{}Bench & \cite{wang2026mmasafetybench} & \textbf{T2}\, Safety & Visual semantic hijacking safety across 5 multimodal agent domains & T & \ruletag{Rule} \llmtag{LLM} & \colltag{Coll} \ruletag{Rule} \llmtag{LLM} & \humantag{Human} \ruletag{Rule} \\ \hline
MMU\allowbreak{}Bench & \cite{li2024MMUBench} & \textbf{T2}\, Machine Unlearning & Unlearning test across efficacy, generality, specificity, fluency, diversity, membership inference attack, and jailbreak attacks & T & \ruletag{Rule} \llmtag{LLM} & \colltag{Coll} \llmtag{LLM} & \humantag{Human} \llmtag{LLM} \\ \hline
Mo\allowbreak{}Ho\allowbreak{}Bench & \cite{zhu2025moho} & \textbf{T2}\, Trustworthiness & Unanswerable Visual Question Answering & T & \ruletag{Rule} \llmtag{LLM} & \colltag{Coll} \llmtag{LLM} & \humantag{Human} \llmtag{LLM} \\ \hline
MOSSBench & \cite{li2024mossbench} & \textbf{T2}\, Safety & Oversensitivity and false refusal on benign multimodal queries & T & \ruletag{Rule} \llmtag{LLM} & \humantag{Human} & \humantag{Human} \\ \hline
Multi\allowbreak{}Trust & \cite{zhang2024MultiTrust} & \textbf{T2}\, Trustworthiness & Basic World Understanding, \etc\ (32 total) & T & \ruletag{Rule} \llmtag{LLM} & \colltag{Coll} \humantag{Human} \ruletag{Rule} \llmtag{LLM} & -- \\ \hline
RTVLM & \cite{li2024red} & \textbf{T2}\, Safety & Image misleading, \etc\ (10 total) & T & \humantag{Human} \llmtag{LLM} & \colltag{Coll} \humantag{Human} \ruletag{Rule} \llmtag{LLM} & \humantag{Human} \ruletag{Rule} \\ \hline
Safe\allowbreak{}Bench & \cite{ying2026safebench} & \textbf{T2}\, Safety & Multimodal query safety (23 risk scenarios) & T & \llmtag{LLM} & \llmtag{LLM} & \llmtag{LLM} \\ \hline
SPA-VL & \cite{zhang2025spavl} & \textbf{T2}\, Safety & Safety preference alignment across 6 harmfulness domains (53 subcategories) & T & \llmtag{LLM} & \colltag{Coll} \llmtag{LLM} & \ruletag{Rule} \llmtag{LLM} \\ \hline
Video-Safety\allowbreak{}Bench & \cite{liu2025videosafetybench} & \textbf{T2}\, Safety & Safety evaluation under video-text attacks & T & \ruletag{Rule} \llmtag{LLM} \humantag{Human} & \llmtag{LLM} \ruletag{Rule} & \humantag{Human} \\ \hline
VLGuard & \cite{zong2024vlguard} & \textbf{T2}\, Safety & Safe/unsafe VLM refusal and black-box jailbreak evaluation & T & \ruletag{Rule} \llmtag{LLM} & \colltag{Coll} \humantag{Human} \llmtag{LLM} & \humantag{Human} \\ \hline
VSys\allowbreak{}Bench & \cite{yeo2026vsysbench} & \textbf{T2}\, Trustworthiness & System-message compliance and user-conflict hierarchy in MLLMs & T & \ruletag{Rule} \llmtag{LLM} & \colltag{Coll} \humantag{Human} \llmtag{LLM} & \humantag{Human} \\ \hline
\end{tabularx}\endgroup\end{table*}
\begin{table*}[tp]\centering
\ContinuedFloat
\caption[]{Overview of Trustworthiness and Safety benchmarks (continued).}
\begingroup\linespread{1.03}\fontsize{7.2}{8.1}\selectfont\setlength\tabcolsep{1.25pt}\renewcommand\arraystretch{1.00}
\prepareSurveyTableTags\begin{tabularx}{\textwidth}{@{}>{\raggedright\arraybackslash}p{.133\textwidth}l>{\raggedright\arraybackslash}p{.166\textwidth}>{\raggedright\arraybackslash}X>{\raggedright\arraybackslash}p{.070\textwidth}>{\raggedright\arraybackslash\spaceskip=1pt}p{.089\textwidth}>{\raggedright\arraybackslash\spaceskip=1pt}p{.160\textwidth}>{\raggedright\arraybackslash\spaceskip=1pt}p{.090\textwidth}@{}}
\hline\thickhline\rowcolor{mygray}\textbf{Benchmark} & \textbf{Ref.} & \textbf{Category \& Focus} & \textbf{Task} & \textbf{Output} & \textbf{Evaluation} & \textbf{Construction} & \textbf{Filter} \\ \hline
Aes\allowbreak{}Bias\allowbreak{}Bench & \cite{li2025aesbiasbench} & \textbf{T3}\, Spurious Bias & Aesthetic Perception; Aesthetic Assessment; Aesthetic Empathy & T & \ruletag{Rule} & \colltag{Coll} & -- \\ \hline
Correlation\allowbreak{}QA & \cite{han2024instinctive} & \textbf{T3}\, Spurious Bias & Spurious correlations & T & \ruletag{Rule} & \colltag{Coll} \ruletag{Rule} \llmtag{LLM} & -- \\ \hline
Jailbreak-Audio\allowbreak{}Bench & \cite{cheng2025jailbreakaudiobenc} & \textbf{T3}\, Robustness & Audio jailbreak attacks on audio-language models & T & \llmtag{LLM} & \colltag{Coll} \ruletag{Rule} & \humantag{Human} \llmtag{LLM} \\ \hline
Jail\allowbreak{}Break\allowbreak{}V-28K & \cite{luo2024JailBreakV-28K} & \textbf{T3}\, Robustness & Jailbreak attack & T & \llmtag{LLM} & \colltag{Coll} \humantag{Human} \ruletag{Rule} \llmtag{LLM} & \humantag{Human} \ruletag{Rule} \llmtag{LLM} \\ \hline
MM-Spu\allowbreak{}Bench & \cite{ye2024mm} & \textbf{T3}\, Spurious Bias & Spurious correlations & T & \ruletag{Rule} & \colltag{Coll} \humantag{Human} \llmtag{LLM} & \humantag{Human} \ruletag{Rule} \\ \hline
MMR & \cite{liu2024MMR} & \textbf{T3}\, Robustness & Positive and negative VQA & T & \ruletag{Rule} & \colltag{Coll} \humantag{Human} & -- \\ \hline
Agent\allowbreak{}Safe & \cite{ying2026safe} & \textbf{T4}\, Safety & Embodied agent safety (3 stages) & T & \ruletag{Rule} \llmtag{LLM} & \humantag{Human} \ruletag{Rule} & -- \\ \hline
LOKI & \cite{ye2025loki} & \textbf{T4}\, Trustworthiness & Synthetic data detection, \etc\ (26 subcategories) & T & \ruletag{Rule} \llmtag{LLM} & \colltag{Coll} \humantag{Human} \llmtag{LLM} & \humantag{Human} \\ \hline
M\allowbreak{}Halu\allowbreak{}Bench & \cite{chen2024unified} & \textbf{T4}\, Hallucination Mitigation & Hallucination detection (3 levels) & T & \ruletag{Rule} & \colltag{Coll} \humantag{Human} \llmtag{LLM} & \humantag{Human} \\ \hline
MM-Judge\allowbreak{}Bias & \cite{lee2026mmjudgebias} & \textbf{T4}\, Trustworthiness & Compositional bias sensitivity and stability in MLLM-as-a-Judge (9 biases) & T & \ruletag{Rule} & \colltag{Coll} \humantag{Human} \ruletag{Rule} \llmtag{LLM} & \humantag{Human} \ruletag{Rule} \\ \hline
MM-RLHF & \cite{zhang2025mmrlhf} & \textbf{T4}\, Trustworthiness & Critique-based multimodal reward modeling and safety preference ranking & T & \ruletag{Rule} \llmtag{LLM} & \colltag{Coll} \humantag{Human} & \humantag{Human} \\ \hline
MM\allowbreak{}Fake\allowbreak{}Bench & \cite{liu2025mmfakebench} & \textbf{T4}\, Trustworthiness & Multimodal misinformation detection & T & \ruletag{Rule} & \colltag{Coll} \llmtag{LLM} & \humantag{Human} \ruletag{Rule} \\ \hline
MM\allowbreak{}Safe\allowbreak{}Aware & \cite{wang2025mmsafeaware} & \textbf{T4}\, Safety & Safety Classification & T & \ruletag{Rule} & \colltag{Coll} \humantag{Human} & \humantag{Human} \\ \hline
SHEEP & \cite{mickus2026sheep} & \textbf{T4}\, Hallucination Mitigation & Span-level hallucination labelling & T & \ruletag{Rule} & \colltag{Coll} \humantag{Human} \llmtag{LLM} & \humantag{Human} \\ \hline
SHIELD & \cite{shi2024shield} & \textbf{T4}\, Trustworthiness & face anti-spoofing face forgery detection & T & \ruletag{Rule} & \colltag{Coll} \ruletag{Rule} & -- \\ \hline
VL-Reward\allowbreak{}Bench & \cite{li2025vlrewardbench} & \textbf{T4}\, Hallucination Mitigation & Judging preference between two candidate responses & T & \ruletag{Rule} & \colltag{Coll} \llmtag{LLM} & \humantag{Human} \llmtag{LLM} \\ \hline
\end{tabularx}\endgroup\end{table*}

%% file: sec/trustworthiness/bg.tex
\subsection{Taxonomy by Safety Risk and Evaluator Role}
\label{sec:trustworthiness_bg}

As MLLMs move into open-world and safety-critical deployments, perceptual and generative fluency must be paired with verifiable reliability. Distinct from the synthesis deliverables in \s\ref{sec:generation} (and from closed-loop safety execution in \s\ref{sec:application}), trustworthiness and safety benchmarks evaluate whether models remain factually grounded in sensory evidence, aligned with safety policies, calibrated under uncertainty, capable of selective unlearning, and resilient against adversarial prompts, spurious correlations, or demographic biases. Fig.~\ref{fig:map-trustworthiness} and Table~\ref{tbl:trustworthiness} divide all 48 benchmarks into four classes separating \textit{generator-side reliability} (\textbf{T1--T3}, 38 benchmarks) from \textit{evaluator, guardrail, and authenticity auditing} (\textbf{T4}, 10 benchmarks): hallucination and factual grounding (\textbf{T1}, 18), safety alignment, calibration, and unlearning (\textbf{T2}, 14), adversarial robustness, spurious correlations, and demographic bias (\textbf{T3}, 6), and evaluator, guardrail, and authenticity auditing (\textbf{T4}, 10).

\subsubsection{Hallucination and Factual Grounding}
\label{sec:t1_hallucination}
Hallucination refers to responses containing claims unsupported by or contradictory to the multimodal input~\cite{rohrbach2018object,li2023pope,ji2023survey,bai2024hallucination,yin2023survey,liu2024survey}. Standard taxonomies distinguish \textit{object hallucination} (fabricating non-existent entities~\cite{lovenia2023negative,li2023pope}), \textit{attribute hallucination} (misstating object color, count, or state~\cite{fu2024mme,wang2023llm}), and \textit{relation hallucination} (misrepresenting spatial or action links among objects~\cite{zheng2024reefknot}). Among the 18 \textbf{T1} benchmarks, image-grounded object, dialogue, and relation errors are probed by THRONE~\cite{kaul2024throne} (free-form object claims), ROPE~\cite{chen2024multi} (multi-object probing), MRHal-Bench~\cite{zhang2024automated} (multi-round drift), and Reefknot~\cite{zheng2024reefknot} (relational links). False premises, abstention, and controlled visual edits are tested by VHTest~\cite{huang2024VHTest}, HaloQuest~\cite{wang2024haloquest}, VQAv2-IDK~\cite{cha2024visually}, Hallu-PI~\cite{ding2024hallu}, BEAF~\cite{ye2024beaf}, and AutoHallusion~\cite{wu2024autohallusion}. Iterative and factual consistency suites include MMECeption~\cite{cao2023genception} (annotation-free coherence), RH-Bench~\cite{liu2025rhbench} (reasoning-chain drift), SimpleVQA~\cite{cheng2025simplevqa} (atomic factual QA), and CrossCheck-Bench~\cite{tian2025crosscheckbench} (cross-modal contradiction resolution). Extending beyond static images, VidHalluc~\cite{li2024vidhalluc} examines temporal hallucinations across action, sequence, and scene transitions; MCR-BENCH~\cite{wang2025audiotextdisagree} exposes text dominance under conflicting audio--text inputs; AV-ConfuseBench~\cite{ye2025eyesearsdisagree} tests hallucinated sounds for silent visual objects; and AVHBench~\cite{kim2025avhbench} evaluates bidirectional video-driven audio and audio-driven video hallucinations.

\subsubsection{Safety Alignment, Calibration, and Unlearning}
\label{sec:t2_safety}
Fourteen benchmarks (\textbf{T2}) evaluate normative alignment, epistemic honesty, and privacy-preserving forgetting:

\begin{itemize}
 \item \textbf{Multimodal Safety and Red-Teaming.} Harmful-request refusal and red-teaming are evaluated across 23 risk scenarios in SafeBench~\cite{ying2026safebench}, four trustworthiness axes in RTVLM~\cite{li2024red}, five bilingual safety dimensions in MLLMGuard~\cite{gu2024mllmguard}, and safety preference pairs in VLGuard~\cite{zong2024vlguard} and SPA-VL~\cite{zhang2025spavl}. Boundary and dynamic threat suites cover exaggerated refusal on benign images (MOSSBench~\cite{li2024mossbench}), vision-grounded system-prompt compliance under conflicting user turns (VSysBench~\cite{yeo2026vsysbench}), temporal harmful intent in video (Video-SafetyBench~\cite{liu2025videosafetybench}), and visual semantic hijacking in agent trajectories (MMA-SafetyBench~\cite{wang2026mmasafetybench}).

 \item \textbf{Holistic Trustworthiness, Honesty, and Calibration.} MultiTrust~\cite{zhang2024MultiTrust} unifies 32 tasks spanning truthfulness, safety, robustness, fairness, and privacy. MoHoBench~\cite{zhu2025moho} evaluates honesty via appropriate refusal across four types of visually unanswerable queries, and JUS~\cite{groot2024jus} tests whether verbalized confidence aligns with empirical accuracy on hard visual questions.

 \item \textbf{Multimodal Machine Unlearning.} Consolidating visual unlearning across the survey, MMUBench~\cite{li2024MMUBench} and MLLMU-Bench~\cite{liu2025mllmubench} test whether MLLMs selectively erase target visual concepts or private personal profiles while preserving general utility and resisting membership-inference or jailbreak recovery.
\end{itemize}

\subsubsection{Adversarial Robustness, Spurious Correlations, and Demographic Bias}
\label{sec:t3_robustness}
Six benchmarks (\textbf{T3}) test stability under deceptive prompts, cross-modal jailbreaks, shortcut correlations, and demographic biases. MMR~\cite{liu2024MMR} pairs positive visual questions with misleading leading prompts; JailBreakV-28K~\cite{luo2024JailBreakV-28K} measures transferability of textual and visual adversarial jailbreaks; and Jailbreak-AudioBench~\cite{cheng2025jailbreakaudiobenc} perturbs tone, speed, accent, and background noise in spoken harmful queries. For spurious correlations and social fairness, CorrelationQA~\cite{han2024instinctive} and MM-SpuBench~\cite{ye2024mm} test whether co-occurring visual--textual priors override core visual evidence, while AesBiasBench~\cite{li2025aesbiasbench} audits demographic stereotyping in aesthetic assessment.

\subsubsection{Evaluator, Guardrail, and Authenticity Auditing}
\label{sec:t4_diagnostics}
Ten benchmarks (\textbf{T4}) evaluate models acting as \textit{discriminative safety guardrails, factuality/bias-audited evaluators, or media-authenticity detectors} over supplied inputs and candidate outputs---whereas general capability-scoring judges (MLLM-as-a-Judge~\cite{chen2024mllmasajudge}), reasoning-step PRMs, and image-editing reward models sit in \textbf{R4} (\cf~\s\ref{sec:r4_judgment}). For factuality, alignment preference, and judge-bias auditing, MHaluBench~\cite{chen2024unified} and SHEEP~\cite{mickus2026sheep} benchmark claim- and span-level hallucination detectors, VL-RewardBench~\cite{li2025vlrewardbench} and MM-RLHF~\cite{zhang2025mmrlhf} score scalar and critique-based reward models on hallucination and safety preference pairs, and MM-JudgeBias~\cite{lee2026mmjudgebias} diagnoses nine compositional biases and perturbation instabilities in MLLM judges. For safety guardrails and media authenticity, MMSafeAware~\cite{wang2025mmsafeaware} tests multimodal hazard classification balanced against benign over-refusal, AgentSafe~\cite{ying2026safe} audits offline hazard recognition across embodied perception, planning, and execution prompts, and SHIELD~\cite{shi2024shield}, LOKI~\cite{ye2025loki}, and MMFakeBench~\cite{liu2025mmfakebench} benchmark face anti-spoofing, synthetic-media detection, and multimodal misinformation verification.

%% file: sec/trustworthiness/task.tex
\subsection{Diagnostic Metrics and Analytical Synthesis}
\label{sec:trustworthiness_task}

Trustworthiness and safety suites replace raw task accuracy with diagnostic metrics that penalize fabrication, overconfidence, unsafe compliance, over-sensitivity, and shortcut reliance (Table~\ref{tbl:trustworthiness}):

\begin{itemize}
 \item \textbf{Hallucination and Abstention Metrics (T1).} Object fabrication in descriptions is measured by Caption Hallucination Assessment with Image Relevance (CHAIR)~\cite{rohrbach2018object} and class-conditioned recall/precision in THRONE~\cite{kaul2024throne}; iterative semantic drift is captured by annotation-free GC@T in MMECeption~\cite{cao2023genception}; unified discriminative/generative hallucination is scored by Halr~\cite{zheng2024reefknot}; and reasoning-length visual drift is quantified by RH-AUC in RH-Bench~\cite{liu2025rhbench}. Under controlled object removal, BEAF~\cite{ye2024beaf} decomposes behavior into True Understanding (TU), Ignorance (IG), Stubbornness (SB), and Indecision (ID), while VQAv2-IDK~\cite{cha2024visually} and SimpleVQA~\cite{cheng2025simplevqa} explicitly credit abstention on unanswerable or uncertain queries.

 \item \textbf{Safety, Calibration, and Unlearning Metrics (T2, T3).} Safety and jailbreak suites report Attack Success Rate (ASR), refusal accuracy paired with benign-query helpfulness, multi-LLM jury consensus in SafeBench~\cite{ying2026safebench}, Llama Guard 3 screening in Jailbreak-AudioBench~\cite{cheng2025jailbreakaudiobenc}, and threshold-calibrated judge confidence (RJScore) in Video-SafetyBench~\cite{liu2025videosafetybench}. Calibration in JUS~\cite{groot2024jus} compares verbalized confidence against accuracy, while unlearning suites jointly score forgetting efficacy, retained-set generality, fluency, and adversarial extraction resistance~\cite{li2024MMUBench,liu2025mllmubench}.

 \item \textbf{Evaluator, Guardrail, and Authenticity Metrics (T4).} Because \textbf{T4} models inspect fixed candidate outputs or media artifacts, evaluators score span- or claim-level F1 in MHaluBench and SHEEP~\cite{chen2024unified,mickus2026sheep}, pairwise preference accuracy in VL-RewardBench~\cite{li2025vlrewardbench}, unsafe recall balanced against benign false-alarm rates in MMSafeAware~\cite{wang2025mmsafeaware}, and forgery/misinformation classification F1 with anomaly explanations in SHIELD, LOKI, and MMFakeBench~\cite{shi2024shield,ye2025loki,liu2025mmfakebench}.
\end{itemize}

\textbf{Analytical synthesis.} Separating \textbf{T1--T3} from \textbf{T4} resolves a recurring construct ambiguity (\s\ref{sec:evolution}): a generative model (\textbf{T1}) can lower object-hallucination precision errors simply by shortening captions unless recall is tracked jointly~\cite{kaul2024throne}, whereas a discriminative evaluator or authenticity detector (\textbf{T4}) inspects fixed inputs and must balance error recall against false refusals or false alarms~\cite{wang2025mmsafeaware,li2024mossbench}. Horizontal comparisons also show a sharp trade-off between jailbreak defense and benign utility: safety-tuned MLLMs that reduce visual/audio Attack Success Rates often exhibit elevated false-refusal rates on benign look-alike prompts (MOSSBench~\cite{li2024mossbench}) or fail when harmful intent is distributed across benign video frames (Video-SafetyBench~\cite{liu2025videosafetybench}). Across the corpus, Trustworthiness and Safety relies on LLM judges in 50.0\% of suites (24/48; 56.7\% with LLM construction vs.\ 38.9\% without, $+17.8$ pp; \s\ref{sec:crossstats}).

%% file: sec/application.tex
\section{Application Benchmarks}
\label{sec:application}
\begin{figure*}[tp]
\centering\chaptergraphic[width=\textwidth]{application_map}
\caption{Taxonomy and representative examples of Application benchmarks ($N=99$) spanning Regime~I interactive agents (\textbf{A1--A2}, 56) and Regime~II specialist verticals (\textbf{A3--A7}, 43). Illustrated tasks cover GUI grounding and game agency in ScreenSpot~\cite{cheng2024screenspot} and BALROG, bimanual manipulation in RoboTwin~2.0~\cite{chen2026robotwin20}, clinical QA in MedXpertQA~\cite{zuo2025medxpertqa}, cross-view remote sensing in UrBench~\cite{zhou2025urbench}, materials science in MaCBench~\cite{alampara2025macbench}, multimodal theory of mind in MuMA-ToM~\cite{shi2025mumatom}, and financial chart analysis in FinMME~\cite{luo2025finmme}.}\label{fig:map-application}
\end{figure*}

\input{sec/misc/tbl_application}

\input{sec/application/bg}

\input{sec/application/task}

%% file: sec/misc/tbl_application.tex
\input{sec/misc/tbl_application_clean}

%% file: sec/misc/tbl_application_clean.tex
\begin{table*}[tp]\centering
\caption{Overview of Application benchmarks ($N=99$), spanning Regime~I interactive agents (\textbf{A1--A2}, 56) and Regime~II specialist verticals (\textbf{A3--A7}, 43). Endpoint codes denote static answers (\textbf{Q}), target locations (\textbf{L}), generated deliverables (\textbf{D}), predicted actions or trajectories (\textbf{P}), and verified environment outcomes (\textbf{E}), with the first code marking the primary track; output formats follow Table~\ref{tbl:understanding}.}
\label{tbl:application}
\begingroup\linespread{0.97}\fontsize{7.2}{8.1}\selectfont\setlength\tabcolsep{1.25pt}\renewcommand\arraystretch{1.00}
\prepareSurveyTableTags\begin{tabularx}{\textwidth}{@{}>{\raggedright\arraybackslash}p{.130\textwidth}l>{\raggedright\arraybackslash}p{.119\textwidth}l>{\raggedright\arraybackslash}X>{\raggedright\arraybackslash}p{.065\textwidth}>{\raggedright\arraybackslash\spaceskip=1pt}p{.089\textwidth}>{\raggedright\arraybackslash\spaceskip=1pt}p{.160\textwidth}>{\raggedright\arraybackslash\spaceskip=1pt}p{.090\textwidth}@{}}
\hline\thickhline
\rowcolor{mygray}\textbf{Benchmark} & \textbf{Ref.} & \textbf{Domain \& Focus} & \textbf{Endpoint} & \textbf{Task} & \textbf{Output} & \textbf{Evaluation} & \textbf{Construction} & \textbf{Filter} \\ \hline
Agent\allowbreak{}Net\allowbreak{}Bench & \cite{wang2025agentnetbench} & \textbf{A1}\, Computer use & P & Next desktop action on recorded steps & DF & \ruletag{Rule} & \humantag{Human} & \humantag{Human} \ruletag{Rule} \\ \cline{1-9}
Agent\allowbreak{}Vid\allowbreak{}Bench & \cite{an2026agentvidbench} & \textbf{A1}\, Web search & Q & Multi-hop agentic video QA with evidence-acquisition trajectory scoring & DF FREE & \ruletag{Rule} \llmtag{LLM} & \colltag{Coll} \humantag{Human} \llmtag{LLM} & \humantag{Human} \\ \cline{1-9}
Android\allowbreak{}World & \cite{rawles2025androidworld} & \textbf{A1}\, Mobile UI & E & Android tasks (116 tasks) & DF & \ruletag{Rule} & \humantag{Human} \ruletag{Rule} \llmtag{LLM} & \humantag{Human} \\ \cline{1-9}
BALROG & \cite{paglieri2025balrog} & \textbf{A1}\, Games & E & Agentic reasoning on games, \etc\ (6 environments) & DF & \ruletag{Rule} & \colltag{Coll} & -- \\ \cline{1-9}
Browse\allowbreak{}Comp-VL & \cite{geng2026browsecompvl} & \textbf{A1}\, Web search & Q & Multi-hop multimodal information seeking & DF & \llmtag{LLM} & \colltag{Coll} \llmtag{LLM} & \humantag{Human} \ruletag{Rule} \llmtag{LLM} \\ \cline{1-9}
CRAB & \cite{xu2024crab} & \textbf{A1}\, Computer use & E & Cross-environment agent tasks (120 tasks) & DF & \ruletag{Rule} & \humantag{Human} \ruletag{Rule} \llmtag{LLM} & \humantag{Human} \\ \cline{1-9}
Ferret-UI & \cite{you2024Ferret-UI} & \textbf{A1}\, Mobile UI & Q, L & Icon recognition, \etc\ (14 mobile UI tasks) & DF FREE & \ruletag{Rule} & \colltag{Coll} \llmtag{LLM} \ruletag{Rule} & \ruletag{Rule} \\ \cline{1-9}
Mind2Web 2 & \cite{gou2025mind2web2} & \textbf{A1}\, Web search & Q & Long-horizon agentic search & FREE & \llmtag{LLM} & \humantag{Human} & \humantag{Human} \\ \cline{1-9}
MMIn\allowbreak{}A & \cite{tian2025mmina} & \textbf{A1}\, Web & E & Multihop compositional tasks across evolving multimodal websites (1,050 tasks) & DF & \ruletag{Rule} \llmtag{LLM} & \colltag{Coll} \humantag{Human} & \humantag{Human} \\ \cline{1-9}
OS\allowbreak{}World & \cite{xie2024osworld} & \textbf{A1}\, Desktop & E & Open-ended computer tasks & DF & \ruletag{Rule} & \colltag{Coll} \humantag{Human} & \humantag{Human} \\ \cline{1-9}
Real\allowbreak{}X-Bench & \cite{hong2026realxbench} & \textbf{A1}\, Web search & Q & Perception, search and reasoning QA & DF & \ruletag{Rule} & \colltag{Coll} \humantag{Human} & \humantag{Human} \ruletag{Rule} \\ \cline{1-9}
Screen\allowbreak{}PR & \cite{fan2024read} & \textbf{A1}\, GUI & Q & Screen point-and-read description & FREE & \humantag{Human} \ruletag{Rule} \llmtag{LLM} & \colltag{Coll} \humantag{Human} \ruletag{Rule} \llmtag{LLM} & \humantag{Human} \ruletag{Rule} \\ \cline{1-9}
Screen\allowbreak{}Spot & \cite{cheng2024screenspot} & \textbf{A1}\, GUI & L & GUI element grounding & DF & \ruletag{Rule} & \humantag{Human} & -- \\ \cline{1-9}
Screen\allowbreak{}Spot-Pro & \cite{li2025screenspotpro} & \textbf{A1}\, GUI & L & Professional high-resolution GUI grounding & DF & \ruletag{Rule} & \humantag{Human} & \humantag{Human} \\ \cline{1-9}
Video\allowbreak{}Game\allowbreak{}QA-Bench & \cite{taesiri2025videogameqabench} & \textbf{A1}\, Game testing & Q, D & Visual unit testing, \etc\ (9 tasks) & BI DF FREE & \ruletag{Rule} \llmtag{LLM} & \colltag{Coll} \humantag{Human} \ruletag{Rule} \llmtag{LLM} & \humantag{Human} \\ \cline{1-9}
Visual\allowbreak{}Web\allowbreak{}Arena & \cite{koh2024visualwebarena} & \textbf{A1}\, Web & E & 910 realistic visually grounded web tasks & DF & \ruletag{Rule} \llmtag{LLM} & \colltag{Coll} \humantag{Human} & \ruletag{Rule} \\ \cline{1-9}
WASP & \cite{evtimov2025wasp} & \textbf{A1}\, Web & E & Web-agent prompt-injection security & DF & \ruletag{Rule} \llmtag{LLM} & \colltag{Coll} \humantag{Human} \ruletag{Rule} & \humantag{Human} \\ \cline{1-9}
Web\allowbreak{}Voyager & \cite{he2024webvoyager} & \textbf{A1}\, Web & E & Web agent tasks (643 tasks) & DF & \humantag{Human} \llmtag{LLM} & \colltag{Coll} \humantag{Human} \llmtag{LLM} & \humantag{Human} \ruletag{Rule} \\ \cline{1-9}
Work\allowbreak{}Arena & \cite{drouin2024workarena} & \textbf{A1}\, Enterprise UI & E & Enterprise web tasks (33 tasks) & DF & \ruletag{Rule} & \humantag{Human} \ruletag{Rule} \llmtag{LLM} & -- \\ \cline{1-9}
Bench2Drive & \cite{jia2024bench2drive} & \textbf{A2}\, Driving & E & Closed-loop end-to-end autonomous driving across 220 CARLA v2 routes & DF & \ruletag{Rule} & \colltag{Coll} \ruletag{Rule} & \ruletag{Rule} \\ \cline{1-9}
DOM benchmark & \cite{xie2026dombenchmark} & \textbf{A2}\, Manipulation & E & Dynamic object manipulation (9 sub-dimensions) & DF & \ruletag{Rule} & \colltag{Coll} \ruletag{Rule} & -- \\ \cline{1-9}
DOMINO & \cite{fang2026domino} & \textbf{A2}\, Manipulation & E & Dynamic manipulation (35 tasks) & DF & \ruletag{Rule} & \ruletag{Rule} & \ruletag{Rule} \\ \cline{1-9}
Drive\allowbreak{}Bench & \cite{xie2025drivebench} & \textbf{A2}\, Driving & Q & Driving QA under corruptions (17 settings) & MC FREE & \ruletag{Rule} \llmtag{LLM} & \colltag{Coll} \ruletag{Rule} & \ruletag{Rule} \llmtag{LLM} \\ \cline{1-9}
Drive\allowbreak{}QA & \cite{wei2025driveqa} & \textbf{A2}\, Driving & Q & Traffic regulation, signage and right-of-way driving knowledge test & MC & \ruletag{Rule} & \colltag{Coll} \ruletag{Rule} \llmtag{LLM} & \humantag{Human} \ruletag{Rule} \\ \cline{1-9}
Embodied\allowbreak{}Bench & \cite{yang2025embodiedbench} & \textbf{A2}\, Embodied tasks & E & Goal completion after household actions & DF & \ruletag{Rule} & \colltag{Coll} \humantag{Human} \ruletag{Rule} \llmtag{LLM} & \humantag{Human} \\ \cline{1-9}
Impromptu VLA & \cite{chi2025impromptuvla} & \textbf{A2}\, Driving & P, Q & Driving QA and trajectory prediction & DF & \ruletag{Rule} & \colltag{Coll} \ruletag{Rule} \llmtag{LLM} & \humantag{Human} \ruletag{Rule} \llmtag{LLM} \\ \cline{1-9}
Instruct\allowbreak{}Part & \cite{wan2025instructpart} & \textbf{A2}\, Manipulation & L & Mask of the part needed for a task & DF & \ruletag{Rule} & \colltag{Coll} \humantag{Human} \llmtag{LLM} & \humantag{Human} \\ \cline{1-9}
Inter\allowbreak{}VLA & \cite{xu2025perceiving} & \textbf{A2}\, Human motion & P & Future hand trajectory from video and instruction & DF & \ruletag{Rule} & \colltag{Coll} \llmtag{LLM} & \humantag{Human} \\ \cline{1-9}
I\allowbreak{}Ref-VLA & \cite{zhang2025irefvla} & \textbf{A2}\, Indoor scenes & L, Q & Referential object-grounding & DF FREE & \ruletag{Rule} & \colltag{Coll} \humantag{Human} & \humantag{Human} \\ \cline{1-9}
IS-Bench & \cite{lu2026isbench} & \textbf{A2}\, Household tasks & E & Safety-aware household task planning & DF & \ruletag{Rule} \llmtag{LLM} & \colltag{Coll} \humantag{Human} \llmtag{LLM} & \humantag{Human} \\ \cline{1-9}
L-CALVIN & \cite{fan2025lcalvin} & \textbf{A2}\, Manipulation & E & Long-horizon manipulation (34 base tasks) & DF & \ruletag{Rule} & \colltag{Coll} \ruletag{Rule} & \ruletag{Rule} \\ \cline{1-9}
LHPR-VLN & \cite{song2025lhprvln} & \textbf{A2}\, Navigation & E & Long-horizon vision-language navigation & DF & \ruletag{Rule} & \colltag{Coll} \ruletag{Rule} \llmtag{LLM} & -- \\ \cline{1-9}
LIBERO-CF & \cite{fang2026liberocf} & \textbf{A2}\, Manipulation & E & Counterfactual language following (50 tasks) & DF & \ruletag{Rule} & \colltag{Coll} \humantag{Human} & -- \\ \cline{1-9}
\hline\end{tabularx}\endgroup
\end{table*}
\begin{table*}[tp]\centering
\ContinuedFloat
\caption[]{Overview of Application benchmarks (continued).}
\begingroup\linespread{0.97}\fontsize{7.2}{8.1}\selectfont\setlength\tabcolsep{1.25pt}\renewcommand\arraystretch{1.00}
\prepareSurveyTableTags\begin{tabularx}{\textwidth}{@{}>{\raggedright\arraybackslash}p{.130\textwidth}l>{\raggedright\arraybackslash}p{.119\textwidth}l>{\raggedright\arraybackslash}X>{\raggedright\arraybackslash}p{.065\textwidth}>{\raggedright\arraybackslash\spaceskip=1pt}p{.089\textwidth}>{\raggedright\arraybackslash\spaceskip=1pt}p{.160\textwidth}>{\raggedright\arraybackslash\spaceskip=1pt}p{.090\textwidth}@{}}
\hline\thickhline
\rowcolor{mygray}\textbf{Benchmark} & \textbf{Ref.} & \textbf{Domain \& Focus} & \textbf{Endpoint} & \textbf{Task} & \textbf{Output} & \textbf{Evaluation} & \textbf{Construction} & \textbf{Filter} \\ \hline
LIBERO-Safety & \cite{cui2026liberosafety} & \textbf{A2}\, Manipulation & E & Manipulation safety, \etc\ (75 tasks) & DF & \ruletag{Rule} & \humantag{Human} \ruletag{Rule} \llmtag{LLM} & \humantag{Human} \\ \cline{1-9}
Mine\allowbreak{}Any\allowbreak{}Build & \cite{wei2025mineanybuild} & \textbf{A2}\, Virtual construction & D, Q & Executable building blueprint, mental-rotation questions & MC DF FREE & \ruletag{Rule} \llmtag{LLM} & \colltag{Coll} \humantag{Human} \llmtag{LLM} & \humantag{Human} \ruletag{Rule} \\ \cline{1-9}
MV-Bench & \cite{zhang2025mvbench} & \textbf{A2}\, Manipulation & E & Multi-view VLA control & DF & \ruletag{Rule} & \colltag{Coll} \ruletag{Rule} & -- \\ \cline{1-9}
Omni\allowbreak{}Drive & \cite{wang2025omnidrive} & \textbf{A2}\, Driving & P, Q & Driving QA and open-loop planning & DF FREE & \ruletag{Rule} & \colltag{Coll} \ruletag{Rule} \llmtag{LLM} & \humantag{Human} \ruletag{Rule} \\ \cline{1-9}
Open\allowbreak{}EQA & \cite{OpenEQA2023} & \textbf{A2}\, Indoor scenes & Q & Embodied QA (7 categories) & FREE & \llmtag{LLM} & \colltag{Coll} \humantag{Human} & \humantag{Human} \\ \cline{1-9}
Pola\allowbreak{}Ri\allowbreak{}S & \cite{jain2025polaris} & \textbf{A2}\, Manipulation & E & Real-to-sim robot policy evaluation & DF & \ruletag{Rule} & \humantag{Human} \ruletag{Rule} \llmtag{LLM} & -- \\ \cline{1-9}
RL4VLA & \cite{liu2025bench} & \textbf{A2}\, Manipulation & E & VLA pick-and-place generalization (3 dimensions) & DF & \ruletag{Rule} & \colltag{Coll} \ruletag{Rule} \llmtag{LLM} & -- \\ \cline{1-9}
Robo\allowbreak{}Casa365 & \cite{nasiriany2026robocasa365} & \textbf{A2}\, Household robotics & E & Household mobile manipulation (365 tasks) & DF & \ruletag{Rule} & \colltag{Coll} \humantag{Human} \ruletag{Rule} \llmtag{LLM} & -- \\ \cline{1-9}
Robo\allowbreak{}Inter-VQA & \cite{li2026robointervqa} & \textbf{A2}\, Manipulation & Q & Embodied spatial and temporal QA & BI MC DF FREE & \ruletag{Rule} & \colltag{Coll} \humantag{Human} \ruletag{Rule} \llmtag{LLM} & \humantag{Human} \\ \cline{1-9}
Robo\allowbreak{}Lab & \cite{yang2026robolab} & \textbf{A2}\, Manipulation & E & Pick-and-place competency (120 tasks) & DF & \ruletag{Rule} & \colltag{Coll} \humantag{Human} \ruletag{Rule} & -- \\ \cline{1-9}
Robo\allowbreak{}MME & \cite{dai2026robomme} & \textbf{A2}\, Manipulation & E & Memory-dependent manipulation (16 tasks) & DF & \ruletag{Rule} & \ruletag{Rule} & \ruletag{Rule} \\ \cline{1-9}
Robo\allowbreak{}Twin & \cite{mu2025robotwin} & \textbf{A2}\, Manipulation & E & Generative digital-twin dual-arm robotic manipulation benchmark & DF & \ruletag{Rule} & \humantag{Human} \ruletag{Rule} \llmtag{LLM} & \humantag{Human} \ruletag{Rule} \\ \cline{1-9}
Robo\allowbreak{}Twin 2.0 & \cite{chen2026robotwin20} & \textbf{A2}\, Manipulation & E & Bimanual manipulation (50 dual-arm tasks) & DF & \ruletag{Rule} & \colltag{Coll} \humantag{Human} \ruletag{Rule} \llmtag{LLM} & \humantag{Human} \ruletag{Rule} \\ \cline{1-9}
Theory of Space & \cite{zhang2026theoryofspace} & \textbf{A2}\, Spatial exploration & Q & Active spatial exploration, \etc\ (9 tasks) & DF & \ruletag{Rule} & \ruletag{Rule} & -- \\ \cline{1-9}
UAV-Need-Help & \cite{wang2025uavneedhelp} & \textbf{A2}\, Aerial navigation & E & Assistant-guided UAV object search & DF & \ruletag{Rule} & \humantag{Human} \llmtag{LLM} & \humantag{Human} \ruletag{Rule} \\ \cline{1-9}
UAV-ON & \cite{xiao2025uavon} & \textbf{A2}\, Aerial navigation & E & Aerial object goal navigation & DF & \ruletag{Rule} & \ruletag{Rule} \llmtag{LLM} & \humantag{Human} \\ \cline{1-9}
VA\allowbreak{}Bench & \cite{yuan2026vabench} & \textbf{A2}\, Manipulation & L, P & Point affordance and visual trace prediction & DF & \ruletag{Rule} \llmtag{LLM} & \colltag{Coll} \humantag{Human} & -- \\ \cline{1-9}
Visual\allowbreak{}Agent\allowbreak{}Bench & \cite{liu2024VisualAgentBench} & \textbf{A2}\, Agent environments & E & Household task completion in OmniGibson (selected track) & DF & \ruletag{Rule} & \colltag{Coll} \humantag{Human} \ruletag{Rule} \llmtag{LLM} & \humantag{Human} \ruletag{Rule} \\ \cline{1-9}
VLA-Arena & \cite{zhang2026vlaarena} & \textbf{A2}\, Manipulation & E & Safety, \etc\ (4 dimensions, 170 tasks) & DF & \ruletag{Rule} & \humantag{Human} \ruletag{Rule} & -- \\ \cline{1-9}
VLA-REPLICA & \cite{huang2026vlareplica} & \textbf{A2}\, Manipulation & E & Reproducible low-cost real-world VLA robotic manipulation suite & DF & \ruletag{Rule} & \humantag{Human} & \humantag{Human} \\ \cline{1-9}
VLA\allowbreak{}Bench & \cite{zhang2025vlabench} & \textbf{A2}\, Manipulation & E & Language-conditioned manipulation (100 tasks) & DF & \ruletag{Rule} & \colltag{Coll} \humantag{Human} \ruletag{Rule} \llmtag{LLM} & \ruletag{Rule} \\ \cline{1-9}
VLN-CE-Isaac & \cite{cheng2024vlnceisaac} & \textbf{A2}\, Navigation & E & Legged-robot vision-language navigation & DF & \ruletag{Rule} & \colltag{Coll} & \ruletag{Rule} \\ \cline{1-9}
World\allowbreak{}Gym & \cite{quevedo2026worldgym} & \textbf{A2}\, Manipulation & E & World-model VLA policy evaluation & DF & \llmtag{LLM} & \colltag{Coll} \llmtag{LLM} & -- \\ \cline{1-9}
Asclepius & \cite{wang2024asclepius} & \textbf{A3}\, Medicine & Q & 15 medical specialties multi-modal QA & BI MC FREE & \ruletag{Rule} \llmtag{LLM} & \colltag{Coll} \humantag{Human} \ruletag{Rule} & \humantag{Human} \\ \cline{1-9}
CARES & \cite{xia2024cares} & \textbf{A3}\, Medicine & Q & Med-LVLM trustworthiness (5 dimensions) & BI MC FREE & \ruletag{Rule} \llmtag{LLM} & \colltag{Coll} \ruletag{Rule} \llmtag{LLM} & \humantag{Human} \llmtag{LLM} \\ \cline{1-9}
Chest\allowbreak{}Agent\allowbreak{}Bench & \cite{fallahpour2025chestagentbench} & \textbf{A3}\, Chest radiology & Q & Chest X-ray reasoning (7 categories) & MC & \ruletag{Rule} & \colltag{Coll} \llmtag{LLM} & \llmtag{LLM} \\ \cline{1-9}
Dining\allowbreak{}Bench & \cite{jin2026diningbench} & \textbf{A3}\, Food and nutrition & Q & Dish classification and nutrition estimation & MC DF FREE & \ruletag{Rule} \llmtag{LLM} & \colltag{Coll} \llmtag{LLM} & \humantag{Human} \ruletag{Rule} \llmtag{LLM} \\ \cline{1-9}
GMAI-MM\allowbreak{}Bench & \cite{chen2024GMAI-MMBench} & \textbf{A3}\, Medicine & Q & 18 clinical-related tasks & MC & \ruletag{Rule} & \colltag{Coll} \humantag{Human} \ruletag{Rule} \llmtag{LLM} & \humantag{Human} \ruletag{Rule} \\ \cline{1-9}
Med\allowbreak{}Agent\allowbreak{}Board & \cite{zhu2025medagentboard} & \textbf{A3}\, Medicine & Q, D & Medical QA, \etc\ (4 task categories) & MC DF FREE & \humantag{Human} \ruletag{Rule} \llmtag{LLM} & \colltag{Coll} \humantag{Human} \llmtag{LLM} & \humantag{Human} \\ \cline{1-9}
Medi\allowbreak{}Confusion & \cite{sepehri2025mediconfusion} & \textbf{A3}\, Medical imaging & Q & Paired visual-confusion reliability probing in radiology VQA & MC & \ruletag{Rule} & \colltag{Coll} \humantag{Human} \ruletag{Rule} & \humantag{Human} \\ \cline{1-9}
Med\allowbreak{}Xpert\allowbreak{}QA & \cite{zuo2025medxpertqa} & \textbf{A3}\, Medicine & Q & Expert-level medical QA, \etc\ (3 task categories) & MC & \ruletag{Rule} & \colltag{Coll} \llmtag{LLM} & \humantag{Human} \ruletag{Rule} \llmtag{LLM} \\ \cline{1-9}
MTB\allowbreak{}Bench & \cite{vasilev2025mtbbench} & \textbf{A3}\, Oncology & Q & Diagnostic classification; Spatial biomarker interpretation & BI MC & \ruletag{Rule} & \colltag{Coll} \humantag{Human} \llmtag{LLM} & \humantag{Human} \\ \cline{1-9}
\hline\end{tabularx}\endgroup
\end{table*}
\begin{table*}[tp]\centering
\ContinuedFloat
\caption[]{Overview of Application benchmarks (continued).}
\begingroup\linespread{0.97}\fontsize{7.2}{8.1}\selectfont\setlength\tabcolsep{1.25pt}\renewcommand\arraystretch{1.00}
\prepareSurveyTableTags\begin{tabularx}{\textwidth}{@{}>{\raggedright\arraybackslash}p{.130\textwidth}l>{\raggedright\arraybackslash}p{.119\textwidth}l>{\raggedright\arraybackslash}X>{\raggedright\arraybackslash}p{.065\textwidth}>{\raggedright\arraybackslash\spaceskip=1pt}p{.089\textwidth}>{\raggedright\arraybackslash\spaceskip=1pt}p{.160\textwidth}>{\raggedright\arraybackslash\spaceskip=1pt}p{.090\textwidth}@{}}
\hline\thickhline
\rowcolor{mygray}\textbf{Benchmark} & \textbf{Ref.} & \textbf{Domain \& Focus} & \textbf{Endpoint} & \textbf{Task} & \textbf{Output} & \textbf{Evaluation} & \textbf{Construction} & \textbf{Filter} \\ \hline
Omni\allowbreak{}Brain\allowbreak{}Bench & \cite{peng2026omnibrainbench} & \textbf{A3}\, Medical imaging & Q & 15-modality clinical brain imaging VQA across 15 multi-stage tasks & MC FREE & \ruletag{Rule} \llmtag{LLM} & \colltag{Coll} \humantag{Human} \llmtag{LLM} & \humantag{Human} \\ \cline{1-9}
Omni\allowbreak{}Med\allowbreak{}VQA & \cite{hu2024omnimedvqa} & \textbf{A3}\, Medical imaging & Q & Modality recognition, \etc\ (5 question types) & MC & \ruletag{Rule} & \colltag{Coll} \ruletag{Rule} \llmtag{LLM} & \humantag{Human} \ruletag{Rule} \\ \cline{1-9}
Slide\allowbreak{}Bench & \cite{chen2025slidebench} & \textbf{A3}\, Pathology & Q, D & Pathology captioning, \etc & MC FREE & \ruletag{Rule} \llmtag{LLM} & \colltag{Coll} \ruletag{Rule} \llmtag{LLM} & \humantag{Human} \llmtag{LLM} \\ \cline{1-9}
Agro\allowbreak{}Bench & \cite{shinoda2025agrobench} & \textbf{A4}\, Agriculture & Q & Disease and pest identification, \etc\ (7 tasks) & MC & \ruletag{Rule} & \colltag{Coll} \humantag{Human} & \humantag{Human} \\ \cline{1-9}
CDDM & \cite{liu2024cddm} & \textbf{A4}\, Agriculture & Q & Crop disease diagnosis QA & DF FREE & \ruletag{Rule} \llmtag{LLM} & \colltag{Coll} \humantag{Human} \llmtag{LLM} & -- \\ \cline{1-9}
Earth-Bench & \cite{feng2026earthbench} & \textbf{A4}\, Earth observation & Q & Tool-augmented Earth observation (14 tasks) & MC DF & \ruletag{Rule} & \colltag{Coll} \humantag{Human} & \humantag{Human} \\ \cline{1-9}
LHRS-Bench & \cite{muhtar2024lhrs} & \textbf{A4}\, Remote sensing & Q & Recognition, \etc\ (5 dimensions) & MC & \ruletag{Rule} & \colltag{Coll} \humantag{Human} & -- \\ \cline{1-9}
Ur\allowbreak{}Bench & \cite{zhou2025urbench} & \textbf{A4}\, Urban scenes & Q & Geo-localization, \etc\ (14 urban tasks) & MC DF & \ruletag{Rule} & \colltag{Coll} \humantag{Human} \ruletag{Rule} \llmtag{LLM} & \humantag{Human} \ruletag{Rule} \\ \cline{1-9}
VRS\allowbreak{}Bench & \cite{li2024vrsbench} & \textbf{A4}\, Remote sensing & L, Q, D & Remote sensing captioning, \etc\ (3 tasks) & BI DF FREE & \ruletag{Rule} \llmtag{LLM} & \colltag{Coll} \llmtag{LLM} & \humantag{Human} \ruletag{Rule} \\ \cline{1-9}
Blender\allowbreak{}Gym & \cite{gu2025blendergym} & \textbf{A5}\, 3D design & D & Blender code editing (5 tasks) & DF & \ruletag{Rule} & \colltag{Coll} \humantag{Human} & -- \\ \cline{1-9}
CORE-Bench & \cite{siegel2024corebench} & \textbf{A5}\, Computational science & Q & Computational reproducibility (270 tasks) & DF & \ruletag{Rule} & \colltag{Coll} \humantag{Human} & \humantag{Human} \ruletag{Rule} \\ \cline{1-9}
Design\allowbreak{}QA & \cite{doris2024DesignQA} & \textbf{A5}\, Engineering design & Q & Design QA & BI DF FREE & \ruletag{Rule} & \colltag{Coll} \humantag{Human} \ruletag{Rule} & \humantag{Human} \\ \cline{1-9}
DS\allowbreak{}Bench & \cite{jing2025dsbench} & \textbf{A5}\, Data science & Q, D & Data analysis and modeling & MC DF & \ruletag{Rule} \llmtag{LLM} & \colltag{Coll} & \ruletag{Rule} \\ \cline{1-9}
Ma\allowbreak{}C\allowbreak{}Bench & \cite{alampara2025macbench} & \textbf{A5}\, Scientific research & Q & Table extraction, \etc\ (9 task categories) & MC DF & \ruletag{Rule} & \colltag{Coll} \humantag{Human} & \humantag{Human} \\ \cline{1-9}
Mech\allowbreak{}VQA & \cite{kou2026mechvqa} & \textbf{A5}\, Mechanical drawings & Q & Mechanical drawing recognition and reasoning & DF FREE & \llmtag{LLM} & \colltag{Coll} \humantag{Human} \ruletag{Rule} \llmtag{LLM} & \humantag{Human} \llmtag{LLM} \\ \cline{1-9}
MLR-Bench & \cite{chen2025mlrbench} & \textbf{A5}\, Auto-research & D & Open-ended ML auto-research from ideation and experiments to manuscripts (201 tasks) & FREE & \llmtag{LLM} & \colltag{Coll} \humantag{Human} \llmtag{LLM} & \humantag{Human} \\ \cline{1-9}
MMAD & \cite{jiang2025mmad} & \textbf{A5}\, Manufacturing & Q & 7 subtasks for industrial anomaly detection & MC & \ruletag{Rule} & \colltag{Coll} \llmtag{LLM} & \humantag{Human} \\ \cline{1-9}
MM\allowbreak{}Art-Bench & \cite{lin2025mmartbench} & \textbf{A5}\, Photo editing & D & Instruction-based photo retouching & FREE & \humantag{Human} \ruletag{Rule} \llmtag{LLM} & \colltag{Coll} \humantag{Human} \ruletag{Rule} \llmtag{LLM} & \humantag{Human} \ruletag{Rule} \\ \cline{1-9}
Mo\allowbreak{}Sci\allowbreak{}Bench & \cite{liu2026moscibench} & \textbf{A5}\, Auto-research & Q & End-to-end multimodal scientific discovery and hypothesis verification (7 modalities) & MC DF & \ruletag{Rule} & \colltag{Coll} \humantag{Human} \llmtag{LLM} & \humantag{Human} \ruletag{Rule} \\ \cline{1-9}
Paper\allowbreak{}Bench & \cite{starace2025paperbench} & \textbf{A5}\, Auto-research & D & Autonomous replication of ICML papers, codebases and plots (8,316 rubrics) & DF FREE & \ruletag{Rule} \llmtag{LLM} & \colltag{Coll} \humantag{Human} & \humantag{Human} \\ \cline{1-9}
Science\allowbreak{}Agent\allowbreak{}Bench & \cite{chen2025scienceagentbench} & \textbf{A5}\, Auto-research & D & Data-driven scientific discovery and visual plot generation (102 tasks) & DF & \ruletag{Rule} \llmtag{LLM} & \colltag{Coll} \humantag{Human} & \humantag{Human} \ruletag{Rule} \\ \cline{1-9}
Sci\allowbreak{}FI\allowbreak{}Bench & \cite{roberts2024scifibench} & \textbf{A5}\, Scientific research & Q & Scientific figure-to-caption and caption-to-figure interpretation & MC & \ruletag{Rule} & \colltag{Coll} \humantag{Human} \ruletag{Rule} \llmtag{LLM} & \humantag{Human} \ruletag{Rule} \llmtag{LLM} \\ \cline{1-9}
SWE-bench Multimodal & \cite{yang2025swebenchmultimodal} & \textbf{A5}\, Software & D & GitHub issue resolution with images & DF & \ruletag{Rule} & \colltag{Coll} & \humantag{Human} \ruletag{Rule} \\ \cline{1-9}
AICD & \cite{jia2025aicd} & \textbf{A6}\, Copyright & Q & AI artwork copyright judgment & BI FREE & \humantag{Human} \ruletag{Rule} & \colltag{Coll} \llmtag{LLM} & \humantag{Human} \\ \cline{1-9}
Art\allowbreak{}Mentor & \cite{zheng2025artmentor} & \textbf{A6}\, Art education & Q & Artwork entity recognition and review & DF FREE & \humantag{Human} \ruletag{Rule} & \colltag{Coll} \humantag{Human} \llmtag{LLM} & -- \\ \cline{1-9}
Edu\allowbreak{}Vid\allowbreak{}QA & \cite{ray2025eduvidqa} & \textbf{A6}\, Education & Q & Lecture video student QA & FREE & \ruletag{Rule} \llmtag{LLM} & \colltag{Coll} \humantag{Human} \llmtag{LLM} & \humantag{Human} \ruletag{Rule} \llmtag{LLM} \\ \cline{1-9}
Essay\allowbreak{}Judge & \cite{su2025essayjudge} & \textbf{A6}\, Education & Q & Trait-specific Essay Scoring & DF & \ruletag{Rule} & \colltag{Coll} \humantag{Human} & \humantag{Human} \\ \cline{1-9}
Human\allowbreak{}Sense & \cite{qin2025humansense} & \textbf{A6}\, Human behavior & Q & Person Appearance, Action Recognition, \etc\ (15 total tasks) & MC & \ruletag{Rule} & \colltag{Coll} \humantag{Human} \ruletag{Rule} \llmtag{LLM} & \humantag{Human} \\ \cline{1-9}
MM-Soc & \cite{jin2024mm-soc} & \textbf{A6}\, Social media & Q & Misinformation detection, \etc\ (10 total) & BI MC DF FREE & \ruletag{Rule} & \colltag{Coll} \llmtag{LLM} & \ruletag{Rule} \\ \cline{1-9}
MM\allowbreak{}To\allowbreak{}M-QA & \cite{jin2024mmtomqa} & \textbf{A6}\, Social cognition & Q & Belief and goal inference QA & MC & \ruletag{Rule} & \ruletag{Rule} \llmtag{LLM} & \ruletag{Rule} \\ \cline{1-9}
Mu\allowbreak{}MA-To\allowbreak{}M & \cite{shi2025mumatom} & \textbf{A6}\, Social cognition & Q & Multi-agent belief and goal inference QA & MC & \ruletag{Rule} & \ruletag{Rule} \llmtag{LLM} & -- \\ \cline{1-9}
E\allowbreak{}Com-Bench & \cite{wang2025ecombench} & \textbf{A7}\, Customer service & E & E-commerce customer-support task completion & DF & \ruletag{Rule} & \colltag{Coll} \humantag{Human} \llmtag{LLM} & \humantag{Human} \\ \cline{1-9}
Fin\allowbreak{}MME & \cite{luo2025finmme} & \textbf{A7}\, Finance & Q & Financial chart perception and reasoning & MC DF & \ruletag{Rule} & \colltag{Coll} \humantag{Human} \llmtag{LLM} & \humantag{Human} \ruletag{Rule} \llmtag{LLM} \\ \cline{1-9}
Multi\allowbreak{}Fin\allowbreak{}Ben & \cite{peng2026multifinben} & \textbf{A7}\, Finance & Q & Financial IE, \etc\ (36 datasets) & DF FREE & \humantag{Human} \ruletag{Rule} & \colltag{Coll} \humantag{Human} \ruletag{Rule} \llmtag{LLM} & \humantag{Human} \ruletag{Rule} \\ \cline{1-9}
\hline\end{tabularx}\endgroup
\end{table*}

%% file: sec/application/bg.tex
\subsection{Taxonomy by Agentic vs. Specialist Vertical Regimes}
\label{sec:application_bg}

Application benchmarks evaluate MLLMs in end-user and professional workflows integrating perception, planning, and domain knowledge. Listing all 99 Application benchmarks (Fig.~\ref{fig:map-application} and Table~\ref{tbl:application}) as flat vertical topics conceals a structural split: \textbf{Regime~I: Interactive GUI and Embodied Agents} (\textbf{A1--A2}, 56 benchmarks) routinely execute actions and verify environment state transitions (34 of all 35 primary \textbf{E} tracks; 34/56 or 60.7\% within Regime~I), whereas \textbf{Regime~II: Knowledge-Intensive Specialist Verticals} (\textbf{A3--A7}, 43 benchmarks) predominantly score static answers or generated deliverables (35 primary \textbf{Q} tracks, 6 \textbf{D} tracks, 1 \textbf{L} track in UrBench~\cite{zhou2025urbench}, and 1 \textbf{E} track in ECom-Bench~\cite{wang2025ecombench}). \s\ref{sec:application_task} pairs these seven operational areas with orthogonal track endpoints (\textbf{Q/L/D/P/E}).

\subsubsection{Regime I: Interactive GUI and Embodied Agents}
\label{sec:app_group1}

\noindent\textbf{GUI, Web, and Game Agents (19 records, A1).}
Digital-agent benchmarks evaluate visual--textual operation across web, mobile, desktop, and game environments. On web platforms, VisualWebArena~\cite{koh2024visualwebarena}, WebVoyager~\cite{he2024webvoyager}, WorkArena~\cite{drouin2024workarena}, and MMInA~\cite{tian2025mmina} execute single-site and multi-hop compositional tasks on self-hosted, live, and enterprise websites, while WASP~\cite{evtimov2025wasp} tests robustness against in-page prompt injections. On mobile and desktop interfaces, Ferret-UI~\cite{you2024Ferret-UI} and AndroidWorld~\cite{rawles2025androidworld} target mobile apps, OSWorld~\cite{xie2024osworld} and AgentNetBench~\cite{wang2025agentnetbench} evaluate online and offline desktop operation, and CRAB~\cite{xu2024crab} spans cross-device workflows. Supporting interface control, ScreenSpot~\cite{cheng2024screenspot} and ScreenSpot-Pro~\cite{li2025screenspotpro} isolate GUI element localization up to high-resolution specialist software, while ScreenPR~\cite{fan2024read} evaluates point-referred UI descriptions. In multimodal web research, RealX-Bench~\cite{hong2026realxbench}, BrowseComp-VL~\cite{geng2026browsecompvl}, Mind2Web~2~\cite{gou2025mind2web2}, and AgentVidBench~\cite{an2026agentvidbench} combine visual browsing, multi-hop video/web evidence acquisition, and trajectory-verified synthesis. In games, VideoGameQA-Bench~\cite{taesiri2025videogameqabench} probes visual glitch detection and unit testing, and BALROG~\cite{paglieri2025balrog} measures long-horizon agentic progression.

\noindent\textbf{Embodied AI and Autonomous Driving (37 records, A2).}
Embodied benchmarks span indoor spatial exploration, robotic manipulation, navigation, physical world models, safety, and autonomous driving:
\begin{itemize}
 \item \textit{Embodied Perception, Planning, and Spatial Exploration.} OpenEQA~\cite{OpenEQA2023} tests indoor open-vocabulary QA from episodic history or active exploration; InterVLA~\cite{xu2025perceiving} predicts 3D hand-interaction trajectories; VisualAgentBench~\cite{liu2024VisualAgentBench} and EmbodiedBench~\cite{yang2025embodiedbench} evaluate high-level planning alongside low-level control; IRef-VLA~\cite{zhang2025irefvla} probes interactive referential disambiguation; MineAnyBuild~\cite{wei2025mineanybuild} scores executable 3D block blueprints; and Theory of Space~\cite{zhang2026theoryofspace} tests active spatial exploration via externalized cognitive maps.
 \item \textit{VLA Manipulation, Affordances, and Simulators.} Vision-language-action (VLA) policy suites evaluate long-horizon manipulation (VLABench~\cite{zhang2025vlabench}), generalization axes (RL4VLA~\cite{liu2025bench}), camera-viewpoint shifts (MV-Bench~\cite{zhang2025mvbench}), counterfactual instructions in familiar scenes (LIBERO-CF~\cite{fang2026liberocf}), episodic memory (RoboMME~\cite{dai2026robomme}), sequential task chains (L-CALVIN~\cite{fan2025lcalvin}), dynamic moving objects (DOM benchmark~\cite{xie2026dombenchmark}, DOMINO~\cite{fang2026domino}), and learned video world models (WorldGym~\cite{quevedo2026worldgym}). Scalable simulation testbeds include bimanual perturbation suites (RoboTwin~\cite{mu2025robotwin} and RoboTwin~2.0~\cite{chen2026robotwin20}), mobile household kitchens (RoboCasa365~\cite{nasiriany2026robocasa365}), graded skill-and-safety arenas (VLA-Arena~\cite{zhang2026vlaarena}, RoboLab~\cite{yang2026robolab}), real-to-sim video reconstruction (PolaRiS~\cite{jain2025polaris}), and low-cost reproducible real-world hardware setups (VLA-REPLICA~\cite{huang2026vlareplica}). Complementing closed-loop rollouts, InstructPart~\cite{wan2025instructpart}, RoboInter-VQA~\cite{li2026robointervqa}, and VABench~\cite{yuan2026vabench} score part masks, grasp poses, affordance points, and visual traces.
 \item \textit{Navigation, Embodied Safety, and Autonomous Driving.} Vision-language navigation covers multi-stage indoor routes (LHPR-VLN~\cite{song2025lhprvln}), legged locomotion (VLN-CE-Isaac~\cite{cheng2024vlnceisaac}), and aerial UAV search (UAV-Need-Help~\cite{wang2025uavneedhelp}, UAV-ON~\cite{xiao2025uavon}). Embodied safety is tested by IS-Bench~\cite{lu2026isbench} and LIBERO-Safety~\cite{cui2026liberosafety}. In autonomous driving, Bench2Drive~\cite{jia2024bench2drive} benchmarks closed-loop end-to-end driving across 220 interactive CARLA~v2 routes, while OmniDrive~\cite{wang2025omnidrive}, DriveBench~\cite{xie2025drivebench}, DriveQA~\cite{wei2025driveqa}, and Impromptu VLA~\cite{chi2025impromptuvla} evaluate counterfactual 3D planning, sensor-corruption robustness, traffic-rule mastery, and unstructured-road trajectories.
\end{itemize}

\subsubsection{Regime II: Knowledge-Intensive Specialist Verticals}
\label{sec:app_group2}

\noindent\textbf{Healthcare and Clinical Medicine (12 records, A3).}
Clinical suites cover general and specialty medical imaging QA in Asclepius~\cite{wang2024asclepius}, GMAI-MMBench~\cite{chen2024GMAI-MMBench}, MedXpertQA~\cite{zuo2025medxpertqa}, OmniMedVQA~\cite{hu2024omnimedvqa}, and whole-slide pathology in SlideBench~\cite{chen2025slidebench}. Multi-stage clinical decision workflows include chest X-ray reasoning in ChestAgentBench~\cite{fallahpour2025chestagentbench}, longitudinal molecular tumor boards in MTBBench~\cite{vasilev2025mtbbench}, multi-agent clinical collaboration in MedAgentBoard~\cite{zhu2025medagentboard}, and 15-modality neuroimaging interpretation in OmniBrainBench~\cite{peng2026omnibrainbench}. Clinical reliability and nutrition suites further test visual-confusion robustness in MediConfusion~\cite{sepehri2025mediconfusion}, five-axis clinical trustworthiness in CARES~\cite{xia2024cares}, and dietary estimation in DiningBench~\cite{jin2026diningbench}.

\noindent\textbf{Remote Sensing and Earth Observation (6 records, A4).}
Earth-observation and agricultural suites cover satellite interpretation in LHRS-Bench~\cite{muhtar2024lhrs} and VRSBench~\cite{li2024vrsbench}, tool-augmented spectral and RGB analysis in Earth-Bench~\cite{feng2026earthbench}, cross-view street-to-satellite localization in UrBench~\cite{zhou2025urbench}, and crop pathology and weed recognition in AgroBench~\cite{shinoda2025agrobench} and CDDM~\cite{liu2024cddm}.

\noindent\textbf{AI for Science, Auto-Research, and Engineering (14 records, A5).}
Spanning automated scientific discovery (``auto-research''), laboratory analysis, and visual software/CAD engineering, benchmarks in \textbf{A5} couple multimodal paper and figure comprehension with executable code experimentation~\cite{seo2026paper2code}. In automated scientific research, MoSciBench~\cite{liu2026moscibench} evaluates end-to-end hypothesis verification across six disciplines and seven data modalities via executable analysis scripts, ScienceAgentBench~\cite{chen2025scienceagentbench} scores data-driven discovery scripts and generated scientific plots, and PaperBench~\cite{starace2025paperbench} tests from-scratch replication of ICML Spotlight and Oral papers across 8,316 rubric criteria. Open-ended ML and data-science workflows are further tested by MLR-Bench~\cite{chen2025mlrbench} (201 ideation-to-manuscript tasks), CORE-Bench~\cite{siegel2024corebench} (repository reproducibility), and DSBench~\cite{jing2025dsbench} (competitive data-science modeling). In domain science and industrial engineering, SciFIBench~\cite{roberts2024scifibench} probes scientific figure--caption interpretation, MaCBench~\cite{alampara2025macbench} evaluates multimodal chemistry and materials workflows, DesignQA~\cite{doris2024DesignQA} tests engineering rule compliance, BlenderGym~\cite{gu2025blendergym} and MMArt-Bench~\cite{lin2025mmartbench} evaluate 3D Blender scene editing and instruction-guided photo retouching, MMAD~\cite{jiang2025mmad} and MechVQA~\cite{kou2026mechvqa} cover few-shot industrial anomaly inspection and mechanical drawing QA, and SWE-bench Multimodal~\cite{yang2025swebenchmultimodal} benchmarks visual software issue patching.

\noindent\textbf{Education and Social Cognition (8 records, A6).}
Educational and socio-legal suites evaluate multimodal student essay grading (EssayJudge~\cite{su2025essayjudge}), lecture-video QA (EduVidQA~\cite{ray2025eduvidqa}), children's art critique (ArtMentor~\cite{zheng2025artmentor}), social-media tasks (MM-Soc~\cite{jin2024mm-soc}), empathetic human understanding (HumanSense~\cite{qin2025humansense}), multimodal theory of mind (MMToM-QA~\cite{jin2024mmtomqa}, MuMA-ToM~\cite{shi2025mumatom}), and AI-artwork copyright analysis (AICD~\cite{jia2025aicd}).

\noindent\textbf{Finance and E-Commerce (3 records, A7).}
Financial and commercial benchmarks span chart-intensive financial reasoning in FinMME~\cite{luo2025finmme}, multilingual vision--audio--text finance in MultiFinBen~\cite{peng2026multifinben}, and customer-service workflow execution in ECom-Bench~\cite{wang2025ecombench}, the sole Regime~II benchmark whose primary track scores an environment outcome.

%% file: sec/application/task.tex
\subsection{Orthogonal Track Endpoints, Protocols, and Synthesis}
\label{sec:application_task}

Operational areas (\textbf{A1--A7}) specify the deployment domain, whereas a separate axis specifies the scored object. Table~\ref{tbl:application} crosses every Application record with an orthogonal \textbf{track endpoint code} across all 99 primary tracks:

\begin{itemize}
 \item \textbf{Environment Outcome (E, 35 primary tracks).} Agents act inside a partially observable Markov decision process (POMDP) $E = (S, A, \Omega, T)$ with observations $o_t \in \Omega$, actions $a_t \in A$, and transitions $T: S \times A \to S$. Evaluators inspect post-action system states across OS/web environments (OSWorld~\cite{xie2024osworld}, WorkArena~\cite{drouin2024workarena}, AndroidWorld~\cite{rawles2025androidworld}, WebVoyager~\cite{he2024webvoyager}, WASP~\cite{evtimov2025wasp}) and graph-structured sub-goals (CRAB~\cite{xu2024crab}). Embodied simulators report success rate (SR), oracle success rate (OSR), and success weighted by path length (SPL) for navigation~\cite{cheng2024vlnceisaac,song2025lhprvln}, rollout completion, consecutive-chain length, and hazard mitigation for manipulation~\cite{zhang2025vlabench,chen2026robotwin20,fan2025lcalvin,fang2026liberocf,lu2026isbench,cui2026liberosafety,yang2025embodiedbench}, and sim-to-real rank correlation~\cite{jain2025polaris,quevedo2026worldgym}.

 \item \textbf{Action or Trajectory Prediction (P, 4 primary tracks).} Open-loop protocols score predicted actions or trajectories against logged references without environment feedback: multi-valid desktop action matching in AgentNetBench~\cite{wang2025agentnetbench}, L2 displacement and collision rates in OmniDrive and Impromptu VLA~\cite{wang2025omnidrive,chi2025impromptuvla}, and MAE/RMSE/DTW on manipulation traces~\cite{yuan2026vabench,li2026robointervqa}.

 \item \textbf{Generated Deliverable (D, 7 primary tracks).} Models produce a complete deliverable verified externally: executable discovery scripts, replicated ML codebases, and full research manuscripts in ScienceAgentBench, PaperBench, and MLR-Bench~\cite{chen2025scienceagentbench,starace2025paperbench,chen2025mlrbench}, 3D Minecraft blueprints in MineAnyBuild~\cite{wei2025mineanybuild}, edited Blender scenes in BlenderGym~\cite{gu2025blendergym}, retouched images in MMArt-Bench~\cite{lin2025mmartbench}, and repository patches verified by unit tests in SWE-bench Multimodal~\cite{yang2025swebenchmultimodal}.

 \item \textbf{Target Localization (L, 6 primary tracks).} Pre-action grounding tracks measure geometric accuracy directly: GUI click-in-box hit rates (ScreenSpot, ScreenSpot-Pro~\cite{cheng2024screenspot,li2025screenspotpro}), functional part masks (InstructPart~\cite{wan2025instructpart}), 3D referential disambiguation (IRef-VLA~\cite{zhang2025irefvla}), and affordance points~\cite{yuan2026vabench,li2026robointervqa}.

 \item \textbf{Answer or Assessment (Q, 47 primary tracks).} Nearly half of Application tracks score static textual or multiple-choice answers, including clinical diagnostic QA~\cite{zuo2025medxpertqa,hu2024omnimedvqa,xia2024cares}, corrupted-sensor driving QA~\cite{xie2025drivebench}, 1-shot industrial inspection~\cite{jiang2025mmad}, cognitive-map accuracy~\cite{zhang2026theoryofspace}, hypothesis verification and prediction-interval checks on scientific experiments (MoSciBench, CORE-Bench~\cite{liu2026moscibench,siegel2024corebench}), quadratic weighted kappa on student essays (EssayJudge~\cite{su2025essayjudge}), hallucination-penalized financial QA (FinMME~\cite{luo2025finmme}), and Agent-as-a-Judge web-research verification~\cite{geng2026browsecompvl,gou2025mind2web2}.
\end{itemize}

\textbf{Analytical synthesis.} Crossing operational regime with endpoint exposes both a capability gap and an evaluation gap (\s\ref{sec:protocols}). Horizontally, original benchmark evaluations show substantial headroom once high-resolution interfaces or multi-step execution are enforced: the best baseline models achieved 18.9\% click accuracy on professional desktop layouts in ScreenSpot-Pro~\cite{li2025screenspotpro}, 12.24\% task success on OSWorld~\cite{xie2024osworld} (versus 72.36\% human performance), and 12\% issue resolution on SWE-bench Multimodal~\cite{yang2025swebenchmultimodal}, with PaperBench~\cite{starace2025paperbench} likewise documenting wide agent--human gaps in end-to-end paper replication. Structurally, Regime~I (\textbf{A1--A2}) supplies 34 of all 35 primary \textbf{E} tracks (34/56, 60.7\%), whereas 35 of 43 Regime~II specialist benchmarks (\textbf{A3--A7}, 81.4\%) stop at static \textbf{Q} answers and only 1/43 (2.3\%) verifies an environment state transition---a $26.1$-fold rate gap (descriptive odds ratio $64.9$). Blueprint~2 in \s\ref{sec:patterns} outlines transactional domain sandboxes to close this specialist verification gap.

%% file: sec/dataset_and_eval.tex
\section{Dataset Construction and Quality Control}
\label{sec:holistic_evaluation}

Across all five evaluation pillars, a benchmark's diagnostic ceiling is set before any model is scored: biases, shortcuts, or label errors introduced during item synthesis and filtering propagate directly into leaderboard rankings.

\subsection{Literature Retrieval, Scope Boundaries, and Coding Protocol}\label{sec:corpus_protocol}
To construct the 358-benchmark corpus analyzed in Table~\ref{tbl:understanding} through Table~\ref{tbl:application} and \s\ref{sec:cross}, we queried DBLP, ACL Anthology, OpenReview, IEEE/CVF Open Access, PMLR, SpringerLink, Semantic Scholar, and Google Scholar through November~2026, combining evaluation terms (\texttt{benchmark}, \texttt{evaluation}, \texttt{arena}, \texttt{testbed}, \texttt{suite}) with multimodal keywords (\texttt{multimodal}, \texttt{vision-language}, \texttt{MLLM}, \texttt{VLM}, \texttt{video}, \texttt{audio-visual}, \texttt{3D}, \texttt{embodied}, \texttt{VLA}, \texttt{GUI agent}) and forward/backward citation snowballing from prior surveys~\cite{li2024benchmarkreview,fu2024mmesurvey,zhang2025lmmevalreview,zhu2026agentreview}. Included studies must (i)~appear in peer-reviewed conference proceedings, archival journals, or refereed conference workshops/findings (specifically, two workshop records are included: JUS~\cite{groot2024jus} at ACL TrustNLP and AgentVidBench~\cite{an2026agentvidbench} at the ICML~2026 FAGEN Workshop; unrefereed preprints are excluded), (ii)~have initial public release in 2024--2026 (dated by the earlier of preprint or proceedings publication, as in CDDM~\cite{liu2024cddm}), and (iii)~release a named evaluation benchmark testing MLLMs, unified generators, VLA policies, or multimodal evaluators on at least one non-text modality. Dual-purpose alignment or continual-tuning releases shipping a dedicated test split (VLGuard~\cite{zong2024vlguard}, SPA-VL~\cite{zhang2025spavl}, MM-RLHF~\cite{zhang2025mmrlhf}, CoIN~\cite{chen2024coin}, UCIT~\cite{guo2025ucit}), frontier exams with a dedicated multimodal visual subset (Humanity's Last Exam~\cite{phan2025humanityslastexam}), compositional world/video synthesis benchmarks probing physical or semantic reasoning (ChronoMagic-Bench~\cite{yuan2024chronomagicbench}, T2V-CompBench~\cite{sun2025t2vcompbench}, WorldScore~\cite{duan2025worldscore}), and LMMs-Eval's~\cite{zhang2025lmmseval} live pruned evaluation subset are retained under this rule, whereas model papers with auxiliary test splits are excluded. Because keyword-and-snowball retrieval yields a structured sample rather than an exhaustive PRISMA census, several eligible concurrent benchmarks (\eg, MEGA-Bench~\cite{chen2025megabench}, alongside concurrent video, agent, and search suites such as MMBench-Video, Spider2-V, and TOMATO) fall outside our inventory.

Each included suite is encoded into a standardized \textbf{18-field feature matrix} (released with source links and coding rules in our companion repository): four identifier and taxonomy fields (\texttt{id}, \texttt{domain}, \texttt{track}, \texttt{application\_primary}), three sensory and response fields (\texttt{input\_media}, \texttt{output}, \texttt{output\_channel}), three pipeline multi-tags (\texttt{construction}, \texttt{filter}, \texttt{evaluation} over \texttt{Coll}/\texttt{Human}/\texttt{Rule}/\texttt{LLM}), three protocol fields (\texttt{setting}, \texttt{execution}, \texttt{granularity}), and five bibliometric fields (\texttt{venue}, \texttt{venue\_series}, \texttt{year}, \texttt{public\_record\_date}, \texttt{citations}). Coding follows positive-documentation rules calibrated on a 30-benchmark boundary pilot, recording undocumented attributes as \texttt{NR} rather than negative labels. Automated scripts cross-check bibliographic metadata and overview-table tags across all 358 records, whereas pipeline and protocol coding was performed by a single curator without multi-annotator Cohen's $\kappa$; statistical tests in \s\ref{sec:cross} are therefore presented as structured descriptive and exploratory analyses. Across this matrix, 315 of 358 benchmarks (88.0\%) combine two or more construction mechanisms within hybrid pipelines.

Overall, 318/358 benchmarks (88.8\%) reuse or curate pre-existing media, 230 (64.2\%) incorporate human authoring, 211 (58.9\%) use LLM-assisted generation, and 135 (37.7\%) employ programmatic or rule-based synthesis, whereas only 38 (10.6\%) synthesize or capture all stimuli entirely from scratch (two records leave construction unspecified). On the quality-control side, 273 benchmarks (76.3\%) document human verification, 137 (38.3\%) deterministic rule-based filtering, and 64 (17.9\%) model-in-the-loop screening, while 54 (15.1\%) report no explicit filtering stage. Scoring tags document deterministic rules in 312 records (87.2\%), LLM judges in 135 (37.7\%), and human raters in 20 (5.6\%). Pillar requirements shape these choices: manual authoring peaks in Understanding (70.1\%) and Reasoning (68.5\%), programmatic synthesis peaks in simulator-rich Application suites (47.5\%) and rises across cohorts from 39.2\% in 2024 to 44.3\% in 2026, and LLM-assisted construction peaks in Trustworthiness and Safety (62.5\%) to scale adversarial and counterfactual probing.

\subsection{Dataset Collection}\label{sec:dataset_collection}

Each construction paradigm balances ecological realism, ground-truth exactness, contamination resistance, and annotation cost:

\begin{itemize}
 \item \textbf{Media Reuse vs.\ Bespoke Capture~(88.8\% reuse; 64.2\% human authoring).} Re-annotating public web images, academic figures, or video repositories amortizes acquisition cost, yet risks \textit{pre-training contamination}: newly written questions still fail to test fresh perception when models have memorized the underlying figure, chart, or clip during web-scale pre-training. Domain experts therefore increasingly pair private, high-resolution scene capture or post-cutoff sources with expert-authored queries~\cite{zhang2025mmerealworld,yang2026mmsibench}, reaching nearly 1,000 subject-matter specialists in Humanity's Last Exam~\cite{phan2025humanityslastexam}. Manual authoring scales linearly in expert hours and remains susceptible to human phrasing regularities, such as asymmetric distractor lengths or spatial question templates that vision-blind predictors can exploit~\cite{brown2026vsibenchdebiased}.

 \item \textbf{Automated Rules and Programmatic Synthesis~(37.7\% of corpus; 44.3\% in 2026).} Procedural scripts, physics simulators, and rendering engines instantiate parameterized test cases with exact, noise-free ground truth: VideoNIAH~\cite{zhao2024VideoNIAH} embeds synthetic visual needles at controlled temporal depths, CV-Bench~\cite{tong2024cvbench} and VSI-Bench~\cite{yang2025vsibench} derive 2D/3D geometric queries from scene graphs and indoor scans, DynaMath~\cite{zou2025dynamath} regenerates mathematical diagrams with fresh numerical seeds to defeat memorization, and RoboTwin~2.0~\cite{chen2026robotwin20} perturbs clutter, lighting, and table geometry across bimanual tasks. The main limitation is the \textit{exactness--realism trade-off}: synthetic scenes risk narrow rendering distributions and template regularities unless both visual assets and linguistic templates are held out across splits.

 \item \textbf{LLM- and Diffusion-Assisted Synthesis~(58.9\% of corpus).} Generative models lower the cost of drafting reasoning queries, hard distractors, and controlled visual counterfactuals. CharXiv~\cite{wang2024charxiv} refines GPT-4V reasoning drafts via human annotators, while MMMU-Pro~\cite{yue2025mmmupro} expands multiple-choice options from four to ten using GPT-4o and Claude to suppress random-guess baselines. On the visual side, NegBench~\cite{alhamoud2025negbench}, AutoHallusion~\cite{wu2024autohallusion}, and Hallu-PI~\cite{ding2024hallu} synthesize object-level counterfactuals or perturbations, and MM-NIAH~\cite{wang2024MM-NIAH} generates needle illustrations via DALL-E~3~\cite{betker2023dalle}. Model-assisted synthesis introduces \textit{generator priors and editing artifacts}: inpainted seams, diffusion textures, or LLM-typical distractor phrasing can create cues that penalize out-of-family models or let shortcut detectors bypass semantic understanding.
\end{itemize}

\subsection{Quality Control}\label{sec:quality_control}

To prevent synthesis noise and unimodal shortcuts from corrupting evaluation, modern pipelines layer three filtering mechanisms, each with distinct blind spots:

\begin{itemize}
 \item \textbf{Manual Expert Adjudication~(76.3\% of corpus).} Human reviewers audit factual correctness, visual legibility, answer uniqueness, and cultural or clinical validity~\cite{zhang2025mmerealworld,xu2026visulogic,fu2025ocrbenchv2}, peaking at 87.7\% in Reasoning benchmarks where multi-step derivations require step-by-step verification. Manual review is documented in 85.8\% of LLM-constructed benchmarks versus 62.6\% of non-LLM-constructed ones (\s\ref{sec:crossstats}), often enforcing multi-annotator consensus~\cite{zhang2025mmerealworld,yang2026mmsibench}. The remaining 30/211 LLM-constructed benchmarks (14.2\%) without documented human review risk retaining hallucinated premises or flawed reference answers as gold labels.

 \item \textbf{Deterministic and Embedding-Based Filtering~(38.3\% of corpus).} Automated pipelines enforce structural hygiene, deduplication, and content safety: II-Bench~\cite{liu2024ii} prunes duplicate and text-dominated images via perceptual hashing and OCR ratios, CharXiv~\cite{wang2024charxiv} removes near-duplicate charts using SigLIP embeddings, CLIP similarity margins screen contrastive pairs~\cite{kil2024CompBench,zhao2024VideoNIAH}, and CoMM~\cite{chen2024comm} filters NSFW media via a LAION-2B MLP classifier~\cite{gadre2024datacomp,schuhmann2022laion} (97.4\% accuracy). However, because contrastive vision--language encoders are often insensitive to fine-grained viewpoint, negation, and relational swaps~\cite{tong2024mmvp}, overly strict CLIP or SigLIP similarity thresholds can inadvertently discard informative hard-negative pairs.

 \item \textbf{Model-in-the-Loop Adversarial Screening~(17.9\% of corpus; rising from 14.7\% in 2024 to 20.3\% in 2026).} Frontier models increasingly screen candidate pools for visual necessity and difficulty calibration~\cite{chen2024MMStar}. MMStar~\cite{chen2024MMStar}, MMMU-Pro~\cite{yue2025mmmupro}, and Video-MME~\cite{fu2024Video-MME} execute text-only or frame-blind solvers to purge items solvable from language priors alone, while Humanity's Last Exam~\cite{phan2025humanityslastexam} retains only candidate items that defeat frontier MLLMs prior to final expert sign-off. Adversarial model filtering carries two caveats: (i)~\textit{filter-model selection bias}, whereby discarding items solved by filter model $M_F$ selectively depresses scores for $M_F$ and related model families; and (ii)~\textit{label-noise concentration}, whereby selecting solely on model failure enriches ambiguous or mislabeled items unless paired with expert adjudication. Twenty-eight benchmarks (7.8\%) deploy LLMs across item construction, model-based filtering, and response judging (\s\ref{sec:cross}).
\end{itemize}

%% file: sec/cross_benchmark.tex
\section{Validity of Multimodal Benchmark Comparisons}\label{sec:cross}
\input{sec/misc/comparison_floats}
When a desktop agent saves a file, the final screen confirms that the goal state was reached, yet reveals almost nothing about whether the agent localized the right icon immediately or recovered from repeated misclicks. More broadly, every multimodal benchmark projects a high-dimensional interaction between a model, a sensory stream, and a task protocol onto a low-dimensional leaderboard score. Whenever distinct underlying mechanisms, shortcut strategies, or error-recovery trajectories map to the same headline number, benchmark comparisons suffer from \textit{score equifinality}. Drawing on our five-pillar taxonomy (\s\ref{sec:understanding}--\s\ref{sec:application}) and the standardized 18-field feature matrix over all 358 peer-reviewed benchmarks (\s\ref{sec:holistic_evaluation}), we analyze comparison validity across four coupled levels: construct identifiability (\s\ref{sec:evolution}), protocol causality and aggregation stability (\s\ref{sec:protocols}), cross-stage pipeline coupling (\s\ref{sec:crossstats}), and ecosystem-level selection incentives (\s\ref{sec:patterns}).

Whereas prior reviews catalog evaluation suites by target capabilities, task formats, or software toolkits~\cite{li2024benchmarkreview,fu2024mmesurvey,zhang2025lmmevalreview}, we partition all 358 peer-reviewed 2024--2026 benchmarks (102 in 2024, 177 in 2025, and 79 in partial 2026 through November~2026; each suite counts once, dated by earliest public release) under the mutually exclusive primary-pillar rules in \s\ref{sec:intro} across Understanding (107), Reasoning (73), Generation (31), Trustworthiness and Safety (48), and Application (99) (Table~\ref{tab:cohort-counts}). Sensitivity checks reassigning multi-capability boundary records---such as shifting CoMM~\cite{chen2024comm} between Generation and Understanding, grouping clinical reliability suites in \textbf{A3} with Trustworthiness, or moving offline agent-safety and direct code/poster synthesis suites (MMA-SafetyBench~\cite{wang2026mmasafetybench}, AgentSafe~\cite{ying2026safe}, OmniGIRL~\cite{guo2025omnigirl}, BlenderBench~\cite{yin2026blenderbench}, Paper2Poster~\cite{pang2025paper2poster}) into Application ($\mathrm{OR}_{\mathrm{MH}}=1.39$, $p=0.16$)---leave all cohort trends, protocol gaps, and coupling signs unchanged. Because tags record explicitly documented procedures (with arrival setting, execution loop, and scoring granularity left unspecified in 33, 255, and 210 records where static QA papers omit execution-loop details), reported protocol shares represent positive lower bounds.
\CohortCountsTable

\subsection{Construct Validity and Causal Identifiability}\label{sec:evolution}
\FamilyRevisionsFigure
\FamilyRevisionsTable
\DesignControlsTable
Construct validity asks whether a score difference isolates the target multimodal capability against simpler confounding explanations. Across the nine predecessor--successor benchmark families in Table~\ref{tab:evolution} and Fig.~\ref{fig:familypaths}(a), revisions intervene on three axes: expanding domain and interface diversity (OCRBench, Q-Bench, ChartQA, ScreenSpot), blocking unimodal priors or policy memorization (MMMU, VSI-Bench, LIBERO), and perturbing live interaction dynamics (Full-Duplex-Bench, RoboTwin). Lowering saturated score ceilings and isolating specific failure modes require different experimental designs.

\textbf{Insight 1. Multi-factor benchmark upgrades lower score ceilings but confound error attribution, whereas single-factor counterfactuals on matched scenes achieve causal identifiability.} Relative to its 90.50\% vendor-reported score on ChartQA~\cite{masry2022chartqa}, Claude~3.5 Sonnet reaches 55.81\% under CoT ($-34.69$ percentage points, pp) and 43.58\% under direct prompting ($-46.92$ pp) on ChartQAPro~\cite{masry2025chartqapro}. Beyond prompt-protocol differences across reports, ChartQAPro shifts chart provenance, multi-panel layouts, reasoning templates, and unanswerable abstention items simultaneously, so the aggregate drop cannot separate visual parsing limits from miscalibrated refusal. By contrast, LIBERO-CF holds the physical scene layout fixed while altering only the language goal: on matched rollouts, the $\pi_{0.5}$ policy achieves 13.2\% success on the requested counterfactual instruction while executing the memorized training trajectory in 60.9\% of trials~\cite{fang2026liberocf}, cleanly isolating goal-blind action memorization from motor incompetence (Fig.~\ref{fig:familypaths}(b)).

Because unimodal and distributional shortcuts arise at multiple levels, different debiasing designs intercept non-overlapping confounds (Table~\ref{tab:patterns}). Zero-shot blind-LLM filtering in MMMU-Pro~\cite{yue2025mmmupro} removes items solvable from pre-trained world knowledge, whereas VSI-Bench-Debiased fits a vision-blind predictor on held-out folds to expose dataset-specific question regularities that zero-shot prompting misses~\cite{brown2026vsibenchdebiased}. Minimal contrast pairs in MMVP and NaturalBench target lucky guessing on binary or multiple-choice items~\cite{tong2024mmvp,li2024naturalbench}: at a 50\% single-item accuracy rate, answering only one member of every pair yields 0\% pair consistency, whereas grounded perception on half the pairs yields 50\%. Construct hygiene also requires separating \textit{generator-side reliability} (\textbf{T1--T3}, 38 benchmarks) from \textit{evaluator-side diagnostics} (\textbf{T4}, 10 benchmarks) in \s\ref{sec:trustworthiness}: measuring whether a model hallucinates during open-ended generation (\textbf{T1}, 18 suites, where terse responses can inflate precision unless paired with recall~\cite{kaul2024throne}) probes a distinct capability from auditing whether a model detects span-level errors or ranks candidate preferences without bias (\textbf{T4}~\cite{chen2024unified,mickus2026sheep,li2025vlrewardbench,lee2026mmjudgebias}). A parallel boundary separates static hazard classification from online hazard avoidance during physical execution~\cite{wang2025mmsafeaware,lu2026isbench}.

\subsection{Protocol Causality, Sandbox Bottlenecks, and Rank Stability}\label{sec:protocols}
\ProtocolsTable
\ProtocolCoverageFigure
Even when two benchmarks probe related domains, observation access and credit assignment alter what a score can substantiate (Table~\ref{tab:protocols}). OpenEQA evaluates identical household questions under recorded video versus active exploration~\cite{OpenEQA2023}, varying sensory control while keeping the scored object an answer (\textbf{Q}). OSWorld instead grades post-action desktop states (\textbf{E})~\cite{xie2024osworld}, granting identical task credit to an immediate one-step solution and a multi-step recovery trajectory unless first-action errors are logged separately.

\textbf{Offline hindsight and the streaming deficit.} Across all 358 benchmarks, still images appear in 243 suites (67.9\%), video in 73 (20.4\%), audio in 27 (7.5\%), and native 3D scenes in six (1.7\%; Fig.~\ref{fig:corpuscoverage}(a)). Comparing pure synthesis (\textbf{G1--G3}) alongside trustworthiness (\textbf{T1--T4}) highlights a rapid rise in synthesis suites: whereas 2024 releases favored trustworthiness over synthesis by 18 to 5, pure Generation more than tripled to 17 benchmarks in 2025 (and 9 in partial 2026) alongside visual-to-code rendering, controllable editing, and duplex speech (Table~\ref{tab:cohort-counts}). Over the same venue-year cohorts, documented static fixed-input protocols decline from 93/102 (91.2\%) in 2024 to 137/177 (77.4\%) in 2025 and 53/79 (67.1\%) in 2026, while closed-loop execution rises from 6/102 (5.9\%) and 14/177 (7.9\%) to 13/79 (16.5\%; Fig.~\ref{fig:corpuscoverage}(b)). Real-time causal constraints remain scarce: among 73 video benchmarks, 66 grant offline full-clip access and only four enforce streaming decision deadlines~\cite{li2025ovobench,zeng2025odvbench,xun2025rtv,xu2026infstreamseval}; among 27 audio benchmarks, 24 use offline clips and only two test live streaming turn-taking~\cite{lin2026fullduplexbenchv15,lin2026fullduplexbenchv2} (Fig.~\ref{fig:corpuscoverage}(c)). Offline video QA permits \textit{future-frame leakage}: when a query is nominally due at frame $t$, supplying frame $t+1$ exposes outcomes that have not yet occurred (Fig.~\ref{fig:familypaths}(c)), conflating retrospective summarization with causal anticipation.

\ApplicationOutputsFigure
\PipelineFigure
\textbf{Insight 2. Despite rapid Application growth, outcome-verified evaluation hits a resettable-sandbox bottleneck outside digital and embodied simulators.} Application benchmarks expand from 19/102 (18.6\%) in 2024 to 54/177 (30.5\%) in 2025 and 26/79 (32.9\%) in 2026 by venue year, or from 40/195 (20.5\%) to 46/133 (34.6\%) and 13/30 (43.3\%) by first public release year (Fig.~\ref{fig:application-outputs}(b)). Across all 99 Application suites, 47 score static answers (\textbf{Q}), 35 environment states (\textbf{E}), seven generated deliverables (\textbf{D}, bolstered by executable codebases and discovery scripts in auto-research~\cite{chen2025scienceagentbench,starace2025paperbench,chen2025mlrbench}), six spatial targets (\textbf{L}), and four open-loop action predictions (\textbf{P}). Outcome verification divides sharply between regimes: Regime~I (\textbf{A1--A2}) accounts for 34 of all 35 primary environment-outcome tracks in Application. Within Regime~I, 34/56 benchmarks (60.7\%) verify environment states (\textbf{E}), compared with 1/43 (2.3\%) across the five specialist verticals in Regime~II (\textbf{A3--A7})---a $26.1$-fold gap in \textbf{E}-track share (descriptive odds ratio $64.9$; Fig.~\ref{fig:application-outputs}(a)), with 35/43 Regime~II suites (81.4\%) grading static answers alone. This $26$-fold rate gap reflects sandbox availability: operating systems, web browsers, and robotics simulators provide deterministic state snapshots, zero-cost rollbacks, and executable state predicates, whereas medicine, finance, Earth observation, and wet-lab sciences lack standardized transactional sandboxes (aside from customer-service state checks in ECom-Bench~\cite{wang2025ecombench} and containerized code execution in computational auto-research~\cite{liu2026moscibench,starace2025paperbench}). Equating specialist exam accuracy (\textbf{Q}) with deployment readiness (\textbf{E}) conceals irreversible side effects, tool-invocation errors, and multi-step state corruption.

\textbf{Insight 3. Scalar leaderboard means mask non-uniform model-by-protocol interactions and violate rank invariance under minor prompt, metric, or weight shifts.} As a fine-grained empirical audit across 21 models evaluated on ChartQAPro under direct versus chain-of-thought (CoT) prompting, the cohort median score shift is near zero ($-0.52$ pp), tempting practitioners to treat prompting choice as benign. Heterogeneous model-by-prompt gains reverse pairwise ordering in 27/210 model pairs (12.9\%), including 16/179 pairs (8.9\%) separated by more than 2 pp under both settings~\cite{masry2025chartqapro}: Gemini Flash~2.0 outperforms Claude~3.5 Sonnet under direct prompting (46.85\% vs.\ 43.58\%), whereas CoT reverses the lead (53.66\% vs.\ 55.81\%). Aggregate scores exhibit parallel sensitivity to subtask weights (Fig.~\ref{fig:application-outputs}(c)): reallocating 9.4 pp of weight from fact-checking to hypothetical questions under direct prompting lets Gemini Flash~1.5 tie Flash~2.0 (26.2 pp under CoT). Similar rank shifts appear across other pillars. In Understanding, moving from standard text options to vision-embedded screenshots in MMMU-Pro~\cite{yue2025mmmupro} produces non-uniform model drops of 16.8--26.9 pp, and replacing marginal item accuracy with paired consistency in NaturalBench~\cite{li2024naturalbench} causes 40--55 pp drops that reorder models with comparable single-item scores. In Reasoning and Agentic suites, MME-CoT~\cite{jiang2025mmecot} shows CoT hurting perception-heavy subsets while boosting math subsets, and enforcing online causal cutoffs in OVO-Bench~\cite{li2025ovobench} or varying step budgets (15 vs.\ 50 steps) in OSWorld~\cite{xie2024osworld} likewise shifts model rankings. Narrow scalar margins without prompt-, horizon-, and stratum-sensitivity bounds cannot establish general dominance.

\subsection{The Scoring Trilemma and Cross-Stage Pipeline Coupling}\label{sec:crossstats}
Why has LLM-as-a-Judge proliferated to 135/358 benchmarks (37.7\%) despite known biases? Cross-tabulating output formats against scoring mechanisms exposes a three-way tension among \textit{open-ended expressiveness}, \textit{deterministic verifiability}, and \textit{low-friction harness scalability}. Pure multiple-choice and binary benchmarks ($n=78$) achieve 97.4\% (76/78) deterministic rule-based scoring (invoking an LLM judge in only 2.6\%) and trivial deployment, sacrificing open-ended expressiveness to closed-option elimination shortcuts. Conversely, unconstrained free-form text ($n=21$) and generated visual/audio/interleaved media ($n=18$) preserve expressiveness and low engineering overhead, but sacrifice deterministic verifiability as LLM-judge reliance rises to 85.7\% (18/21) and 83.3\% (15/18). Closed-loop environment execution ($n=33$) secures both open-ended action freedom and 90.9\% (30/33) deterministic rule-based state verification (with 27.3\% pairing rules with auxiliary LLM judges), but pays the third cost: bespoke simulator and container harnesses that constrain benchmark supply (\s\ref{sec:patterns}).

\textbf{Insight 4. Within-pillar LLM-judge rates are consistently higher in LLM-constructed benchmarks, though stratified coupling is not statistically significant; the 85 dual-LLM suites still warrant self-preference auditing.} Across the corpus, LLM judging appears in 85/211 benchmarks built with LLM assistance versus 50/147 without it (40.3\% vs.\ 34.0\%, $+6.3$ pp pooled; $+7.2$ pp excluding two unspecified records). Decomposing the $+6.3$ pp pooled gap separates $+7.6$ pp of within-pillar difference from $-1.3$ pp of pillar-composition offset. Pure Generation exhibits the highest judge reliance (23/31, 74.2\%; 80.0\% with LLM construction vs.\ 68.8\% without, $+11.2$ pp), with positive descriptive gaps across Trustworthiness ($+17.8$ pp; 17/30 vs.\ 7/18), Understanding ($+15.0$ pp; 26/64 vs.\ 11/43), and Application ($+4.7$ pp; 19/59 vs.\ 11/40; Fig.~\ref{fig:pipeline-association}(a)). Reasoning's $-7.8$ pp reversal (11/43 vs.\ 10/30) reflects track composition: LLM judging concentrates in outcome-based reasoning (\textbf{R1}, 17/50) and CoT process evaluation (\textbf{R3}, 3/3), whereas visually grounded reasoning (\textbf{R2}, 0/9) uses geometric IoU/distance metrics and 10 of 11 reflective critique and editing suites (\textbf{R4}, 1/11) pair LLM item synthesis with deterministic verification; excluding \textbf{R4} narrows the Reasoning gap to $-2.1$ pp (10/32 vs.\ 10/30; Fig.~\ref{fig:pipeline-association}(b)). Stratified across all five pillars via Cochran--Mantel--Haenszel (CMH) analysis, the pooled association is positive but does not reach statistical significance ($\mathrm{OR}_{\mathrm{MH}}=1.41$, 95\% CI $0.90\text{--}2.23$, $p=0.14$; in a post-hoc exploratory check excluding deterministically rule-defined \textbf{R2}/\textbf{R4} tracks, $n=338$, $\mathrm{OR}_{\mathrm{MH}}=1.56$, 95\% CI $0.98\text{--}2.49$, $p=0.062$).

Even without a statistically significant odds ratio, the absolute prevalence of dual-LLM pipelines (85/358 benchmarks, 23.7\% of the corpus; Fig.~\ref{fig:pipeline-association}(c)) creates concrete evaluation exposure alongside a documentation caveat. Benchmark authors document human verification more often in LLM-constructed suites than in non-LLM ones (181/211, 85.8\% vs.\ 92/147, 62.6\%; $\mathrm{OR}_{\mathrm{MH}}=3.87$, 95\% CI $2.28\text{--}6.55$, $\chi^2_{\mathrm{CMH}}=26.72$, $p=2.3\times 10^{-7}$), partly reflecting \textit{documentation reporting bias}: 2025--2026 conference checklists and reviewer scrutiny pressure LLM-synthesized suites to disclose human sign-off more explicitly than manually curated benchmarks. Moreover, human screening checks \textit{item validity during construction} rather than \textit{evaluator neutrality at test time}. Among the 85 dual-LLM suites, \textit{same-family coupling} (\eg, GPT-4/GPT-4o drafting items and grading responses) risks direct family-specific self-preference, whereas \textit{cross-family pipelines} (\eg, Claude/Gemini drafting paired with a GPT-4o judge) mitigate single-vendor affinity while sharing broader LLM verbosity and formatting priors~\cite{lee2026mmjudgebias}. Because many papers omit exact model snapshots across stages, independent multi-family judge auditing remains necessary.

\subsection{Ecosystem Incentives, Citation Attention, and Design Blueprints}\label{sec:patterns}
\VenueCoverageFigure
\BenchmarkSelectionTable
\textbf{Insight 5. Interactive and process-scored benchmarks attract roughly double the monthly citation rate yet represent only 13.4\% of corpus supply, leaving static leaderboards dominant in volume.} Across 285 records in seven major conference series (with the remaining 73 spanning 32 peer-reviewed NLP, AI, multimedia, robotics, speech, and journal venues including EMNLP, AAAI, NAACL, ACM~MM, COLM, RSS/CoRL/ICRA, and \textit{TMLR}/\textit{TPAMI}), disciplinary norms partition what gets measured: computer vision venues (CVPR, ICCV, ECCV) publish 6 of the 9 visually grounded Reasoning suites (\textbf{R2}) but document closed-loop execution or process scoring in just 5/87 papers (5.7\%), whereas machine learning venues (NeurIPS, ICLR, ICML) contribute 13 of the 18 embodied outcome benchmarks (\textbf{A2}-\textbf{E}) and 23/150 closed-loop or process-scored suites (15.3\%, reaching 7/18 or 38.9\% at ICML; Fig.~\ref{fig:author-venues}(a)). Semantic Scholar citation counts (merged across preprint and proceedings records on November~18, 2026, with a three-month minimum age floor for monthly rates) show steep concentration: three conference editions (NeurIPS~2024, CVPR~2025, ACL~2024; 93/285 records, 32.6\%) capture 47.0\% of conference citations (22,015/46,873; Fig.~\ref{fig:author-venues}(b)), and the top 10\% of benchmarks (36/358) absorb 47.0\% of all corpus citations (25,175/53,596). All 310 static item-level benchmarks together account for 81.1\% of corpus citations (43,458/53,596), and the 12 static suites inside the top 15 alone capture 28.2\% of all static citations (12,258/43,458; 79.6\% of top-15 citations).

Age-normalized citation velocity contrasts per-benchmark attention against supply volume: the $n_1=48$ dynamic benchmarks incorporating streaming deadlines, closed-loop execution, or step-level process scoring (13.4\% of supply) achieve a median of 6.02 citations/month (IQR $3.86\text{--}12.36$) versus 3.04 (IQR $1.39\text{--}6.47$) across the $n_2=310$ static benchmarks ($1.98\times$; two-sided Mann--Whitney $U=10{,}237.5$, $z=4.19$, rank-biserial $r_{\mathrm{rb}}=0.38$, $p=2.8\times 10^{-5}$). Across five subgroup comparisons with Holm--Bonferroni multiplicity adjustment ($p_{\mathrm{Holm}}$), dynamic suites lead within 2024 ($n=6$ vs.\ $96$; $16.61$ vs.\ $3.37$, $4.94\times$, $p=0.0043$, $p_{\mathrm{Holm}}=0.013$), 2025 ($n=21$ vs.\ $156$; $7.07$ vs.\ $3.01$, $2.35\times$, $p=0.0023$, $p_{\mathrm{Holm}}=0.0092$), partial 2026 ($n=21$ vs.\ $58$; $4.77$ vs.\ $2.87$, $1.66\times$, $p=0.030$, $p_{\mathrm{Holm}}=0.060$), and NeurIPS/ICLR/ICML ($n=26$ vs.\ $124$; $8.08$ vs.\ $3.78$, $2.14\times$, $p=0.0004$, $p_{\mathrm{Holm}}=0.0020$; vs.\ $4.88$ vs.\ $3.06$, $p=0.076$, $p_{\mathrm{Holm}}=0.076$ in CVPR/ICCV/ECCV/ACL, $n=13$ vs.\ $122$). Age-floor sensitivity checks confirm that the three-month minimum denominator does not manufacture this gap: a six-month floor yields $2.00\times$ ($p=3.4\times 10^{-5}$), unfloored age yields $1.98\times$ ($p=2.5\times 10^{-5}$), and excluding suites younger than six months ($n=46$ vs.\ $303$) widens the ratio to $2.15\times$ ($6.59$ vs.\ $3.07$, $r_{\mathrm{rb}}=0.40$, $p=1.5\times 10^{-5}$). In an OLS regression of $\log(1+\text{citations/month})$ controlling for cohort year, ML-conference venue, Application-pillar membership, public GitHub repository release, and non-image modality (video/audio/3D), dynamic protocols retain an associational $1.76\times$ multiplier (95\% CI $1.35\text{--}2.31$, $t=4.13$, $p=3.6\times 10^{-5}$). Because citation velocity reflects scholarly interest and can co-vary with unobserved factors such as institutional visibility or agent-topic momentum, it does not directly measure model post-training targets. Nevertheless, packaging Dockerized OS images, physics simulators, and streaming harnesses imposes clear supply-side engineering costs, leaving 86.6\% of available benchmarks static and low-friction to run. Model evaluations should therefore pair an endpoint-matched interactive test (\eg, OSWorld~\cite{xie2024osworld}, \textbf{E}) with a controlled component diagnostic (\eg, ScreenSpot-Pro~\cite{li2025screenspotpro}, \textbf{L}; Table~\ref{tab:selection}).

\textbf{Insight 6. Remedying the corpus's three main validity deficits requires matching instrumentation to each measurement bottleneck.} Only 4/73 video benchmarks enforce causal streaming deadlines, only 1/43 specialist Application benchmarks verifies downstream environment state changes, and 85/358 benchmarks couple LLM construction with LLM judging. Addressing the first two gaps requires new protocol harnesses, whereas the third can be audited directly on existing model outputs:

\textbf{Blueprint 1. Causal boundary probing on diverging video streams.} Construct paired video streams sharing identical prefixes before diverging into opposite outcomes, with independent annotators marking the earliest outcome-resolving frame interval $[t_0, t_1]$. Evaluate models under matched frame budgets in both streaming and offline modes, decomposing performance into premature-guessing rate prior to $t_0$, causal accuracy within $[t_0, t_1]$, and detection latency after $t_1$.

\textbf{Blueprint 2. Transactional domain sandboxes for specialist applications.} Replace static specialist QA with containerized, snapshot-resettable domain artifacts (\eg, an executable financial workbook, clinical order sandbox, or GIS layer editor). Require models first to predict the analytical impact (\textbf{Q}) and then to execute the state update (\textbf{E}), grading deterministic post-edit invariants under held-out perturbations alongside zero-collateral-damage checks on unrelated fields.

\textbf{Blueprint 3. Cross-family blind auditing of coupled LLM judges.} Sample visually grounded responses from the 85 dual-LLM benchmarks and construct controlled triplets: the original response, a human-verified meaning-preserving paraphrase, and a fluent single-fact visual corruption. Blind candidate judges to model provenance and rescore across distinct model families to quantify style/verbosity bias, corruption-detection recall, and judge-induced rank reversals.

\textbf{Pre-release validity checklist.} To reduce score ambiguity before leaderboard deployment, benchmark authors should document six items per scored track:
\begin{enumerate}
\item \textbf{Construct and shortcut baseline.} State the target capability, scored object, plausible unimodal or memorization shortcut, and empirical blind-solver or counterfactual floor.
\item \textbf{Causal evidence boundary.} Log accessible frames, sensory channels, external tools, and latency or action budgets permitted when each decision is emitted.
\item \textbf{Credit assignment and trajectory decomposition.} Define terminal success, partial credit, and error-recovery rules, separating first-action accuracy from post-error recovery.
\item \textbf{Cross-stage pipeline independence.} Disclose media provenance, authoring/filtering/judging model versions, and human--judge agreement stratified by model family.
\item \textbf{Rank-invariance sensitivity.} Report paired model rank stability across prompt templates, subtask-weight perturbations, and alternative judge families.
\item \textbf{Resettable reproducibility artifacts.} Release dataset and simulator hashes, track denominators, deterministic scoring scripts, rendering parameters, and environment-reset checks.
\end{enumerate}

%% file: sec/misc/comparison_floats.tex
\newcommand{\FamilyRevisionsTable}{%
\begin{table*}[tp]\centering\normalcolor
\caption{Comparison of nine representative predecessor--successor benchmark families across revision axes, design shifts, and residual confounds.}\label{tab:evolution}
\begingroup\footnotesize\linespread{1.02}\selectfont\renewcommand{\arraystretch}{1.14}\setlength{\tabcolsep}{4.2pt}
\begin{tabularx}{\textwidth}{@{}>{\hsize=1.02\hsize\raggedright\arraybackslash}Xc>{\hsize=0.72\hsize\raggedright\arraybackslash}X>{\hsize=1.38\hsize\raggedright\arraybackslash}Xc>{\hsize=0.88\hsize\raggedright\arraybackslash}X@{}}\hline\thickhline
\rowcolor{mygray}\textbf{Benchmark Evolution} & \textbf{Pillar} & \textbf{Revision Axis} & \textbf{Core Design Modification} & \textbf{Factor Shift} & \textbf{Residual Confound / Scope}\\\hline
OCRBench~\cite{liu2024ocrbench} $\rightarrow$ v2~\cite{fu2025ocrbenchv2} & \textbf{U1} & \multirow{4}{*}{Coverage expansion} & 10,000 human-verified QA pairs covering 23 tasks and 31 scenarios & Multi-factor & Subtask-weight sensitivity\\
Q-Bench~\cite{wu23q-bench} $\rightarrow$ Q-Bench+~\cite{zhang2024q-bench} & \textbf{U1} & & 1,999 image pairs for relative quality assessment & Multi-factor & Pair presentation-order bias\\
ChartQA~\cite{masry2022chartqa} $\rightarrow$ Pro~\cite{masry2025chartqapro} & \textbf{R1} & & Multi-panel charts, infographics \& abstention & Multi-factor & Joint layout \& refusal shift\\
ScreenSpot~\cite{cheng2024screenspot} $\rightarrow$ Pro~\cite{li2025screenspotpro} & \textbf{A1} & & 4K desktop UIs across 23 professional apps & Multi-factor & Single-step localization only\\\cline{1-6}
MMMU~\cite{yue2024mmmu} $\rightarrow$ MMMU-Pro~\cite{yue2025mmmupro} & \textbf{U1} & \multirow{3}{*}{Shortcut removal} & 10-option MC \& image-embedded text queries & Multi-factor & Joint option \& OCR shift\\
VSI-Bench~\cite{yang2025vsibench} $\rightarrow$ Debiased~\cite{brown2026vsibenchdebiased} & \textbf{U3} & & Held-out blind-solver question filtering & Single-factor & Blind-solver capacity bound\\
LIBERO~\cite{liu2023libero} $\rightarrow$ LIBERO-CF~\cite{fang2026liberocf} & \textbf{A2} & & Counterfactual instructions on matched scenes & Single-factor & Simulated tabletop scope\\\cline{1-6}
Full-Duplex v1.5~\cite{lin2026fullduplexbenchv15} $\rightarrow$ v2~\cite{lin2026fullduplexbenchv2} & \textbf{G3} & \multirow{2}{*}{Trial dynamics} & Examiner-paced multi-turn speech workflows & Multi-factor & Examiner-pacing coupling\\
RoboTwin~\cite{mu2025robotwin} $\rightarrow$ 2.0~\cite{chen2026robotwin20} & \textbf{A2} & & 5-axis domain randomization across 50 tasks & Factorial & Sim-to-real transfer gap\\\hline
\end{tabularx}\endgroup
\end{table*}%
}

\newcommand{\FamilyRevisionsFigure}{%
\begin{figure*}[tp]\centering\normalcolor
\includegraphics[width=\textwidth]{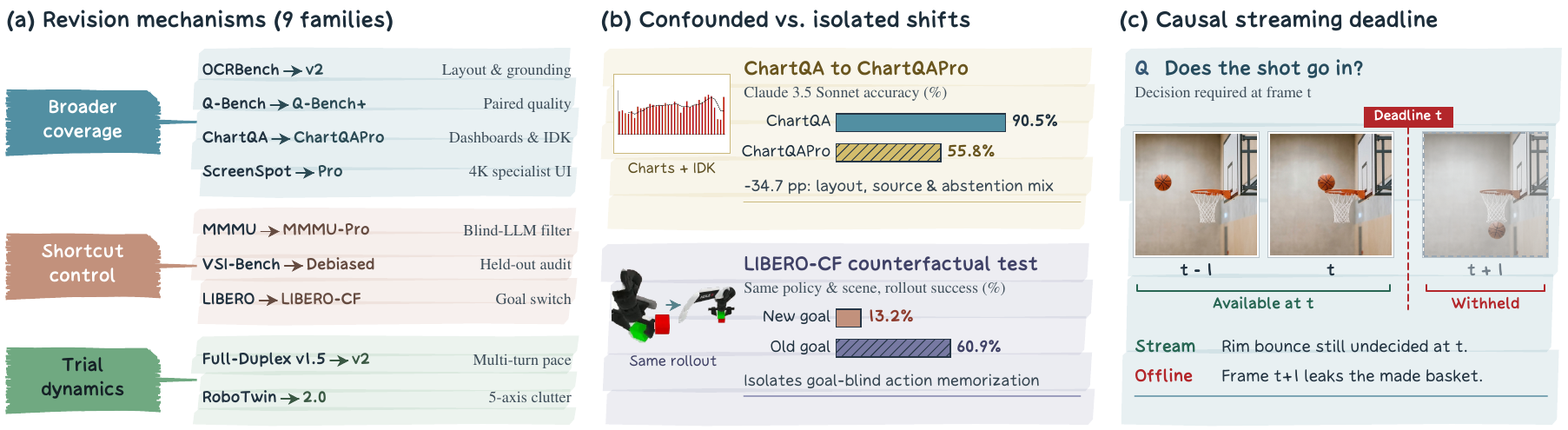}
\caption{Benchmark revision mechanisms and decision-time evidence boundaries. (a) Nine predecessor--successor families grouped by primary revision axis. (b) Multi-factor score drop from ChartQA to ChartQAPro versus single-factor counterfactual goal switching in LIBERO-CF~\cite{masry2025chartqapro,fang2026liberocf}. (c) Causal frame-access boundary at streaming deadline $t$ versus offline retrospective access to future frame $t+1$.}\label{fig:familypaths}
\end{figure*}%
}

\newcommand{\ProtocolsTable}{%
\begin{table*}[tp]\centering\normalcolor
\caption{Latent score ambiguities (equifinality) across multimodal evaluation protocols and corresponding diagnostic controls.}\label{tab:protocols}
\begingroup\footnotesize\linespread{1.02}\selectfont\renewcommand{\arraystretch}{1.14}\setlength{\tabcolsep}{4.0pt}
\begin{tabularx}{\textwidth}{@{}>{\hsize=0.77\hsize\raggedright\arraybackslash}X>{\hsize=0.79\hsize\raggedright\arraybackslash}X>{\hsize=1.03\hsize\raggedright\arraybackslash}X>{\hsize=1.03\hsize\raggedright\arraybackslash}X>{\hsize=1.38\hsize\raggedright\arraybackslash}X@{}}\hline\thickhline
\rowcolor{mygray}\textbf{Protocol} & \textbf{Standard Metric} & \textbf{Masked Failure Mode} & \textbf{Disaggregated Control} & \textbf{Representative Suites}\\\hline
Streaming video & Full-clip accuracy & Acausal future-frame leakage & Causal deadline accuracy \& delay & OVO-Bench~\cite{li2025ovobench}, ODV-Bench~\cite{zeng2025odvbench}\\
Grounded reasoning & Final QA accuracy & Correct guess from wrong region & Joint answer--IoU consistency & Visual CoT~\cite{shao2024VisualCoT}, TreeBench~\cite{wang2026treebench}\\
Media synthesis & Visual quality score & Causal or factual rule violation & Semantic vs.\ perceptual split & WISE~\cite{niu2026wise}, RISEBench~\cite{zhao2025risebench}\\
Active exploration & Episode QA accuracy & Navigation vs.\ VQA conflation & Visibility-conditioned replay & OpenEQA~\cite{OpenEQA2023}\\
Closed-loop policy & Terminal state pass & Trial-and-error recovery loops & First-step \& recovery rate split & OSWorld~\cite{xie2024osworld}, EmbodiedBench~\cite{yang2025embodiedbench}\\
Full-duplex speech & Transcript QA score & Barge-in \& turn-taking delay & Acoustic timing \& recall split & Full-Duplex v1.5~\cite{lin2026fullduplexbenchv15}, v2~\cite{lin2026fullduplexbenchv2}\\\hline
\end{tabularx}\endgroup
\end{table*}%
}

\newcommand{\ProtocolCoverageFigure}{%
\begin{figure*}[tp]\centering\normalcolor
\includegraphics[width=\textwidth]{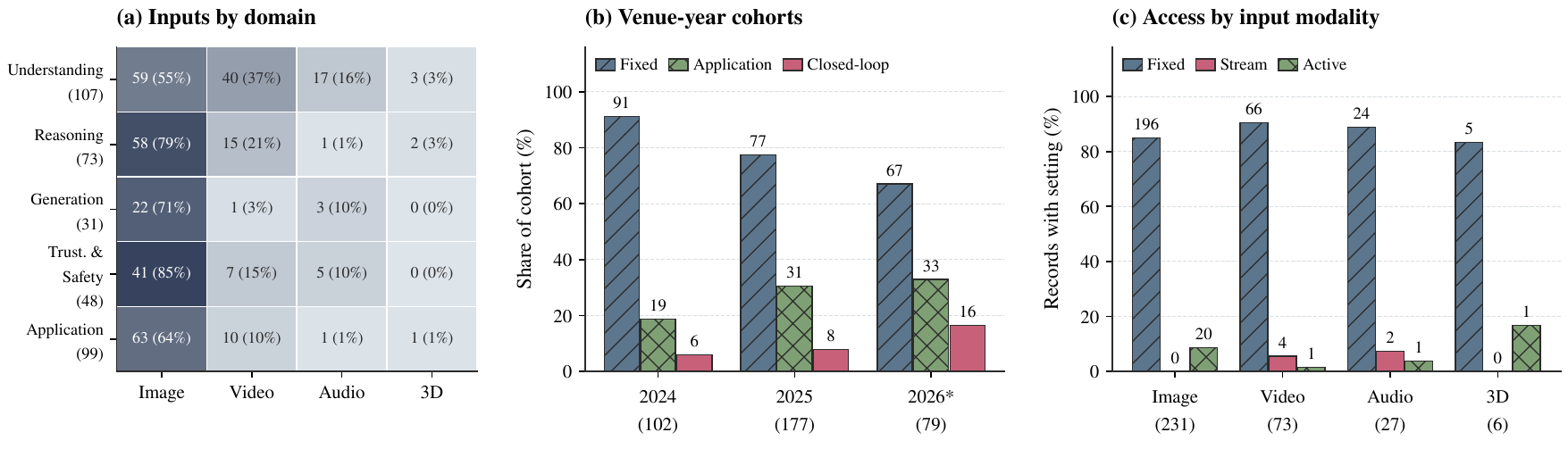}
\caption{Input modality coverage and documented evaluation protocols across the corpus ($N=358$). (a) Modality counts and shares within each evaluation pillar. (b) Protocol shares across venue-year cohorts ($2026^*$ through November~2026). (c) Protocol counts by non-text input modality, where Image ($n=231$) denotes still-image-only suites excluding 12 benchmarks that also include video, audio, or 3D inputs; protocol tags are non-exclusive and reflect explicitly documented settings.}\label{fig:corpuscoverage}
\end{figure*}%
}

\newcommand{\DesignControlsTable}{%
\begin{table*}[tp]\centering\normalcolor
\caption{Experimental control strategies for mitigating unimodal shortcuts and evaluation confounds.}\label{tab:patterns}
\begingroup\footnotesize\linespread{1.02}\selectfont\renewcommand{\arraystretch}{1.14}\setlength{\tabcolsep}{4.0pt}
\begin{tabularx}{\textwidth}{@{}>{\hsize=0.86\hsize\raggedright\arraybackslash}X>{\hsize=0.79\hsize\raggedright\arraybackslash}X>{\hsize=1.02\hsize\raggedright\arraybackslash}X>{\hsize=0.85\hsize\raggedright\arraybackslash}X>{\hsize=1.48\hsize\raggedright\arraybackslash}X@{}}\hline\thickhline
\rowcolor{mygray}\textbf{Control Strategy} & \textbf{Target Confound} & \textbf{Core Mechanism} & \textbf{Residual Risk / Cost} & \textbf{Representative Suites}\\\hline
Minimal contrast pairs & Random-guess luck & Paired minimal-attribute swap & Unintended edit artifacts & MMVP~\cite{tong2024mmvp}, NaturalBench~\cite{li2024naturalbench}\\
Counterfactual editing & Scene memorization & Context-fixed object/goal edit & Inpainting / physics seams & AutoHallusion~\cite{wu2024autohallusion}, LIBERO-CF~\cite{fang2026liberocf}\\
Blind-modality ablation & Language-only priors & Vision-stripped solver baseline & Solver-family dependence & MMStar~\cite{chen2024MMStar}, MMMU-Pro~\cite{yue2025mmmupro}\\
Step \& region grounding & Unfaithful reasoning & Intermediate box/step check & Alternative valid paths & Visual CoT~\cite{shao2024VisualCoT}, TreeBench~\cite{wang2026treebench}\\
Programmatic synthesis & Static item leakage & Symbolic template instantiation & Structural template bias & CV-Bench~\cite{tong2024cvbench}, DynaMath~\cite{zou2025dynamath}\\
Render-and-compare & Code surface mismatch & Executable output visual diff & Compiler \& font drift & ChartMimic~\cite{yang2025chartmimic}, Design2Code~\cite{si2025design2code}\\
State-predicate check & Trajectory rigidity & Post-action OS/sim invariants & Unchecked side effects & OSWorld~\cite{xie2024osworld}, AndroidWorld~\cite{rawles2025androidworld}\\\hline
\end{tabularx}\endgroup
\end{table*}%
}

\newcommand{\ApplicationOutputsFigure}{%
\begin{figure*}[tp]\centering\normalcolor
\includegraphics[width=\textwidth]{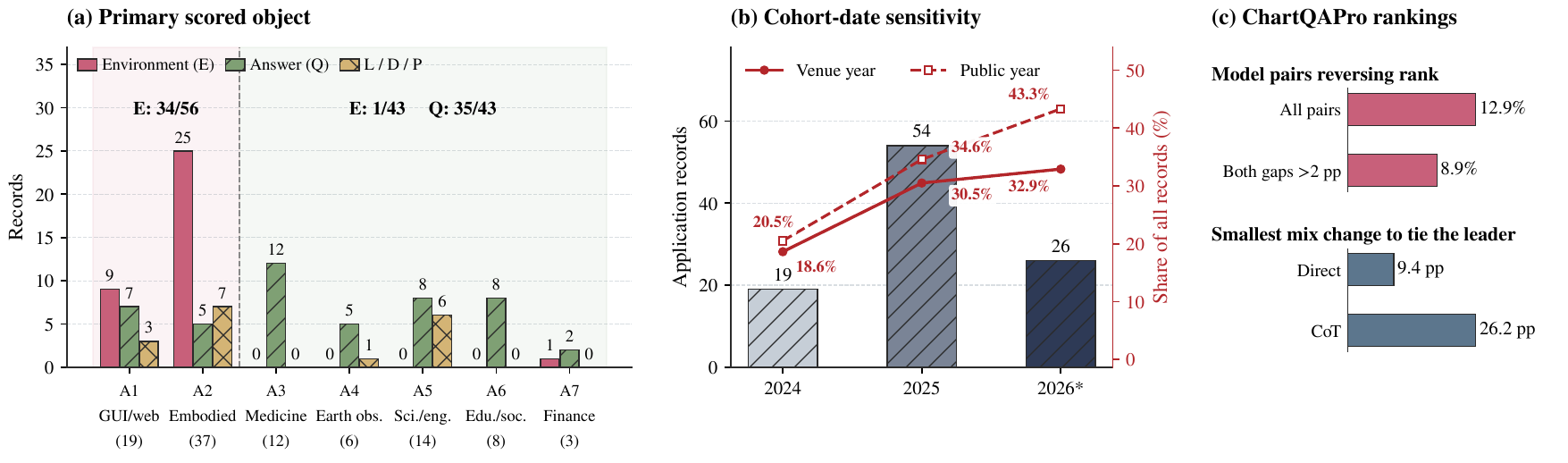}
\vspace{-2em}
\caption{Application benchmark distribution by scored endpoint, release cohort, and ranking sensitivity. (a) Primary scored endpoints across interactive agentic domains (\textbf{A1--A2}) and specialist verticals (\textbf{A3--A7}), where \textbf{E}, \textbf{Q}, \textbf{L}, \textbf{D}, and \textbf{P} denote environment outcomes, static answers, target locations, generated deliverables, and predicted actions. (b) Application benchmark counts by venue year and cohort shares under venue-year versus initial-release dating ($2026^*$ through November~2026). (c) Prompt-induced pairwise rank reversals and minimum task-weight shifts required to tie the leader on ChartQAPro.}\label{fig:application-outputs}
\end{figure*}%
}

\newcommand{\BenchmarkSelectionTable}{%
\begin{table*}[tp]\centering\normalcolor
\caption{Recommended pairings of component diagnostics and endpoint-matched benchmarks for controlled model comparison.}\label{tab:selection}
\begingroup\footnotesize\linespread{1.02}\selectfont\renewcommand{\arraystretch}{1.14}\setlength{\tabcolsep}{4.5pt}
\begin{tabularx}{\textwidth}{@{}>{\hsize=0.88\hsize\raggedright\arraybackslash}X>{\hsize=1.06\hsize\raggedright\arraybackslash}X>{\hsize=1.02\hsize\raggedright\arraybackslash}X>{\hsize=0.92\hsize\raggedright\arraybackslash}X>{\hsize=1.12\hsize\raggedright\arraybackslash}X@{}}\hline\thickhline
\rowcolor{mygray}\textbf{Evaluation Scenario} & \textbf{Component Diagnostic} & \textbf{End-to-End Suite} & \textbf{Standardized Control} & \textbf{Paired Metrics}\\\hline
Perception vs.\ STEM & MMVP~\cite{tong2024mmvp} (paired VQA) & MMMU-Pro~\cite{yue2025mmmupro} (10-way QA) & Matched prompt format & Pair consistency \& QA accuracy\\
Offline vs.\ live video & VSI-Bench~\cite{yang2025vsibench} (offline 3D) & OVO-Bench~\cite{li2025ovobench} (streaming) & Fixed frame budget & Spatial, stream accuracy \& delay\\
Visual-to-code & ChartMimic~\cite{yang2025chartmimic} (plot scripts) & Design2Code~\cite{si2025design2code} (DOM/CSS) & Pinned runtime \& viewport & Compile rate \& visual fidelity\\
GUI grounding vs.\ agency & ScreenSpot-Pro~\cite{li2025screenspotpro} (4K, \textbf{L}) & OSWorld~\cite{xie2024osworld} (live OS, \textbf{E}) & Fixed scaffold \& step cap & Click hit rate \& task success\\\hline
\end{tabularx}\endgroup
\end{table*}%
}

\newcommand{\PipelineFigure}{%
\begin{figure*}[tp]\centering\normalcolor
\includegraphics[width=\textwidth]{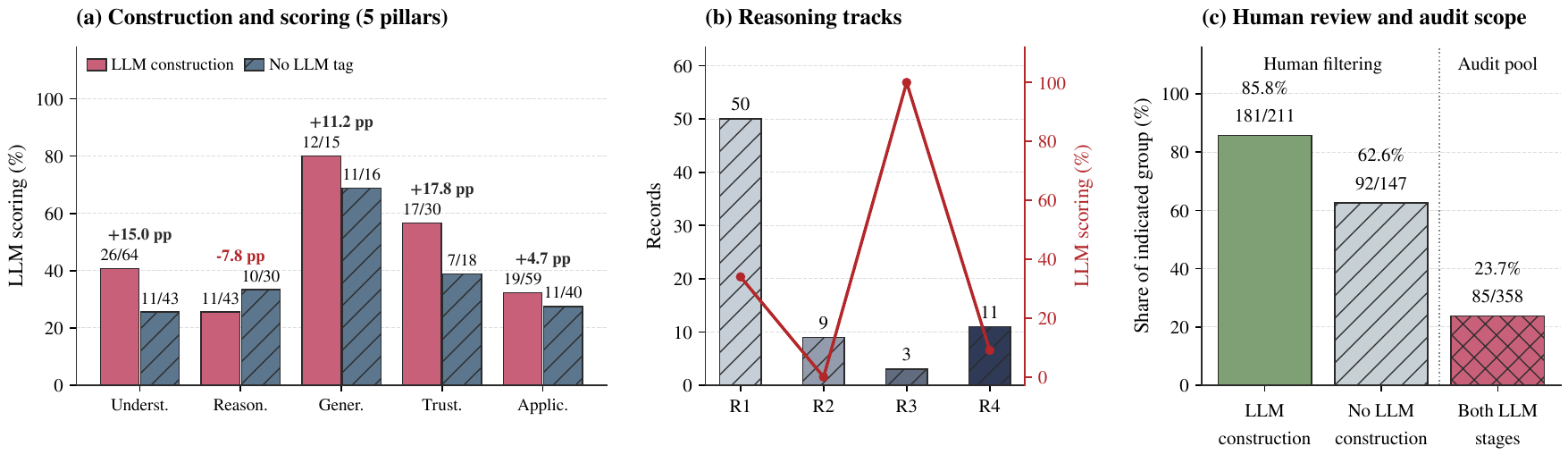}
\vspace{-2em}
\caption{Cross-stage relationships among LLM-assisted construction, LLM-based scoring, and human verification ($N=358$). (a) Share of benchmarks using LLM judges with versus without LLM-assisted construction across the five pillars. (b) Benchmark counts and LLM-judge shares across Reasoning sub-pillars (\textbf{R1--R4}). (c) Documented human filtering rates by construction group alongside the overall share of dual-LLM pipelines.}\label{fig:pipeline-association}
\end{figure*}%
}

\newcommand{\CohortCountsTable}{%
\begin{table*}[tp]\centering\normalcolor
\caption{Distribution of the surveyed benchmarks ($N=358$) across the five evaluation pillars, venue-year cohorts, LLM pipeline stages, quality-control filters, and documented evaluation protocols.}\label{tab:cohort-counts}
\begingroup\footnotesize\linespread{1.02}\selectfont\renewcommand{\arraystretch}{1.14}\setlength{\tabcolsep}{3.8pt}
\begin{tabularx}{\textwidth}{@{}>{\raggedright\arraybackslash}Xrrrrp{1pt}rrrp{1pt}rrrp{1pt}rrrr@{}}\hline\thickhline
\rowcolor{mygray} & & \multicolumn{3}{c}{\cellcolor{mygray}\textbf{Venue-Year Cohort}} & & \multicolumn{3}{c}{\cellcolor{mygray}\textbf{LLM Pipeline}} & & \multicolumn{3}{c}{\cellcolor{mygray}\textbf{Quality Filter}} & & \multicolumn{4}{c@{}}{\cellcolor{mygray}\textbf{Evaluation Protocol}}\\
\cline{3-5}\cline{7-9}\cline{11-13}\cline{15-18}\noalign{\vskip 0.4pt}
\rowcolor{mygray}\multirow{-2}{*}{\textbf{Evaluation Pillar}} & \multirow{-2}{*}{\textbf{Total}} & \textbf{2024} & \textbf{2025} & \textbf{2026} & & \textbf{Build} & \textbf{Judge} & \textbf{Both} & & \textbf{Human} & \textbf{Rule} & \textbf{LLM} & & \textbf{Static} & \textbf{Stream} & \textbf{Closed} & \textbf{Process}\\\hline
Understanding & 107 & 37 & 53 & 17 & & 64 & 37 & 26 & & 87 & 38 & 24 & & 99 & 4 & 0 & 0\\
Reasoning & 73 & 23 & 33 & 17 & & 43 & 21 & 11 & & 64 & 27 & 12 & & 65 & 0 & 0 & 6\\
Generation & 31 & 5 & 17 & 9 & & 15 & 23 & 12 & & 23 & 19 & 6 & & 26 & 2 & 1 & 2\\
Trustworthiness \& Safety & 48 & 18 & 20 & 10 & & 30 & 24 & 17 & & 33 & 13 & 10 & & 46 & 0 & 0 & 0\\
Application & 99 & 19 & 54 & 26 & & 59 & 30 & 19 & & 66 & 40 & 12 & & 47 & 0 & 32 & 3\\\hline
\textbf{Corpus Total} & \textbf{358} & \textbf{102} & \textbf{177} & \textbf{79} & & \textbf{211} & \textbf{135} & \textbf{85} & & \textbf{273} & \textbf{137} & \textbf{64} & & \textbf{283} & \textbf{6} & \textbf{33} & \textbf{11}\\\hline
\end{tabularx}\endgroup
\end{table*}%
}

\newcommand{\VenueCoverageFigure}{%
\begin{figure*}[tp]\centering\normalcolor
\includegraphics[width=\textwidth]{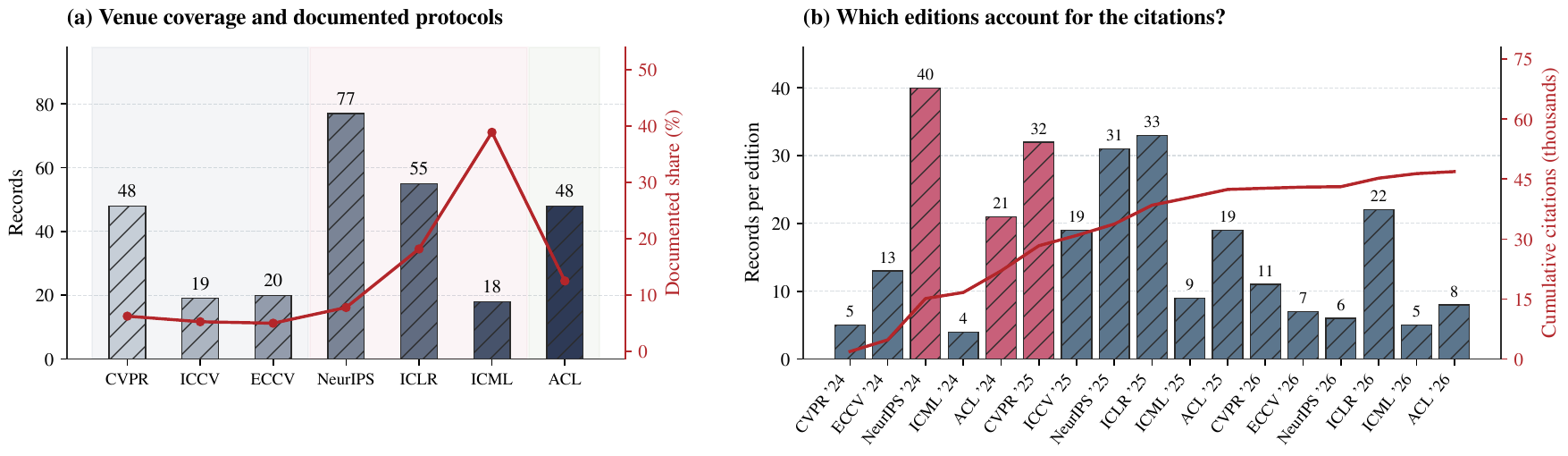}
\caption{Distribution of 285 benchmarks across seven major conference series. (a) Benchmark counts per venue and share documenting closed-loop execution or process scoring. (b) Benchmark counts and cumulative citations across conference editions, highlighting the three highest-citation editions.}\label{fig:author-venues}
\end{figure*}%
}

%% file: sec/direction.tex
\section{Open Challenges and Future Research Directions}
\label{sec:direction}

Synthesizing the empirical gaps across our five-pillar inventory (\s\ref{sec:understanding}--\s\ref{sec:cross}) points to five concrete research priorities, each anchored to a measured deficit in the 358-benchmark corpus:

\noindent\textbf{1.~Causal Streaming and Deadline-Bounded Evaluation.} Only four of 73 video benchmarks in our corpus enforce streaming frame arrival, and two of 27 audio benchmarks test live streaming speech (\s\ref{sec:protocols}); offline full-clip access allows models to exploit future frames unavailable at decision time. Streaming video suites query models while frames arrive~\cite{li2025ovobench,zeng2025odvbench,xu2026infstreamseval}, and full-duplex spoken-dialogue benchmarks measure turn-taking and response latency~\cite{lin2026fullduplexbenchv15,lin2026fullduplexbenchv2}. Next-generation benchmarks can jointly report accuracy, premature-commitment rate, and post-evidence delay around annotated outcome-resolving frames (Blueprint~1), calibrating deadlines to deployment risk.

\noindent\textbf{2.~Verified State-Change Sandboxes and Autonomous Multimodal Research.} Whereas 34 of 56 interactive and embodied Application benchmarks (\textbf{A1--A2}, 60.7\%) score verified environment outcomes (\textbf{E}), 35 of 43 specialist benchmarks (\textbf{A3--A7}, 81.4\%) evaluate static answers (\textbf{Q}) and only 1/43 (2.3\%) verifies an environment state change ($26.1\times$ rate gap; descriptive odds ratio $64.9$; \s\ref{sec:protocols}). Recent automated scientific research (``auto-research'') benchmarks execute code inside containers over multimodal papers, figures, and raw datasets~\cite{chen2025scienceagentbench,starace2025paperbench,chen2025mlrbench,liu2026moscibench,seo2026paper2code}, though open-ended ML research grading still relies on LLM review rubrics that can reward plausible prose over reproducible experimental gains. Closing this gap across specialist domains requires coupling diagnostic queries with resettable execution sandboxes such as clinical EHRs, wet-lab digital twins, engineering CAD/codebases, and financial workbooks (Blueprint~2). Embodied benchmarks are moving from static question answering towards closed-loop evaluation of VLA policies in simulation~\cite{zhang2026vlaarena,yang2026robolab}, in real-to-sim reconstructions~\cite{jain2025polaris}, and in learned world models~\cite{quevedo2026worldgym}. Measuring how reliably simulated and world-model rollouts transfer to physical hardware requires matched real-robot evaluation alongside safety verification~\cite{lu2026isbench,cui2026liberosafety}.

\noindent\textbf{3.~Decoupled Judge Auditing, Contamination Defense, and Anchor Versioning.} In our inventory, 85/358 benchmarks deploy LLMs in both construction and scoring, pure Generation reaches 80.0\% LLM-judge reliance under LLM-assisted construction, and ChartQAPro demonstrates that prompt choice alone reverses 12.9\% of model pairs (\s\ref{sec:protocols}--\s\ref{sec:crossstats}). The reliability of the judges has therefore become an evaluation target in itself~\cite{chen2024mllmasajudge}, and some benchmarks remove questions that frontier models already solve or that can be answered without the visual input~\cite{phan2025humanityslastexam,yue2025mmmupro}. Future releases should audit judges across independent model families using verified paraphrases and factual perturbations (Blueprint~3), screen dynamic splits for contamination~\cite{zhang2026kbe}, and preserve frozen anchor subsets across version upgrades so capability gains remain separable from rubric drift.

\noindent\textbf{4.~Cross-Modal Representational Consistency and Any-to-Any Evaluation.} Still images appear in 243/358 benchmarks while audio (27) and native 3D (6) remain sparse, and pure multimodal synthesis (31 benchmarks) has only recently begun bridging understanding and generation. Recent benchmarks have started in this direction: some evaluate unified models that generate or edit images using world knowledge and reasoning~\cite{niu2026wise,zhao2025risebench}, and others evaluate omni-modal models that take audio and video together~\cite{hong2026worldsense,li2026omnivideobench}. Future suites should test \textit{cross-modal representational consistency} directly: checking whether physical, spatial, or factual constraints answered correctly in \textbf{R1/R2} hold when the same model synthesizes visual or spoken artifacts in \textbf{G2/G3} without introducing cross-modal hallucinations (\textbf{T1}).

\noindent\textbf{5.~Recursive Self-Improvement (RSI) and Collaborative Human-in-the-Loop Evolution.} Nearly all surveyed benchmarks evaluate frozen model checkpoints in single-shot episodes, leaving multi-episode recursive self-improvement (RSI)~\cite{deng2025selfimprovementsurvey} and interactive human--AI co-adaptation~\cite{bekris2026clara,zhang2026harness} largely untested. Initial reflective benchmarks (\textbf{R4}) test two-turn visual critique and step-level process verification~\cite{wu2025visco,wang2025vicrit,wang2026visualprm}, while spoken turn-taking~\cite{arora2025talkingturns} and examiner-paced speech suites~\cite{lin2026fullduplexbenchv2} probe live conversational adaptation. Next-generation benchmarks should evaluate \textit{closed-loop self-evolution}: testing whether multimodal agents diagnose visual-execution failures, synthesize verifiable training curricula without reward hacking or mode collapse, request human disambiguation on underspecified goals, and preserve safety and calibration across iterative self-updates.

%% file: sec/conclusion.tex
\section{Conclusion}
\label{sec:conclusions}

This survey reviewed 358 peer-reviewed multimodal benchmarks published between 2024 and 2026 across five evaluation pillars: \textbf{understanding} (107), \textbf{reasoning} (73), \textbf{generation} (31), \textbf{trustworthiness and safety} (48), and \textbf{application} (99). Recent benchmarks extend evaluation from static, single-pass recognition to streaming video, visually grounded and self-improving reasoning, unified multimodal synthesis, and closed-loop GUI, embodied, and scientific agents. Our cross-corpus audit shows that fair model comparison depends as much on experimental controls as on task difficulty: unimodal shortcuts, acausal evidence access, missing transactional sandboxes in specialist domains, and cross-stage coupling between LLM synthesis and judging can all confound leaderboard scores. By combining a five-pillar taxonomy and dataset-pipeline audit with three benchmark design blueprints, a six-item pre-release checklist, and five corpus-grounded research directions, this survey provides practical guidance for building and interpreting next-generation multimodal evaluations.

%% file: references.bib
@inproceedings{chen2024unified,
  author = {Xiang Chen and Chenxi Wang and Yida Xue and Ningyu Zhang and Xiaoyan Yang and Qiang Li and Yue Shen and Lei Liang and Jinjie Gu and Huajun Chen},
  title = {Unified Hallucination Detection for Multimodal Large Language Models},
  booktitle = {Proceedings of the Annual Meeting of the Association for Computational Linguistics},
  pages = {3235--3252},
  year = {2024},
  doi = {10.18653/v1/2024.acl-long.178}
}

@inproceedings{alayrac2022flamingo,
  author = {Alayrac, Jean-Baptiste and Donahue, Jeff and Luc, Pauline and Miech, Antoine and Barr, Iain and Hasson, Yana and Lenc, Karel and Mensch, Arthur and Millican, Katherine and Reynolds, Malcolm and others},
  title = {Flamingo: a Visual Language Model for Few-Shot Learning},
  booktitle = {Advances in Neural Information Processing Systems},
  year = {2022}
}

@inproceedings{zhou2024your,
  author = {Zhou, Zihao and Liu, Shudong and Ning, Maizhen and Liu, Wei and Wang, Jindong and Wong, Derek F and Huang, Xiaowei and Wang, Qiufeng and Huang, Kaizhu},
  title = {Is Your Model Really A Good Math Reasoner? {Evaluating} Mathematical Reasoning with Checklist},
  booktitle = {Proceedings of the International Conference on Learning Representations},
  year = {2025}
}

@article{zhu2024multi,
  author = {Zhu, Zichen and Xu, Yang and Chen, Lu and Yang, Jingkai and Ma, Yichuan and Sun, Yiming and Wen, Hailin and Liu, Jiaqi and Cai, Jinyu and Ma, Yingzi and others},
  title = {{MULTI}: Multimodal Understanding Leaderboard with Text and Images},
  journal = {Science China Information Sciences},
  volume = {68},
  number = {10},
  year = {2025},
  doi = {10.1007/s11432-024-4602-x}
}

@inproceedings{he2024cmmu,
  author = {He, Zheqi and Wu, Xinya and Zhou, Pengfei and Xuan, Richeng and Liu, Guang and Yang, Xi and Zhu, Qiannan and Huang, Hua},
  title = {{CMMU}: A Benchmark for {Chinese} Multi-modal Multi-type Question Understanding and Reasoning},
  booktitle = {Proceedings of the International Joint Conference on Artificial Intelligence},
  pages = {830--838},
  year = {2024}
}

@inproceedings{yue2024mmmu,
  author = {Yue, Xiang and Ni, Yuansheng and Zhang, Kai and Zheng, Tianyu and Liu, Ruoqi and Zhang, Ge and Stevens, Samuel and Jiang, Dongfu and Ren, Weiming and Sun, Yuxuan and others},
  title = {{MMMU}: A Massive Multi-Discipline Multimodal Understanding and Reasoning Benchmark for Expert {AGI}},
  booktitle = {Proceedings of the IEEE/CVF Conference on Computer Vision and Pattern Recognition},
  pages = {9556--9567},
  year = {2024},
  doi = {10.1109/CVPR52733.2024.00913}
}

@inproceedings{liang2024scemqa,
  author = {Liang, Zhenwen and Guo, Kehan and Liu, Gang and Guo, Taicheng and Zhou, Yujun and Yang, Tianyu and Jiao, Jiajun and Pi, Renjie and Zhang, Jipeng and Zhang, Xiangliang},
  title = {{SceMQA}: A Scientific College Entrance Level Multimodal Question Answering Benchmark},
  booktitle = {Proceedings of the Annual Meeting of the Association for Computational Linguistics},
  pages = {109--119},
  year = {2024},
  doi = {10.18653/v1/2024.acl-short.11}
}

@inproceedings{zhang2024mathverse,
  author = {Zhang, Renrui and Jiang, Dongzhi and Zhang, Yichi and Lin, Haokun and Guo, Ziyu and Qiu, Pengshuo and Zhou, Aojun and Lu, Pan and Chang, Kai-Wei and Gao, Peng and others},
  title = {{MathVerse}: Does Your Multi-modal {LLM} Truly See the Diagrams in Visual Math Problems?},
  booktitle = {Proceedings of the European Conference on Computer Vision},
  pages = {169--186},
  year = {2024},
  doi = {10.1007/978-3-031-73242-3_10}
}

@inproceedings{wang2024measuring,
  author = {Ke Wang and Junting Pan and Weikang Shi and Zimu Lu and Houxing Ren and Aojun Zhou and Mingjie Zhan and Hongsheng Li},
  title = {Measuring Multimodal Mathematical Reasoning with {MATH-Vision} Dataset},
  booktitle = {Advances in Neural Information Processing Systems (Datasets and Benchmarks Track)},
  year = {2024}
}

@article{yin2023survey,
  author = {Yin, Shukang and Fu, Chaoyou and Zhao, Sirui and Li, Ke and Sun, Xing and Xu, Tong and Chen, Enhong},
  title = {A Survey on Multimodal Large Language Models},
  journal = {National Science Review},
  volume = {11},
  number = {12},
  year = {2024},
  doi = {10.1093/nsr/nwae403}
}

@inproceedings{pramanick2024spiqa,
  author = {Shraman Pramanick and Rama Chellappa and Subhashini Venugopalan},
  title = {{SPIQA}: A Dataset for Multimodal Question Answering on Scientific Papers},
  booktitle = {Advances in Neural Information Processing Systems (Datasets and Benchmarks Track)},
  year = {2024}
}

@inproceedings{li2024densefusion,
  author = {Li, Xiaotong and Zhang, Fan and Diao, Haiwen and Wang, Yueze and Wang, Xinlong and Duan, Ling-Yu},
  title = {{DenseFusion-1M}: Merging Vision Experts for Comprehensive Multimodal Perception},
  booktitle = {Advances in Neural Information Processing Systems (Datasets and Benchmarks Track)},
  year = {2024}
}

@inproceedings{gu2024mllmguard,
  author = {Gu, Tianle and Zhou, Zeyang and Huang, Kexin and Liang, Dandan and Wang, Yixu and Zhao, Haiquan and Yao, Yuanqi and Qiao, Xingge and Wang, Keqing and Yang, Yujiu and others},
  title = {{MLLMGuard}: A Multi-dimensional Safety Evaluation Suite for Multimodal Large Language Models},
  booktitle = {Advances in Neural Information Processing Systems (Datasets and Benchmarks Track)},
  year = {2024}
}

@inproceedings{zhu2024multimodal,
  author = {Zhu, Wanrong and Hessel, Jack and Awadalla, Anas and Gadre, Samir Yitzhak and Dodge, Jesse and Fang, Alex and Yu, Youngjae and Schmidt, Ludwig and Wang, William Yang and Choi, Yejin},
  title = {Multimodal {C4}: An Open, Billion-scale Corpus of Images Interleaved With Text},
  booktitle = {Advances in Neural Information Processing Systems},
  year = {2023}
}

@inproceedings{chen2024comm,
  author = {Chen, Wei and Li, Lin and Yang, Yongqi and Wen, Bin and Yang, Fan and Gao, Tingting and Wu, Yu and Chen, Long},
  title = {{CoMM}: A Coherent Interleaved Image-Text Dataset for Multimodal Understanding and Generation},
  booktitle = {Proceedings of the IEEE/CVF Conference on Computer Vision and Pattern Recognition},
  pages = {8073--8082},
  year = {2025},
  doi = {10.1109/CVPR52734.2025.00756}
}

@inproceedings{masry2022chartqa,
  author = {Masry, Ahmed and Do, Xuan Long and Tan, Jia Qing and Joty, Shafiq and Hoque, Enamul},
  title = {{ChartQA}: A Benchmark for Question Answering about Charts with Visual and Logical Reasoning},
  booktitle = {Findings of the Association for Computational Linguistics: ACL 2022},
  pages = {2263--2279},
  year = {2022},
  doi = {10.18653/v1/2022.findings-acl.177}
}

@inproceedings{wang2024charxiv,
  author = {Wang, Zirui and Xia, Mengzhou and He, Luxi and Chen, Howard and Liu, Yitao and Zhu, Richard and Liang, Kaiqu and Wu, Xindi and Liu, Haotian and Malladi, Sadhika and others},
  title = {{CharXiv}: Charting Gaps in Realistic Chart Understanding in Multimodal {LLMs}},
  booktitle = {Advances in Neural Information Processing Systems (Datasets and Benchmarks Track)},
  year = {2024}
}

@inproceedings{fu2024mme,
  author = {Chaoyou Fu and Peixian Chen and Yunhang Shen and Yulei Qin and Mengdan Zhang and Xu Lin and Jinrui Yang and Xiawu Zheng and Ke Li and Xing Sun and Yunsheng Wu and Rongrong Ji and Caifeng Shan and Ran He},
  title = {{MME}: A Comprehensive Evaluation Benchmark for Multimodal Large Language Models},
  booktitle = {Advances in Neural Information Processing Systems (Datasets and Benchmarks Track)},
  year = {2025}
}

@inproceedings{liu2023mmbench,
  author = {Liu, Yuan and Duan, Haodong and Zhang, Yuanhan and Li, Bo and Zhang, Songyang and Zhao, Wangbo and Yuan, Yike and Wang, Jiaqi and He, Conghui and Liu, Ziwei and others},
  title = {{MMBench}: Is Your Multi-modal Model an All-Around Player?},
  booktitle = {Proceedings of the European Conference on Computer Vision},
  pages = {216--233},
  year = {2024},
  doi = {10.1007/978-3-031-72658-3_13}
}

@inproceedings{qian2024mia,
  author = {Qian, Yusu and Ye, Hanrong and Fauconnier, Jean-Philippe and Grasch, Peter and Yang, Yinfei and Gan, Zhe},
  title = {{MIA-Bench}: Towards Better Instruction Following Evaluation of Multimodal {LLMs}},
  booktitle = {Proceedings of the International Conference on Learning Representations},
  year = {2025}
}

@inproceedings{lu1codexglue,
  author = {Lu, Shuai and Guo, Daya and Ren, Shuo and Huang, Junjie and Svyatkovskiy, Alexey and Blanco, Ambrosio and Clement, Colin and Drain, Dawn and Jiang, Daxin and Tang, Duyu and others},
  title = {{CodeXGLUE}: A Machine Learning Benchmark Dataset for Code Understanding and Generation},
  booktitle = {Proceedings of the Neural Information Processing Systems Track on Datasets and Benchmarks},
  year = {2021}
}

@inproceedings{yun2024web2code,
  author = {Yun, Sukmin and Lin, Haokun and Thushara, Rusiru and Bhat, Mohammad Qazim and Wang, Yongxin and Jiang, Zutao and Deng, Mingkai and Wang, Jinhong and Tao, Tianhua and Li, Junbo and others},
  title = {{Web2Code}: A Large-scale Webpage-to-Code Dataset and Evaluation Framework for Multimodal {LLMs}},
  booktitle = {Advances in Neural Information Processing Systems (Datasets and Benchmarks Track)},
  year = {2024}
}

@inproceedings{wu2024plot2code,
  author = {Wu, Chengyue and Liang, Zhixuan and Ge, Yixiao and Guo, Qiushan and Lu, Zeyu and Wang, Jiahao and Shan, Ying and Luo, Ping},
  title = {{Plot2Code}: A Comprehensive Benchmark for Evaluating Multi-modal Large Language Models in Code Generation from Scientific Plots},
  booktitle = {Findings of the Association for Computational Linguistics: NAACL 2025},
  pages = {3006--3028},
  year = {2025},
  doi = {10.18653/v1/2025.findings-naacl.164}
}

@inproceedings{li2024mmcode,
  author = {Kaixin Li and Yuchen Tian and Qisheng Hu and Ziyang Luo and Zhiyong Huang and Jing Ma},
  title = {{MMCode}: Benchmarking Multimodal Large Language Models for Code Generation with Visually Rich Programming Problems},
  booktitle = {Findings of the Association for Computational Linguistics: EMNLP 2024},
  pages = {736--783},
  year = {2024},
  doi = {10.18653/v1/2024.findings-emnlp.42}
}

@inproceedings{liu2024ii,
  author = {Liu, Ziqiang and Fang, Feiteng and Feng, Xi and Du, Xinrun and Zhang, Chenhao and Wang, Zekun and Bai, Yuelin and Zhao, Qixuan and Fan, Liyang and Gan, Chengguang and others},
  title = {{II-Bench}: An Image Implication Understanding Benchmark for Multimodal Large Language Models},
  booktitle = {Advances in Neural Information Processing Systems (Datasets and Benchmarks Track)},
  year = {2024}
}

@inproceedings{ye2024mm,
  author = {Wenqian Ye and Bohan Liu and Guangtao Zheng and Di Wang and Xu Cao and Yunsheng Ma and Bolin Lai and James M. Rehg and Aidong Zhang},
  title = {{MM-SpuBench}: Towards Better Understanding of Spurious Biases in Multimodal {LLMs}},
  booktitle = {Proceedings of the ACM SIGKDD Conference on Knowledge Discovery and Data Mining},
  pages = {2854--2865},
  year = {2026},
  doi = {10.1145/3770854.3785678}
}

@inproceedings{zhou2024mlvu,
  author = {Junjie Zhou and Yan Shu and Bo Zhao and Boya Wu and Zhengyang Liang and Shitao Xiao and Minghao Qin and Xi Yang and Yongping Xiong and Bo Zhang and Tiejun Huang and Zheng Liu},
  title = {{MLVU}: Benchmarking Multi-task Long Video Understanding},
  booktitle = {Proceedings of the IEEE/CVF Conference on Computer Vision and Pattern Recognition},
  pages = {13691--13701},
  year = {2025},
  doi = {10.1109/CVPR52734.2025.01278}
}

@inproceedings{fu2024blink,
  author = {Fu, Xingyu and Hu, Yushi and Li, Bangzheng and Feng, Yu and Wang, Haoyu and Lin, Xudong and Roth, Dan and Smith, Noah A and Ma, Wei-Chiu and Krishna, Ranjay},
  title = {{BLINK}: Multimodal Large Language Models Can See but Not Perceive},
  booktitle = {Proceedings of the European Conference on Computer Vision},
  pages = {148--166},
  year = {2024},
  doi = {10.1007/978-3-031-73337-6_9}
}

@inproceedings{wang2024multimodal,
  author = {Wang, Hengyi and Shi, Haizhou and Tan, Shiwei and Qin, Weiyi and Wang, Wenyuan and Zhang, Tunyu and Nambi, Akshay and Ganu, Tanuja and Wang, Hao},
  title = {Multimodal Needle in a Haystack: Benchmarking Long-Context Capability of Multimodal Large Language Models},
  booktitle = {Proceedings of the Conference of the Nations of the Americas Chapter of the Association for Computational Linguistics: Human Language Technologies},
  pages = {3221--3241},
  year = {2025},
  doi = {10.18653/v1/2025.naacl-long.166}
}

@inproceedings{song2024milebench,
  author = {Song, Dingjie and Chen, Shunian and Chen, Guiming Hardy and Yu, Fei and Wan, Xiang and Wang, Benyou},
  title = {{MileBench}: Benchmarking {MLLMs} in Long Context},
  booktitle = {Proceedings of the Conference on Language Modeling},
  year = {2024}
}

@inproceedings{li2022mmcoqa,
  author = {Li, Yongqi and Li, Wenjie and Nie, Liqiang},
  title = {{MMCoQA}: Conversational Question Answering over Text, Tables, and Images},
  booktitle = {Proceedings of the Annual Meeting of the Association for Computational Linguistics},
  pages = {4220--4231},
  year = {2022},
  doi = {10.18653/v1/2022.acl-long.290}
}

@inproceedings{wu2star,
  author = {Wu, Bo and Yu, Shoubin and Chen, Zhenfang and Tenenbaum, Joshua B and Gan, Chuang},
  title = {{STAR}: A Benchmark for Situated Reasoning in Real-World Videos},
  booktitle = {Proceedings of the Neural Information Processing Systems Track on Datasets and Benchmarks},
  year = {2021}
}

@inproceedings{fan2024read,
  author = {Fan, Yue and Ding, Lei and Kuo, Ching-Chen and Jiang, Shan and Zhao, Yang and Guan, Xinze and Yang, Jie and Zhang, Yi and Wang, Xin Eric},
  title = {Read Anywhere Pointed: Layout-aware {GUI} Screen Reading with Tree-of-Lens Grounding},
  booktitle = {Proceedings of the Conference on Empirical Methods in Natural Language Processing},
  pages = {9503--9522},
  year = {2024},
  doi = {10.18653/v1/2024.emnlp-main.533}
}

@article{street2024llms,
  author = {Street, Winnie and Siy, John Oliver and Keeling, Geoff and Baranes, Adrien and Barnett, Benjamin and McKibben, Michael and Kanyere, Tatenda and Lentz, Alison and Dunbar, Robin IM and others},
  title = {{LLMs} achieve adult human performance on higher-order theory of mind tasks},
  journal = {Frontiers in Human Neuroscience},
  volume = {19},
  year = {2026},
  doi = {10.3389/fnhum.2025.1633272}
}

@inproceedings{li2024facial,
  author = {Li, Yifan and Dao, Anh and Bao, Wentao and Tan, Zhen and Chen, Tianlong and Liu, Huan and Kong, Yu},
  title = {Facial Affective Behavior Analysis with Instruction Tuning},
  booktitle = {Proceedings of the European Conference on Computer Vision},
  pages = {165--186},
  year = {2024},
  doi = {10.1007/978-3-031-72649-1_10}
}

@inproceedings{chen2024coin,
  author = {Chen, Cheng and Zhu, Junchen and Luo, Xu and Shen, Hengtao and Gao, Lianli and Song, Jingkuan},
  title = {{CoIN}: A Benchmark of Continual Instruction Tuning for Multimodel Large Language Models},
  booktitle = {Advances in Neural Information Processing Systems (Datasets and Benchmarks Track)},
  year = {2024}
}

@inproceedings{balog2017deepcoder,
  author = {Balog, M and Gaunt, AL and Brockschmidt, M and Nowozin, S and Tarlow, D},
  title = {{DeepCoder}: Learning to Write Programs},
  booktitle = {Proceedings of the International Conference on Learning Representations},
  year = {2017}
}

@misc{chen2021evaluating,
  author = {Chen, Mark and Tworek, Jerry and Jun, Heewoo and Yuan, Qiming and Pinto, Henrique Ponde De Oliveira and Kaplan, Jared and Edwards, Harri and Burda, Yuri and Joseph, Nicholas and Brockman, Greg and others},
  title = {Evaluating Large Language Models Trained on Code},
  howpublished = {arXiv preprint arXiv:2107.03374},
  year = {2021}
}

@misc{zhou2023instruction,
  author = {Zhou, Jeffrey and Lu, Tianjian and Mishra, Swaroop and Brahma, Siddhartha and Basu, Sujoy and Luan, Yi and Zhou, Denny and Hou, Le},
  title = {Instruction-Following Evaluation for Large Language Models},
  howpublished = {arXiv preprint arXiv:2311.07911},
  year = {2023}
}

@misc{achiam2023gpt,
  author = {Achiam, Josh and Adler, Steven and Agarwal, Sandhini and Ahmad, Lama and Akkaya, Ilge and Aleman, Florencia Leoni and Almeida, Diogo and Altenschmidt, Janko and Altman, Sam and Anadkat, Shyamal and others},
  title = {{GPT-4} Technical Report},
  howpublished = {arXiv preprint arXiv:2303.08774},
  year = {2023}
}

@inproceedings{qin2024infobench,
  author = {Qin, Yiwei and Song, Kaiqiang and Hu, Yebowen and Yao, Wenlin and Cho, Sangwoo and Wang, Xiaoyang and Wu, Xuansheng and Liu, Fei and Liu, Pengfei and Yu, Dong},
  title = {{InFoBench}: Evaluating Instruction Following Ability in Large Language Models},
  booktitle = {Findings of the Association for Computational Linguistics: ACL 2024},
  pages = {13025--13048},
  year = {2024},
  doi = {10.18653/v1/2024.findings-acl.772}
}

@inproceedings{rohrbach2018object,
  author = {Rohrbach, Anna and Hendricks, Lisa Anne and Burns, Kaylee and Darrell, Trevor and Saenko, Kate},
  title = {Object Hallucination in Image Captioning},
  booktitle = {Proceedings of the Conference on Empirical Methods in Natural Language Processing},
  pages = {4035--4045},
  year = {2018},
  doi = {10.18653/v1/d18-1437}
}

@article{ji2023survey,
  author = {Ji, Ziwei and Lee, Nayeon and Frieske, Rita and Yu, Tiezheng and Su, Dan and Xu, Yan and Ishii, Etsuko and Bang, Ye Jin and Madotto, Andrea and Fung, Pascale},
  title = {Survey of Hallucination in Natural Language Generation},
  journal = {ACM Computing Surveys},
  volume = {55},
  number = {12},
  pages = {248:1--248:38},
  year = {2023},
  doi = {10.1145/3571730}
}

@misc{bai2024hallucination,
  author = {Bai, Zechen and Wang, Pichao and Xiao, Tianjun and He, Tong and Han, Zongbo and Zhang, Zheng and Shou, Mike Zheng},
  title = {Hallucination of Multimodal Large Language Models: A Survey},
  howpublished = {arXiv preprint arXiv:2404.18930},
  year = {2024}
}

@inproceedings{zheng2024reefknot,
  author = {Kening Zheng and Junkai Chen and Yibo Yan and Xin Zou and Huiyu Zhou and Xuming Hu},
  title = {Reefknot: A Comprehensive Benchmark for Relation Hallucination Evaluation, Analysis and Mitigation in Multimodal Large Language Models},
  booktitle = {Findings of the Association for Computational Linguistics: ACL 2025},
  pages = {6193--6212},
  year = {2025},
  doi = {10.18653/v1/2025.findings-acl.322}
}

@inproceedings{zhang2024automated,
  author = {Mengxi Zhang and Wenhao Wu and Yu Lu and Yuxin Song and Kang Rong and Huanjin Yao and Jianbo Zhao and Fanglong Liu and Haocheng Feng and Jingdong Wang and Yifan Sun},
  title = {Automated Multi-level Preference for {MLLMs}},
  booktitle = {Advances in Neural Information Processing Systems},
  year = {2024}
}

@inproceedings{ding2024hallu,
  author = {Ding, Peng and Wu, Jingyu and Kuang, Jun and Ma, Dan and Cao, Xuezhi and Cai, Xunliang and Chen, Shi and Chen, Jiajun and Huang, Shujian},
  title = {{Hallu-PI}: Evaluating Hallucination in Multi-modal Large Language Models within Perturbed Inputs},
  booktitle = {Proceedings of the ACM International Conference on Multimedia},
  pages = {10707--10715},
  year = {2024},
  doi = {10.1145/3664647.3681251}
}

@inproceedings{wang2024haloquest,
  author = {Wang, Zhecan and Bingham, Garrett and Yu, Adams and Le, Quoc and Luong, Thang and Ghiasi, Golnaz},
  title = {{HaloQuest}: A Visual Hallucination Dataset for Advancing Multimodal Reasoning},
  booktitle = {Proceedings of the European Conference on Computer Vision},
  pages = {288--304},
  year = {2024},
  doi = {10.1007/978-3-031-72980-5_17}
}

@inproceedings{chen2024multi,
  author = {Chen, Xuweiyi and Ma, Ziqiao and Zhang, Xuejun and Xu, Sihan and Qian, Shengyi and Yang, Jianing and Fouhey, David F and Chai, Joyce},
  title = {Multi-Object Hallucination in Vision-Language Models},
  booktitle = {Advances in Neural Information Processing Systems},
  year = {2024}
}

@inproceedings{ye2024beaf,
  author = {Ye-Bin, Moon and Hyeon-Woo, Nam and Choi, Wonseok and Oh, Tae-Hyun},
  title = {{BEAF}: Observing {BEfore-AFter} Changes to Evaluate Hallucination in Vision-Language Models},
  booktitle = {Proceedings of the European Conference on Computer Vision},
  pages = {232--248},
  year = {2024},
  doi = {10.1007/978-3-031-73247-8_14}
}

@inproceedings{lovenia2023negative,
  author = {Lovenia, Holy and Dai, Wenliang and Cahyawijaya, Samuel and Ji, Ziwei and Fung, Pascale},
  title = {Negative Object Presence Evaluation ({NOPE}) to Measure Object Hallucination in Vision-Language Models},
  booktitle = {Proceedings of the Workshop on Advances in Language and Vision Research (ALVR)},
  pages = {37--58},
  year = {2024},
  doi = {10.18653/v1/2024.alvr-1.4}
}

@misc{liu2024survey,
  author = {Liu, Hanchao and Xue, Wenyuan and Chen, Yifei and Chen, Dapeng and Zhao, Xiutian and Wang, Ke and Hou, Liping and Li, Rongjun and Peng, Wei},
  title = {A Survey on Hallucination in Large Vision-Language Models},
  howpublished = {arXiv preprint arXiv:2402.00253},
  year = {2024}
}

@inproceedings{wu2024autohallusion,
  author = {Wu, Xiyang and Guan, Tianrui and Li, Dianqi and Huang, Shuaiyi and Liu, Xiaoyu and Wang, Xijun and Xian, Ruiqi and Shrivastava, Abhinav and Huang, Furong and Boyd-Graber, Jordan Lee and others},
  title = {{AutoHallusion}: Automatic Generation of Hallucination Benchmarks for Vision-Language Models},
  booktitle = {Findings of the Association for Computational Linguistics: EMNLP 2024},
  pages = {8395--8419},
  year = {2024},
  doi = {10.18653/v1/2024.findings-emnlp.493}
}

@inproceedings{kaul2024throne,
  author = {Kaul, Prannay and Li, Zhizhong and Yang, Hao and Dukler, Yonatan and Swaminathan, Ashwin and Taylor, CJ and Soatto, Stefano},
  title = {{THRONE}: An Object-Based Hallucination Benchmark for the Free-Form Generations of Large Vision-Language Models},
  booktitle = {Proceedings of the IEEE/CVF Conference on Computer Vision and Pattern Recognition},
  pages = {27218--27228},
  year = {2024},
  doi = {10.1109/CVPR52733.2024.02571}
}

@inproceedings{li2024eyes,
  author = {Yian Li and Wentao Tian and Yang Jiao and Tianwen Qian and Na Zhao and Bin Zhu and Jingjing Chen and Yu-Gang Jiang},
  title = {Look Before You Decide: Prompting Active Deduction of {MLLMs} for Assumptive Reasoning},
  booktitle = {Proceedings of the ACM International Conference on Multimedia},
  pages = {2713--2722},
  year = {2025},
  doi = {10.1145/3746027.3754720}
}

@misc{chatgpt,
  author = {{OpenAI}},
  title = {Introducing {ChatGPT}},
  howpublished = {\url{https://openai.com/blog/chatgpt}},
  year = {2022}
}

@misc{team2023gemini,
  author = {{Gemini Team} and Anil, Rohan and Borgeaud, Sebastian and Wu, Yonghui and Alayrac, Jean-Baptiste and Yu, Jiahui and Soricut, Radu and Schalkwyk, Johan and Dai, Andrew M and Hauth, Anja and others},
  title = {Gemini: A Family of Highly Capable Multimodal Models},
  howpublished = {arXiv preprint arXiv:2312.11805},
  year = {2023}
}

@inproceedings{liu2024VisualAgentBench,
  author = {Xiao Liu and Tianjie Zhang and Yu Gu and Iat Long Iong and Xixuan Song and Yifan Xu and Shudan Zhang and Hanyu Lai and Jiadai Sun and Xinyue Yang and Yu Yang and Zehan Qi and Shuntian Yao and Xueqiao Sun and Siyi Cheng and Qinkai Zheng and Hao Yu and Hanchen Zhang and Wenyi Hong and Ming Ding and Lihang Pan and Xiaotao Gu and Aohan Zeng and Zhengxiao Du and Chan Hee Song and Yu Su and Yuxiao Dong and Jie Tang},
  title = {{VisualAgentBench}: Towards Large Multimodal Models as Visual Foundation Agents},
  booktitle = {Proceedings of the International Conference on Learning Representations},
  year = {2025}
}

@inproceedings{chen2024GMAI-MMBench,
  author = {Pengcheng Chen and Jin Ye and Guoan Wang and Yanjun Li and Zhongying Deng and Wei Li and Tianbin Li and Haodong Duan and Ziyan Huang and Yanzhou Su and Benyou Wang and Shaoting Zhang and Bin Fu and Jianfei Cai and Bohan Zhuang and Eric J Seibel and Junjun He and Yu Qiao},
  title = {{GMAI-MMBench}: A Comprehensive Multimodal Evaluation Benchmark Towards General Medical {AI}},
  booktitle = {Advances in Neural Information Processing Systems (Datasets and Benchmarks Track)},
  year = {2024}
}

@inproceedings{meng2024MMIU,
  author = {Fanqing Meng and Jin Wang and Chuanhao Li and Quanfeng Lu and Hao Tian and Jiaqi Liao and Xizhou Zhu and Jifeng Dai and Yu Qiao and Ping Luo and Kaipeng Zhang and Wenqi Shao},
  title = {{MMIU}: Multimodal Multi-image Understanding for Evaluating Large Vision-Language Models},
  booktitle = {Proceedings of the International Conference on Learning Representations},
  year = {2025}
}

@inproceedings{deng2009imagenet,
  author = {Deng, Jia and Dong, Wei and Socher, Richard and Li, Li-Jia and Li, Kai and Fei-Fei, Li},
  title = {{ImageNet}: A large-scale hierarchical image database},
  booktitle = {Proceedings of the IEEE Conference on Computer Vision and Pattern Recognition},
  pages = {248--255},
  year = {2009},
  doi = {10.1109/CVPR.2009.5206848}
}

@article{krishna2017visual,
  author = {Krishna, Ranjay and Zhu, Yuke and Groth, Oliver and Johnson, Justin and Hata, Kenji and Kravitz, Joshua and Chen, Stephanie and Kalantidis, Yannis and Li, Li-Jia and Shamma, David A and others},
  title = {Visual Genome: Connecting Language and Vision Using Crowdsourced Dense Image Annotations},
  journal = {International Journal of Computer Vision},
  volume = {123},
  number = {1},
  pages = {32--73},
  year = {2017},
  doi = {10.1007/s11263-016-0981-7}
}

@inproceedings{koh2024visualwebarena,
  author = {Koh, Jing Yu and Lo, Robert and Jang, Lawrence and Duvvur, Vikram and Lim, Ming Chong and Huang, Po-Yu and Neubig, Graham and Zhou, Shuyan and Salakhutdinov, Ruslan and Fried, Daniel},
  title = {{VisualWebArena}: Evaluating Multimodal Agents on Realistic Visual Web Tasks},
  booktitle = {Proceedings of the Annual Meeting of the Association for Computational Linguistics},
  pages = {881--905},
  year = {2024},
  doi = {10.18653/v1/2024.acl-long.50}
}

@article{Jiang2024MANTISIM,
  author = {Dongfu Jiang and Xuan He and Huaye Zeng and Cong Wei and Max W.F. Ku and Qian Liu and Wenhu Chen},
  title = {{MANTIS}: Interleaved Multi-Image Instruction Tuning},
  journal = {Transactions on Machine Learning Research},
  year = {2024}
}

@inproceedings{weck2024MuChoMusic,
  author = {Benno Weck and Ilaria Manco and Emmanouil Benetos and Elio Quinton and George Fazekas and Dmitry Bogdanov},
  title = {{MuChoMusic}: Evaluating Music Understanding in Multimodal Audio-Language Models},
  booktitle = {Proceedings of the International Society for Music Information Retrieval Conference},
  pages = {825--833},
  year = {2024},
  doi = {10.5281/zenodo.14877459}
}

@inproceedings{kil2024CompBench,
  author = {Jihyung Kil and Zheda Mai and Justin Lee and Zihe Wang and Kerrie Cheng and Lemeng Wang and Ye Liu and Arpita Chowdhury and Wei-Lun Chao},
  title = {{MLLM-CompBench}: A Comparative Reasoning Benchmark for Multimodal {LLMs}},
  booktitle = {Advances in Neural Information Processing Systems (Datasets and Benchmarks Track)},
  year = {2024}
}

@inproceedings{OpenEQA2023,
  author = {Majumdar, Arjun and Ajay, Anurag and Zhang, Xiaohan and Putta, Pranav and Yenamandra, Sriram and Henaff, Mikael and Silwal, Sneha and Mcvay, Paul and Maksymets, Oleksandr and Arnaud, Sergio and Yadav, Karmesh and Li, Qiyang and Newman, Ben and Sharma, Mohit and Berges, Vincent and Zhang, Shiqi and Agrawal, Pulkit and Bisk, Yonatan and Batra, Dhruv and Kalakrishnan, Mrinal and Meier, Franziska and Paxton, Chris and Sax, Sasha and Rajeswaran, Aravind},
  title = {{OpenEQA}: Embodied Question Answering in the Era of Foundation Models},
  booktitle = {Proceedings of the IEEE/CVF Conference on Computer Vision and Pattern Recognition},
  pages = {16488--16498},
  year = {2024},
  doi = {10.1109/CVPR52733.2024.01560}
}

@inproceedings{xu2024crab,
  author = {Xu, Tianqi and Chen, Linyao and Wu, Dai-Jie and Chen, Yanjun and Zhang, Zecheng and Yao, Xiang and Xie, Zhiqiang and Chen, Yongchao and Liu, Shilong and Qian, Bochen and others},
  title = {{CRAB}: Cross-environment Agent Benchmark for Multimodal Language Model Agents},
  booktitle = {Findings of the Association for Computational Linguistics: ACL 2025},
  pages = {21607--21647},
  year = {2025},
  doi = {10.18653/v1/2025.findings-acl.1113}
}

@inproceedings{chen2024rextime,
  author = {Chen, Jr-Jen and Liao, Yu-Chien and Lin, Hsi-Che and Yu, Yu-Chu and Chen, Yen-Chun and Wang, Yu-Chiang Frank},
  title = {{ReXTime}: A Benchmark Suite for Reasoning-Across-Time in Videos},
  booktitle = {Advances in Neural Information Processing Systems (Datasets and Benchmarks Track)},
  year = {2024}
}

@inproceedings{zheng2024judging,
  author = {Zheng, Lianmin and Chiang, Wei-Lin and Sheng, Ying and Zhuang, Siyuan and Wu, Zhanghao and Zhuang, Yonghao and Lin, Zi and Li, Zhuohan and Li, Dacheng and Xing, Eric and others},
  title = {Judging {LLM-as-a-Judge} with {MT-Bench} and Chatbot Arena},
  booktitle = {Advances in Neural Information Processing Systems},
  year = {2023}
}

@inproceedings{li2024mossbench,
  author = {Li, Xirui and Zhou, Hengguang and Wang, Ruochen and Zhou, Tianyi and Cheng, Minhao and Hsieh, Cho-Jui},
  title = {Is Your Multimodal Language Model Oversensitive to Safe Queries?},
  booktitle = {Proceedings of the International Conference on Learning Representations},
  year = {2025}
}

@inproceedings{tong2024cvbench,
  author = {Shengbang Tong and Ellis Brown and Penghao Wu and Sanghyun Woo and Manoj Middepogu and Sai Charitha Akula and Jihan Yang and Shusheng Yang and Adithya Iyer and Xichen Pan and Austin Wang and Rob Fergus and Yann LeCun and Saining Xie},
  title = {Cambrian-1: A Fully Open, Vision-Centric Exploration of Multimodal {LLMs}},
  booktitle = {Advances in Neural Information Processing Systems},
  year = {2024}
}

@inproceedings{chakraborty2024ADLMCQ,
  author = {Dominick Reilly and Rajatsubhra Chakraborty and Arkaprava Sinha and Manish Kumar Govind and Pu Wang and Fran{\c c}ois Br{\'e}mond and Le Xue and Srijan Das},
  title = {{LLAVIDAL}: A Large {LAnguage} {VIsion} Model for Daily Activities of Living},
  booktitle = {Proceedings of the IEEE/CVF Conference on Computer Vision and Pattern Recognition},
  pages = {24297--24308},
  year = {2025},
  doi = {10.1109/CVPR52734.2025.02263}
}

@inproceedings{wang2024MuirBench,
  author = {Fei Wang and Xingyu Fu and James Y. Huang and Zekun Li and Qin Liu and Xiaogeng Liu and Mingyu Derek Ma and Nan Xu and Wenxuan Zhou and Kai Zhang and Tianyi Lorena Yan and Wenjie Jacky Mo and Hsiang-Hui Liu and Pan Lu and Chunyuan Li and Chaowei Xiao and Kai-Wei Chang and Dan Roth and Sheng Zhang and Hoifung Poon and Muhao Chen},
  title = {{MuirBench}: A Comprehensive Benchmark for Robust Multi-image Understanding},
  booktitle = {Proceedings of the International Conference on Learning Representations},
  year = {2025}
}

@inproceedings{liu2024MMR,
  author = {Yexin Liu and Zhengyang Liang and Yueze Wang and Xianfeng Wu and Feilong Tang and Muyang He and Jian Li and Zheng Liu and Harry Yang and Sernam Lim and Bo Zhao},
  title = {Unveiling the Ignorance of {MLLMs}: Seeing Clearly, Answering Incorrectly},
  booktitle = {Proceedings of the IEEE/CVF Conference on Computer Vision and Pattern Recognition},
  pages = {9087--9097},
  year = {2025},
  doi = {10.1109/CVPR52734.2025.00849}
}

@inproceedings{amaduzzi2024LLaNA,
  author = {Andrea Amaduzzi and Pierluigi Zama Ramirez and Giuseppe Lisanti and Samuele Salti and Luigi Di Stefano},
  title = {{LLaNA}: Large Language and {NeRF} Assistant},
  booktitle = {Advances in Neural Information Processing Systems},
  year = {2024}
}

@inproceedings{cao2024VCog-Bench,
  author = {Xu Cao and Bolin Lai and Wenqian Ye and Yunsheng Ma and Joerg Heintz and Jintai Chen and Jianguo Cao and James M. Rehg},
  title = {What is the Visual Cognition Gap between Humans and Multimodal {LLMs}?},
  booktitle = {Proceedings of the Conference on Language Modeling},
  year = {2025}
}

@inproceedings{zhao2024VideoNIAH,
  author = {Zijia Zhao and Haoyu Lu and Yuqi Huo and Yifan Du and Tongtian Yue and Longteng Guo and Bingning Wang and Weipeng Chen and Jing Liu},
  title = {Needle In A Video Haystack: A Scalable Synthetic Evaluator for Video {MLLMs}},
  booktitle = {Proceedings of the International Conference on Learning Representations},
  year = {2025}
}

@inproceedings{he2024MMWorld,
  author = {Xuehai He and Weixi Feng and Kaizhi Zheng and Yujie Lu and Wanrong Zhu and Jiachen Li and Yue Fan and Jianfeng Wang and Linjie Li and Zhengyuan Yang and Kevin Lin and William Yang Wang and Lijuan Wang and Xin Eric Wang},
  title = {{MMWorld}: Towards Multi-discipline Multi-faceted World Model Evaluation in Videos},
  booktitle = {Proceedings of the International Conference on Learning Representations},
  year = {2025}
}

@inproceedings{zhang2024MultiTrust,
  author = {Yichi Zhang and Yao Huang and Yitong Sun and Chang Liu and Zhe Zhao and Zhengwei Fang and Yifan Wang and Huanran Chen and Xiao Yang and Xingxing Wei and Hang Su and Yinpeng Dong and Jun Zhu},
  title = {{MultiTrust}: A Comprehensive Benchmark Towards Trustworthy Multimodal Large Language Models},
  booktitle = {Advances in Neural Information Processing Systems (Datasets and Benchmarks Track)},
  year = {2024}
}

@inproceedings{pi2024image,
  author = {Pi, Renjie and Zhang, Jianshu and Zhang, Jipeng and Pan, Rui and Chen, Zhekai and Zhang, Tong},
  title = {Image Textualization: An Automatic Framework for Generating Rich and Detailed Image Descriptions},
  booktitle = {Advances in Neural Information Processing Systems (Datasets and Benchmarks Track)},
  year = {2024}
}

@inproceedings{wang2024MM-NIAH,
  author = {Weiyun Wang and Shuibo Zhang and Yiming Ren and Yuchen Duan and Tiantong Li and Shuo Liu and Mengkang Hu and Zhe Chen and Kaipeng Zhang and Lewei Lu and Xizhou Zhu and Ping Luo and Yu Qiao and Jifeng Dai and Wenqi Shao and Wenhai Wang},
  title = {Needle In A Multimodal Haystack},
  booktitle = {Advances in Neural Information Processing Systems (Datasets and Benchmarks Track)},
  year = {2024}
}

@inproceedings{zhang2024muie,
  author = {Zhang, Meishan and Fei, Hao and Wang, Bin and Wu, Shengqiong and Cao, Yixin and Li, Fei and Zhang, Min},
  title = {Recognizing Everything from All Modalities at Once: Grounded Multimodal Universal Information Extraction},
  booktitle = {Findings of the Association for Computational Linguistics: ACL 2024},
  pages = {14498--14511},
  year = {2024},
  doi = {10.18653/v1/2024.findings-acl.863}
}

@inproceedings{sun2024parrot,
  author = {Sun, Hai-Long and Zhou, Da-Wei and Li, Yang and Lu, Shiyin and Yi, Chao and Chen, Qing-Guo and Xu, Zhao and Luo, Weihua and Zhang, Kaifu and Zhan, De-Chuan and others},
  title = {Parrot: Multilingual Visual Instruction Tuning},
  booktitle = {Proceedings of the International Conference on Machine Learning},
  year = {2025}
}

@inproceedings{fu2024Video-MME,
  author = {Chaoyou Fu and Yuhan Dai and Yongdong Luo and Lei Li and Shuhuai Ren and Renrui Zhang and Zihan Wang and Chenyu Zhou and Yunhang Shen and Mengdan Zhang and Peixian Chen and Yanwei Li and Shaohui Lin and Sirui Zhao and Ke Li and Tong Xu and Xiawu Zheng and Enhong Chen and Caifeng Shan and Ran He and Xing Sun},
  title = {{Video-MME}: The First-Ever Comprehensive Evaluation Benchmark of Multi-modal {LLMs} in Video Analysis},
  booktitle = {Proceedings of the IEEE/CVF Conference on Computer Vision and Pattern Recognition},
  pages = {24108--24118},
  year = {2025},
  doi = {10.1109/CVPR52734.2025.02245}
}

@inproceedings{li2024MMUBench,
  author = {Jiaqi Li and Qianshan Wei and Chuanyi Zhang and Guilin Qi and Miaozeng Du and Yongrui Chen and Sheng Bi and Fan Liu},
  title = {Single Image Unlearning: Efficient Machine Unlearning in Multimodal Large Language Models},
  booktitle = {Advances in Neural Information Processing Systems},
  year = {2024}
}

@inproceedings{cheng2024spatialrgpt,
  author = {Cheng, An-Chieh and Yin, Hongxu and Fu, Yang and Guo, Qiushan and Yang, Ruihan and Kautz, Jan and Wang, Xiaolong and Liu, Sifei},
  title = {{SpatialRGPT}: Grounded Spatial Reasoning in Vision-Language Models},
  booktitle = {Advances in Neural Information Processing Systems},
  year = {2024}
}

@inproceedings{hengyuan2024lova3,
  author = {Zhao, Henry Hengyuan and Zhou, Pan and Gao, Difei and Shou, Mike Zheng},
  title = {{LOVA3}: Learning to Visual Question Answering, Asking and Assessment},
  booktitle = {Advances in Neural Information Processing Systems},
  year = {2024}
}

@inproceedings{tang2024mtvqa,
  author = {Jingqun Tang and Qi Liu and Yongjie Ye and Jinghui Lu and Shu Wei and An-Lan Wang and Chunhui Lin and Hao Feng and Zhen Zhao and Yanjie Wang and others},
  title = {{MTVQA}: Benchmarking Multilingual Text-Centric Visual Question Answering},
  booktitle = {Findings of the Association for Computational Linguistics: ACL 2025},
  pages = {7748--7763},
  year = {2025},
  doi = {10.18653/v1/2025.findings-acl.404}
}

@inproceedings{wang2024SOK-Bench,
  author = {Andong Wang and Bo Wu and Sunli Chen and Zhenfang Chen and Haotian Guan and Wei-Ning Lee and Li Erran Li and Chuang Gan},
  title = {{SOK-Bench}: A Situated Video Reasoning Benchmark with Aligned Open-World Knowledge},
  booktitle = {Proceedings of the IEEE/CVF Conference on Computer Vision and Pattern Recognition},
  pages = {13384--13394},
  year = {2024},
  doi = {10.1109/CVPR52733.2024.01271}
}

@inproceedings{yingmmt,
  author = {Ying, Kaining and Meng, Fanqing and Wang, Jin and Li, Zhiqian and Lin, Han and Yang, Yue and Zhang, Hao and Zhang, Wenbo and Lin, Yuqi and Liu, Shuo and others},
  title = {{MMT-Bench}: A Comprehensive Multimodal Benchmark for Evaluating Large Vision-Language Models Towards Multitask {AGI}},
  booktitle = {Proceedings of the International Conference on Machine Learning},
  pages = {57116--57198},
  year = {2024}
}

@inproceedings{jiang2024MARVEL,
  author = {Yifan Jiang and Jiarui Zhang and Kexuan Sun and Zhivar Sourati and Kian Ahrabian and Kaixin Ma and Filip Ilievski and Jay Pujara},
  title = {{MARVEL}: Multidimensional Abstraction and Reasoning through Visual Evaluation and Learning},
  booktitle = {Advances in Neural Information Processing Systems (Datasets and Benchmarks Track)},
  year = {2024}
}

@article{doris2024DesignQA,
  author = {Anna C. Doris and Daniele Grandi and Ryan Tomich and Md Ferdous Alam and Mohammadmehdi Ataei and Hyunmin Cheong and Faez Ahmed},
  title = {{DesignQA}: A Multimodal Benchmark for Evaluating Large Language Models' Understanding of Engineering Documentation},
  journal = {Journal of Computing and Information Science in Engineering},
  volume = {25},
  number = {2},
  year = {2025},
  doi = {10.1115/1.4067333}
}

@inproceedings{liu2024VisualWebBench,
  author = {Junpeng Liu and Yifan Song and Bill Yuchen Lin and Wai Lam and Graham Neubig and Yuanzhi Li and Xiang Yue},
  title = {{VisualWebBench}: How Far Have Multimodal {LLMs} Evolved in Web Page Understanding and Grounding?},
  booktitle = {Proceedings of the Conference on Language Modeling},
  year = {2024}
}

@inproceedings{you2024Ferret-UI,
  author = {Keen You and Haotian Zhang and Eldon Schoop and Floris Weers and Amanda Swearngin and Jeffrey Nichols and Yinfei Yang and Zhe Gan},
  title = {{Ferret-UI}: Grounded Mobile {UI} Understanding with Multimodal {LLMs}},
  booktitle = {Proceedings of the European Conference on Computer Vision},
  pages = {240--255},
  year = {2024},
  doi = {10.1007/978-3-031-73039-9_14}
}

@inproceedings{luo2024JailBreakV-28K,
  author = {Weidi Luo and Siyuan Ma and Xiaogeng Liu and Xiaoyu Guo and Chaowei Xiao},
  title = {{JailBreakV}: A Benchmark for Assessing the Robustness of {MultiModal} Large Language Models against Jailbreak Attacks},
  booktitle = {Proceedings of the Conference on Language Modeling},
  year = {2024}
}

@inproceedings{lin2024MDVP-Bench,
  author = {Weifeng Lin and Xinyu Wei and Ruichuan An and Peng Gao and Bocheng Zou and Yulin Luo and Siyuan Huang and Shanghang Zhang and Hongsheng Li},
  title = {Draw-and-Understand: Leveraging Visual Prompts to Enable {MLLMs} to Comprehend What You Want},
  booktitle = {Proceedings of the International Conference on Learning Representations},
  year = {2025}
}

@inproceedings{chen2024MMStar,
  author = {Lin Chen and Jinsong Li and Xiaoyi Dong and Pan Zhang and Yuhang Zang and Zehui Chen and Haodong Duan and Jiaqi Wang and Yu Qiao and Dahua Lin and Feng Zhao},
  title = {Are We on the Right Way for Evaluating Large Vision-Language Models?},
  booktitle = {Advances in Neural Information Processing Systems},
  year = {2024}
}

@inproceedings{shao2024VisualCoT,
  author = {Hao Shao and Shengju Qian and Han Xiao and Guanglu Song and Zhuofan Zong and Letian Wang and Yu Liu and Hongsheng Li},
  title = {Visual {CoT}: Advancing Multi-Modal Language Models with a Comprehensive Dataset and Benchmark for Chain-of-Thought Reasoning},
  booktitle = {Advances in Neural Information Processing Systems (Datasets and Benchmarks Track)},
  year = {2024}
}

@inproceedings{zong2024VL-ICLBench,
  author = {Yongshuo Zong and Ondrej Bohdal and Timothy Hospedales},
  title = {{VL-ICL} Bench: The Devil in the Details of Multimodal In-Context Learning},
  booktitle = {Proceedings of the International Conference on Learning Representations},
  year = {2025}
}

@inproceedings{wang2024CRPE,
  author = {Weiyun Wang and Yiming Ren and Haowen Luo and Tiantong Li and Chenxiang Yan and Zhe Chen and Wenhai Wang and Qingyun Li and Lewei Lu and Xizhou Zhu and Yu Qiao and Jifeng Dai},
  title = {The All-Seeing Project {V2}: Towards General Relation Comprehension of the Open World},
  booktitle = {Proceedings of the European Conference on Computer Vision},
  pages = {471--490},
  year = {2024},
  doi = {10.1007/978-3-031-73414-4_27}
}

@inproceedings{huang2024VLKEB,
  author = {Han Huang and Haitian Zhong and Tao Yu and Qiang Liu and Shu Wu and Liang Wang and Tieniu Tan},
  title = {{VLKEB}: A Large Vision-Language Model Knowledge Editing Benchmark},
  booktitle = {Advances in Neural Information Processing Systems (Datasets and Benchmarks Track)},
  year = {2024}
}

@inproceedings{zhang2024mc-mke,
  author = {Zhang, Junzhe and Zhang, Huixuan and Yin, Xunjian and Huang, Baizhou and Zhang, Xu and Hu, Xinyu and Wan, Xiaojun},
  title = {{MC-MKE}: A Fine-Grained Multimodal Knowledge Editing Benchmark Emphasizing Modality Consistency},
  booktitle = {Findings of the Association for Computational Linguistics: ACL 2025},
  pages = {17430--17445},
  year = {2025},
  doi = {10.18653/v1/2025.findings-acl.896}
}

@inproceedings{liu2024TempCompass,
  author = {Yuanxin Liu and Shicheng Li and Yi Liu and Yuxiang Wang and Shuhuai Ren and Lei Li and Sishuo Chen and Xu Sun and Lu Hou},
  title = {{TempCompass}: Do Video {LLMs} Really Understand Videos?},
  booktitle = {Findings of the Association for Computational Linguistics: ACL 2024},
  pages = {8731--8772},
  year = {2024},
  doi = {10.18653/v1/2024.findings-acl.517}
}

@inproceedings{nguyen2024oscar,
  author = {Nguyen, Nguyen and Bi, Jing and Vosoughi, Ali and Tian, Yapeng and Fazli, Pooyan and Xu, Chenliang},
  title = {{OSCaR}: Object State Captioning and State Change Representation},
  booktitle = {Findings of the Association for Computational Linguistics: NAACL 2024},
  pages = {3565--3576},
  year = {2024},
  doi = {10.18653/v1/2024.findings-naacl.226}
}

@inproceedings{huang2024VHTest,
  author = {Huang, Wen and Liu, Hongbin and Guo, Minxin and Gong, Neil Zhenqiang},
  title = {Visual Hallucinations of Multi-modal Large Language Models},
  booktitle = {Findings of the Association for Computational Linguistics: ACL 2024},
  pages = {9614--9631},
  year = {2024},
  doi = {10.18653/v1/2024.findings-acl.573}
}

@article{cao2023genception,
  author = {Lele Cao and Valentin Buchner and Zineb Senane and Fangkai Yang},
  title = {{GenCeption}: Evaluate Vision {LLMs} with Unlabeled Unimodal Data},
  journal = {Computer Speech \& Language},
  volume = {93},
  pages = {101785},
  year = {2025},
  doi = {10.1016/j.csl.2025.101785}
}

@inproceedings{luo2024codis,
  author = {Luo, Fuwen and Chen, Chi and Wan, Zihao and Kang, Zhaolu and Yan, Qidong and Li, Yingjie and Wang, Xiaolong and Wang, Siyu and Wang, Ziyue and Mi, Xiaoyue and others},
  title = {{CODIS}: Benchmarking Context-Dependent Visual Comprehension for Multimodal Large Language Models},
  booktitle = {Proceedings of the Annual Meeting of the Association for Computational Linguistics},
  pages = {10639--10659},
  year = {2024},
  doi = {10.18653/v1/2024.acl-long.573}
}

@article{xia2024chartx,
  author = {Renqiu Xia and Hancheng Ye and Xiangchao Yan and Qi Liu and Hongbin Zhou and Zijun Chen and Botian Shi and Junchi Yan and Bo Zhang},
  title = {{ChartX} and {ChartVLM}: A Versatile Benchmark and Foundation Model for Complicated Chart Reasoning},
  journal = {IEEE Transactions on Image Processing},
  volume = {34},
  pages = {7436--7447},
  year = {2025},
  doi = {10.1109/TIP.2025.3607618}
}

@inproceedings{chen2024mcub,
  author = {Chen, Chi and Du, Yiyang and Fang, Zheng and Wang, Ziyue and Luo, Fuwen and Li, Peng and Yan, Ming and Zhang, Ji and Huang, Fei and Sun, Maosong and others},
  title = {Model Composition for Multimodal Large Language Models},
  booktitle = {Proceedings of the Annual Meeting of the Association for Computational Linguistics},
  pages = {11246--11262},
  year = {2024},
  doi = {10.18653/v1/2024.acl-long.606}
}

@inproceedings{jin2024mm-soc,
  author = {Jin, Yiqiao and Choi, Minje and Verma, Gaurav and Wang, Jindong and Kumar, Srijan},
  title = {{MM-Soc}: Benchmarking Multimodal Large Language Models in Social Media Platforms},
  booktitle = {Findings of the Association for Computational Linguistics: ACL 2024},
  pages = {6192--6210},
  year = {2024},
  doi = {10.18653/v1/2024.findings-acl.370}
}

@inproceedings{li2024mike,
  author = {Li, Jiaqi and Du, Miaozeng and Zhang, Chuanyi and Chen, Yongrui and Hu, Nan and Qi, Guilin and Jiang, Haiyun and Cheng, Siyuan and Tian, Bozhong},
  title = {{MIKE}: A New Benchmark for Fine-grained Multimodal Entity Knowledge Editing},
  booktitle = {Findings of the Association for Computational Linguistics: ACL 2024},
  pages = {5018--5029},
  year = {2024},
  doi = {10.18653/v1/2024.findings-acl.298}
}

@inproceedings{wang2024asclepius,
  author = {Jie Liu and Wenxuan Wang and Yihang Su and Jingyuan Huang and Yudi Zhang and Cheng-Yi Li and Wenting Chen and Xiaohan Xing and Kao-Jung Chang and Linlin Shen and others},
  title = {Asclepius: A Spectrum Evaluation Benchmark for Medical Multi-Modal Large Language Models},
  booktitle = {Proceedings of the Annual Meeting of the Association for Computational Linguistics},
  pages = {24181--24201},
  year = {2025},
  doi = {10.18653/v1/2025.acl-long.1178}
}

@inproceedings{cha2024visually,
  author = {Cha, Sungguk and Lee, Jusung and Lee, Younghyun and Yang, Cheoljong},
  title = {Visually Dehallucinative Instruction Generation},
  booktitle = {Proceedings of the IEEE International Conference on Acoustics, Speech and Signal Processing},
  pages = {5510--5514},
  year = {2024},
  doi = {10.1109/ICASSP48485.2024.10446658}
}

@article{zhang2024q-bench,
  author = {Zhang, Zicheng and Wu, Haoning and Zhang, Erli and Zhai, Guangtao and Lin, Weisi},
  title = {{Q-Bench+}: A Benchmark for Multi-Modal Foundation Models on Low-Level Vision From Single Images to Pairs},
  journal = {IEEE Transactions on Pattern Analysis and Machine Intelligence},
  volume = {46},
  number = {12},
  pages = {10404--10418},
  year = {2024},
  doi = {10.1109/TPAMI.2024.3445770}
}

@inproceedings{yang2024air,
  author = {Yang, Qian and Xu, Jin and Liu, Wenrui and Chu, Yunfei and Jiang, Ziyue and Zhou, Xiaohuan and Leng, Yichong and Lv, Yuanjun and Zhao, Zhou and Zhou, Chang and others},
  title = {{AIR-Bench}: Benchmarking Large Audio-Language Models via Generative Comprehension},
  booktitle = {Proceedings of the Annual Meeting of the Association for Computational Linguistics},
  pages = {1979--1998},
  year = {2024},
  doi = {10.18653/v1/2024.acl-long.109}
}

@article{shi2024shield,
  author = {Shi, Yichen and Gao, Yuhao and Lai, Yingxin and Wang, Hongyang and Feng, Jun and He, Lei and Wan, Jun and Chen, Changsheng and Yu, Zitong and Cao, Xiaochun},
  title = {{SHIELD}: An Evaluation Benchmark for Face Spoofing and Forgery Detection with Multimodal Large Language Models},
  journal = {Visual Intelligence},
  volume = {3},
  number = {1},
  year = {2025},
  doi = {10.1007/s44267-025-00079-w}
}

@inproceedings{han2024instinctive,
  author = {Han, Tianyang and Lian, Qing and Pan, Rui and Pi, Renjie and Zhang, Jipeng and Diao, Shizhe and Lin, Yong and Zhang, Tong},
  title = {The Instinctive Bias: Spurious Images lead to Illusion in {MLLMs}},
  booktitle = {Proceedings of the Conference on Empirical Methods in Natural Language Processing},
  pages = {16163--16177},
  year = {2024},
  doi = {10.18653/v1/2024.emnlp-main.904}
}

@inproceedings{muhtar2024lhrs,
  author = {Muhtar, Dilxat and Li, Zhenshi and Gu, Feng and Zhang, Xueliang and Xiao, Pengfeng},
  title = {{LHRS-Bot}: Empowering Remote Sensing with {VGI-Enhanced} Large Multimodal Language Model},
  booktitle = {Proceedings of the European Conference on Computer Vision},
  pages = {440--457},
  year = {2024},
  doi = {10.1007/978-3-031-72904-1_26}
}

@inproceedings{li2024red,
  author = {Li, Mukai and Li, Lei and Yin, Yuwei and Ahmed, Masood and Liu, Zhenguang and Liu, Qi},
  title = {Red Teaming Visual Language Models},
  booktitle = {Findings of the Association for Computational Linguistics: ACL 2024},
  pages = {3326--3342},
  year = {2024},
  doi = {10.18653/v1/2024.findings-acl.198}
}

@inproceedings{tong2024mmvp,
  author = {Tong, Shengbang and Liu, Zhuang and Zhai, Yuexiang and Ma, Yi and LeCun, Yann and Xie, Saining},
  title = {Eyes Wide Shut? {Exploring} the Visual Shortcomings of Multimodal {LLMs}},
  booktitle = {Proceedings of the IEEE/CVF Conference on Computer Vision and Pattern Recognition},
  pages = {9568--9578},
  year = {2024},
  doi = {10.1109/CVPR52733.2024.00914}
}

@inproceedings{wang2024mm-sap,
  author = {Wang, Yuhao and Liao, Yusheng and Liu, Heyang and Liu, Hongcheng and Wang, Yanfeng and Wang, Yu},
  title = {{MM-SAP}: A Comprehensive Benchmark for Assessing Self-Awareness of Multimodal Large Language Models in Perception},
  booktitle = {Proceedings of the Annual Meeting of the Association for Computational Linguistics},
  pages = {9192--9205},
  year = {2024},
  doi = {10.18653/v1/2024.acl-long.498}
}

@inproceedings{wang2024mementos,
  author = {Wang, Xiyao and Zhou, Yuhang and Liu, Xiaoyu and Lu, Hongjin and Xu, Yuancheng and He, Feihong and Yoon, Jaehong and Lu, Taixi and Liu, Fuxiao and Bertasius, Gedas and Bansal, Mohit and Yao, Huaxiu and Huang, Furong},
  title = {Mementos: A Comprehensive Benchmark for Multimodal Large Language Model Reasoning over Image Sequences},
  booktitle = {Proceedings of the Annual Meeting of the Association for Computational Linguistics},
  pages = {416--442},
  year = {2024},
  doi = {10.18653/v1/2024.acl-long.25}
}

@misc{wang2023llm,
  author = {Wang, Junyang and Wang, Yuhang and Xu, Guohai and Zhang, Jing and Gu, Yukai and Jia, Haitao and Yan, Ming and Zhang, Ji and Sang, Jitao},
  title = {{AMBER}: An {LLM-free} Multi-dimensional Benchmark for {MLLMs} Hallucination Evaluation},
  howpublished = {arXiv preprint arXiv:2311.07397},
  year = {2023}
}

@inproceedings{romero2024cvqa,
  author = {David Romero and Chenyang Lyu and Haryo Akbarianto Wibowo and Santiago G{\'o}ngora and Aishik Mandal and Sukannya Purkayastha and Jes{\'u}s-Germ{\'a}n Ortiz-Barajas and Emilio Villa-Cueva and Jinheon Baek and Soyeong Jeong and others},
  title = {{CVQA}: Culturally-diverse Multilingual Visual Question Answering Benchmark},
  booktitle = {Advances in Neural Information Processing Systems (Datasets and Benchmarks Track)},
  year = {2024}
}

@inproceedings{alwajih2024peacock,
  author = {Alwajih, Fakhraddin and Nagoudi, El Moatez Billah and Bhatia, Gagan and Mohamed, Abdelrahman and Abdul-Mageed, Muhammad},
  title = {Peacock: A Family of {Arabic} Multimodal Large Language Models and Benchmarks},
  booktitle = {Proceedings of the Annual Meeting of the Association for Computational Linguistics},
  pages = {12753--12776},
  year = {2024},
  doi = {10.18653/v1/2024.acl-long.689}
}

@inproceedings{wu23q-bench,
  author = {Wu, Haoning and Zhang, Zicheng and Zhang, Erli and Chen, Chaofeng and Liao, Liang and Wang, Annan and Li, Chunyi and Sun, Wenxiu and Yan, Qiong and Zhai, Guangtao and others},
  title = {{Q-Bench}: A Benchmark for General-Purpose Foundation Models on Low-level Vision},
  booktitle = {Proceedings of the International Conference on Learning Representations},
  year = {2024}
}

@inproceedings{song2024moviechat,
  author = {Song, Enxin and Chai, Wenhao and Wang, Guanhong and Zhang, Yucheng and Zhou, Haoyang and Wu, Feiyang and Chi, Haozhe and Guo, Xun and Ye, Tian and Zhang, Yanting and others},
  title = {{MovieChat}: From Dense Token to Sparse Memory for Long Video Understanding},
  booktitle = {Proceedings of the IEEE/CVF Conference on Computer Vision and Pattern Recognition},
  pages = {18221--18232},
  year = {2024},
  doi = {10.1109/CVPR52733.2024.01725}
}

@inproceedings{li2023pope,
  author = {Li, Yifan and Du, Yifan and Zhou, Kun and Wang, Jinpeng and Zhao, Xin and Wen, Ji-Rong},
  title = {Evaluating Object Hallucination in Large Vision-Language Models},
  booktitle = {Proceedings of the Conference on Empirical Methods in Natural Language Processing},
  pages = {292--305},
  year = {2023},
  doi = {10.18653/v1/2023.emnlp-main.20}
}

@article{liu2024ocrbench,
  author = {Yuliang Liu and Zhang Li and Mingxin Huang and Biao Yang and Wenwen Yu and Chunyuan Li and Xucheng Yin and Cheng-Lin Liu and Lianwen Jin and Xiang Bai},
  title = {{OCRBench}: On the Hidden Mystery of {OCR} in Large Multimodal Models},
  journal = {Science China Information Sciences},
  volume = {67},
  number = {12},
  year = {2024},
  doi = {10.1007/s11432-024-4235-6}
}

@inproceedings{zou2024implicitave,
  author = {Zou, Henry Peng and Samuel, Vinay and Zhou, Yue and Zhang, Weizhi and Fang, Liancheng and Song, Zihe and Yu, Philip S and Caragea, Cornelia},
  title = {{ImplicitAVE}: An Open-Source Dataset and Multimodal {LLMs} Benchmark for Implicit Attribute Value Extraction},
  booktitle = {Findings of the Association for Computational Linguistics: ACL 2024},
  pages = {338--354},
  year = {2024},
  doi = {10.18653/v1/2024.findings-acl.20}
}

@misc{betker2023dalle,
  author = {Betker, James and Goh, Gabriel and Jing, Li and Brooks, Tim and Wang, Jianfeng and Li, Linjie and Ouyang, Long and Zhuang, Juntang and Lee, Joyce and Guo, Yufei and others},
  title = {Improving image generation with better captions},
  howpublished = {OpenAI technical report, \url{https://cdn.openai.com/papers/dall-e-3.pdf}},
  year = {2023}
}

@misc{contributors2023opencompass,
  author = {{OpenCompass Contributors}},
  title = {{OpenCompass}: A Universal Evaluation Platform for Foundation Models},
  howpublished = {GitHub repository, \url{https://github.com/open-compass/opencompass}},
  year = {2023}
}

@inproceedings{zhu2024scanreason,
  author = {Chenming Zhu and Tai Wang and Wenwei Zhang and Kai Chen and Xihui Liu},
  title = {{ScanReason}: Empowering {3D} Visual Grounding with Reasoning Capabilities},
  booktitle = {Proceedings of the European Conference on Computer Vision},
  pages = {151--168},
  year = {2024},
  doi = {10.1007/978-3-031-73242-3_9}
}

@inproceedings{qu2024unified,
  author = {Qu, Leigang and Li, Haochuan and Wang, Tan and Wang, Wenjie and Li, Yongqi and Nie, Liqiang and Chua, Tat-Seng},
  title = {{TIGeR}: Unifying Text-to-Image Generation and Retrieval with Large Multimodal Models},
  booktitle = {Proceedings of the International Conference on Learning Representations},
  year = {2025}
}

@inproceedings{heusel2017fid,
  author = {Heusel, Martin and Ramsauer, Hubert and Unterthiner, Thomas and Nessler, Bernhard and Hochreiter, Sepp},
  title = {{GANs} Trained by a Two Time-Scale Update Rule Converge to a Local Nash Equilibrium},
  booktitle = {Advances in Neural Information Processing Systems},
  pages = {6626--6637},
  year = {2017}
}

@inproceedings{salimans2016improved,
  author = {Salimans, Tim and Goodfellow, Ian and Zaremba, Wojciech and Cheung, Vicki and Radford, Alec and Chen, Xi},
  title = {Improved Techniques for Training {GANs}},
  booktitle = {Advances in Neural Information Processing Systems},
  pages = {2226--2234},
  year = {2016}
}

@inproceedings{schuhmann2022laion,
  author = {Schuhmann, Christoph and Beaumont, Romain and Vencu, Richard and Gordon, Cade and Wightman, Ross and Cherti, Mehdi and Coombes, Theo and Katta, Aarush and Mullis, Clayton and Wortsman, Mitchell and others},
  title = {{LAION-5B}: An open large-scale dataset for training next generation image-text models},
  booktitle = {Advances in Neural Information Processing Systems},
  year = {2022}
}

@inproceedings{gadre2024datacomp,
  author = {Gadre, Samir Yitzhak and Ilharco, Gabriel and Fang, Alex and Hayase, Jonathan and Smyrnis, Georgios and Nguyen, Thao and Marten, Ryan and Wortsman, Mitchell and Ghosh, Dhruba and Zhang, Jieyu and others},
  title = {{DataComp}: In search of the next generation of multimodal datasets},
  booktitle = {Advances in Neural Information Processing Systems},
  year = {2023}
}

@inproceedings{lin2014microsoft,
  author = {Lin, Tsung-Yi and Maire, Michael and Belongie, Serge and Hays, James and Perona, Pietro and Ramanan, Deva and Doll{\'a}r, Piotr and Zitnick, C Lawrence},
  title = {Microsoft {COCO}: Common Objects in Context},
  booktitle = {Proceedings of the European Conference on Computer Vision},
  pages = {740--755},
  year = {2014},
  doi = {10.1007/978-3-319-10602-1_48}
}

@inproceedings{sepehri2025mediconfusion,
  author = {Sepehri, Mohammad Shahab and Fabian, Zalan and Soltanolkotabi, Maryam and Soltanolkotabi, Mahdi},
  title = {{MediConfusion}: Can you trust your {AI} radiologist? {Probing} the reliability of multimodal medical foundation models},
  booktitle = {Proceedings of the International Conference on Learning Representations},
  year = {2025}
}

@inproceedings{bai2024tokenseg,
  author = {Bai, Zechen and He, Tong and Mei, Haiyang and Wang, Pichao and Gao, Ziteng and Chen, Joya and Liu, Lei and Zhang, Zheng and Shou, Mike Zheng},
  title = {One Token to Seg Them All: Language Instructed Reasoning Segmentation in Videos},
  booktitle = {Advances in Neural Information Processing Systems},
  year = {2024}
}

@inproceedings{ramakrishnan2025spatialcognition,
  author = {Santhosh Kumar Ramakrishnan and Erik Wijmans and Philipp Kr{\"a}henb{\"u}hl and Vladlen Koltun},
  title = {Does Spatial Cognition Emerge in Frontier Models?},
  booktitle = {Proceedings of the International Conference on Learning Representations},
  year = {2025}
}

@inproceedings{hu2025mragbenchvisioncentricevaluationretrievalaugmented,
  author = {Wenbo Hu and Jia-Chen Gu and Zi-Yi Dou and Mohsen Fayyaz and Pan Lu and Kai-Wei Chang and Nanyun Peng},
  title = {{MRAG-Bench}: Vision-Centric Evaluation for Retrieval-Augmented Multimodal Models},
  booktitle = {Proceedings of the International Conference on Learning Representations},
  year = {2025}
}

@inproceedings{zou2024sportu,
  author = {Haotian Xia and Zhengbang Yang and Junbo Zou and Rhys Tracy and Yuqing Wang and Chi Lu and Christopher Lai and Yanjun He and Xun Shao and Zhuoqing Xie and Yuan-Fang Wang and Weining Shen and Hanjie Chen},
  title = {{SPORTU}: A Comprehensive Sports Understanding Benchmark for Multimodal Large Language Models},
  booktitle = {Proceedings of the International Conference on Learning Representations},
  year = {2025}
}

@inproceedings{jiang2025mmad,
  author = {Xi Jiang and Jian Li and Hanqiu Deng and Yong Liu and Bin-Bin Gao and Yifeng Zhou and Jialin Li and Chengjie Wang and Feng Zheng},
  title = {{MMAD}: A Comprehensive Benchmark for Multimodal Large Language Models in Industrial Anomaly Detection},
  booktitle = {Proceedings of the International Conference on Learning Representations},
  year = {2025}
}

@inproceedings{xu2025mcbench,
  author = {Yunqiu Xu and Linchao Zhu and Yi Yang},
  title = {{MC-Bench}: A Benchmark for Multi-Context Visual Grounding in the Era of {MLLMs}},
  booktitle = {Proceedings of the IEEE/CVF International Conference on Computer Vision},
  pages = {17675--17687},
  year = {2025},
  doi = {10.1109/ICCV51701.2025.01642}
}

@inproceedings{kim2025avhbench,
  author = {Kim Sung-Bin and Oh Hyun-Bin and JungMok Lee and Arda Senocak and Joon Son Chung and Tae-Hyun Oh},
  title = {{AVHBench}: A Cross-Modal Hallucination Benchmark for Audio-Visual Large Language Models},
  booktitle = {Proceedings of the International Conference on Learning Representations},
  year = {2025}
}

@inproceedings{li2025eee-bench,
  author = {Ming Li and Jike Zhong and Tianle Chen and Yuxiang Lai and Konstantinos Psounis},
  title = {{EEE-Bench}: A Comprehensive Multimodal Electrical And Electronics Engineering Benchmark},
  booktitle = {Proceedings of the IEEE/CVF Conference on Computer Vision and Pattern Recognition},
  pages = {13337--13349},
  year = {2025},
  doi = {10.1109/CVPR52734.2025.01245}
}

@inproceedings{tang2025vidcomposition,
  author = {Yolo Y. Tang and Junjia Guo and Hang Hua and Susan Liang and Mingqian Feng and Xinyang Li and Rui Mao and Chao Huang and Jing Bi and Zeliang Zhang and Pooyan Fazli and Chenliang Xu},
  title = {{VidComposition}: Can {MLLMs} Analyze Compositions in Compiled Videos?},
  booktitle = {Proceedings of the IEEE/CVF Conference on Computer Vision and Pattern Recognition},
  pages = {8490--8500},
  year = {2025},
  doi = {10.1109/CVPR52734.2025.00794}
}

@inproceedings{li2025vlrewardbench,
  author = {Lei Li and Yuancheng Wei and Zhihui Xie and Xuqing Yang and Yifan Song and Peiyi Wang and Chenxin An and Tianyu Liu and Sujian Li and Bill Yuchen Lin and Lingpeng Kong and Qi Liu},
  title = {{VL-RewardBench}: A Challenging Benchmark for Vision-Language Generative Reward Models},
  booktitle = {Proceedings of the IEEE/CVF Conference on Computer Vision and Pattern Recognition},
  pages = {24657--24668},
  year = {2025},
  doi = {10.1109/CVPR52734.2025.02296}
}

@inproceedings{liang2025wsillava,
  author = {Yuci Liang and Xinheng Lyu and Wenting Chen and Meidan Ding and Jipeng Zhang and Xiangjian He and Song Wu and Xiaohan Xing and Sen Yang and Xiyue Wang and Linlin Shen},
  title = {{WSI-LLaVA}: A Multimodal Large Language Model for Whole Slide Image},
  booktitle = {Proceedings of the IEEE/CVF International Conference on Computer Vision},
  pages = {22718--22727},
  year = {2025},
  doi = {10.1109/ICCV51701.2025.02109}
}

@inproceedings{zhang2025ohrbench,
  author = {Junyuan Zhang and Qintong Zhang and Bin Wang and Linke Ouyang and Zichen Wen and Ying Li and Ka-Ho Chow and Conghui He and Wentao Zhang},
  title = {{OCR} Hinders {RAG}: Evaluating the Cascading Impact of {OCR} on Retrieval-Augmented Generation},
  booktitle = {Proceedings of the IEEE/CVF International Conference on Computer Vision},
  pages = {17443--17453},
  year = {2025},
  doi = {10.1109/ICCV51701.2025.01620}
}

@inproceedings{li2024vidhalluc,
  author = {Chaoyu Li and Eun Woo Im and Pooyan Fazli},
  title = {{VidHalluc}: Evaluating Temporal Hallucinations in Multimodal Large Language Models for Video Understanding},
  booktitle = {Proceedings of the IEEE/CVF Conference on Computer Vision and Pattern Recognition},
  pages = {13723--13733},
  year = {2025},
  doi = {10.1109/CVPR52734.2025.01281}
}

@inproceedings{ouyang2025punchbench,
  author = {Kun Ouyang and Yuanxin Liu and Shicheng Li and Yi Liu and Hao Zhou and Fandong Meng and Jie Zhou and Xu Sun},
  title = {{PunchBench}: Benchmarking {MLLMs} in Multimodal Punchline Comprehension},
  booktitle = {Proceedings of the Annual Meeting of the Association for Computational Linguistics},
  pages = {986--1008},
  year = {2025},
  doi = {10.18653/v1/2025.acl-long.49}
}

@inproceedings{alhamoud2025negbench,
  author = {Kumail Alhamoud and Shaden Alshammari and Yonglong Tian and Guohao Li and Philip Torr and Yoon Kim and Marzyeh Ghassemi},
  title = {Vision-Language Models Do Not Understand Negation},
  booktitle = {Proceedings of the IEEE/CVF Conference on Computer Vision and Pattern Recognition},
  pages = {29612--29622},
  year = {2025},
  doi = {10.1109/CVPR52734.2025.02757}
}

@inproceedings{zhou2025egotextvqa,
  author = {Sheng Zhou and Junbin Xiao and Qingyun Li and Yicong Li and Xun Yang and Dan Guo and Meng Wang and Tat-Seng Chua and Angela Yao},
  title = {{EgoTextVQA}: Towards Egocentric Scene-Text Aware Video Question Answering},
  booktitle = {Proceedings of the IEEE/CVF Conference on Computer Vision and Pattern Recognition},
  pages = {3363--3373},
  year = {2025},
  doi = {10.1109/CVPR52734.2025.00319}
}

@inproceedings{wang2025mmsafeaware,
  author = {Wenxuan Wang and Xiaoyuan Liu and Kuiyi Gao and Jen-Tse Huang and Youliang Yuan and Pinjia He and Shuai Wang and Zhaopeng Tu},
  title = {Can't See the Forest for the Trees: Benchmarking Multimodal Safety Awareness for Multimodal {LLMs}},
  booktitle = {Proceedings of the Annual Meeting of the Association for Computational Linguistics},
  pages = {16993--17006},
  year = {2025},
  doi = {10.18653/v1/2025.acl-long.832}
}

@inproceedings{su2025essayjudge,
  author = {Jiamin Su and Yibo Yan and Fangteng Fu and Han Zhang and Jingheng Ye and Xiang Liu and Jiahao Huo and Huiyu Zhou and Xuming Hu},
  title = {{EssayJudge}: A Multi-Granular Benchmark for Assessing Automated Essay Scoring Capabilities of Multimodal Large Language Models},
  booktitle = {Findings of the Association for Computational Linguistics: ACL 2025},
  pages = {6363--6389},
  year = {2025},
  doi = {10.18653/v1/2025.findings-acl.329}
}

@inproceedings{du2025mmke-bench,
  author = {Yuntao Du and Kailin Jiang and Zhi Gao and Chenrui Shi and Zilong Zheng and Siyuan Qi and Qing Li},
  title = {{MMKE-Bench}: A Multimodal Editing Benchmark for Diverse Visual Knowledge},
  booktitle = {Proceedings of the International Conference on Learning Representations},
  year = {2025}
}

@inproceedings{yilmaz2025voila,
  author = {Nilay Yilmaz and Maitreya Patel and Yiran Lawrence Luo and Tejas Gokhale and Chitta Baral and Suren Jayasuriya and Yezhou Yang},
  title = {{VOILA}: Evaluation of {MLLMs} For Perceptual Understanding and Analogical Reasoning},
  booktitle = {Proceedings of the International Conference on Learning Representations},
  year = {2025}
}

@inproceedings{burgess2025microvqa,
  author = {James Burgess and Jeffrey J. Nirschl and Laura Bravo-S{\'a}nchez and Alejandro Lozano and Sanket Rajan Gupte and Jesus G. Galaz-Montoya and Yuhui Zhang and Yuchang Su and Disha Bhowmik and Zachary Coman and Sarina M. Hasan and Alexandra Johannesson and William D. Leineweber and Malvika G. Nair and Ridhi Yarlagadda and Connor Zuraski and Wah Chiu and Sarah Cohen and Jan N. Hansen and Manuel D. Leonetti and Chad Liu and Emma Lundberg and Serena Yeung-Levy},
  title = {{MicroVQA}: A Multimodal Reasoning Benchmark for Microscopy-Based Scientific Research},
  booktitle = {Proceedings of the IEEE/CVF Conference on Computer Vision and Pattern Recognition},
  pages = {19552--19564},
  year = {2025},
  doi = {10.1109/CVPR52734.2025.01821}
}

@inproceedings{weng2025visnumbench,
  author = {Tengjin Weng and Jingyi Wang and Wenhao Jiang and Zhong Ming},
  title = {{VisNumBench}: Evaluating Number Sense of Multimodal Large Language Models},
  booktitle = {Proceedings of the IEEE/CVF International Conference on Computer Vision},
  pages = {3830--3840},
  year = {2025},
  doi = {10.1109/ICCV51701.2025.00365}
}

@inproceedings{zhang2025irefvla,
  author = {Haochen Zhang and Nader Zantout and Pujith Kachana and Ji Zhang and Wenshan Wang},
  title = {{IRef-VLA}: A Benchmark for Interactive Referential Grounding with Imperfect Language in {3D} Scenes},
  booktitle = {Proceedings of the IEEE International Conference on Robotics and Automation},
  pages = {1677--1683},
  year = {2025},
  doi = {10.1109/ICRA55743.2025.11127464}
}

@inproceedings{yang2025avut,
  author = {Yudong Yang and Jimin Zhuang and Guangzhi Sun and Changli Tang and Yixuan Li and Peihan Li and Yifan Jiang and Wei Li and Zejun Ma and Chao Zhang},
  title = {Audio-centric Video Understanding Benchmark without Text Shortcut},
  booktitle = {Proceedings of the Conference on Empirical Methods in Natural Language Processing},
  pages = {6569--6587},
  year = {2025},
  doi = {10.18653/v1/2025.emnlp-main.333}
}

@inproceedings{wang2025xlrsbench,
  author = {Fengxiang Wang and Hongzhen Wang and Mingshuo Chen and Di Wang and Yulin Wang and Zonghao Guo and Qiang Ma and Long Lan and Wenjing Yang and Jing Zhang and Zhiyuan Liu and Maosong Sun},
  title = {{XLRS-Bench}: Could Your Multimodal {LLMs} Understand Extremely Large Ultra-High-Resolution Remote Sensing Imagery?},
  booktitle = {Proceedings of the IEEE/CVF Conference on Computer Vision and Pattern Recognition},
  pages = {14325--14336},
  year = {2025},
  doi = {10.1109/CVPR52734.2025.01336}
}

@inproceedings{xun2025rtv,
  author = {Shuhang Xun and Sicheng Tao and Jungang Li and Yibo Shi and Zhixin Lin and Zhanhui Zhu and Yibo Yan and Hanqian Li and Linghao Zhang and Shikang Wang and Yixin Liu and Hanbo Zhang and Ying Ma and Xuming Hu},
  title = {{RTV-Bench}: Benchmarking {MLLM} Continuous Perception, Understanding and Reasoning through Real-Time Video},
  booktitle = {Advances in Neural Information Processing Systems (Datasets and Benchmarks Track)},
  year = {2025}
}

@article{guo2025omnigirl,
  author = {Lianghong Guo and Wei Tao and Runhan Jiang and Yanlin Wang and Jiachi Chen and Xilin Liu and Yuchi Ma and Mingzhi Mao and Hongyu Zhang and Zibin Zheng},
  title = {{OmniGIRL}: A Multilingual and Multimodal Benchmark for {GitHub} Issue Resolution},
  journal = {Proceedings of the ACM on Software Engineering},
  volume = {2},
  number = {ISSTA},
  pages = {24--46},
  year = {2025},
  doi = {10.1145/3728871}
}

@inproceedings{lei2025godbench,
  author = {Yiming Lei and Chenkai Zhang and Zeming Liu and Haitao Leng and Shaoguo Liu and Tingting Gao and Qingjie Liu and Yunhong Wang},
  title = {{GODBench}: A Benchmark for Multimodal Large Language Models in Video Comment Art},
  booktitle = {Proceedings of the Annual Meeting of the Association for Computational Linguistics},
  pages = {11884--11952},
  year = {2025},
  doi = {10.18653/v1/2025.acl-long.583}
}

@inproceedings{liu2025videosafetybench,
  author = {Xuannan Liu and Zekun Li and Zheqi He and Peipei Li and Shuhan Xia and Xing Cui and Huaibo Huang and Xi Yang and Ran He},
  title = {{Video-SafetyBench}: A Benchmark for Safety Evaluation of Video {LVLMs}},
  booktitle = {Advances in Neural Information Processing Systems (Datasets and Benchmarks Track)},
  year = {2025}
}

@inproceedings{zhao2025chartedit,
  author = {Xuanle Zhao and Xuexin Liu and Haoyue Yang and Xianzhen Luo and Fanhu Zeng and Jianling Li and Qi Shi and Chi Chen},
  title = {{ChartEdit}: How Far Are {MLLMs} From Automating Chart Analysis? {Evaluating} {MLLMs}' Capability via Chart Editing},
  booktitle = {Findings of the Association for Computational Linguistics: ACL 2025},
  pages = {3616--3630},
  year = {2025},
  doi = {10.18653/v1/2025.findings-acl.185}
}

@inproceedings{wan2025instructpart,
  author = {Zifu Wan and Yaqi Xie and Ce Zhang and Zhiqiu Lin and Zihan Wang and Simon Stepputtis and Deva Ramanan and Katia Sycara},
  title = {{InstructPart}: Task-Oriented Part Segmentation with Instruction Reasoning},
  booktitle = {Proceedings of the Annual Meeting of the Association for Computational Linguistics},
  pages = {24202--24227},
  year = {2025},
  doi = {10.18653/v1/2025.acl-long.1179}
}

@inproceedings{wei2025mineanybuild,
  author = {Ziming Wei and Bingqian Lin and Zijian Jiao and Yunshuang Nie and Liang Ma and Yuecheng Liu and Yuzheng Zhuang and Xiaodan Liang},
  title = {{MineAnyBuild}: Benchmarking Spatial Planning for Open-world {AI} Agents},
  booktitle = {Advances in Neural Information Processing Systems (Datasets and Benchmarks Track)},
  year = {2025}
}

@inproceedings{shi2025mme,
  author = {Yang Shi and Huanqian Wang and Wulin Xie and Huanyao Zhang and Lijie Zhao and Yi-Fan Zhang and Xinfeng Li and Chaoyou Fu and Zhuoer Wen and Wenting Liu and Zhuoran Zhang and Xinlong Chen and Bohan Zeng and Sihan Yang and Yushuo Guan and Zhang Zhang and Liang Wang and Haoxuan Li and Zhouchen Lin and Yuanxing Zhang and Pengfei Wan and Haotian Wang and Wenjing Yang},
  title = {{MME-VideoOCR}: Evaluating {OCR-Based} Capabilities of Multimodal {LLMs} in Video Scenarios},
  booktitle = {Advances in Neural Information Processing Systems (Datasets and Benchmarks Track)},
  year = {2025}
}

@inproceedings{grover2025huemanity,
  author = {Rynaa Grover and Jayant Sravan Tamarapalli and Sahiti Yerramilli and Nilay Pande},
  title = {{HueManity}: Probing Fine-Grained Visual Perception in {MLLMs}},
  booktitle = {Proceedings of the International Conference on Pattern Recognition},
  pages = {187--200},
  year = {2026},
  doi = {10.1007/978-3-032-31930-2_13}
}

@inproceedings{zhu2025moho,
  author = {Yanxu Zhu and Shitong Duan and Xiangxu Zhang and Jitao Sang and Peng Zhang and Tun Lu and Xiao Zhou and Jing Yao and Xiaoyuan Yi and Xing Xie},
  title = {{MoHoBench}: Assessing Honesty of Multimodal Large Language Models via Unanswerable Visual Questions},
  booktitle = {Proceedings of the AAAI Conference on Artificial Intelligence},
  pages = {29205--29213},
  year = {2026},
  doi = {10.1609/aaai.v40i34.40159}
}

@inproceedings{xu2025perceiving,
  author = {Liang Xu and Chengqun Yang and Zili Lin and Fei Xu and Yifan Liu and Congsheng Xu and Yiyi Zhang and Jie Qin and Xingdong Sheng and Yunhui Liu and Xin Jin and Yichao Yan and Wenjun Zeng and Xiaokang Yang},
  title = {Perceiving and Acting in First-Person: A Dataset and Benchmark for Egocentric Human-Object-Human Interactions},
  booktitle = {Proceedings of the IEEE/CVF International Conference on Computer Vision},
  pages = {12535--12548},
  year = {2025},
  doi = {10.1109/ICCV51701.2025.01165}
}

@inproceedings{qin2025humansense,
  author = {Zheng Qin and Ruobing Zheng and Yabing Wang and Tianqi Li and Yi Yuan and Jingdong Chen and Le Wang},
  title = {{HumanSense}: From Multimodal Perception to Empathetic Context-Aware Responses Through Reasoning {MLLMs}},
  booktitle = {Proceedings of the AAAI Conference on Artificial Intelligence},
  pages = {24973--24981},
  year = {2026},
  doi = {10.1609/aaai.v40i30.39685}
}

@inproceedings{wang2025audiotextdisagree,
  author = {Cheng Wang and Gelei Deng and Xianglin Yang and Han Qiu and Tianwei Zhang},
  title = {When Audio and Text Disagree: Revealing Text Bias in Large Audio-Language Models},
  booktitle = {Proceedings of the Conference on Empirical Methods in Natural Language Processing},
  pages = {4878--4888},
  year = {2025},
  doi = {10.18653/v1/2025.emnlp-main.246}
}

@inproceedings{wei2025driveqa,
  author = {Maolin Wei and Wanzhou Liu and Eshed Ohn-Bar},
  title = {Passing the Driving Knowledge Test},
  booktitle = {Proceedings of the IEEE/CVF International Conference on Computer Vision},
  year = {2025}
}

@inproceedings{peng2025eyemllm,
  author = {Taiying Peng and Jiacheng Hua and Miao Liu and Feng Lu},
  title = {In the Eye of {MLLM}: Benchmarking Egocentric Video Intent Understanding with Gaze-Guided Prompting},
  booktitle = {Advances in Neural Information Processing Systems (Datasets and Benchmarks Track)},
  year = {2025}
}

@inproceedings{li2025aesbiasbench,
  author = {Kun Li and Lai-Man Po and Hongzheng Yang and Xuyuan Xu and Kangcheng Liu and Yuzhi Zhao},
  title = {{AesBiasBench}: Evaluating Bias and Alignment in Multimodal Language Models for Personalized Image Aesthetic Assessment},
  booktitle = {Proceedings of the Conference on Empirical Methods in Natural Language Processing},
  pages = {7607--7620},
  year = {2025},
  doi = {10.18653/v1/2025.emnlp-main.386}
}

@inproceedings{liu2025assocciam,
  author = {Yifan Liu and Wenkuan Zhao and Shanshan Zhong and Jinghui Qin and Mingfu Liang and Zhongzhan Huang and Wushao Wen},
  title = {{AssoCiAm}: A Benchmark for Evaluating Association Thinking while Circumventing Ambiguity},
  booktitle = {Proceedings of the Conference on Empirical Methods in Natural Language Processing},
  pages = {5203--5219},
  year = {2025},
  doi = {10.18653/v1/2025.emnlp-main.263}
}

@inproceedings{ye2025eyesearsdisagree,
  author = {Qilang Ye and Wei Zeng and Meng Liu and Jie Zhang and Yupeng Hu and Zitong Yu and Yu Zhou},
  title = {When Eyes and Ears Disagree: Can {MLLMs} Discern Audio-Visual Confusion?},
  booktitle = {Proceedings of the AAAI Conference on Artificial Intelligence},
  pages = {11955--11963},
  year = {2026},
  doi = {10.1609/aaai.v40i14.38183}
}

@inproceedings{vasilev2025mtbbench,
  author = {Kiril Vasilev and Alexandre Misrahi and Eeshaan Jain and Phil F. Cheng and Petros Liakopoulos and Olivier Michielin and Michael Moor and Charlotte Bunne},
  title = {{MTBBench}: A Multimodal Sequential Clinical Decision-Making Benchmark in Oncology},
  booktitle = {Advances in Neural Information Processing Systems (Datasets and Benchmarks Track)},
  year = {2025}
}

@inproceedings{tian2025crosscheckbench,
  author = {Baoliang Tian and Yuxuan Si and Jilong Wang and Lingyao Li and Zhongyuan Bao and Zineng Zhou and Tao Wang and Sixu Li and Ziyao Xu and Mingze Wang and Zhouzhuo Zhang and Zhihao Wang and Yike Yun and Ke Tian and Ning Yang and Minghui Qiu},
  title = {{CrossCheck-Bench}: Diagnosing Compositional Failures in Multimodal Conflict Resolution},
  booktitle = {Proceedings of the AAAI Conference on Artificial Intelligence},
  pages = {25887--25895},
  year = {2026},
  doi = {10.1609/aaai.v40i31.39788}
}

@inproceedings{wu2024micbench,
  author = {Haoning Wu and Hanwei Zhu and Zicheng Zhang and Erli Zhang and Chaofeng Chen and Liang Liao and Chunyi Li and Annan Wang and Wenxiu Sun and Qiong Yan and Xiaohong Liu and Guangtao Zhai and Shiqi Wang and Weisi Lin},
  title = {Towards Open-Ended Visual Quality Comparison},
  booktitle = {Proceedings of the European Conference on Computer Vision},
  pages = {360--377},
  year = {2024},
  doi = {10.1007/978-3-031-72646-0_21}
}

@inproceedings{zhang2025qbenchvideo,
  author = {Zicheng Zhang and Ziheng Jia and Haoning Wu and Chunyi Li and Zijian Chen and Yingjie Zhou and Wei Sun and Xiaohong Liu and Xiongkuo Min and Weisi Lin and Guangtao Zhai},
  title = {{Q-Bench-Video}: Benchmark the Video Quality Understanding of {LMMs}},
  booktitle = {Proceedings of the IEEE/CVF Conference on Computer Vision and Pattern Recognition},
  pages = {3229--3239},
  year = {2025},
  doi = {10.1109/CVPR52734.2025.00307}
}

@inproceedings{zhang2025mmerealworld,
  author = {Yifan Zhang and Huanyu Zhang and Haochen Tian and Chaoyou Fu and Shuangqing Zhang and Junfei Wu and Feng Li and Kun Wang and Qingsong Wen and Zhang Zhang and Liang Wang and Rong Jin},
  title = {{MME-RealWorld}: Could Your Multimodal {LLM} Challenge High-Resolution Real-World Scenarios that are Difficult for Humans?},
  booktitle = {Proceedings of the International Conference on Learning Representations},
  year = {2025}
}

@inproceedings{fu2025ocrbenchv2,
  author = {Ling Fu and Zhebin Kuang and Jiajun Song and Mingxin Huang and Biao Yang and Yuzhe Li and Linghao Zhu and Qidi Luo and Xinyu Wang and Hao Lu and Zhang Li and Guozhi Tang and Bin Shan and Chunhui Lin and Qi Liu and Binghong Wu and Hao Feng and Hao Liu and Can Huang and Jingqun Tang and Wei Chen and Lianwen Jin and Yuliang Liu and Xiang Bai},
  title = {{OCRBench} v2: An Improved Benchmark for Evaluating Large Multimodal Models on Visual Text Localization and Reasoning},
  booktitle = {Advances in Neural Information Processing Systems (Datasets and Benchmarks Track)},
  year = {2025}
}

@inproceedings{wang2024hrbench,
  author = {Wenbin Wang and Liang Ding and Minyan Zeng and Xiabin Zhou and Li Shen and Yong Luo and Wei Yu and Dacheng Tao},
  title = {Divide, Conquer and Combine: A Training-Free Framework for High-Resolution Image Perception in Multimodal Large Language Models},
  booktitle = {Proceedings of the AAAI Conference on Artificial Intelligence},
  pages = {7907--7915},
  year = {2025},
  doi = {10.1609/aaai.v39i8.32852}
}

@inproceedings{ouyang2025omnidocbench,
  author = {Linke Ouyang and Yuan Qu and Hongbin Zhou and Jiawei Zhu and Rui Zhang and Qunshu Lin and Bin Wang and Zhiyuan Zhao and Man Jiang and Xiaomeng Zhao and Jin Shi and Fan Wu and Pei Chu and Minghao Liu and Zhenxiang Li and Chao Xu and Bo Zhang and Botian Shi and Zhongying Tu and Conghui He},
  title = {{OmniDocBench}: Benchmarking Diverse {PDF} Document Parsing with Comprehensive Annotations},
  booktitle = {Proceedings of the IEEE/CVF Conference on Computer Vision and Pattern Recognition},
  pages = {24838--24848},
  year = {2025},
  doi = {10.1109/CVPR52734.2025.02313}
}

@inproceedings{lian2025dlcbench,
  author = {Long Lian and Yifan Ding and Yunhao Ge and Sifei Liu and Hanzi Mao and Boyi Li and Marco Pavone and Ming-Yu Liu and Trevor Darrell and Adam Yala and Yin Cui},
  title = {Describe Anything: Detailed Localized Image and Video Captioning},
  booktitle = {Proceedings of the IEEE/CVF International Conference on Computer Vision},
  pages = {21766--21777},
  year = {2025},
  doi = {10.1109/ICCV51701.2025.02021}
}

@inproceedings{xu2025visadr,
  author = {Jiacong Xu and Shao-Yuan Lo and Bardia Safaei and Vishal M. Patel and Isht Dwivedi},
  title = {Towards Zero-Shot Anomaly Detection and Reasoning with Multimodal Large Language Models},
  booktitle = {Proceedings of the IEEE/CVF Conference on Computer Vision and Pattern Recognition},
  pages = {20370--20382},
  year = {2025},
  doi = {10.1109/CVPR52734.2025.01897}
}

@inproceedings{yang2025ccocr,
  author = {Zhibo Yang and Jun Tang and Zhaohai Li and Pengfei Wang and Jianqiang Wan and Humen Zhong and Xuejing Liu and Mingkun Yang and Peng Wang and Shuai Bai and LianWen Jin and Junyang Lin},
  title = {{CC-OCR}: A Comprehensive and Challenging {OCR} Benchmark for Evaluating Large Multimodal Models in Literacy},
  booktitle = {Proceedings of the IEEE/CVF International Conference on Computer Vision},
  pages = {21744--21754},
  year = {2025},
  doi = {10.1109/ICCV51701.2025.02019}
}

@inproceedings{zhang2024imagewikiqa,
  author = {Yuhui Zhang and Alyssa Unell and Xiaohan Wang and Dhruba Ghosh and Yuchang Su and Ludwig Schmidt and Serena Yeung-Levy},
  title = {Why are Visually-Grounded Language Models Bad at Image Classification?},
  booktitle = {Advances in Neural Information Processing Systems},
  year = {2024}
}

@inproceedings{hu2025mtmeur,
  author = {Jinpeng Hu and Hongchang Shi and Chongyuan Dai and Zhuo Li and Peipei Song and Meng Wang},
  title = {Beyond Emotion Recognition: A Multi-Turn Multimodal Emotion Understanding and Reasoning Benchmark},
  booktitle = {Proceedings of the ACM International Conference on Multimedia},
  pages = {5814--5823},
  year = {2025},
  doi = {10.1145/3746027.3755726}
}

@inproceedings{han2025camer,
  author = {Zhiyuan Han and Beier Zhu and Yanlong Xu and Peipei Song and Xun Yang},
  title = {Benchmarking and Bridging Emotion Conflicts for Multimodal Emotion Reasoning},
  booktitle = {Proceedings of the ACM International Conference on Multimedia},
  pages = {5528--5537},
  year = {2025},
  doi = {10.1145/3746027.3754856}
}

@inproceedings{zhang2026mmeemotion,
  author = {Fan Zhang and Zebang Cheng and Chong Deng and Haoxuan Li and Zheng Lian and Qian Chen and Huadai Liu and Wen Wang and Yi-Fan Zhang and Renrui Zhang and Ziyu Guo and Zhihong Zhu and Hao Wu and Haixin Wang and Yefeng Zheng and Xiaojiang Peng and Xian Wu and Kun Wang and Xiangang Li and Jieping Ye and Pheng-Ann Heng},
  title = {{MME-Emotion}: A Holistic Evaluation Benchmark for Emotional Intelligence in Multimodal Large Language Models},
  booktitle = {Proceedings of the International Conference on Learning Representations},
  year = {2026}
}

@inproceedings{liu2025shotbench,
  author = {Hongbo Liu and Jingwen He and Yi Jin and Dian Zheng and Yuhao Dong and Fan Zhang and Ziqi Huang and Yinan He and Weichao Chen and Yu Qiao and Wanli Ouyang and Shengjie Zhao and Ziwei Liu},
  title = {{ShotBench}: Expert-Level Cinematic Understanding in Vision-Language Models},
  booktitle = {Advances in Neural Information Processing Systems},
  year = {2025}
}

@inproceedings{zhang2025lmmseval,
  author = {Kaichen Zhang and Bo Li and Peiyuan Zhang and Fanyi Pu and Joshua Adrian Cahyono and Kairui Hu and Shuai Liu and Yuanhan Zhang and Jingkang Yang and Chunyuan Li and Ziwei Liu},
  title = {{LMMs-Eval}: Reality Check on the Evaluation of Large Multimodal Models},
  booktitle = {Findings of the Association for Computational Linguistics: NAACL 2025},
  pages = {881--916},
  year = {2025},
  doi = {10.18653/v1/2025.findings-naacl.51}
}

@inproceedings{li2024naturalbench,
  author = {Baiqi Li and Zhiqiu Lin and Wenxuan Peng and Nyandwi, Jean de Dieu and Daniel Jiang and Zixian Ma and Simran Khanuja and Ranjay Krishna and Graham Neubig and Deva Ramanan},
  title = {{NaturalBench}: Evaluating Vision-Language Models on Natural Adversarial Samples},
  booktitle = {Advances in Neural Information Processing Systems (Datasets and Benchmarks Track)},
  year = {2024}
}

@inproceedings{nayak2024culturalvqa,
  author = {Shravan Nayak and Kanishk Jain and Rabiul Awal and Siva Reddy and {van Steenkiste}, Sjoerd and Lisa Anne Hendricks and Karolina Stanczak and Aishwarya Agrawal},
  title = {Benchmarking Vision Language Models for Cultural Understanding},
  booktitle = {Proceedings of the Conference on Empirical Methods in Natural Language Processing},
  pages = {5769--5790},
  year = {2024},
  doi = {10.18653/v1/2024.emnlp-main.329}
}

@inproceedings{vayani2025almbench,
  author = {Ashmal Vayani and Dinura Dissanayake and Hasindri Watawana and Noor Ahsan and Nevasini Sasikumar and Omkar Thawakar and Henok Biadglign Ademtew and Yahya Hmaiti and Amandeep Kumar and Kartik Kuckreja and Mykola Maslych and Wafa Al Ghallabi and Mihail Mihaylov and Chao Qin and Abdelrahman M Shaker and Mike Zhang and Mahardika Krisna Ihsani and Amiel Esplana and Monil Gokani and Shachar Mirkin and Harsh Singh and Ashay Srivastava and Endre Hamerlik and Fathinah Asma Izzati and Fadillah Adamsyah Maani and Sebastian Cavada and Jenny Chim and Rohit Gupta and Sanjay Manjunath and Kamila Zhumakhanova and Feno Heriniaina Rabevohitra and Azril Amirudin and Muhammad Ridzuan and Daniya Kareem and Ketan More and Kunyang Li and Pramesh Shakya and Muhammad Saad and Amirpouya Ghasemaghaei and Amirbek Djanibekov and Dilshod Azizov and Branislava Jankovic and Naman Bhatia and Alvaro Cabrera and Johan Obando-Ceron and Olympiah Otieno and Fabian Farestam and Muztoba Rabbani and Sanoojan Baliah and Santosh Sanjeev and Abduragim Shtanchaev and Maheen Fatima and Thao Nguyen and Amrin Kareem and Toluwani Aremu and Nathan Xavier and Amit Bhatkal and Hawau Toyin and Aman Chadha and Hisham Cholakkal and Rao Muhammad Anwer and Michael Felsberg and Jorma Laaksonen and Thamar Solorio and Monojit Choudhury and Ivan Laptev and Mubarak Shah and Salman Khan and Fahad Khan},
  title = {All Languages Matter: Evaluating {LMMs} on Culturally Diverse 100 Languages},
  booktitle = {Proceedings of the IEEE/CVF Conference on Computer Vision and Pattern Recognition},
  pages = {19565--19575},
  year = {2025},
  doi = {10.1109/CVPR52734.2025.01822}
}

@inproceedings{winata2025worldcuisines,
  author = {Genta Indra Winata and Frederikus Hudi and Patrick Amadeus Irawan and David Anugraha and Rifki Afina Putri and Yutong Wang and Adam Nohejl and Ubaidillah Ariq Prathama and Nedjma Ousidhoum and Afifa Amriani and Anar Rzayev and Anirban Das and Ashmari Pramodya and Aulia Adila and Bryan Wilie and Candy Olivia Mawalim and Ching Lam Cheng and Daud Abolade and Emmanuele Chersoni and Enrico Santus and Fariz Ikhwantri and Garry Kuwanto and Hanyang Zhao and Haryo Akbarianto Wibowo and Holy Lovenia and Jan Christian Blaise Cruz and Jan Wira Gotama Putra and Junho Myung and Lucky Susanto and Maria Angelica Riera Machin and Marina Zhukova and Michael Anugraha and Muhammad Farid Adilazuarda and Natasha Santosa and Peerat Limkonchotiwat and Raj Dabre and Rio Alexander Audino and Samuel Cahyawijaya and Shi-Xiong Zhang and Stephanie Yulia Salim and Yi Zhou and Yinxuan Gui and David Ifeoluwa Adelani and En-Shiun Annie Lee and Shogo Okada and Ayu Purwarianti and Alham Fikri Aji and Taro Watanabe and Derry Tanti Wijaya and Alice Oh and Chong-Wah Ngo},
  title = {{WorldCuisines}: A Massive-Scale Benchmark for Multilingual and Multicultural Visual Question Answering on Global Cuisines},
  booktitle = {Proceedings of the Conference of the Nations of the Americas Chapter of the Association for Computational Linguistics: Human Language Technologies},
  pages = {3242--3264},
  year = {2025},
  doi = {10.18653/v1/2025.naacl-long.167}
}

@inproceedings{nie2026personamme,
  author = {Chang Nie and Chaoyou Fu and Yifan Zhang and Haihua Yang and Caifeng Shan},
  title = {{PersonaVLM}: Long-Term Personalized Multimodal {LLMs}},
  booktitle = {Proceedings of the IEEE/CVF Conference on Computer Vision and Pattern Recognition},
  pages = {15000--15009},
  year = {2026}
}

@article{zhang2024vniah,
  author = {Peiyuan Zhang and Kaichen Zhang and Bo Li and Guangtao Zeng and Jingkang Yang and Yuanhan Zhang and Ziyue Wang and Haoran Tan and Chunyuan Li and Ziwei Liu},
  title = {Long Context Transfer from Language to Vision},
  journal = {Transactions on Machine Learning Research},
  year = {2025}
}

@inproceedings{liu2024mmdu,
  author = {Ziyu Liu and Tao Chu and Yuhang Zang and Xilin Wei and Xiaoyi Dong and Pan Zhang and Zijian Liang and Yuanjun Xiong and Yu Qiao and Dahua Lin and Jiaqi Wang},
  title = {{MMDU}: A Multi-Turn Multi-Image Dialog Understanding Benchmark and Instruction-Tuning Dataset for {LVLMs}},
  booktitle = {Advances in Neural Information Processing Systems (Datasets and Benchmarks Track)},
  year = {2024}
}

@inproceedings{du2025mir,
  author = {Hang Du and Jiayang Zhang and Guoshun Nan and Wendi Deng and Zhenyan Chen and Chenyang Zhang and Wang Xiao and Shan Huang and Yuqi Pan and Tao Qi and Sicong Leng},
  title = {From Easy to Hard: The {MIR} Benchmark for Progressive Interleaved Multi-Image Reasoning},
  booktitle = {Proceedings of the IEEE/CVF International Conference on Computer Vision},
  pages = {859--869},
  year = {2025},
  doi = {10.1109/ICCV51701.2025.00088}
}

@inproceedings{zhang2026longvlbench,
  author = {Yuqi Zhang and Cheng Chen and Yuyu Guo and Wenjie Yang and Lingchen Meng and Peng Di and Hang Yu and Zuxuan Wu and Yu-Gang Jiang},
  title = {{VLZip}: Unified Visual and Textual Compression for Interleaved Long-Context Modeling},
  booktitle = {Proceedings of the European Conference on Computer Vision},
  pages = {188--206},
  year = {2026},
  doi = {10.1007/978-3-032-37550-6_11}
}

@inproceedings{hua2026tokenswapbench,
  author = {Andong Hua and Colton Bishop and Igor Mordatch and Arian Hosseini and Jindong Gu and Aleksandra Faust and Rebecca Roelofs and Yao Qin},
  title = {{TokenSwap}: Benchmarking and Reducing the Modality Gap in Multimodal {LLMs}},
  booktitle = {Advances in Neural Information Processing Systems},
  year = {2026}
}

@inproceedings{dong2026demoiclbench,
  author = {Yuhao Dong and Shulin Tian and Shuai Liu and Shuangrui Ding and Yuhang Zang and Xiaoyi Dong and Yuhang Cao and Jiaqi Wang and Ziwei Liu},
  title = {{Demo-ICL}: In-Context Learning for Procedural Video Knowledge Acquisition},
  booktitle = {Findings of the Association for Computational Linguistics: EMNLP 2026},
  pages = {15381--15399},
  year = {2026},
  doi = {10.18653/2026.findings-emnlp.732}
}

@inproceedings{hua2024icca,
  author = {Yilun Hua and Yoav Artzi},
  title = {Talk Less, Interact Better: Evaluating In-context Conversational Adaptation in Multimodal {LLMs}},
  booktitle = {Proceedings of the Conference on Language Modeling},
  year = {2024}
}

@inproceedings{yue2025mmmupro,
  author = {Xiang Yue and Tianyu Zheng and Yuansheng Ni and Yubo Wang and Kai Zhang and Shengbang Tong and Yuxuan Sun and Botao Yu and Ge Zhang and Huan Sun and Yu Su and Wenhu Chen and Graham Neubig},
  title = {{MMMU-Pro}: A More Robust Multi-discipline Multimodal Understanding Benchmark},
  booktitle = {Proceedings of the Annual Meeting of the Association for Computational Linguistics},
  pages = {15134--15186},
  year = {2025},
  doi = {10.18653/v1/2025.acl-long.736}
}

@inproceedings{hu2026videommmu,
  author = {Kairui Hu and Penghao Wu and Fanyi Pu and Wang Xiao and Xiang Yue and Bo Li and Yuanhan Zhang and Ziwei Liu},
  title = {{Video-MMMU}: Evaluating Knowledge Acquisition from Multidisciplinary Professional Videos},
  booktitle = {Proceedings of the Annual Meeting of the Association for Computational Linguistics},
  pages = {27798--27828},
  year = {2026},
  doi = {10.18653/v1/2026.acl-long.1281}
}

@inproceedings{guo2025ucit,
  author = {Haiyang Guo and Fanhu Zeng and Ziwei Xiang and Fei Zhu and Da-Han Wang and Xu-Yao Zhang and Cheng-Lin Liu},
  title = {{HiDe-LLaVA}: Hierarchical Decoupling for Continual Instruction Tuning of Multimodal Large Language Model},
  booktitle = {Proceedings of the Annual Meeting of the Association for Computational Linguistics},
  pages = {13572--13586},
  year = {2025},
  doi = {10.18653/v1/2025.acl-long.666}
}

@inproceedings{zhang2026timelensbench,
  author = {Jun Zhang and Teng Wang and Yuying Ge and Yixiao Ge and Xinhao Li and Limin Wang},
  title = {{TimeLens}: Rethinking Video Temporal Grounding with Multimodal {LLMs}},
  booktitle = {Proceedings of the IEEE/CVF Conference on Computer Vision and Pattern Recognition},
  pages = {10419--10429},
  year = {2026}
}

@inproceedings{yuan2025videoreferbench,
  author = {Yuqian Yuan and Hang Zhang and Wentong Li and Zesen Cheng and Boqiang Zhang and Long Li and Xin Li and Deli Zhao and Wenqiao Zhang and Yueting Zhuang and Jianke Zhu and Lidong Bing},
  title = {{VideoRefer} Suite: Advancing Spatial-Temporal Object Understanding with Video {LLM}},
  booktitle = {Proceedings of the IEEE/CVF Conference on Computer Vision and Pattern Recognition},
  pages = {18970--18980},
  year = {2025},
  doi = {10.1109/CVPR52734.2025.01767}
}

@inproceedings{wu2024longvideobench,
  author = {Haoning Wu and Dongxu Li and Bei Chen and Junnan Li},
  title = {{LongVideoBench}: A Benchmark for Long-context Interleaved Video-Language Understanding},
  booktitle = {Advances in Neural Information Processing Systems (Datasets and Benchmarks Track)},
  year = {2024}
}

@inproceedings{wang2025lvbench,
  author = {Weihan Wang and Zehai He and Wenyi Hong and Yean Cheng and Xiaohan Zhang and Ji Qi and Xiaotao Gu and Shiyu Huang and Bin Xu and Yuxiao Dong and Ming Ding and Jie Tang},
  title = {{LVBench}: An Extreme Long Video Understanding Benchmark},
  booktitle = {Proceedings of the IEEE/CVF International Conference on Computer Vision},
  pages = {22958--22967},
  year = {2025},
  doi = {10.1109/ICCV51701.2025.02131}
}

@inproceedings{chai2025vdc,
  author = {Wenhao Chai and Enxin Song and Yilun Du and Chenlin Meng and Vashisht Madhavan and Omer Bar-Tal and Jenq-Neng Hwang and Saining Xie and Christopher D. Manning},
  title = {{AuroraCap}: Efficient, Performant Video Detailed Captioning and a New Benchmark},
  booktitle = {Proceedings of the International Conference on Learning Representations},
  year = {2025}
}

@inproceedings{zhang2025hivau70k,
  author = {Huaxin Zhang and Xiaohao Xu and Xiang Wang and Jialong Zuo and Xiaonan Huang and Changxin Gao and Shanjun Zhang and Li Yu and Nong Sang},
  title = {{Holmes-VAU}: Towards Long-term Video Anomaly Understanding at Any Granularity},
  booktitle = {Proceedings of the IEEE/CVF Conference on Computer Vision and Pattern Recognition},
  pages = {13843--13853},
  year = {2025},
  doi = {10.1109/CVPR52734.2025.01292}
}

@inproceedings{yang2024vripthard,
  author = {Dongjie Yang and Suyuan Huang and Chengqiang Lu and Xiaodong Han and Haoxin Zhang and Yan Gao and Yao Hu and Hai Zhao},
  title = {Vript: A Video Is Worth Thousands of Words},
  booktitle = {Advances in Neural Information Processing Systems (Datasets and Benchmarks Track)},
  year = {2024}
}

@inproceedings{li2025ovobench,
  author = {Junbo Niu and Yifei Li and Ziyang Miao and Chunjiang Ge and Yuanhang Zhou and Qihao He and Xiaoyi Dong and Haodong Duan and Shuangrui Ding and Rui Qian and Pan Zhang and Yuhang Zang and Yuhang Cao and Conghui He and Jiaqi Wang},
  title = {{OVO-Bench}: How Far is Your {Video-LLMs} from Real-World Online Video Understanding?},
  booktitle = {Proceedings of the IEEE/CVF Conference on Computer Vision and Pattern Recognition},
  pages = {18902--18913},
  year = {2025},
  doi = {10.1109/CVPR52734.2025.01761}
}

@inproceedings{xu2026infstreamseval,
  author = {Ruyi Xu and Guangxuan Xiao and Yukang Chen and Liuning He and Kelly Peng and Yao Lu and Song Han},
  title = {{StreamingVLM}: Real-Time Understanding for Infinite Video Streams},
  booktitle = {Proceedings of the International Conference on Learning Representations},
  year = {2026}
}

@inproceedings{zeng2025odvbench,
  author = {Xiangyu Zeng and Kefan Qiu and Qingyu Zhang and Xinhao Li and Jing Wang and Jiaxin Li and Ziang Yan and Kun Tian and Meng Tian and Xinhai Zhao and Yi Wang and Limin Wang},
  title = {{StreamForest}: Efficient Online Video Understanding with Persistent Event Memory},
  booktitle = {Advances in Neural Information Processing Systems},
  year = {2025}
}

@inproceedings{sakshi2025mmau,
  author = {S Sakshi and Utkarsh Tyagi and Sonal Kumar and Ashish Seth and Ramaneswaran Selvakumar and Oriol Nieto and Ramani Duraiswami and Sreyan Ghosh and Dinesh Manocha},
  title = {{MMAU}: A Massive Multi-Task Audio Understanding and Reasoning Benchmark},
  booktitle = {Proceedings of the International Conference on Learning Representations},
  year = {2025}
}

@inproceedings{ma2025mmar,
  author = {Ziyang Ma and Yinghao Ma and Yanqiao Zhu and Chen Yang and Yi-Wen Chao and Ruiyang Xu and Wenxi Chen and Yuanzhe Chen and Zhuo Chen and Jian Cong and Kai Li and Keliang Li and Siyou Li and Xinfeng Li and Xiquan Li and Zheng Lian and Yuzhe Liang and Minghao Liu and Zhikang Niu and Tianrui Wang and Yuping Wang and Yuxuan Wang and Yihao Wu and Guanrou Yang and Jianwei Yu and Ruibin Yuan and Zhisheng Zheng and Ziya Zhou and Haina Zhu and Wei Xue and Emmanouil Benetos and Kai Yu and Eng-Siong Chng and Xie Chen},
  title = {{MMAR}: A Challenging Benchmark for Deep Reasoning in Speech, Audio, Music, and Their Mix},
  booktitle = {Advances in Neural Information Processing Systems (Datasets and Benchmarks Track)},
  year = {2025}
}

@inproceedings{ghosh2025longaudiobench,
  author = {Sreyan Ghosh and Zhifeng Kong and Sonal Kumar and S Sakshi and Jaehyeon Kim and Wei Ping and Rafael Valle and Dinesh Manocha and Bryan Catanzaro},
  title = {Audio {Flamingo} 2: An Audio-Language Model with Long-Audio Understanding and Expert Reasoning Abilities},
  booktitle = {Proceedings of the International Conference on Machine Learning},
  year = {2025}
}

@inproceedings{wang2026mmsu,
  author = {Dingdong Wang and Junan Li and Jincenzi Wu and Dongchao Yang and Xueyuan Chen and Tianhua Zhang and Helen Meng},
  title = {{MMSU}: A Massive Multi-task Spoken Language Understanding and Reasoning Benchmark},
  booktitle = {Proceedings of the International Conference on Learning Representations},
  year = {2026}
}

@inproceedings{qi20243dmmvet,
  author = {Zekun Qi and Runpei Dong and Shaochen Zhang and Haoran Geng and Chunrui Han and Zheng Ge and Li Yi and Kaisheng Ma},
  title = {{ShapeLLM}: Universal {3D} Object Understanding for Embodied Interaction},
  booktitle = {Proceedings of the European Conference on Computer Vision},
  pages = {214--238},
  year = {2024},
  doi = {10.1007/978-3-031-72775-7_13}
}

@inproceedings{brown2026vsibenchdebiased,
  author = {Ellis Brown and Jihan Yang and Shusheng Yang and Rob Fergus and Saining Xie},
  title = {Benchmark Designers Should ``Train on the Test Set'' to Expose Exploitable Non-Visual Shortcuts},
  booktitle = {Proceedings of the Conference on Language Modeling},
  year = {2026}
}

@inproceedings{hong2026worldsense,
  author = {Jack Hong and Shilin Yan and Jiayin Cai and Xiaolong Jiang and Yao Hu and Weidi Xie},
  title = {{WorldSense}: Evaluating Real-world Omnimodal Understanding for Multimodal {LLMs}},
  booktitle = {Proceedings of the International Conference on Learning Representations},
  year = {2026}
}

@inproceedings{li2026omnivideobench,
  author = {Caorui Li and Yu Chen and Yiyan Ji and Jin Xu and Zhenyu Cui and Shihao Li and Yuanxing Zhang and Wentao Wang and Zhenghao Song and Dingling Zhang and Ying He and Haoxiang Liu and Yuxuan Wang and Qiufeng Wang and Jiafu Tang and Zhenhe Wu and Jiehui Luo and Zhiyu Pan and Weihao Xie and Chenchen Zhang and Zhaohui Wang and Jiayi Tian and Yanghai Wang and Zhe Cao and Minxin Dai and Ke Wang and Runzhe Wen and Yinghao Ma and Yaning Pan and Sungkyun Chang and Termeh Taheri and Haiwen Xia and Christos Plachouras and Emmanouil Benetos and Yizhi Li and Ge Zhang and Jian Yang and Tianhao Peng and Zili Wang and Minghao Liu and Junran Peng and Zhaoxiang Zhang and Jiaheng Liu},
  title = {{OmniVideoBench}: Towards Audio-Visual Understanding Evaluation for Omni {MLLMs}},
  booktitle = {Proceedings of the International Conference on Learning Representations},
  year = {2026}
}

@inproceedings{he2024olympiadbench,
  author = {Chaoqun He and Renjie Luo and Yuzhuo Bai and Shengding Hu and Zhen Leng Thai and Junhao Shen and Jinyi Hu and Xu Han and Yujie Huang and Yuxiang Zhang and Jie Liu and Lei Qi and Zhiyuan Liu and Maosong Sun},
  title = {{OlympiadBench}: A Challenging Benchmark for Promoting {AGI} with {Olympiad}-Level Bilingual Multimodal Scientific Problems},
  booktitle = {Proceedings of the Annual Meeting of the Association for Computational Linguistics},
  pages = {3828--3850},
  year = {2024},
  doi = {10.18653/v1/2024.acl-long.211}
}

@inproceedings{yang2025r1onevisionbench,
  author = {Yi Yang and Xiaoxuan He and Hongkun Pan and Xiyan Jiang and Yan Deng and Xingtao Yang and Haoyu Lu and Dacheng Yin and Fengyun Rao and Minfeng Zhu and Bo Zhang and Wei Chen},
  title = {{R1-Onevision}: Advancing Generalized Multimodal Reasoning Through Cross-Modal Formalization},
  booktitle = {Proceedings of the IEEE/CVF International Conference on Computer Vision},
  pages = {2376--2385},
  year = {2025},
  doi = {10.1109/ICCV51701.2025.00229}
}

@article{phan2025humanityslastexam,
  author = {{Center for AI Safety} and {Scale AI} and {HLE Contributors Consortium}},
  title = {A benchmark of expert-level academic questions to assess {AI} capabilities},
  journal = {Nature},
  volume = {649},
  number = {8099},
  pages = {1139--1146},
  year = {2026},
  doi = {10.1038/s41586-025-09962-4}
}

@inproceedings{zhao2025mmvu,
  author = {Yilun Zhao and Haowei Zhang and Lujing Xie and Tongyan Hu and Guo Gan and Yitao Long and Zhiyuan Hu and Weiyuan Chen and Chuhan Li and Zhijian Xu and Chengye Wang and Ziyao Shangguan and Zhenwen Liang and Yixin Liu and Chen Zhao and Arman Cohan},
  title = {{MMVU}: Measuring Expert-Level Multi-Discipline Video Understanding},
  booktitle = {Proceedings of the IEEE/CVF Conference on Computer Vision and Pattern Recognition},
  pages = {8475--8489},
  year = {2025},
  doi = {10.1109/CVPR52734.2025.00793}
}

@inproceedings{qiao2025wemath,
  author = {Runqi Qiao and Qiuna Tan and Guanting Dong and Minhui Wu and Chong Sun and Xiaoshuai Song and Jiapeng Wang and Zhuoma Gongque and Shanglin Lei and Yifan Zhang and Zhe Wei and Miaoxuan Zhang and Runfeng Qiao and Xiao Zong and Yida Xu and Peiqing Yang and Zhimin Bao and Muxi Diao and Chen Li and Honggang Zhang},
  title = {We-Math: Does Your Large Multimodal Model Achieve Human-like Mathematical Reasoning?},
  booktitle = {Proceedings of the Annual Meeting of the Association for Computational Linguistics},
  pages = {20023--20070},
  year = {2025},
  doi = {10.18653/v1/2025.acl-long.983}
}

@inproceedings{hao2025emma,
  author = {Yunzhuo Hao and Jiawei Gu and Huichen Will Wang and Linjie Li and Zhengyuan Yang and Lijuan Wang and Yu Cheng},
  title = {Can {MLLMs} Reason in Multimodality? {EMMA}: An Enhanced {MultiModal} {ReAsoning} Benchmark},
  booktitle = {Proceedings of the International Conference on Machine Learning},
  year = {2025}
}

@inproceedings{masry2025chartqapro,
  author = {Ahmed Masry and Mohammed Saidul Islam and Mahir Ahmed and Aayush Bajaj and Firoz Kabir and Aaryaman Kartha and Md Tahmid Rahman Laskar and Mizanur Rahman and Shadikur Rahman and Mehrad Shahmohammadi and Megh Thakkar and Md Rizwan Parvez and Enamul Hoque and Shafiq Joty},
  title = {{ChartQAPro}: A More Diverse and Challenging Benchmark for Chart Question Answering},
  booktitle = {Findings of the Association for Computational Linguistics: ACL 2025},
  pages = {19123--19151},
  year = {2025},
  doi = {10.18653/v1/2025.findings-acl.978}
}

@inproceedings{zou2025dynamath,
  author = {Chengke Zou and Xingang Guo and Rui Yang and Junyu Zhang and Bin Hu and Huan Zhang},
  title = {{DynaMath}: A Dynamic Visual Benchmark for Evaluating Mathematical Reasoning Robustness of Vision Language Models},
  booktitle = {Proceedings of the International Conference on Learning Representations},
  year = {2025}
}

@inproceedings{yang2025vsibench,
  author = {Jihan Yang and Shusheng Yang and Anjali W. Gupta and Rilyn Han and Li Fei-Fei and Saining Xie},
  title = {Thinking in Space: How Multimodal Large Language Models See, Remember, and Recall Spaces},
  booktitle = {Proceedings of the IEEE/CVF Conference on Computer Vision and Pattern Recognition},
  pages = {10632--10643},
  year = {2025},
  doi = {10.1109/CVPR52734.2025.00994}
}

@inproceedings{yang2026mmsibench,
  author = {Sihan Yang and Runsen Xu and Yiman Xie and Sizhe Yang and Mo Li and Jingli Lin and Chenming Zhu and Xiaochen Chen and Haodong Duan and Xiangyu Yue and Dahua Lin and Tai Wang and Jiangmiao Pang},
  title = {{MMSI-Bench}: A Benchmark for Multi-Image Spatial Intelligence},
  booktitle = {Proceedings of the International Conference on Learning Representations},
  year = {2026}
}

@inproceedings{zhou2025refspatialbench,
  author = {Enshen Zhou and Jingkun An and Cheng Chi and Yi Han and Shanyu Rong and Chi Zhang and Pengwei Wang and Zhongyuan Wang and Tiejun Huang and Lu Sheng and Shanghang Zhang},
  title = {{RoboRefer}: Towards Spatial Referring with Reasoning in Vision-Language Models for Robotics},
  booktitle = {Advances in Neural Information Processing Systems},
  year = {2025}
}

@inproceedings{song2025robospatial,
  author = {Chan Hee Song and Valts Blukis and Jonathan Tremblay and Stephen Tyree and Yu Su and Stan Birchfield},
  title = {{RoboSpatial}: Teaching Spatial Understanding to {2D} and {3D} Vision-Language Models for Robotics},
  booktitle = {Proceedings of the IEEE/CVF Conference on Computer Vision and Pattern Recognition},
  pages = {15768--15780},
  year = {2025},
  doi = {10.1109/CVPR52734.2025.01470}
}

@inproceedings{ma20253dsrbench,
  author = {Wufei Ma and Haoyu Chen and Guofeng Zhang and Yu-Cheng Chou and Jieneng Chen and {de Melo}, Celso M and Alan Yuille},
  title = {{3DSRBench}: A Comprehensive {3D} Spatial Reasoning Benchmark},
  booktitle = {Proceedings of the IEEE/CVF International Conference on Computer Vision},
  pages = {6924--6934},
  year = {2025},
  doi = {10.1109/ICCV51701.2025.00651}
}

@inproceedings{zhang2025sparbench,
  author = {Jiahui Zhang and Yurui Chen and Yanpeng Zhou and Yueming Xu and Ze Huang and Jilin Mei and Junhui Chen and Yu-Jie Yuan and Xinyue Cai and Guowei Huang and Xingyue Quan and Hang Xu and Li Zhang},
  title = {From Flatland to Space: Teaching Vision-Language Models to Perceive and Reason in {3D}},
  booktitle = {Advances in Neural Information Processing Systems (Datasets and Benchmarks Track)},
  year = {2025}
}

@inproceedings{yan2024revos,
  author = {Cilin Yan and Haochen Wang and Shilin Yan and Xiaolong Jiang and Yao Hu and Guoliang Kang and Weidi Xie and Efstratios Gavves},
  title = {{VISA}: Reasoning Video Object Segmentation via Large Language Models},
  booktitle = {Proceedings of the European Conference on Computer Vision},
  pages = {98--115},
  year = {2024},
  doi = {10.1007/978-3-031-72633-0_6}
}

@inproceedings{xu2026visulogic,
  author = {Weiye Xu and Jiahao Wang and Weiyun Wang and Zhe Chen and Wengang Zhou and Aijun Yang and Lewei Lu and Houqiang Li and Xiaohua Wang and Xizhou Zhu and Wenhai Wang and Jifeng Dai and Jinguo Zhu},
  title = {{VisuLogic}: A Benchmark for Evaluating Visual Reasoning in Multi-modal Large Language Models},
  booktitle = {Proceedings of the International Conference on Learning Representations},
  year = {2026}
}

@inproceedings{chen2024m3cot,
  author = {Qiguang Chen and Libo Qin and Jin Zhang and Zhi Chen and Xiao Xu and Wanxiang Che},
  title = {{M$^3$CoT}: A Novel Benchmark for Multi-Domain Multi-step Multi-modal Chain-of-Thought},
  booktitle = {Proceedings of the Annual Meeting of the Association for Computational Linguistics},
  pages = {8199--8221},
  year = {2024},
  doi = {10.18653/v1/2024.acl-long.446}
}

@inproceedings{cheng2026videoholmes,
  author = {Junhao Cheng and Yuying Ge and Teng Wang and Yixiao Ge and Jing Liao and Ying Shan},
  title = {Video-Holmes: Can {MLLM} Think Like Holmes for Complex Video Reasoning?},
  booktitle = {Proceedings of the European Conference on Computer Vision},
  pages = {539--556},
  year = {2026},
  doi = {10.1007/978-3-032-37531-5_30}
}

@inproceedings{chen2026seedbenchr1,
  author = {Yi Chen and Yuying Ge and Rui Wang and Yixiao Ge and Junhao Cheng and Ying Shan and Xihui Liu},
  title = {{GRPO-CARE}: Consistency-Aware Reinforcement Learning for Multimodal Reasoning},
  booktitle = {Findings of the Association for Computational Linguistics: ACL 2026},
  pages = {4305--4320},
  year = {2026},
  doi = {10.18653/v1/2026.findings-acl.210}
}

@inproceedings{thawakar2025vrcbench,
  author = {Omkar Thawakar and Dinura Dissanayake and Ketan More and Ritesh Thawkar and Ahmed Heakl and Noor Ahsan and Yuhao Li and Mohammed Zumri and Jean Lahoud and Rao Muhammad Anwer and Hisham Cholakkal and Ivan Laptev and Mubarak Shah and Fahad Shahbaz Khan and Salman Khan},
  title = {{LlamaV-o1}: Rethinking Step-by-step Visual Reasoning in {LLMs}},
  booktitle = {Findings of the Association for Computational Linguistics: ACL 2025},
  pages = {24290--24315},
  year = {2025},
  doi = {10.18653/v1/2025.findings-acl.1247}
}

@inproceedings{jiang2025mmecot,
  author = {Dongzhi Jiang and Renrui Zhang and Ziyu Guo and Yanwei Li and Yu Qi and Xinyan Chen and Liuhui Wang and Jianhan Jin and Claire Guo and Shen Yan and Bo Zhang and Chaoyou Fu and Peng Gao and Hongsheng Li},
  title = {{MME-CoT}: Benchmarking Chain-of-Thought in Large Multimodal Models for Reasoning Quality, Robustness, and Efficiency},
  booktitle = {Proceedings of the International Conference on Machine Learning},
  year = {2025}
}

@inproceedings{han2025videoespresso,
  author = {Songhao Han and Wei Huang and Hairong Shi and Le Zhuo and Xiu Su and Shifeng Zhang and Xu Zhou and Xiaojuan Qi and Yue Liao and Si Liu},
  title = {{VideoEspresso}: A Large-Scale Chain-of-Thought Dataset for Fine-Grained Video Reasoning via Core Frame Selection},
  booktitle = {Proceedings of the IEEE/CVF Conference on Computer Vision and Pattern Recognition},
  pages = {26181--26191},
  year = {2025},
  doi = {10.1109/CVPR52734.2025.02438}
}

@inproceedings{wang2026treebench,
  author = {Haochen Wang and Xiangtai Li and Zilong Huang and Anran Wang and Jiacong Wang and Tao Zhang and Jiani Zheng and Sule Bai and Zijian Kang and Jiashi Feng and Zhuochen Wang and Zhaoxiang Zhang},
  title = {Traceable Evidence Enhanced Visual Grounded Reasoning: Evaluation and Method},
  booktitle = {Proceedings of the International Conference on Learning Representations},
  year = {2026}
}

@inproceedings{tong2026videothinkbench,
  author = {Jingqi Tong and Yurong Mou and Hangcheng Li and Mingzhe Li and Yongzhuo Yang and Ming Zhang and Qiguang Chen and Tianyi Liang and Xiaomeng Hu and Yining Zheng and Xinchi Chen and Jun Zhao and Xuanjing Huang and Xipeng Qiu},
  title = {Thinking with Video: Video Generation as a Promising Multimodal Reasoning Paradigm},
  booktitle = {Proceedings of the IEEE/CVF Conference on Computer Vision and Pattern Recognition},
  pages = {41121--41129},
  year = {2026}
}

@inproceedings{shi2026mathcanvasbench,
  author = {Weikang Shi and Aldrich Yu and Rongyao Fang and Houxing Ren and Ke Wang and Aojun Zhou and Changyao Tian and Xinyu Fu and Yuxuan Hu and Zimu Lu and Linjiang Huang and Si Liu and Rui Liu and Hongsheng Li},
  title = {{MathCanvas}: Intrinsic Visual Chain-of-Thought for Multimodal Mathematical Reasoning},
  booktitle = {Proceedings of the Annual Meeting of the Association for Computational Linguistics},
  pages = {27933--27954},
  year = {2026},
  doi = {10.18653/v1/2026.acl-long.1287}
}

@inproceedings{chow2025physbench,
  author = {Wei Chow and Jiageng Mao and Boyi Li and Daniel Seita and Vitor Guizilini and Yue Wang},
  title = {{PhysBench}: Benchmarking and Enhancing Vision-Language Models for Physical World Understanding},
  booktitle = {Proceedings of the International Conference on Learning Representations},
  year = {2025}
}

@inproceedings{zhou2026paibench,
  author = {Fengzhe Zhou and Jiannan Huang and Jialuo Li and Deva Ramanan and Humphrey Shi},
  title = {{PAI-Bench}: A Comprehensive Benchmark For Physical {AI}},
  booktitle = {Proceedings of the IEEE/CVF Conference on Computer Vision and Pattern Recognition},
  pages = {21522--21536},
  year = {2026}
}

@inproceedings{puyin2026quantiphy,
  author = {Li Puyin and Tiange Xiang and Ella Mao and Shirley Wei and Xinye Chen and Adnan Masood and Fei-Fei Li and Ehsan Adeli},
  title = {{QuantiPhy}: A Quantitative Benchmark Evaluating Physical Reasoning Abilities of Vision-Language Models},
  booktitle = {Proceedings of the IEEE/CVF Conference on Computer Vision and Pattern Recognition},
  pages = {33174--33184},
  year = {2026}
}

@inproceedings{chen2024mllmasajudge,
  author = {Dongping Chen and Ruoxi Chen and Shilin Zhang and Yinuo Liu and Yaochen Wang and Huichi Zhou and Qihui Zhang and Yao Wan and Pan Zhou and Lichao Sun},
  title = {{MLLM-as-a-Judge}: Assessing Multimodal {LLM-as-a-Judge} with Vision-Language Benchmark},
  booktitle = {Proceedings of the International Conference on Machine Learning},
  pages = {6562--6595},
  year = {2024}
}

@inproceedings{luo2026editrewardbench,
  author = {Xin Luo and Jiahao Wang and Chenyuan Wu and Shitao Xiao and Xiyan Jiang and Defu Lian and Jiajun Zhang and Dong Liu and Zheng Liu},
  title = {{EditScore}: Unlocking Online {RL} for Image Editing via High-Fidelity Reward Modeling},
  booktitle = {Proceedings of the International Conference on Learning Representations},
  year = {2026}
}

@inproceedings{wu2026editrewardbench,
  author = {Keming Wu and Sicong Jiang and Max Ku and Ping Nie and Minghao Liu and Wenhu Chen},
  title = {{EditReward}: A Human-Aligned Reward Model for Instruction-Guided Image Editing},
  booktitle = {Proceedings of the International Conference on Learning Representations},
  year = {2026}
}

@inproceedings{pang2025paper2poster,
  author = {Wei Pang and Kevin Qinghong Lin and Xiangru Jian and Xi He and Philip Torr},
  title = {{Paper2Poster}: Towards Multimodal Poster Automation from Scientific Papers},
  booktitle = {Advances in Neural Information Processing Systems (Datasets and Benchmarks Track)},
  year = {2025}
}

@inproceedings{liu2024interleavedbench,
  author = {Minqian Liu and Zhiyang Xu and Zihao Lin and Trevor Ashby and Joy Rimchala and Jiaxin Zhang and Lifu Huang},
  title = {Holistic Evaluation for Interleaved Text-and-Image Generation},
  booktitle = {Proceedings of the Conference on Empirical Methods in Natural Language Processing},
  pages = {22002--22016},
  year = {2024},
  doi = {10.18653/v1/2024.emnlp-main.1228}
}

@inproceedings{yu2025mramgbench,
  author = {Qinhan Yu and Zhiyou Xiao and Binghui Li and Zhengren Wang and Chong Chen and Wentao Zhang},
  title = {{MRAMG-Bench}: A Comprehensive Benchmark for Advancing Multimodal Retrieval-Augmented Multimodal Generation},
  booktitle = {Proceedings of the International ACM SIGIR Conference on Research and Development in Information Retrieval},
  pages = {3616--3626},
  year = {2025},
  doi = {10.1145/3726302.3730288}
}

@inproceedings{si2025design2code,
  author = {Chenglei Si and Yanzhe Zhang and Ryan Li and Zhengyuan Yang and Ruibo Liu and Diyi Yang},
  title = {{Design2Code}: Benchmarking Multimodal Code Generation for Automated Front-End Engineering},
  booktitle = {Proceedings of the Conference of the Nations of the Americas Chapter of the Association for Computational Linguistics: Human Language Technologies},
  pages = {3956--3974},
  year = {2025},
  doi = {10.18653/v1/2025.naacl-long.199}
}

@inproceedings{yin2026blenderbench,
  author = {Shaofeng Yin and Jiaxin Ge and Zora Zhiruo Wang and Chenyang Wang and Xiuyu Li and Michael J. Black and Trevor Darrell and Angjoo Kanazawa and Haiwen Feng},
  title = {Vision-as-Inverse-Graphics Agent via Interleaved Multimodal Reasoning},
  booktitle = {Proceedings of the European Conference on Computer Vision},
  pages = {557--575},
  year = {2026},
  doi = {10.1007/978-3-032-37464-6_30}
}

@inproceedings{yang2025chartmimic,
  author = {Cheng Yang and Chufan Shi and Yaxin Liu and Bo Shui and Junjie Wang and Mohan Jing and Linran Xu and Xinyu Zhu and Siheng Li and Yuxiang Zhang and Gongye Liu and Xiaomei Nie and Deng Cai and Yujiu Yang},
  title = {{ChartMimic}: Evaluating {LMM's} Cross-Modal Reasoning Capability via Chart-to-Code Generation},
  booktitle = {Proceedings of the International Conference on Learning Representations},
  year = {2025}
}

@inproceedings{ding2025mmifeval,
  author = {Shengyuan Ding and Shenxi Wu and Xiangyu Zhao and Yuhang Zang and Haodong Duan and Xiaoyi Dong and Pan Zhang and Yuhang Cao and Dahua Lin and Jiaqi Wang},
  title = {{MM-IFEngine}: Towards Multimodal Instruction Following},
  booktitle = {Proceedings of the IEEE/CVF International Conference on Computer Vision},
  pages = {1099--1109},
  year = {2025},
  doi = {10.1109/ICCV51701.2025.00110}
}

@inproceedings{niu2026wise,
  author = {Yuwei Niu and Munan Ning and Mengren Zheng and Weiyang Jin and Bin Lin and Peng Jin and Jiaqi Liao and Chaoran Feng and Fanqing Meng and Kunpeng Ning and Bin Zhu and Li Yuan},
  title = {{WISE}: World Knowledge-Informed Semantic Evaluation for Text-to-Image Generation},
  booktitle = {Proceedings of the International Conference on Machine Learning},
  year = {2026}
}

@inproceedings{zhao2025risebench,
  author = {Xiangyu Zhao and Peiyuan Zhang and Kexian Tang and Xiaorong Zhu and Hao Li and Wenhao Chai and Zicheng Zhang and Renqiu Xia and Guangtao Zhai and Junchi Yan and Hua Yang and Xue Yang and Haodong Duan},
  title = {Envisioning Beyond the Pixels: Benchmarking Reasoning-Informed Visual Editing},
  booktitle = {Advances in Neural Information Processing Systems (Datasets and Benchmarks Track)},
  year = {2025}
}

@inproceedings{feng2026knowgen,
  author = {Kaituo Feng and Manyuan Zhang and Shuang Chen and Yunlong Lin and Kaixuan Fan and Yilei Jiang and Hongyu Li and Dian Zheng and Chenyang Wang and Xiangyu Yue},
  title = {Gen-Searcher: Reinforcing Agentic Search for Image Generation},
  booktitle = {Advances in Neural Information Processing Systems},
  year = {2026}
}

@inproceedings{lin2026fullduplexbenchv2,
  author = {Guan-Ting Lin and Shih-Yun Shan Kuan and Jiatong Shi and Kai-Wei Chang and Siddhant Arora and Shinji Watanabe and Hung-Yi Lee},
  title = {Full-Duplex-Bench-v2: A Multi-Turn Evaluation Framework for Duplex Dialogue Systems with an Automated Examiner},
  booktitle = {Proceedings of the Annual Meeting of the Association for Computational Linguistics},
  pages = {27--36},
  year = {2026},
  doi = {10.18653/v1/2026.acl-short.4}
}

@inproceedings{arora2025talkingturns,
  author = {Siddhant Arora and Zhiyun Lu and Chung-Cheng Chiu and Ruoming Pang and Shinji Watanabe},
  title = {Talking Turns: Benchmarking Audio Foundation Models on Turn-Taking Dynamics},
  booktitle = {Proceedings of the International Conference on Learning Representations},
  year = {2025}
}

@inproceedings{lin2026fullduplexbenchv15,
  author = {Guan-Ting Lin and Shih-Yun Shan Kuan and Qirui Wang and Jiachen Lian and Tingle Li and Shinji Watanabe and Hung-Yi Lee},
  title = {Full-Duplex-Bench v1.5: Evaluating Overlap Handling for Full-Duplex Speech Models},
  booktitle = {Proceedings of the IEEE International Conference on Acoustics, Speech and Signal Processing},
  year = {2026}
}

@inproceedings{liu2025rhbench,
  author = {Zhongxing Xu and Chengzhi Liu and Qingyue Wei and Juncheng Wu and James Y. Zou and Xin Wang and Yuyin Zhou and Sheng Liu},
  title = {More Thinking, Less Seeing? {Assessing} Amplified Hallucination in Multimodal Reasoning Models},
  booktitle = {Advances in Neural Information Processing Systems},
  year = {2025}
}

@inproceedings{mickus2026sheep,
  author = {Timothee Mickus and Claudio Savelli and Eduardo Calo and Emilio Raimond and Stella Frank and Hengyu Luo and Flavio Giobergia and Vincent Segonne and Chuyuan Li and Aman Sinha and Lorenzo Vaiani and Jorg Tiedemann and Raul Vazquez},
  title = {Can Humans Dream of Electric Sheep? {Human-Written} Samples for Fine-Grained Vision-and-Language Hallucination Benchmarking},
  booktitle = {Proceedings of the Conference on Empirical Methods in Natural Language Processing},
  pages = {30009--30033},
  year = {2026},
  doi = {10.18653/2026.emnlp-main.1318}
}

@inproceedings{cheng2025simplevqa,
  author = {Xianfu Cheng and Wei Zhang and Shiwei Zhang and Jian Yang and Xiangyuan Guan and Xianjie Wu and Xiang Li and Ge Zhang and Jiaheng Liu and Yuying Mai and Yutao Zeng and Zhoufutu Wen and Ke Jin and Baorui Wang and Weixiao Zhou and Yunhong Lu and Tongliang Li and Wenhao Huang and Zhoujun Li},
  title = {{SimpleVQA}: Multimodal Factuality Evaluation for Multimodal Large Language Models},
  booktitle = {Proceedings of the IEEE/CVF International Conference on Computer Vision},
  pages = {4637--4646},
  year = {2025},
  doi = {10.1109/ICCV51701.2025.00441}
}

@article{ying2026safebench,
  author = {Zonghao Ying and Aishan Liu and Siyuan Liang and Lei Huang and Jinyang Guo and Wenbo Zhou and Xianglong Liu and Dacheng Tao},
  title = {{SafeBench}: A Safety Evaluation Framework for Multimodal Large Language Models},
  journal = {International Journal of Computer Vision},
  volume = {134},
  number = {1},
  pages = {18},
  year = {2026},
  doi = {10.1007/s11263-025-02613-1}
}

@inproceedings{ying2026safe,
  author = {Zonghao Ying and Le Wang and Yisong Xiao and Jiakai Wang and Yuqing Ma and Jinyang Guo and Zhenfei Yin and Mingchuan Zhang and Aishan Liu and Xianglong Liu},
  title = {{AGENTSAFE}: Benchmarking the Safety of Embodied Agents on Hazardous Instructions},
  booktitle = {Proceedings of the IEEE/CVF Conference on Computer Vision and Pattern Recognition},
  pages = {37664--37673},
  year = {2026}
}

@inproceedings{liu2025mllmubench,
  author = {Zheyuan Liu and Guangyao Dou and Mengzhao Jia and Zhaoxuan Tan and Qingkai Zeng and Yongle Yuan and Meng Jiang},
  title = {Protecting Privacy in Multimodal Large Language Models with {MLLMU-Bench}},
  booktitle = {Proceedings of the Conference of the Nations of the Americas Chapter of the Association for Computational Linguistics: Human Language Technologies},
  pages = {4105--4135},
  year = {2025},
  doi = {10.18653/v1/2025.naacl-long.207}
}

@inproceedings{liu2025mmfakebench,
  author = {Xuannan Liu and Zekun Li and Peipei Li and Huaibo Huang and Shuhan Xia and Xing Cui and Linzhi Huang and Weihong Deng and Zhaofeng He},
  title = {{MMFakeBench}: A Mixed-Source Multimodal Misinformation Detection Benchmark for {LVLMs}},
  booktitle = {Proceedings of the International Conference on Learning Representations},
  year = {2025}
}

@inproceedings{groot2024jus,
  author = {Tobias Groot and Matias Valdenegro-Toro},
  title = {Overconfidence is Key: Verbalized Uncertainty Evaluation in Large Language and Vision-Language Models},
  booktitle = {Proceedings of the Workshop on Trustworthy Natural Language Processing (TrustNLP)},
  pages = {145--171},
  year = {2024},
  doi = {10.18653/v1/2024.trustnlp-1.13}
}

@inproceedings{ye2025loki,
  author = {Junyan Ye and Baichuan Zhou and Zilong Huang and Junan Zhang and Tianyi Bai and Hengrui Kang and Jun He and Honglin Lin and Zihao Wang and Tong Wu and Zhizheng Wu and Yiping Chen and Dahua Lin and Conghui He and Weijia Li},
  title = {{LOKI}: A Comprehensive Synthetic Data Detection Benchmark using Large Multimodal Models},
  booktitle = {Proceedings of the International Conference on Learning Representations},
  year = {2025}
}

@inproceedings{cheng2025jailbreakaudiobenc,
  author = {Hao Cheng and Erjia Xiao and Jing Shao and Yichi Wang and Le Yang and Chao Shen and Philip Torr and Jindong Gu and Renjing Xu},
  title = {{Jailbreak-AudioBench}: In-Depth Evaluation and Analysis of Jailbreak Threats for Large Audio Language Models},
  booktitle = {Advances in Neural Information Processing Systems (Datasets and Benchmarks Track)},
  year = {2025}
}

@inproceedings{xie2024osworld,
  author = {Tianbao Xie and Danyang Zhang and Jixuan Chen and Xiaochuan Li and Siheng Zhao and Ruisheng Cao and Toh Jing Hua and Zhoujun Cheng and Dongchan Shin and Fangyu Lei and Yitao Liu and Yiheng Xu and Shuyan Zhou and Silvio Savarese and Caiming Xiong and Victor Zhong and Tao Yu},
  title = {{OSWorld}: Benchmarking Multimodal Agents for Open-Ended Tasks in Real Computer Environments},
  booktitle = {Advances in Neural Information Processing Systems (Datasets and Benchmarks Track)},
  year = {2024}
}

@inproceedings{cheng2024screenspot,
  author = {Kanzhi Cheng and Qiushi Sun and Yougang Chu and Fangzhi Xu and Yantao Li and Jianbing Zhang and Zhiyong Wu},
  title = {{SeeClick}: Harnessing {GUI} Grounding for Advanced Visual {GUI} Agents},
  booktitle = {Proceedings of the Annual Meeting of the Association for Computational Linguistics},
  pages = {9313--9332},
  year = {2024},
  doi = {10.18653/v1/2024.acl-long.505}
}

@inproceedings{rawles2025androidworld,
  author = {Christopher Rawles and Sarah Clinckemaillie and Yifan Chang and Jonathan Waltz and Gabrielle Lau and Marybeth Fair and Alice Li and William Bishop and Wei Li and Folawiyo Campbell-Ajala and Daniel Toyama and Robert Berry and Divya Tyamagundlu and Timothy Lillicrap and Oriana Riva},
  title = {{AndroidWorld}: A Dynamic Benchmarking Environment for Autonomous Agents},
  booktitle = {Proceedings of the International Conference on Learning Representations},
  year = {2025}
}

@inproceedings{li2025screenspotpro,
  author = {Kaixin Li and Ziyang Meng and Hongzhan Lin and Ziyang Luo and Yuchen Tian and Jing Ma and Zhiyong Huang and Tat-Seng Chua},
  title = {{ScreenSpot-Pro}: {GUI} Grounding for Professional High-Resolution Computer Use},
  booktitle = {Proceedings of the ACM International Conference on Multimedia},
  pages = {8778--8786},
  year = {2025},
  doi = {10.1145/3746027.3755688}
}

@inproceedings{he2024webvoyager,
  author = {Hongliang He and Wenlin Yao and Kaixin Ma and Wenhao Yu and Yong Dai and Hongming Zhang and Zhenzhong Lan and Dong Yu},
  title = {{WebVoyager}: Building an End-to-End Web Agent with Large Multimodal Models},
  booktitle = {Proceedings of the Annual Meeting of the Association for Computational Linguistics},
  pages = {6864--6890},
  year = {2024},
  doi = {10.18653/v1/2024.acl-long.371}
}

@inproceedings{drouin2024workarena,
  author = {Alexandre Drouin and Maxime Gasse and Massimo Caccia and Issam H. Laradji and Manuel Del Verme and Tom Marty and David V{\'a}zquez and Nicolas Chapados and Alexandre Lacoste},
  title = {{WorkArena}: How Capable Are Web Agents at Solving Common Knowledge Work Tasks?},
  booktitle = {Proceedings of the International Conference on Machine Learning},
  pages = {11642--11662},
  year = {2024}
}

@inproceedings{evtimov2025wasp,
  author = {Ivan Evtimov and Arman Zharmagambetov and Aaron Grattafiori and Chuan Guo and Kamalika Chaudhuri},
  title = {{WASP}: Benchmarking Web Agent Security Against Prompt Injection Attacks},
  booktitle = {Advances in Neural Information Processing Systems (Datasets and Benchmarks Track)},
  year = {2025}
}

@inproceedings{wang2025agentnetbench,
  author = {Xinyuan Wang and Bowen Wang and Dunjie Lu and Junlin Yang and Tianbao Xie and Junli Wang and Jiaqi Deng and Xiaole Guo and Yiheng Xu and Chen Henry Wu and Zhennan Shen and Zhuokai Li and Ryan Li and Xiaochuan Li and Junda Chen and Boyuan Zheng and Peihang Li and Fangyu Lei and Ruisheng Cao and Yeqiao Fu and Dongchan Shin and Martin Shin and Jiarui Hu and Yuyan Wang and Jixuan Chen and Yuxiao Ye and Danyang Zhang and Yipu Wang and Heng Wang and Diyi Yang and Victor Zhong and Y. Charles and Zhilin Yang and Tao Yu},
  title = {{OpenCUA}: Open Foundations for Computer-Use Agents},
  booktitle = {Advances in Neural Information Processing Systems},
  year = {2025}
}

@inproceedings{yang2025embodiedbench,
  author = {Rui Yang and Hanyang Chen and Junyu Zhang and Mark Zhao and Cheng Qian and Kangrui Wang and Qineng Wang and Teja Venkat Koripella and Marziyeh Movahedi and Manling Li and Heng Ji and Huan Zhang and Tong Zhang},
  title = {{EmbodiedBench}: Comprehensive Benchmarking Multi-modal Large Language Models for Vision-Driven Embodied Agents},
  booktitle = {Proceedings of the International Conference on Machine Learning},
  year = {2025}
}

@inproceedings{cheng2024vlnceisaac,
  author = {An-Chieh Cheng and Yandong Ji and Zhaojing Yang and Zaitian Gongye and Xueyan Zou and Jan Kautz and Erdem Byk and Hongxu Yin and Sifei Liu and Xiaolong Wang},
  title = {{NaVILA}: Legged Robot Vision-Language-Action Model for Navigation},
  booktitle = {Proceedings of Robotics: Science and Systems},
  year = {2025}
}

@inproceedings{dai2026robomme,
  author = {Yinpei Dai and Hongze Fu and Jayjun Lee and Yuejiang Liu and Haoran Zhang and Jianing Yang and Chelsea Finn and Nima Fazeli and Joyce Chai},
  title = {{RoboMME}: Benchmarking and Understanding Memory for Robotic Generalist Policies},
  booktitle = {Proceedings of the International Conference on Machine Learning},
  year = {2026}
}

@inproceedings{liu2025bench,
  author = {Jijia Liu and Feng Gao and Bingwen Wei and Xinlei Chen and Qingmin Liao and Yi Wu and Chao Yu and Yu Wang},
  title = {What Can {RL} Bring to {VLA} Generalization? {An} Empirical Study},
  booktitle = {Advances in Neural Information Processing Systems},
  year = {2025}
}
